\documentclass[10pt,twocolumn,letterpaper]{article}

\usepackage[pagenumbers]{cvpr}
\usepackage[dvipsnames]{xcolor}
\usepackage{minitoc}
\usepackage{nicematrix}

\usepackage{multirow,bigdelim}

\definecolor{cvprblue}{rgb}{0.21,0.49,0.74}
\usepackage[pagebackref,breaklinks,colorlinks,citecolor=cvprblue]{hyperref}

\newcommand{\ap}[1]{``#1''}

\def\confName{CVPR}
\def\confYear{2024}

\title{\vspace{-2cm}Implicit Style-Content Separation using B-LoRA\vspace{-0.5cm}}

\author{
        Yarden Frenkel$^1$ \hspace{0.05\linewidth}
        Yael Vinker$^1$ \hspace{0.05\linewidth}
        Ariel Shamir$^2$ \hspace{0.05\linewidth}
        Daniel Cohen-Or$^1$
        \\[10pt]
        $^1$Tel Aviv University \hspace{0.05\linewidth} $^2$Reichman University
         \\[7pt]
         \url{https://B-LoRA.github.io/B-LoRA/}
         \\[-20pt]
}

\begin{document}

\doparttoc %
\faketableofcontents %
\twocolumn[{%
\maketitle
\renewcommand\twocolumn[1][]{#1}%
\begin{center}

     \setlength{\abovecaptionskip}{0.15cm}
     \setlength{\belowcaptionskip}{0pt}
    \centering
    \includegraphics[width=0.88\textwidth]{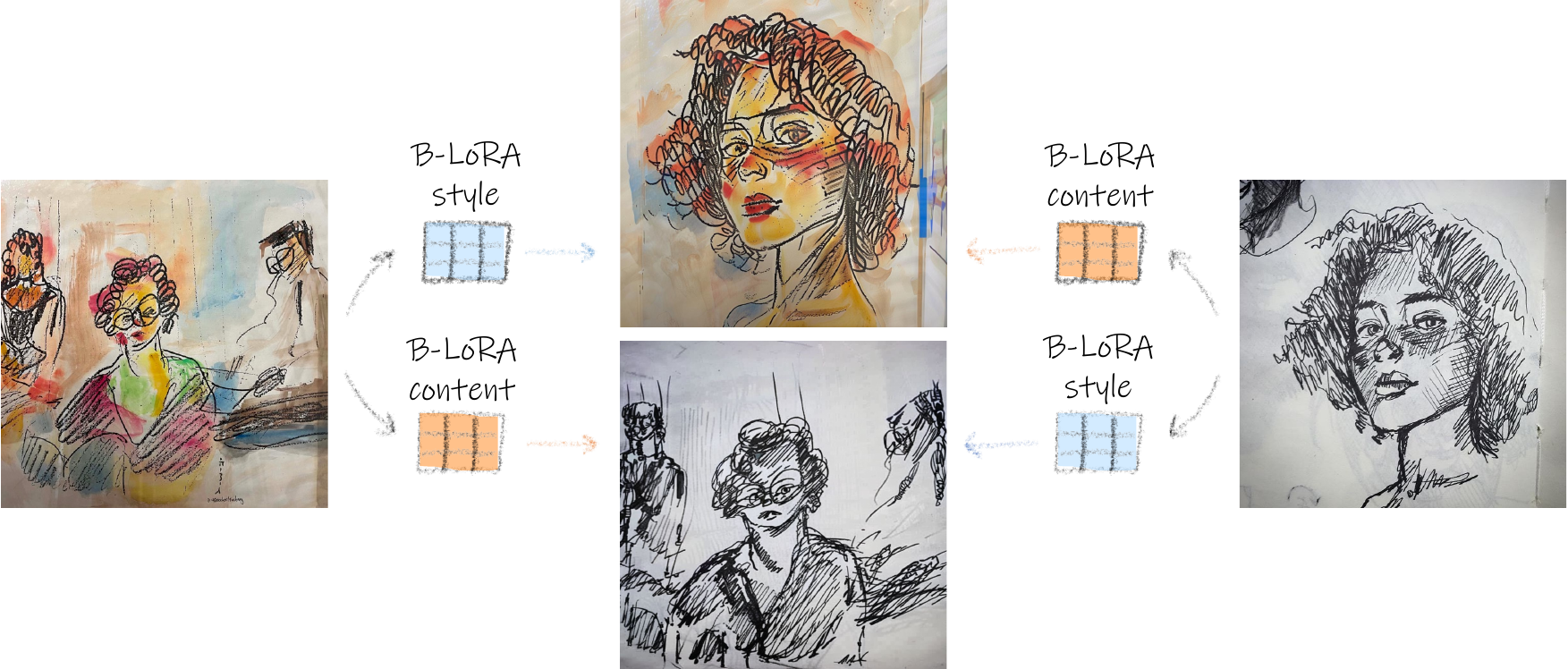}
    \captionsetup{type=figure}\caption{By implicitly decomposing a single image into its style and content representation captured by B-LoRA, we can perform high quality style-content mixing and even swapping the style and content between two stylized images.\copyright The painting on the left is by Judith Kondor Mochary.}    
    \label{fig:teaser}
\end{center}
}]

\maketitle

\begin{abstract} \vspace{-9pt}
Image stylization involves manipulating the visual appearance and texture (style) of an image while preserving its underlying objects, structures, and concepts (content). The separation of style and content is essential for manipulating the image's style independently from its content, ensuring a harmonious and visually pleasing result. Achieving this separation requires a deep understanding of both the visual and semantic characteristics of images, often necessitating the training of specialized models or employing heavy optimization.
In this paper, we introduce B-LoRA, a method that leverages LoRA (Low-Rank Adaptation) to \textit{implicitly} separate the style and content components of a \textit{single} image, facilitating various image stylization tasks. 
By analyzing the architecture of SDXL combined with LoRA, we find that jointly learning the LoRA weights of two specific blocks (referred to as B-LoRAs) achieves style-content separation that cannot be achieved by training each B-LoRA independently.
Consolidating the training into only two blocks and separating style and content allows for significantly improving style manipulation and overcoming overfitting issues often associated with model fine-tuning. 
Once trained, the two B-LoRAs can be used as independent components to allow various image stylization tasks, including image style transfer, text-based image stylization, consistent style generation, and style-content mixing.

\end{abstract}

\begin{figure*}
    \centering
    \includegraphics[width=1\linewidth]{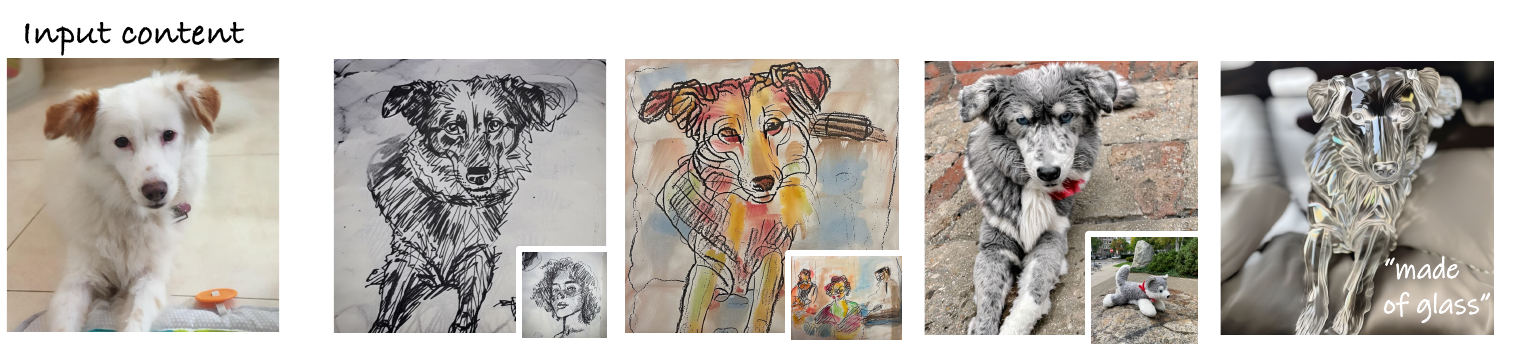}
    \vspace{-0.6cm}
    \caption{Examples of image stylization generated with our approach. The content image is shown on the left. We show here three results of image style transfer based on a reference style, one (on the right) based on a guiding text prompt. Note that our method requires only a single image, and preserves the image's content and structure well while applying the desired style.}
    \vspace{-0.3cm}
    \label{fig:teaser2}
\end{figure*}

\section{Introduction}~\label{sec:intro}
Image stylization is a well-established task in computer vision, and has been actively researched for many years \cite{ImageAnalogies,Gatys2016}.
This task involves changing the style of an image following some style reference, which can be text-based or image-based while preserving its content. 
Content refers to the semantic information and structure of the image, while style often refers to visual features and patterns such as colors and textures \cite{Freeman96}.
Image style manipulation is a highly challenging task, since style and content are strongly connected, leading to an inherent trade-off between style transformation and content preservation.
On the other hand, many style manipulation tasks require a clear separation between style and content within an image. 

In this paper, we present B-LoRA, a method for style-content separation of any given image. Our method distills the style and content from a single image to support various style manipulation applications.

In the realm of recent advancements in large language-vision models, existing approaches utilize the strong visual-semantic priors embedded within these models to facilitate style manipulation tasks. 
Common techniques involve fine-tuning a pre-trained text-to-image model to account for a new style or content \cite{dreambooth23,LoRA21,SVDiff2023,arar2024Palp}. However, fine-tuned models often suffer from the inherent trade-off between style transformation and content preservation as they are prone to over-fitting.  
Unlike these methods, we unify the learning of style and content components by separating them per image (see \Cref{fig:teaser}). This separation is performed by fitting a light-weight adapter (B-LoRA) that is less prone to over-fitting issues, and enables task flexibility, allowing for both text-based and reference style image conditions. 

Our method utilizes LoRA (Low Rank Adaptation) \cite{LoRA21}, which has emerged as a popular approach due to its high-quality results and space-time efficiency. LoRA incorporates optimizing external low-rank weight matrices for the attention layers of the base model, while the pretrained model weights remain ``frozen''. After training, these matrices define the adapted model that can be used for the desired task.
LoRA is often utilized for image stylization by fine-tuning the base model with respect to a set of images that can either represent the desired style or the desired content.

Specifically, we use LoRA with Stable Diffusion XL (SDXL) \cite{sdxl2023}, a recently introduced text-to-image diffusion model renowned for its powerful style learning capabilities. Through detailed analysis of various layers within SDXL and their effect on the adaptation procedure, we made a surprising discovery: two specific transformer blocks can be used to separate the style and content of an input image, and to easily control them distinctly in generated images. For clarification, in this paper, we define a block as a sequence of 10 consecutive attention layers.

Therefore, when provided with a \textit{single} input image, we jointly optimize the LoRA weights corresponding to these two distinct transformer blocks with the objective of reconstructing the given image based on a provided text prompt. 
Since we only optimize the LoRA weights of these two transformer blocks, we refer to them as ``B-LoRAs''. 
The crucial aspect is that these B-LoRAs are trained on a single image only, yet they successfully disentangle its style and content, thereby circumventing the notorious overfitting problem associated with common LoRA techniques. Our technique benefits from the innate style-content disentanglement within the layers of the architecture.
Another advantage of our method is that the B-LoRAs can be easily used as separate components, allowing various challenging style manipulation tasks without requiring any additional training or fine-tuning. In particular, we demonstrate style transfer, text-guided style manipulation and consistent style-conditioned image generation (see \Cref{fig:teaser2}).

We note that recent attempts have been made to combine trained LoRAs of style and content to a unified model \cite{ziplora23}. This approach requires a new optimization process for each style-content combination.
This is both time-consuming and raises challenges in achieving an effective trade-off between style transformation and content preservation.
In contrast, 
our trained B-LoRAs can be easily re-plugged into a pre-trained model combined with other learned blocks from other reference images, without any further training.

We provide extensive evaluation of our method showing its advantages compared to alternative approaches that are often designed to achieve one of these tasks.
Our method provides a practical and simple way for image stylization that can be broadly used with existing models.
\vspace{10pt}

\section{Related Work} \vspace{5pt}

\paragraph{\textbf{Style Transfer}}
Image style transfer is a longstanding challenge in computer vision \cite{Efros2001ImageQF, ImageAnalogies},
aimed at altering the style of an image based on a given reference.
With the progress of deep learning research, Neural Style Transfer (NST) approaches rely on deep features extracted from pre-trained networks to merge content and style \cite{Gatys2016,NeuralST2017,Johnson2016PerceptualLF}.
Subsequent GAN-based \cite{GAN2014} techniques were proposed to transfer images across domains, using either paired \cite{ImagetoImage2016} or unpaired \cite{CycleGAN2017,CrossDomainCD2020,ContrastiveL2020} image sets, yet they require domain-specific datasets and training.

Recent advancements in language-vision models and diffusion models have revolutionized the field of image stylization. 
Leveraging the vast knowledge encoded in pre-trained language-vision models, modern approaches explore zero-shot image stylization and editing \cite{DMokadyHAPC23,
Li2023StyleDiffusionPI, Yang2023ZeroShotCL, Chen2023TrainingFreeLC, Epstein2023DiffusionSF, Couairon2022DiffEditDS, Parmar2023ZeroshotIT, Avrahami2022BlendedLD}, where images are manipulated without additional fine-tuning or data adaptation by intervening in the generation process. 
Prompt-to-Prompt \cite{Hertz2022PrompttoPromptIE} proposes an approach to edit generated images by manipulating their cross-attention maps. In Plug-and-Play \cite{PlugandPlay2022} the appearance of a content image is manipulated with respect to a given text prompt by adjusting spatial features from the guidance image via the self-attention mechanism. Cross Image Attention (CIA) \cite{CrossImageA2023} presents a method to modify the image appearance based on a reference image through alterations in cross-attention mechanisms. While these approaches effectively transform the appearance of the content image, they may encounter challenges in transferring appearance between subjects with differing semantics.

StyleAligned \cite{styleHertz23} utilizes attention features sharing combined with the AdaIN mechanism \cite{AdaIN2017} to achieve style alignment between a sequence of generated images. However, the method is not explicitly designed to control the content of the generated image, potentially resulting in style image structure leakage. Similarly, the lack of style-content separation is also evident in encoder-based methods, such as IP-Adapter \cite{IPAdapter2023}. InstantStyle \cite{InstantStyle2024} is a concurrent work to ours, aiming to improve IP-Adapter for image stylization tasks by injecting the CLIP embedding of the style image into specific blocks within SDXL. In our work, we decompose the style and the content and learn a separate representation for each.

\paragraph{\textbf{Text-to-Image Personalization}}
In another line of work \cite{TI22, dreambooth23, SVDiff2023, NeTIf23, p+23,arar2024Palp}, optimization techniques are proposed to extend pre-trained Text-to-Image models to support the generation of novel visual concepts, including both style and content, based on a small set of input images with the same concept. This allows utilizing the rich semantic-visual prior of pre-trained models for customized tasks such as producing images of a desired style.
Existing methods employ either token optimization techniques \cite{TI22,InST2022, Prospect23, p+23, treeVinker23, MATTE2023}, fine-tuning the model's weights \cite{dreambooth23}, or a combination of both \cite{NeTIf23,arar2024Palp, avrahami2023chosen, avrahami2023bas}. 
Token optimization requires longer training times and often results in sub-optimal reconstruction. While model fine-tuning provides better reconstruction, it consumes substantial memory and tends to overfit.
To address the memory inefficiency, and to facilitate more efficient model fine-tuning, Parameter Efficient Fine-Tuning (PEFT) approaches have been proposed \cite{Houlsby2019ParameterEfficientTL, LoRA21, customdiffusion2022}. 
StyleDrop \cite{sohn2023styledrop} utilizes Muse \cite{Muse2023} as a base model, and adjusts its styles to align with a reference image. StyleDrop trains a lightweight adapter layer at the end of each attention block within the transformer model. However, similar to StyleAligned \cite{styleHertz23}, their approach is designed for style adaptation, but for content preservation, another optimization is required.
Among existing PEFT methods, Low-Rank Adaptation (LoRA) \cite{LoRA21} is a popular fine-tuning technique, widely used by researchers and practitioners for its versatility and high-quality results. 

\paragraph{\textbf{LoRA for Image Stylization}}
LoRA is often used for image stylization by fine-tuning a model to produce images of a desired style. Commonly, a LoRA is trained on a set of images, and then it is combined with control methods such as stylistic Concept-Sliders \cite{ConceptSlider2023} or ControlNet \cite{Zhang2023AddingCC, CtrlLoRA}, along with a text prompt to condition the generated image content. 
While LoRA-based approaches have demonstrated significant abilities in capturing style and content, two separate LoRA models are required for this task, and there is no trivial way to combine them. 
A common na\"ive approach is to combine two LoRAs by directly interpolating their weights \cite{DBlora}, relying on a manual search for the desired coefficients. 
Alternative approaches \cite{mixofshow2023, orthogonal2023} propose an optimization-based strategy to find the optimal coefficients for such a combination. However, they focus on combining two objects and not on image stylization tasks.

Recently, Shah et al. introduced ZipLoRA \cite{ziplora23}, proposing to merge two individual LoRAs trained for style and content into a new 'zipped' LoRA by learning mixing coefficients for their columns. 
This work is closely related to ours, as we also mix LoRA weights trained on different images to facilitate image stylization. 
However, ZipLoRA requires an additional optimization stage for each new combination of content and style, thereby restricting the flexibility of reusing trained LoRA weights, which is LoRA's primary advantage.
In contrast, our approach allows for the direct reuse of learned styles and contents without additional training, enhancing efficiency and versatility. Moreover, we demonstrate that our implicit approach is more robust to challenging styles and contents.

\vspace{10pt}
\section{Preliminaries}

\paragraph{\textbf{SDXL Architecture}}
\begin{figure}[t]%
  \centering\includegraphics[width=0.98\linewidth]{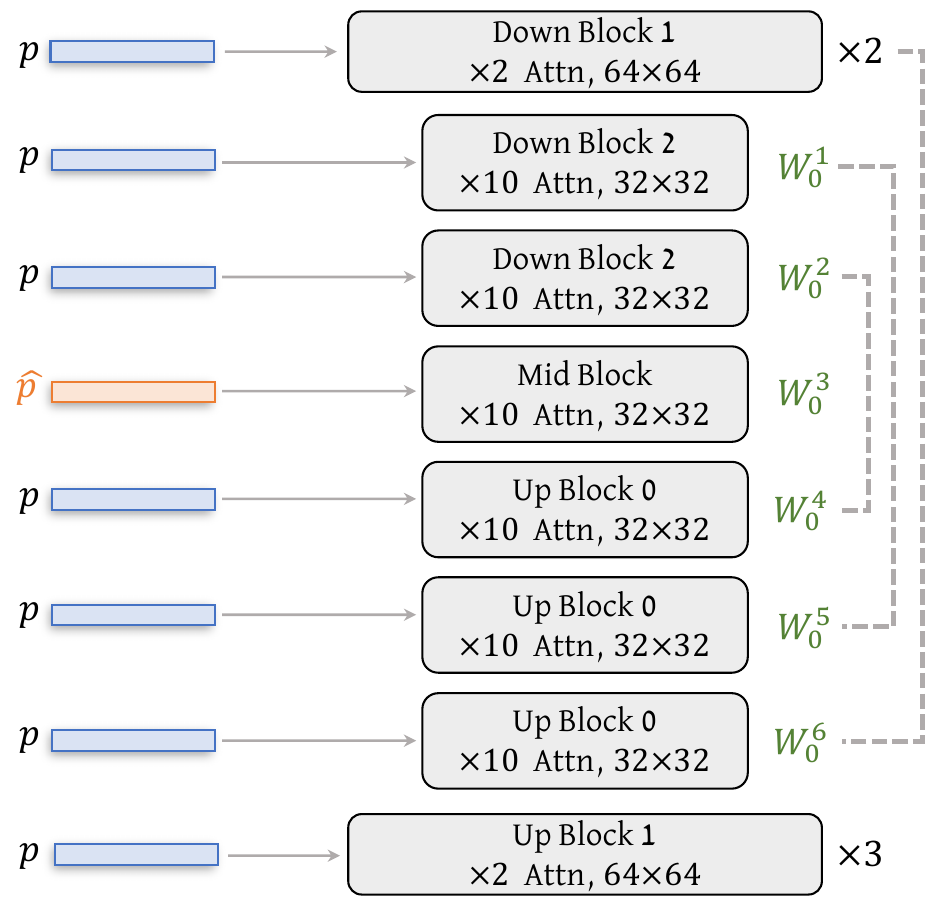}
  \caption{Illustration of SDXL architecture and our text-based analysis. To examine the effect of the i'th transformer block on the generated image, we inject a different text prompt $\hat{p}$ to it, while $p$ is injected into all other blocks.}
  \label{fig:prompt_inject_illustration}\vspace{-10pt}
\end{figure} 

\begin{figure*}[h]
    \centering
    \includegraphics[width=0.81\linewidth]{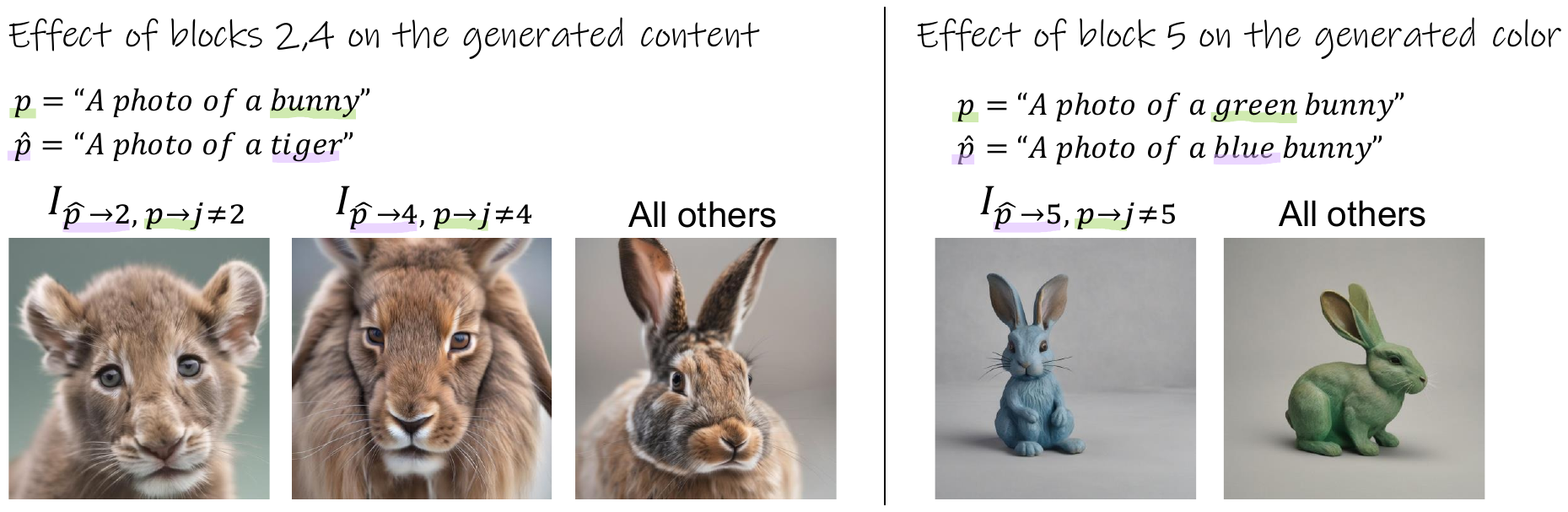}
    \caption{Prompt injection effect on the generated image. On the left, we demonstrate how blocks 2 and 4 affect the content in the generated image (turning into a tiger), whereas the rightmost image shows that injecting $\hat{p}$ to a block $i \neq 2,4$ has no effect on the generated image. On the right we show how the fifth block controls the generated image's color.}
    \vspace{-0.3cm}
    \label{fig:injection_vis}
\end{figure*}

\begin{figure*}[t]
    \centering
    \includegraphics[width=0.8\linewidth]{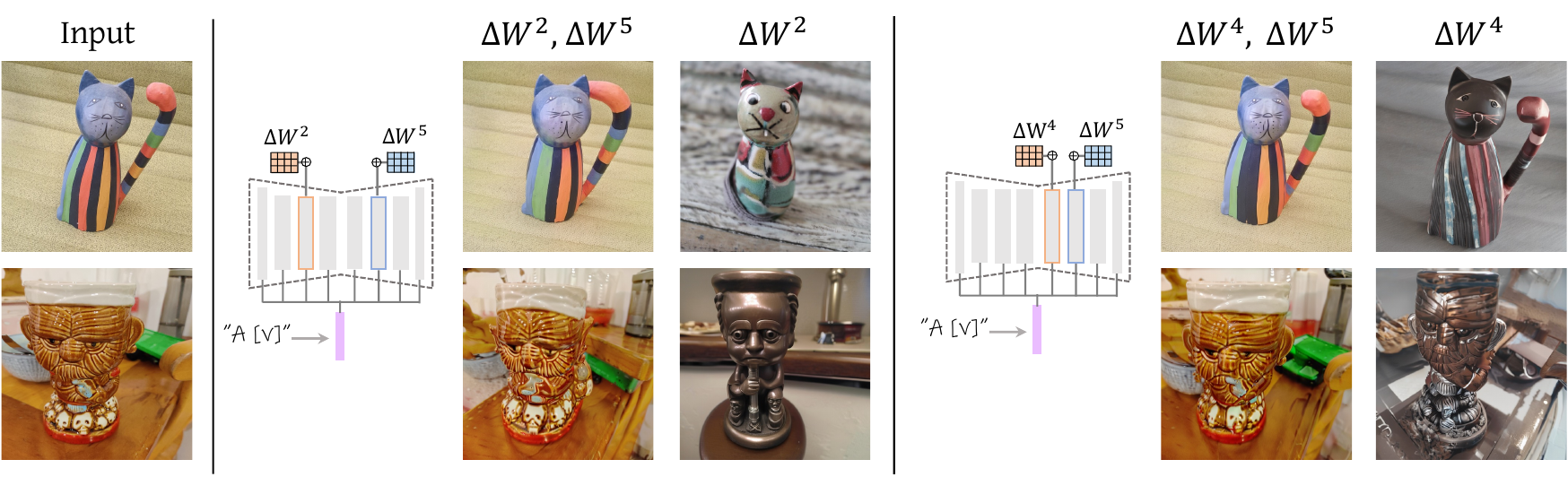}
    \caption{Comparison of training B-LoRAs for the input images shown on the left for $W_0^2,W_0^5$ (middle) and $W_0^4,W_0^5$ (right).
    For each pair of trained LoRA weights, we show the results of applying both together (to reconstruct the input image) and applying the content layer separately (i.e. using only $\Delta W^2$ and $\Delta W^4$). The results demonstrate that $\Delta W^4$ better captures the fine details of the input object.}
    \vspace{-0.5cm}
    \label{fig:lora_subset_training}
\end{figure*}
In our work, we utilize the recently introduced publicly available text-to-image Stable Diffusion XL (SDXL) \cite{sdxl2023}, which is an upgraded version of the known Stable Diffusion \cite{stabledifussion2021}. Both models are types of latent diffusion models (LDM), where the diffusion process is applied in the latent space of a pre-trained image autoencoder. 
The SDXL architecture leverages a three times larger UNet backbone compared to Stable Diffusion. The UNet consists of a total number of $70$ attention layers. Each layer consists of a cross and self-attention. These attention layers are often referred to as attention blocks. In this paper, for clarity, we refer to them as \textit{layers} so they are not confused with the larger transformer \textit{blocks} we optimize.
These attention layers are divided into $11$ transformer blocks where the first two and last three blocks are comprised of four and six attention layers, respectively. The six inner blocks consist of 10 attention layers each, as illustrated in \Cref{fig:prompt_inject_illustration}).

Text condition generation is also extended in SDXL in the following way: given a text prompt $y$, it is encoded twice, with both OpenCLIP ViT-bigG \cite{Openclip2021} and CLIP ViT-L \cite{CLIP2021}. The resulting embeddings are then concatenated to define the conditioning encoding $c$. 
Then this text embedding is fed into the cross-attention layers of the network, following the attention mechanism \cite{Attention2017}.

Specifically, in each layer, the deep spatial features $x$ are projected to a query matrix $Q = l_Q(x)$, and the textual embedding is projected to a key matrix $K = l_K(c)$ and a value matrix $V = l_V(c)$ via learned linear projections $l_Q, l_K, l_V$.
The attention maps are then defined by:

\begin{equation}
    A_t = Softmax(\frac{QK^T}{\sqrt{d}})V,
\end{equation}
where $d$ is the projection dimension of the keys and queries.

\paragraph{\textbf{LoRA}} \label{sec:lora}
Low-Rank Adaptation \cite{LoRA21} is a method for efficiently fine-tuning large pre-trained models for specific tasks or domains. LoRA has emerged as a very popular approach for fine-tuning pre-trained text-to-image diffusion models \cite{DBlora} due to its high-quality results and efficiency.

Let us denote the weights of a pre-trained text-to-image diffusion model with $W_0$, and the learned residuals after fine-tuning the model for a specific task with $\Delta W$.
The key idea in LoRA is that $\Delta W\in\mathbb{R}^{m\times n}$ can be decomposed into two low intrinsic rank matrices $B\in\mathbb{R}^{m\times r}$ and $A\in\mathbb{R}^{r\times n}$, such that $\Delta W=BA$, and the rank $r << \min(m,n)$.
During training, the original model weights $W_0$ remain frozen, and only $A$ and $B$ are updated.
Thus, by the end of the training, we can obtain the tuned model weights by using $W = W_0 + \Delta W$.

LoRA is commonly used in text-to-image diffusion models only in the cross and self-attention layers.
As discussed, the attention mechanism in each layer relies on four projection matrices: $l_Q$, $l_K$, $l_V$, and $l_{\text{out}}$. The LoRA weights $\Delta W_Q$, $\Delta W_K$, $\Delta W_V$, and $\Delta W_{\text{out}}$ are optimized for each of these pre-trained matrices. We denote by $\Delta W$ the LoRA weights of all four matrices.

\section{Method}
Our objective is to decouple the style and content aspects of an input image $I$ into separate components, enabling both text-based and image-based stylization applications.
Our approach harnesses the capabilities of a pre-trained SDXL text-to-image generation model \cite{sdxl2023}, known for its robustness in capturing stylistic features \cite{ziplora23}. We conduct an analysis of the SDXL architecture to gain insight into the contributions of individual layers to either the style or the content of the generated image. Guided by our observations, we employ LoRA \cite{LoRA21} to train update matrices of only two specific transformer blocks within the SDXL model. These matrices capture the representation of the content and the style of the input image and they suffice to facilitate a number of image stylization tasks.

\subsection{SDXL Architecture Analysis}
\label{sec:SDXLAnalysis}
Similar to previous works ~\cite{p+23,MATTE2023} we examine the effect of different layers within the base text-to-image model on the generated image.
We adopt a similar approach to the one proposed in Voynov et al. \cite{p+23}. The key idea is to inject a different text prompt into the cross-attention layers of one of the transformer blocks within SDXL. Then examine the similarity between the different prompts and the resulting image. 
If we only change the input prompt corresponding to the i'th block, and the i'th block dominates a certain quality of the generated image, this will be apparent in the resulting image. 
Specifically, we examine six intermediate transformer blocks $\{W_0^1, .. W_0^6\}$ of SDXL, each containing $10$ attention layers (see \Cref{fig:prompt_inject_illustration}). These layers have been selected based on previous works \cite{p+23, MATTE2023}, which demonstrate that they are most likely to affect the important visual properties of the generated images.

We define two random sets of text prompts $P_{content}$ and $P_{style}$ describing different objects with different colors. The prompts in $P_{content}$ are defined by placing random objects in the template text \ap{A photo of a {\small\textless}object{\small\textgreater}}. For $P_{style}$ we use the template \ap{A photo of a {\small\textless}color{\small\textgreater} \  {\small\textless}object{\small\textgreater}}. The random objects and colors are generated with ChatGPT. Note that color is used as a proxy for style since we use CLIP \cite{CLIP2021} to evaluate results (as will be described next), and we found CLIP to be a better indicator for changes in color than changes in style. 
We sample a pair of prompts $(p, \hat{p})\in P_{content}$ and $(p, \hat{p})\in P_{style}$ such that $p\neq \hat{p}$.
For each pair $(\hat{p},p)$, we generate an image $I_{\hat{p}\rightarrow i, p \rightarrow j\neq i}$ by injecting the embedding of $\hat{p}$ to $W_0^i$ while injecting the embedding of $p$ to all other layers $W_0^j, j\neq i$ (depicted in \Cref{fig:prompt_inject_illustration}). This is performed for each of the six transformer blocks we target, yielding six images per pair.

Next, to measure the effect of injecting $\hat{p}$ into the i'th block on the generated image, we estimate the following similarity score:
\vspace{-3pt}
\begin{equation}
    \mathcal{C}(I_{\hat{p}\rightarrow i, p \rightarrow j}, \hat{p}) = sim(C_I(I_{\hat{p}\rightarrow i, p \rightarrow j}), C_T(\hat{p})),   
    \label{eq:clip_score}
\end{equation}
where $C_I(I_{\hat{p}\rightarrow i, p \rightarrow j})$ and $C_T(\hat{p})$ are the CLIP image embedding of the generated image, and the CLIP text embedding of the prompt, respectively. $sim(x,y)=\frac{x\cdot y}{||x||\cdot ||y||}$ indicates the cosine similarity between the clip embeddings.

In total, we examined 400 pairs of content and style prompts and averaged the scores of each layer.
The three topmost layers that show similarity to one type of prompt are $W_0^2$  and $W_0^4$ which dominate the content of the generated image, and $W_0^5$ which dominates its color.
We visually demonstrate these conclusions in \Cref{fig:injection_vis}. 
On the left, we show the effect of blocks 2 and 4 on the generated content. Note that $I_{\hat{p} \rightarrow 2, p \rightarrow j}$ and $I_{\hat{p} \rightarrow 4, p \rightarrow j}$ demonstrate that when \ap{A photo of a tiger} is injected to only one block (2 or 4), while \ap{A photo of a bunny} is injected to the rest of the blocks, the generated images depict a tiger, while in all other options, the generated image will depict a bunny. Similarly, on the right we show the effect of block 5 on the generated image's color.

\vspace{-2.5pt}
\subsection{LoRA-Based Separation with B-LoRA}\vspace{-2pt}
While the observations above apply to a \textit{generated} image, our goal is to examine if the layers we locate could be useful in capturing the content and style of a \textit{given} input image $I$.
To fine-tune the model to generate variations of our given image we utilize the LoRA \cite{LoRA21} approach.

Let us denote the frozen weights of our base pre-trained SDXL model with $W_0$ and the learned residual matrices for each block with $\Delta W^i$.
We follow the default settings of DreamBooth LoRA \cite{DBlora} to finetune the model to reconstruct the given input image $I$.

\begin{figure*}[t]
    \vspace{-11pt}
    \centering
    \includegraphics[width=0.96\linewidth]{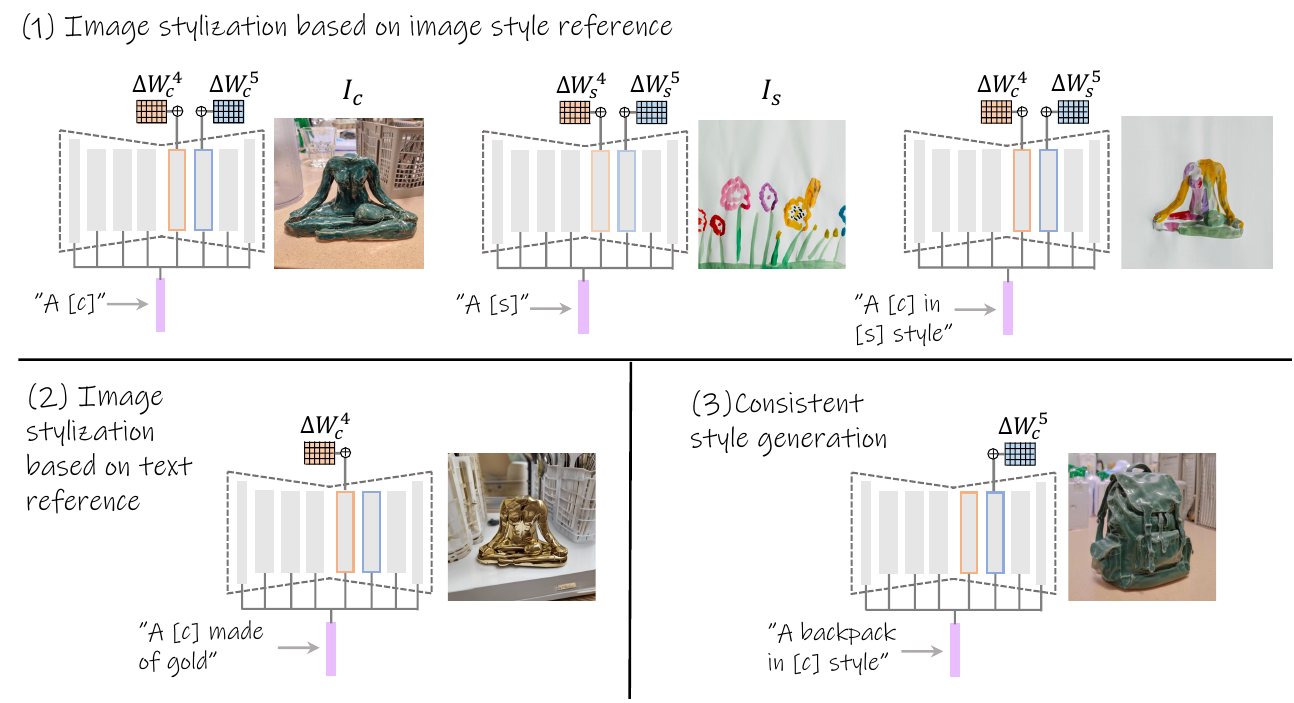}
    \caption{B-LoRA for Image Stylization. (1) To stylize a given content image $I_c$ w.r.t an given style image reference $I_s$, we train our B-LoRAs for both images and then combine $\Delta W_c^4$ and $\Delta W_s^5$ to a single adapted model. (2) For text-based stylization we simply plug only the trained $\Delta W_c^4$ to adapt the model and then use the desired text prompt during inference. (3) The learned style weights $\Delta W_c^5$ can be also used as is to adjust the backbone model to produce images with the style of $I_c$.}
    \vspace{-7pt}
    \label{fig:apps_all}\vspace{-10pt}
\end{figure*}

However, instead of optimizing the LoRA weights of all eleven blocks (as usually done), we conduct two experiments, where in the first experiment we optimize the pair \{$\Delta W^2, \Delta W^5$\}, and in the second experiment we optimize $\{\Delta W^4, \Delta W^5\}$ (as we found $W_0^2$ and $W_0^4$ to dominate the content, and $W_0^5$ to dominate the color). In addition, we use a general prompt \ap{A [v]} during training to prevent the model from being explicitly guided to capture either the image's style or content. This process and example results are depicted in \Cref{fig:lora_subset_training}. 
As can be seen, we find that the best combination to optimize in terms of 1.\ Achieving a full reconstruction of the input concept, and 2.\ Capturing the input image's content, are $\Delta W^4, \Delta W^5$.
Note that using the deeper layers of the UNet $\Delta W^4$, rather than $\Delta W^2$ during the LoRA training process, aligns with the goal of preserving finer details in the output image, as demonstrated in \cite{PlugandPlay2022}.
We provide ablation and analysis of the effect of other layers and specific parts within them, as well as the effect of using different text prompts in the supplementary material.
We call such a training scheme \emph{B-LoRA}, as it only trains two transformer Blocks instead of the full weights. 
Hence, apart from the style-content separation abilities such a method also reduces storage requirements by $70\%$.

\vspace{-3.5pt}
\subsection{B-LoRA for Image Stylization}\vspace{-3pt}
\label{sec:apps}
Combining the insights from the above analyses, we now describe the B-LoRA training approach.
Given an input image $I$, we only fine-tune the LoRA weights $\Delta W^4, \Delta W^5$ with the objective of reconstructing the image, w.r.t a general text prompt \ap{A [v]}.
Besides increasing efficiency, we find that by training only these two layers, we can achieve an \textbf{implicit} style-content decomposition, where $\Delta W^4$ captures the content and $\Delta W^5$ captures the style. 

Once we find these update matrices, we can easily use them by updating the corresponding block weights of the pre-trained SDXL model for style manipulation applications as described next and demonstrated in \Cref{fig:apps_all}.
\vspace{-0.2cm}
\paragraph{Image stylization based on image style reference}
Given two input images $I_c, I_s$ depicting the desired content and style respectively, we use the process described above to learn their corresponding B-LoRA weights: $\Delta W_c^4, \Delta W_c^5$ for $I_c$ and $\Delta W_s^4, \Delta W_s^5$ for $I_s$.
We then directly use $\Delta W_c^4$ and $\Delta W_s^5$ to update the transformer blocks $W_0^4$ and $W_0^5$ of the pre-trained network.
For the inference process, we use the prompt \ap{A [c] in [s] style}, as illustrated at the top of \Cref{fig:apps_all}.
\vspace{-0.2cm}
\paragraph{Text-based image stylization}
By omitting $\Delta W_c^5$ (capturing the style of $I_c$) and only using $\Delta W_c^4$ to update the weights of the pre-trained model, we get a personalized model that is adapted to only the content of $I_c$. To manipulate the style of $I_c$ with text-based guidance, we simply inject the desired text into the adapted layers during inference (see \Cref{fig:apps_all} bottom-left).
Note that because the style and content are separated and encoded in different blocks, our approach allows challenging style manipulations.

\begin{figure*}[h!]
    \centering
    \setlength{\tabcolsep}{1.5pt}
    {\small
    \begin{NiceTabular}{c | @{\hspace{0.1cm}}c c c c c}
        \diagbox{Input \\ Content}{Style \quad} &
        \hspace{0.11cm}
        \includegraphics[width=0.158\textwidth]{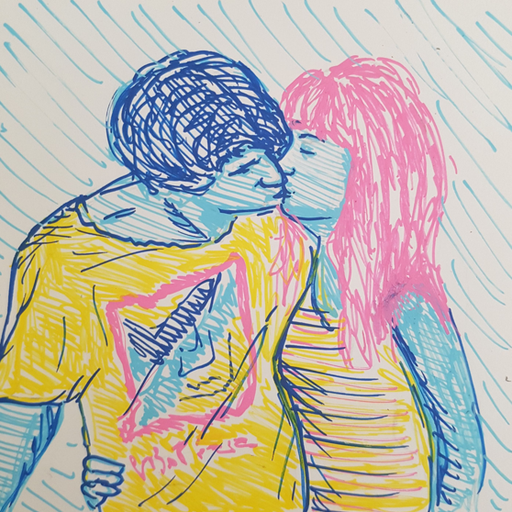} &
        \includegraphics[width=0.158\textwidth]{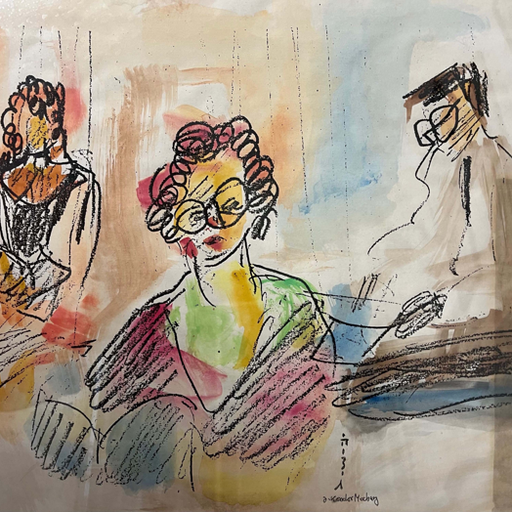} &
        \includegraphics[width=0.158\textwidth]{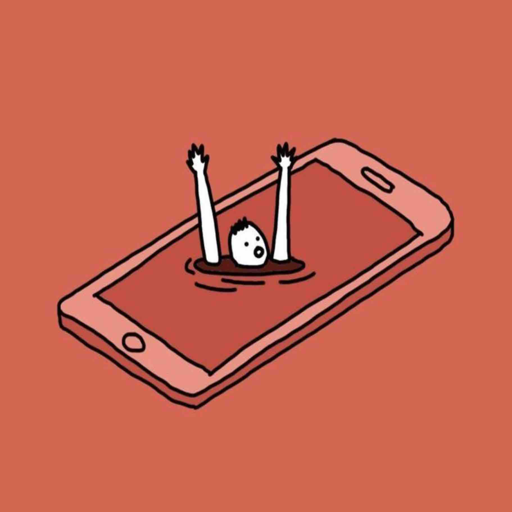} &
        \includegraphics[width=0.158\textwidth]{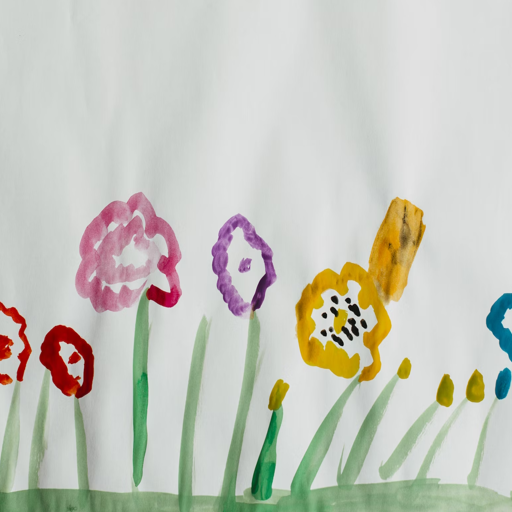} &
        \includegraphics[width=0.158\textwidth]{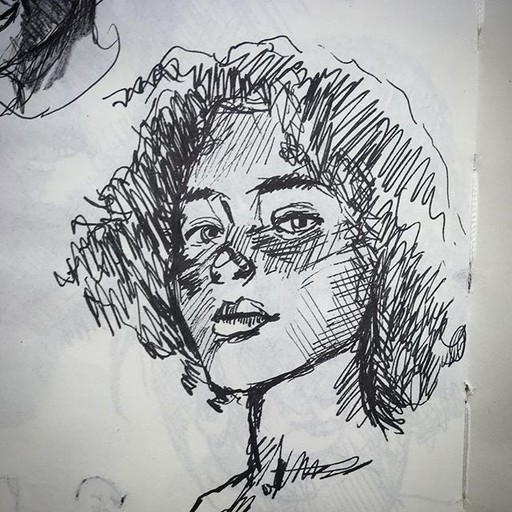} \\
        \midrule

        \includegraphics[width=0.158\textwidth]{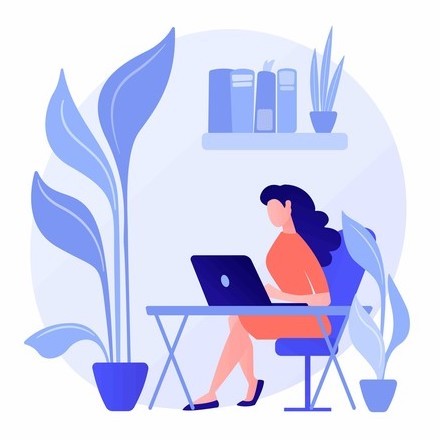} &
        \hspace{0.11cm}
        \includegraphics[width=0.158\textwidth]{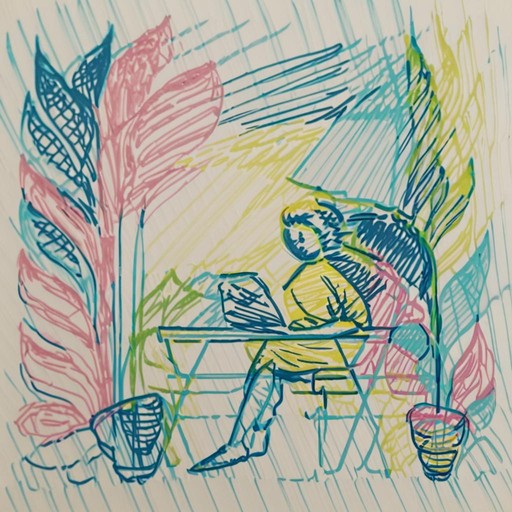} &
        \includegraphics[width=0.158\textwidth]{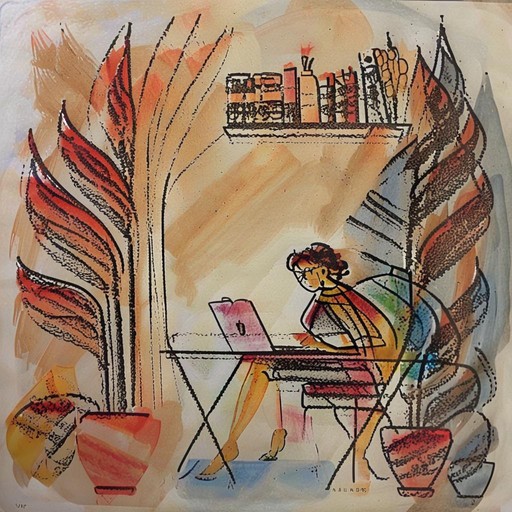} &
        \includegraphics[width=0.158\textwidth]{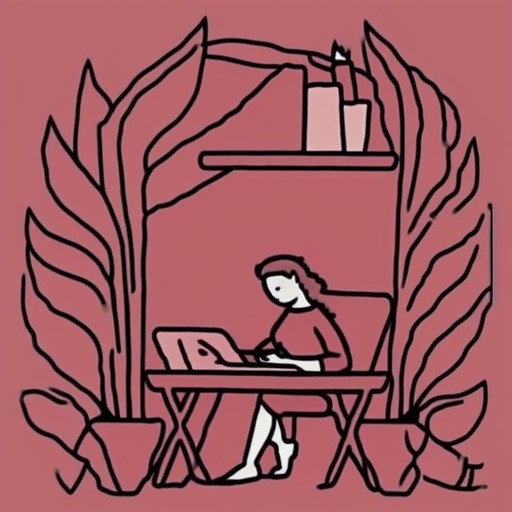} &
        \includegraphics[width=0.158\textwidth]{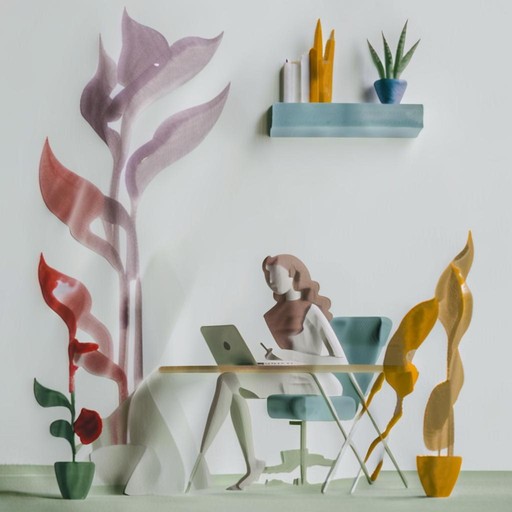} &
        \includegraphics[width=0.158\textwidth]{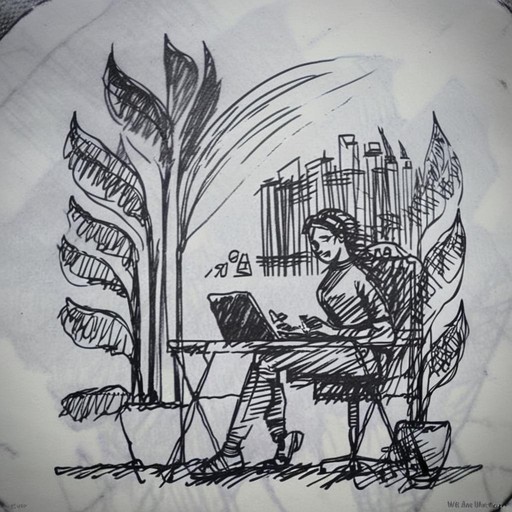} \\

        \includegraphics[width=0.158\textwidth]{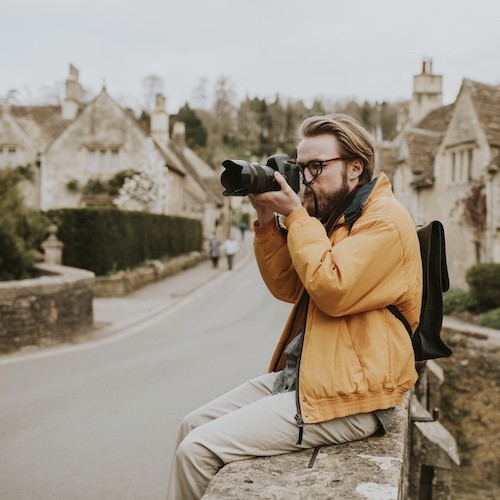} &
        \hspace{0.11cm}
        \includegraphics[width=0.158\textwidth]{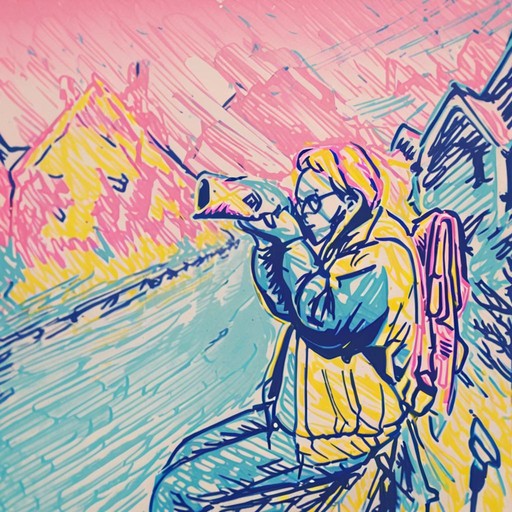} &
        \includegraphics[width=0.158\textwidth]{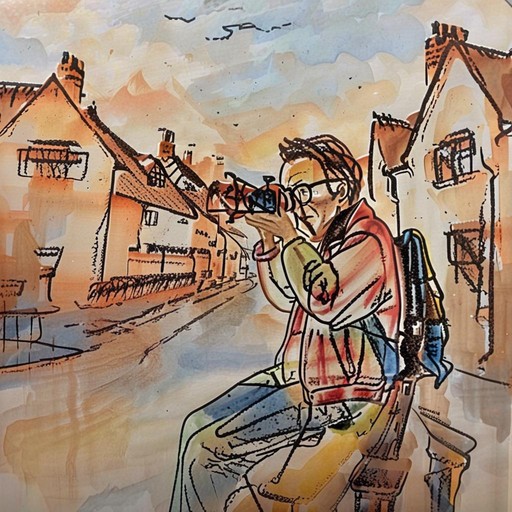} &
        \includegraphics[width=0.158\textwidth]{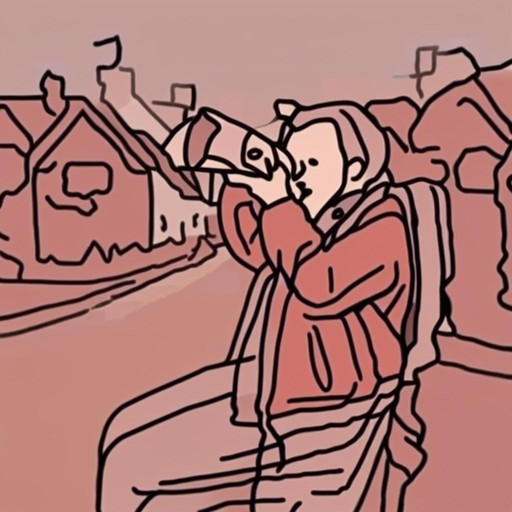} &
        \includegraphics[width=0.158\textwidth]{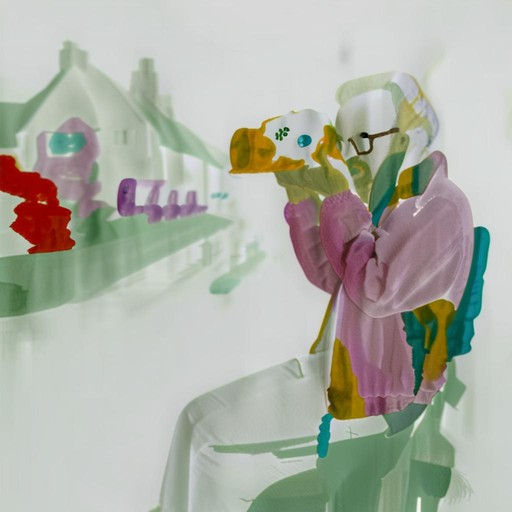} &
        \includegraphics[width=0.158\textwidth]{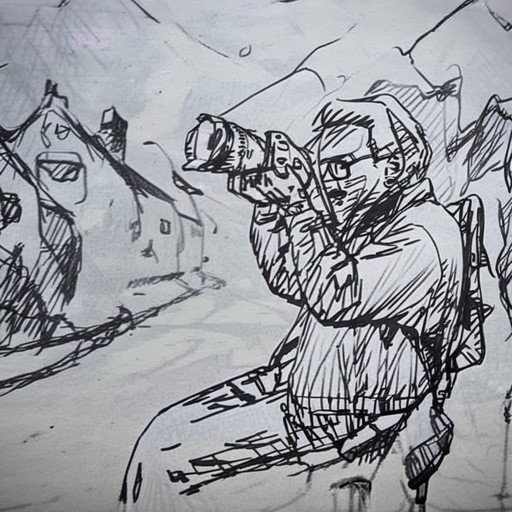} \\

        \midrule
        \vspace{1cm}
        \begin{tabular}[c]{@{}c@{}} Input \\  Content \end{tabular} &  \ap{Armored} & \ap{On fire} & \ap{Ice}& \begin{tabular}[c]{@{}c@{}} ``Sketch \\  cartoon'' \end{tabular} & \ap{Neon lights}\\
        \includegraphics[width=0.158\textwidth]{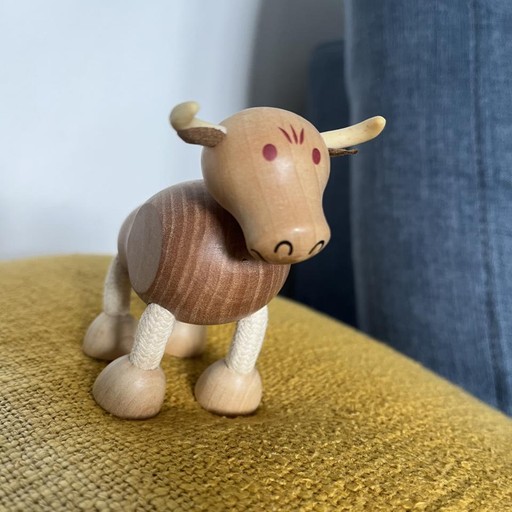} &
        \hspace{0.11cm}
        \includegraphics[width=0.158\textwidth]{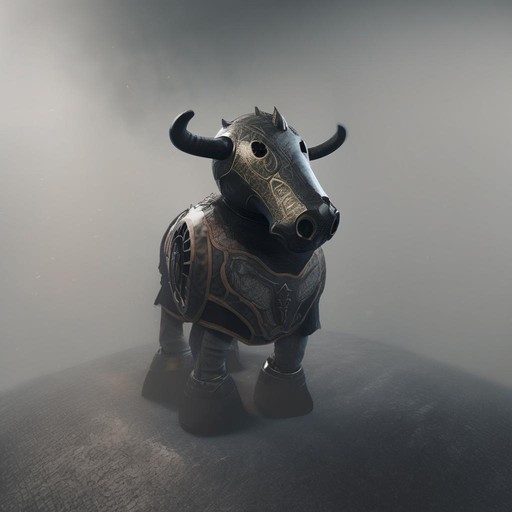} &
        \includegraphics[width=0.158\textwidth]{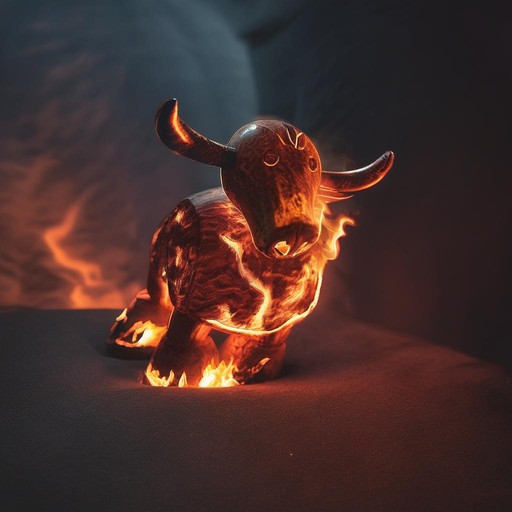} &
        \includegraphics[width=0.158\textwidth]{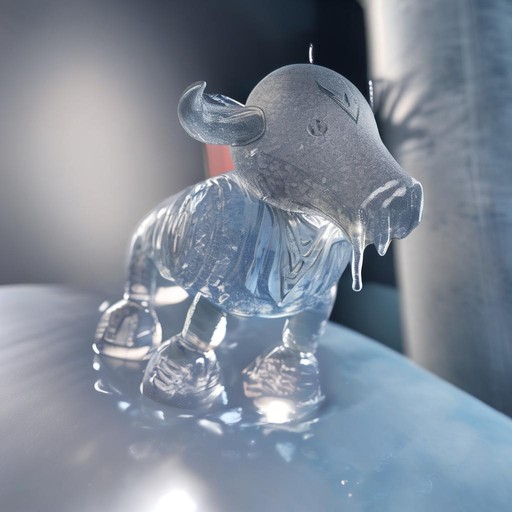} &
        \includegraphics[width=0.158\textwidth]{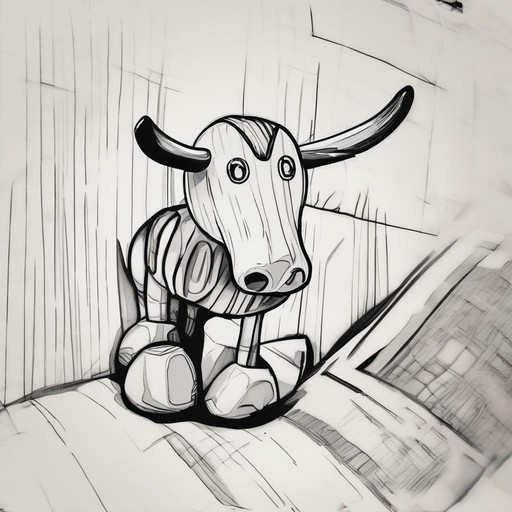} &
        \includegraphics[width=0.158\textwidth]{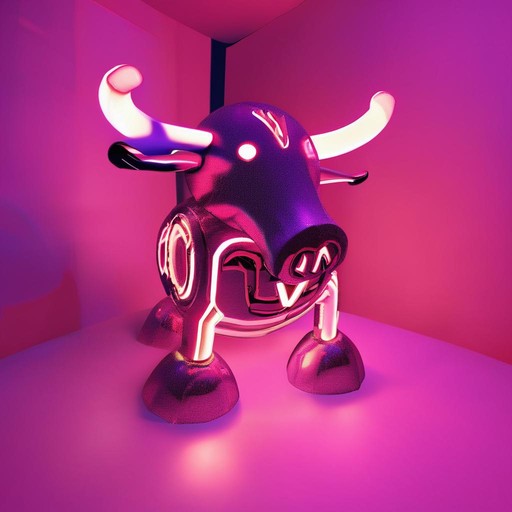} \\
        \midrule
        \vspace{1cm}
        Input Style & ``Lion'' & ``Guitar''& ``Car''& ``Sneakers''& ``Bicycle''\\
        \includegraphics[width=0.158\textwidth]{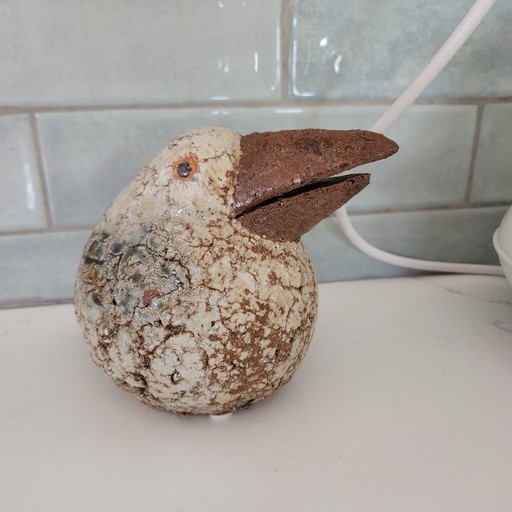} &
        \hspace{0.11cm}
        \includegraphics[width=0.158\textwidth]{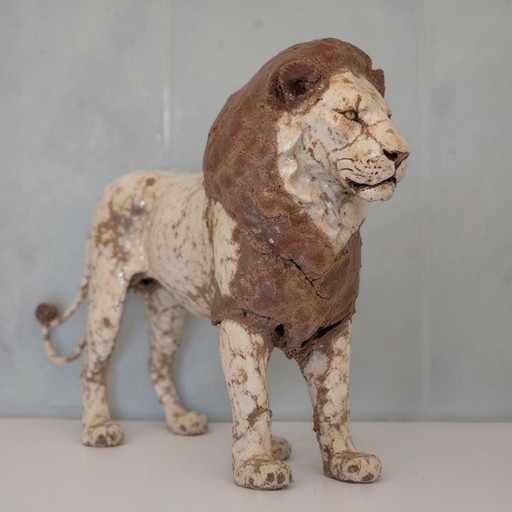} &
        \includegraphics[width=0.158\textwidth]{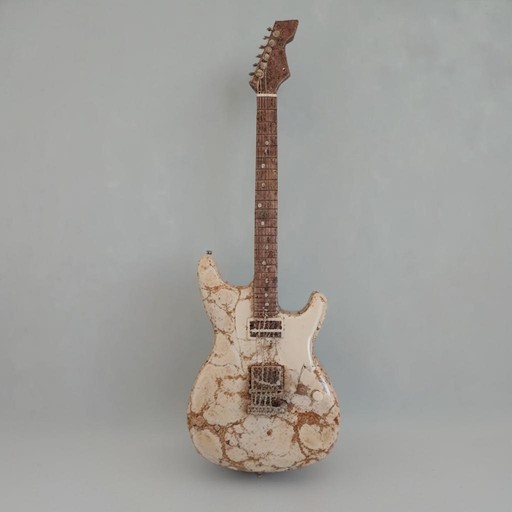} &
        \includegraphics[width=0.158\textwidth]{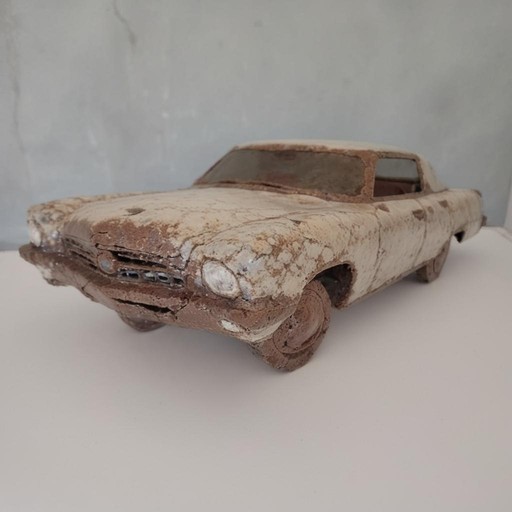} &
        \includegraphics[width=0.158\textwidth]{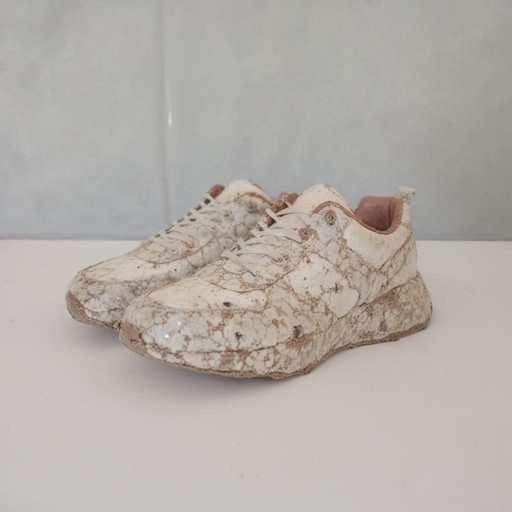} &
        \includegraphics[width=0.158\textwidth]{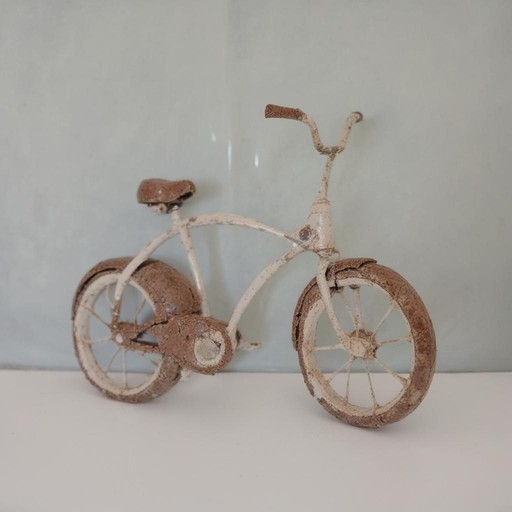} \\
    \end{NiceTabular}
    }
    \vspace{-0.2cm}
    \caption{Results produced by our method for three image stylization tasks. Rows 1-3: image style transfer. Our method can operate on scene images and extract content from a stylized image. Fourth row: text-based image stylization applied to the content image reference on the left. Note how the pose and identity are preserved well. Last row: consistent style generation, where that style is extracted from the image on the left and used to generate new objects. In this row, we use 
$\alpha =1.1$ to enhance the style effect.}
    \vspace{-0.2cm}
    \label{fig:main_res}
\end{figure*}
\paragraph{Consistent style generation}
Lastly, in a similar manner, one can adapt the model for a specific style provided in $I_s$ by excluding $\Delta W_s^4$ and using only $\Delta W_s^5$. This results in a model adapted to the desired style, and one can use text-based conditions to generate any content with the desired style (see \Cref{fig:apps_all} bottom-right).

\subsection{Implementation details}
We train the B-LoRA weights on SDXL v1.0 \cite{sdxl2023} while keeping both the model weights and text encoders frozen during the fine-tuning process. All LoRA training was performed on a single image. We utilize the Adam optimizer with a learning rate of $5e-5$. For data augmentations, we only use center cropping during training. We set the LoRA weights rank to $r=64$ and use the prompt \ap{A [v]} for 1000 optimization steps, requiring approximately 10 minutes per image on a single A100 GPU. Note that while other methods typically train LoRA for 400 steps to mitigate overfitting concerns, this was not an issue in our case.

\begin{figure*}[ht]
    \centering
    \setlength{\tabcolsep}{1.5pt}
    {\small
    \begin{tabular}{l c c @{\hspace{0.17cm}} | @{\hspace{0.17cm}}c c c c @{\hspace{0.17cm}} | @{\hspace{0.17cm}}c}
        
    ~ &  Content & Style & DB-LoRA & ZipLoRA 
         & \begin{tabular}[c]{@{}c@{}} StyleDrop \\ SDXL \end{tabular}
         & \begin{tabular}[c]{@{}c@{}} Style- \\ Aligned \end{tabular}
          & Ours \\
          
         \multirow{4}{*}{\rotatebox{90}{Multiple Content Images}} &  
        \includegraphics[width=0.122\textwidth]{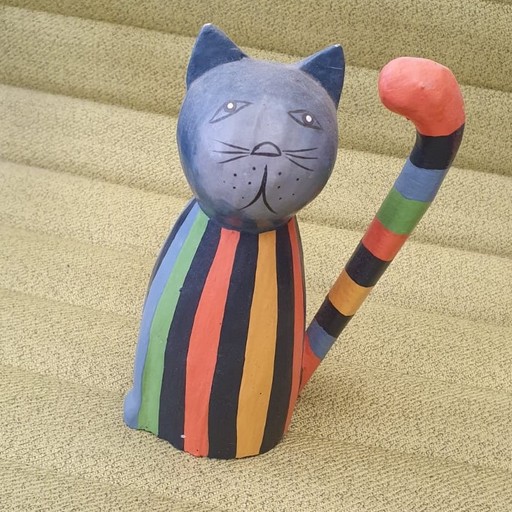} &
        \includegraphics[width=0.125\textwidth]{temp_figs/style_images/kiss.png} &
        
        \includegraphics[width=0.125\textwidth]{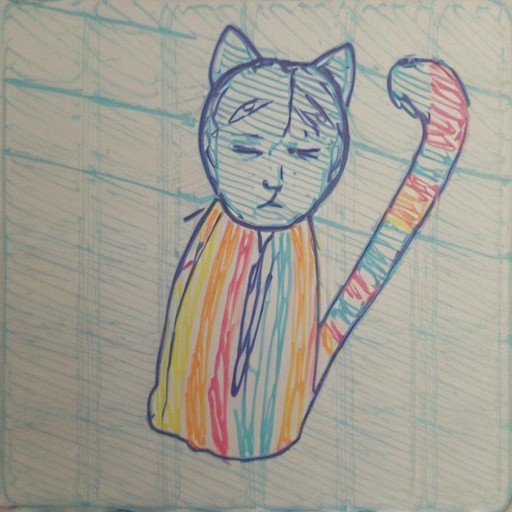} &
        \includegraphics[width=0.125\textwidth]{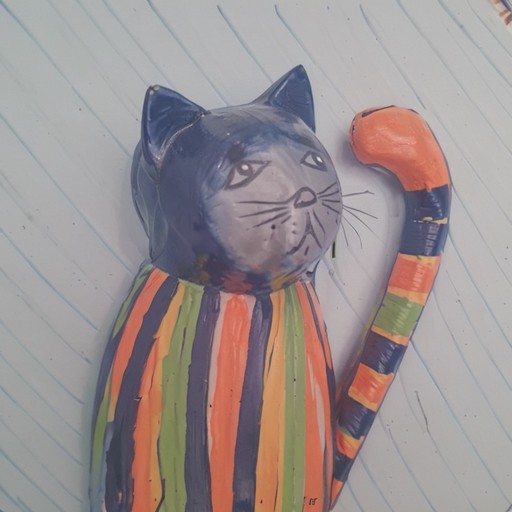} &
        \includegraphics[width=0.125\textwidth]{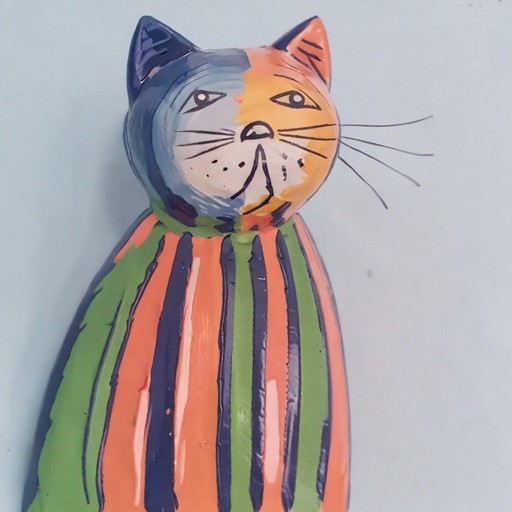} &
        \includegraphics[width=0.125\textwidth]{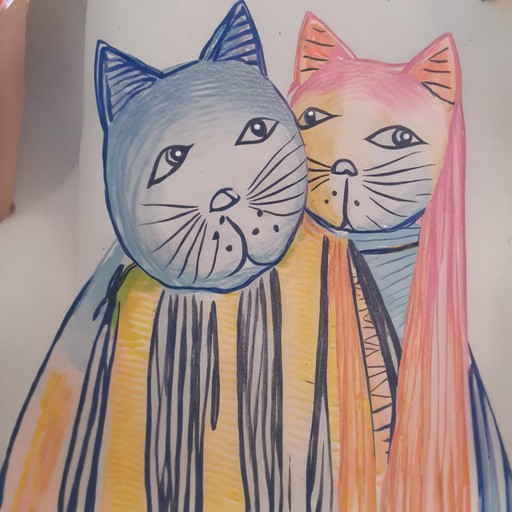} &
        \includegraphics[width=0.125\textwidth]{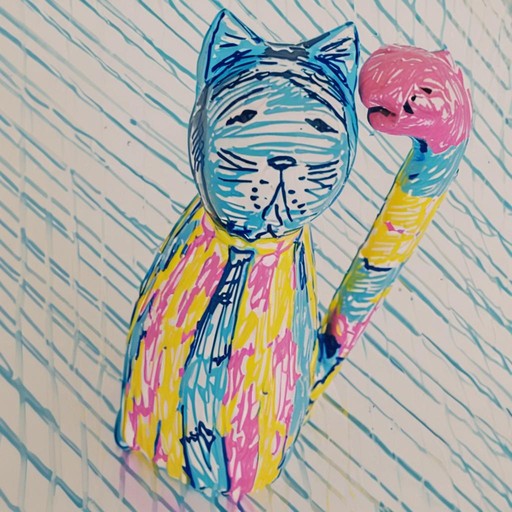} \\

        ~ & \includegraphics[width=0.125\textwidth]{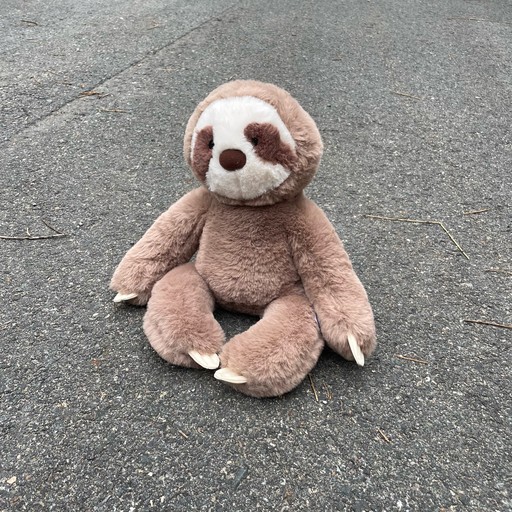} &
        \includegraphics[width=0.125\textwidth]{temp_figs/style_images/drawing3.png} &
        \includegraphics[width=0.125\textwidth]{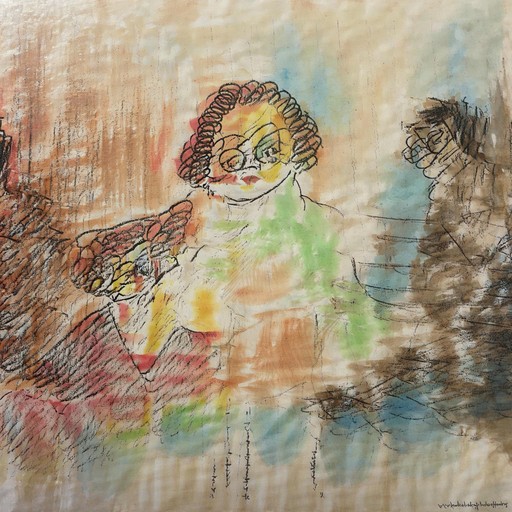} &
        \includegraphics[width=0.125\textwidth]{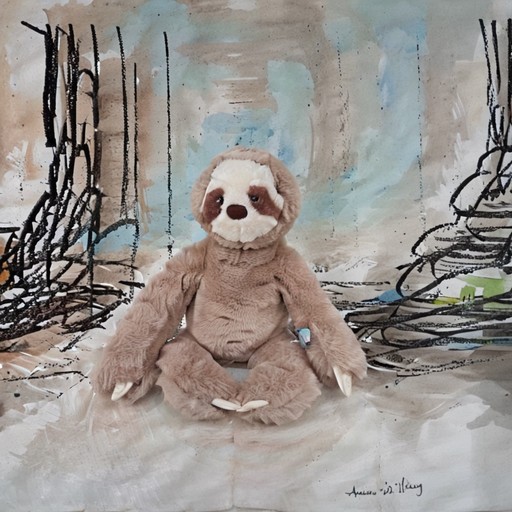} &
        \includegraphics[width=0.125\textwidth]{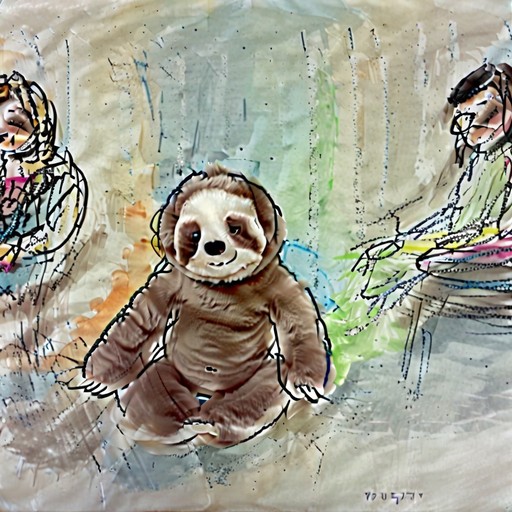} &
        \includegraphics[width=0.125\textwidth]{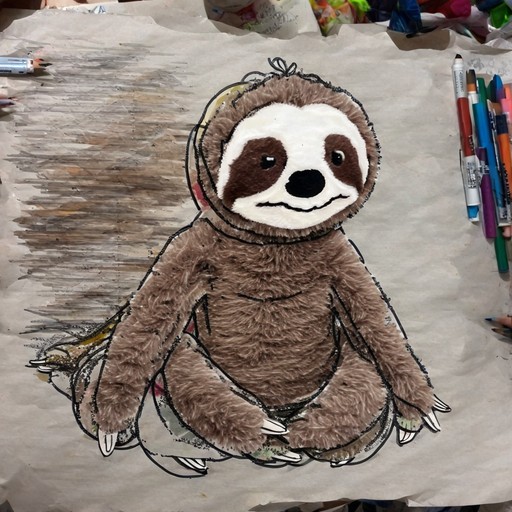} &
        \includegraphics[width=0.125\textwidth]{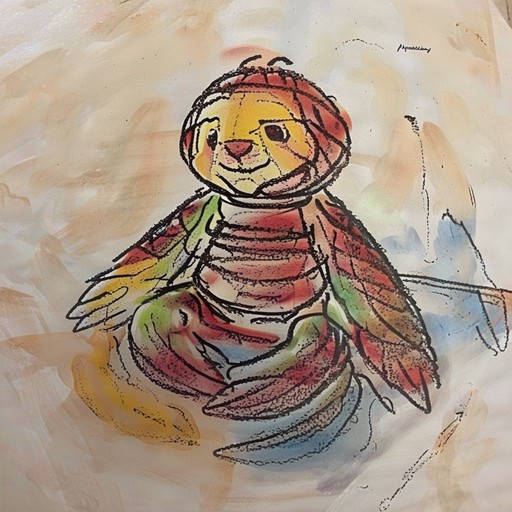} \\

        ~ & \includegraphics[width=0.125\textwidth]{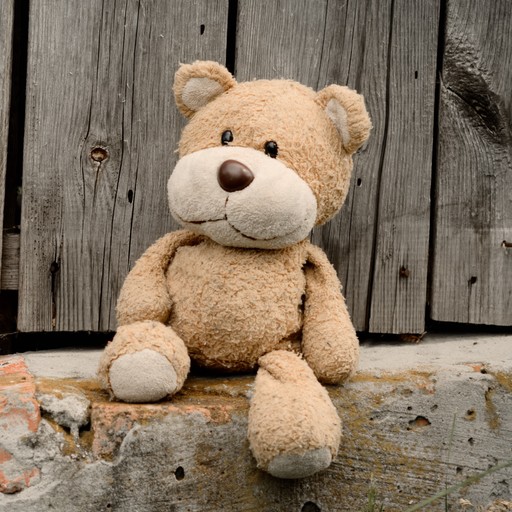} &
        \includegraphics[width=0.125\textwidth]{temp_figs/style_images/pen_sketch.jpeg} &
        \includegraphics[width=0.125\textwidth]{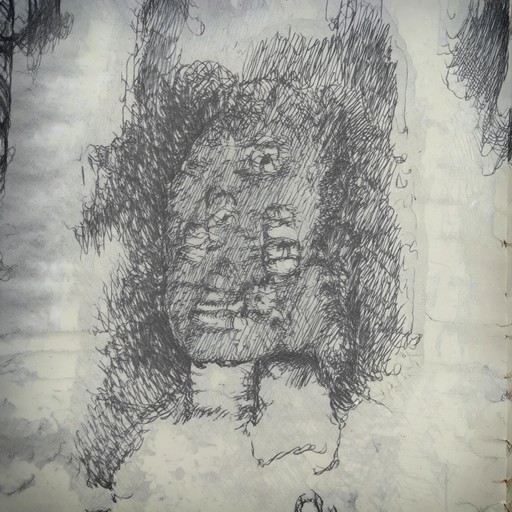} &
        \includegraphics[width=0.125\textwidth]{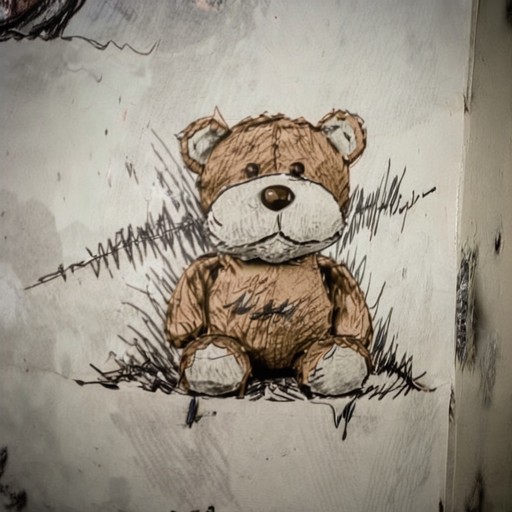} &
        \includegraphics[width=0.125\textwidth]{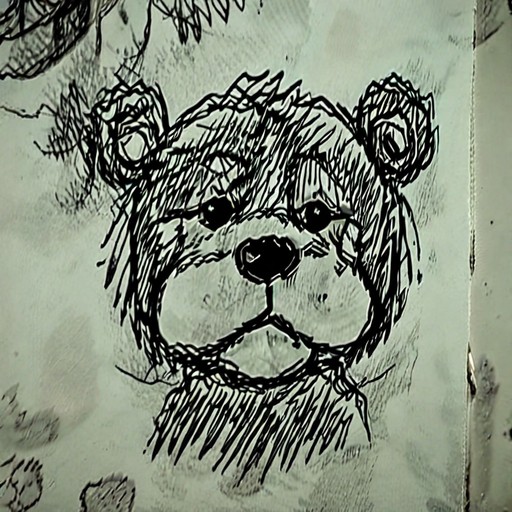} &
        \includegraphics[width=0.125\textwidth]{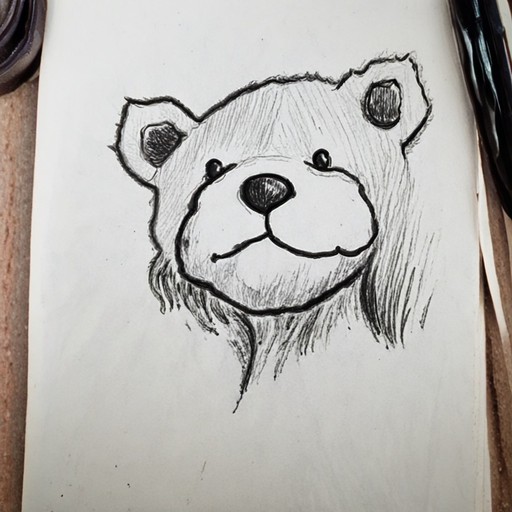} &
        \includegraphics[width=0.125\textwidth]{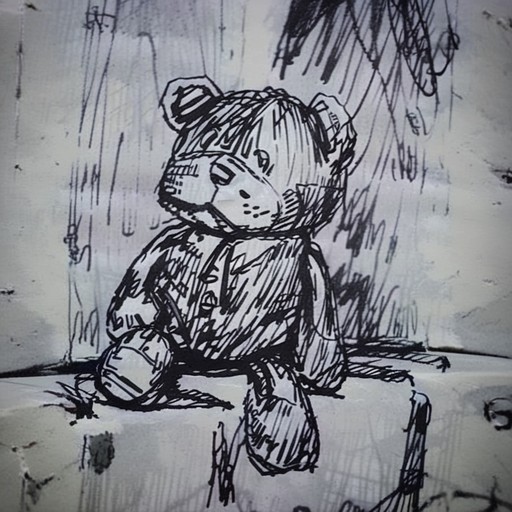} \\
        
        ~ & \includegraphics[width=0.125\textwidth]{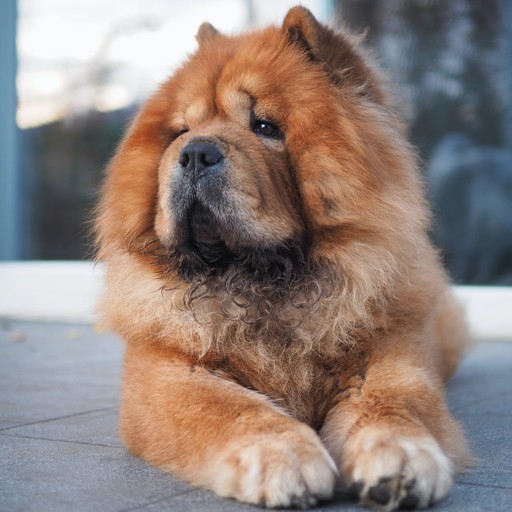} &
        \includegraphics[width=0.125\textwidth]{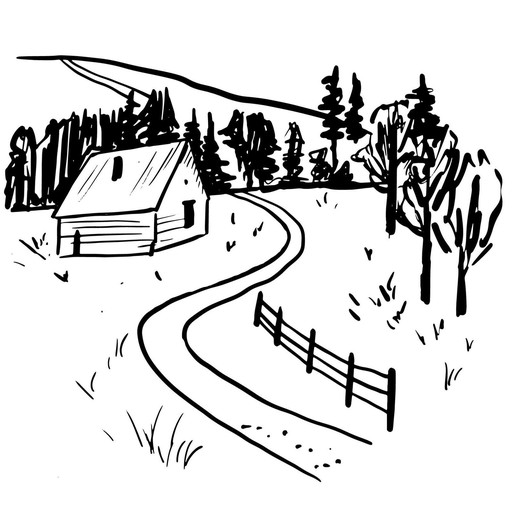} &
        \includegraphics[width=0.125\textwidth]{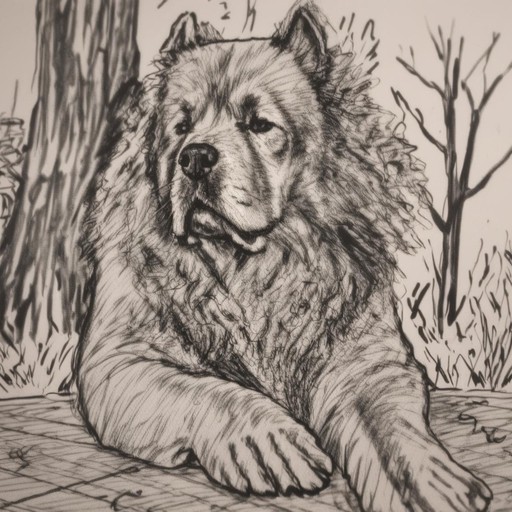} &
        \includegraphics[width=0.125\textwidth]{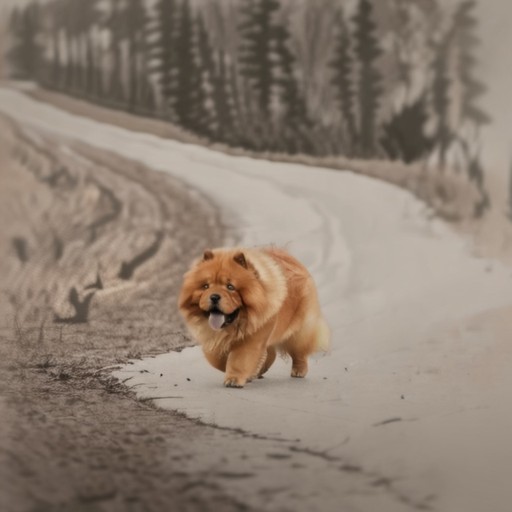} &
        \includegraphics[width=0.125\textwidth]{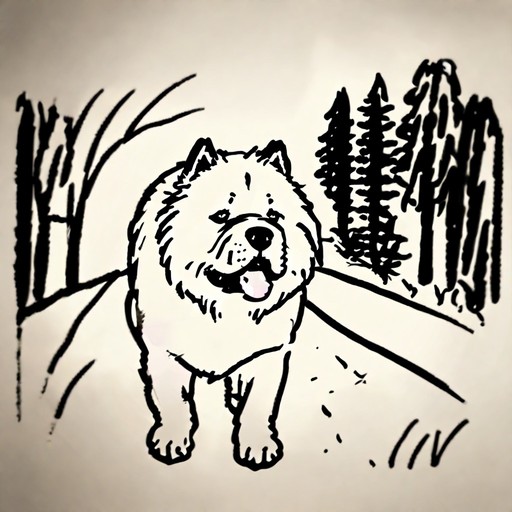} &
        \includegraphics[width=0.125\textwidth]{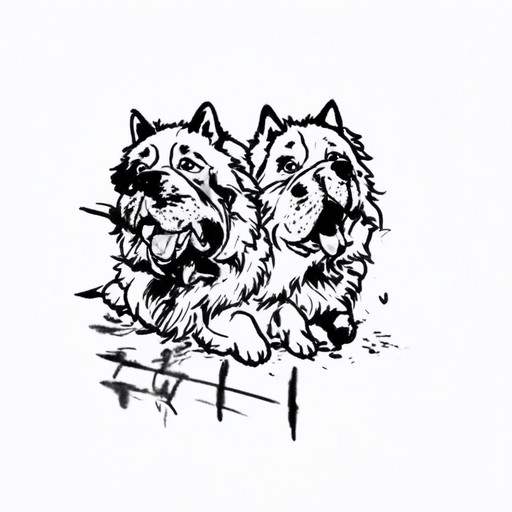} &
        \includegraphics[width=0.125\textwidth]{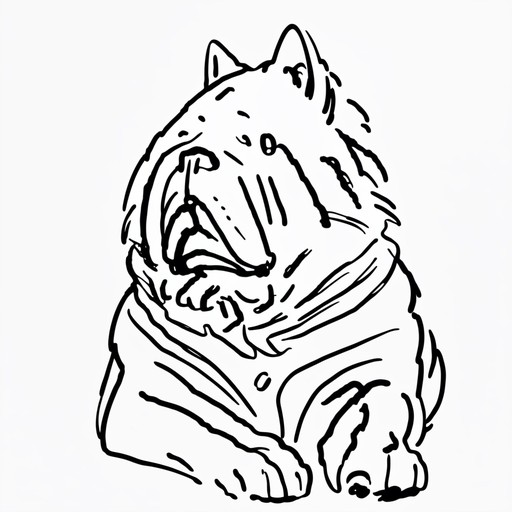} \\
        \midrule
        \raisebox{0.5cm}{\rotatebox[origin=l]{90}{Single}} &
        \includegraphics[width=0.125\textwidth]{temp_figs/content_images/dog2.jpg} &
        \includegraphics[width=0.125\textwidth]{temp_figs/style_images/ink_sketch.jpeg} &
        \includegraphics[width=0.125\textwidth]{temp_figs/Fig_7/ctrl_dog2.jpg} &
        \includegraphics[width=0.125\textwidth]{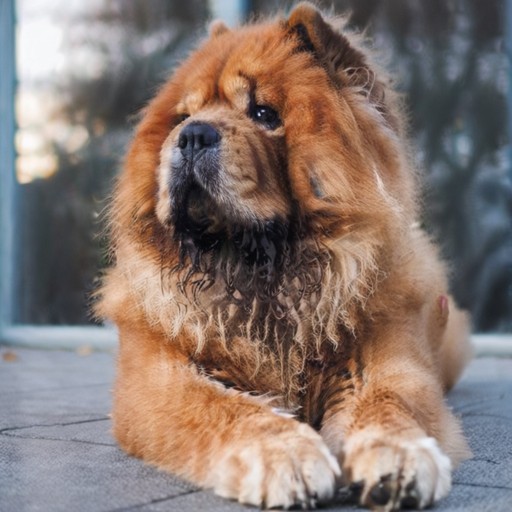} &
        \includegraphics[width=0.125\textwidth]{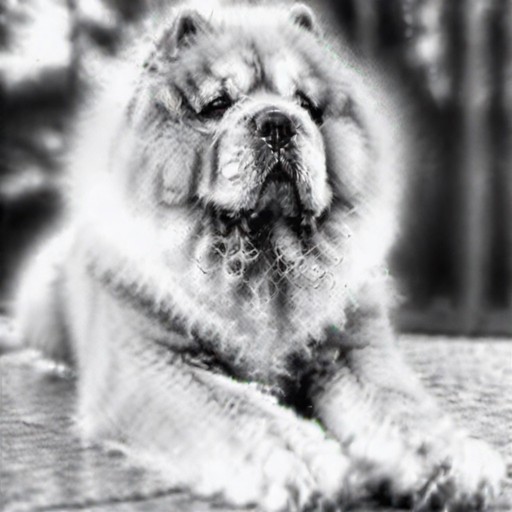} &
        \includegraphics[width=0.125\textwidth]{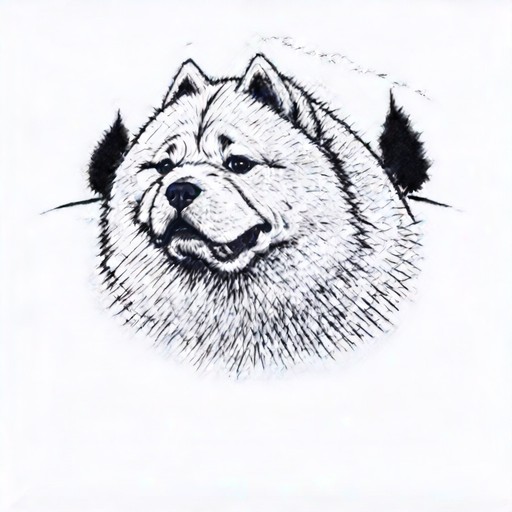} &
        \includegraphics[width=0.125\textwidth]{temp_figs/Fig_7/ours_dog2.jpeg} \\
    \end{tabular}
    }   
    \vspace{-0.11cm}
    \caption{Comparison with alternative approaches. The input style and content references are shown on the left, where multiple content images were used for alternative methods. In the last row, we applied other approaches to a single content image. ZipLoRA tends to overfit the content, and thus struggles with depicting the desired style. StyleDrop also struggles to preserve the content when trained on multiple images. In the case of a single content image (last row), both methods preserve the content but lose the style. StyleAligned preserves the style well; however, it tends to include semantic content originating in the \emph{style} image, such as creating a couple in row 1. Additional comparisons to InstantStyle \cite{InstantStyle2024} are provided in the supplementary material.}
    \vspace{-0.4cm}
    \label{fig:qual_comparison}
\end{figure*}

\section{Results}
To produce the various results of our approach we optimized our B-LoRAs ($\Delta W^4, \Delta W^5$) once for each image and then plugged either one of them or both of them (depending on the application) at inference time to receive image stylization without any further optimization or fine-tuning.

We present some qualitative results of the three applications discussed in \Cref{sec:apps} in \Cref{fig:main_res}.
In the first two rows of \Cref{fig:main_res}, our method manages to transfer the style of the image references (top row) while preserving the content of the input image on the left.
Notable, this can be done for challenging content inputs such as stylized images (first row) and images of whole scenes (second row).
Our method is robust to many types of different styles and manages to preserve the essence of the content reference even in very abstract styles such as the one depicted in the third style column.
In the third row, we show examples of text-based image stylization. As can be seen with our implicit style-content separation, the content of the input object is preserved well while the style is governed by the desired text prompt.
In the last row, we demonstrate how our method can be used for consistent style generation where only the B-LoRA weights of the style are used. Observe that the object's style is well preserved across all text-based generated images.
Please refer to the supplementary material for many more examples. %

\begin{table*}[t]
\caption{Quantitative comparison. We measure the average cosine similarity between the DINO features of the output image and the reference style and content. Our method performs best at adapting to the style without overfitting the content image.}
\vspace{-0.2cm}
\setlength{\tabcolsep}{5pt}
\renewcommand{\arraystretch}{1}
    \centering
    \begin{tabular}{c|cccccc}
    \toprule
        ~ & Input &StyleDrop & StyleAligned & ZipLoRA & DB-LoRA & Ours \\
        \midrule
        \begin{tabular}[c]{@{}c@{}} Style \\  Transfer \end{tabular} & \begin{tabular}[c]{@{}c@{}} Multiple \\ Single \end{tabular} & \begin{tabular}[c]{@{}c@{}} $0.826\pm{0.07}$ \\  $0.790\pm{0.06}$ \end{tabular} & \begin{tabular}[c]{@{}c@{}} $0.855\pm{0.05}$ \\  $0.829\pm{0.05}$ \end{tabular} & \begin{tabular}[c]{@{}c@{}} $0.796\pm{0.07}$  \\  $0.782\pm{0.05}$ \end{tabular}& $0.863\pm{0.06}$ & $\boldsymbol{0.881}\pm{0.05}$ \\
        \midrule
        Content & \begin{tabular}[c]{@{}c@{}} Multiple \\  Single \end{tabular} & \begin{tabular}[c]{@{}c@{}} $0.817\pm{0.06}$ \\  $0.874\pm{0.08}$ \end{tabular} & \begin{tabular}[c]{@{}c@{}} $0.779\pm{0.05}$ \\  $0.792\pm{0.06}$ \end{tabular} & \begin{tabular}[c]{@{}c@{}} $\boldsymbol{0.841}\pm{0.05}$  \\  $\boldsymbol{0.933}\pm{0.05}$ \end{tabular}& $0.769\pm{0.05}$ & $0.790\pm{0.05}$ \\
        \bottomrule
    \end{tabular}
    \label{tab:dino_scores}
\end{table*}
\subsection{Comparisons}
We next compare our method with alternative approaches, both qualitatively and quantitatively. Note that since we rely on SDXL as our backbone model, for a fair comparison we applied alternative approaches on SDXL as well. As a na\"ive baseline we employ DB-LoRA \cite{DBlora} (fine-tuned for style) with a ControlNet \cite{Zhang2023AddingCC} for content conditioning. We additionally compare to three recent approaches for image stylization that rely on the prior of large pre-trained text-to-image models, namely, ZipLoRA \cite{ziplora23}, StyleDrop \cite{sohn2023styledrop}, and StyleAligned \cite{styleHertz23}.
StyleAligned is applied using the author's official implementation. With the lack of official implementations for StyleDrop and ZipLoRA, we implemented StyleDrop on SDXL (as described in \cite{styleHertz23}), and utilized a non-official implementation of ZipLoRA \cite{unofficialZipLoRA}.

Note that for content preservation, all three alternative methods require \emph{multiple} content image examples, while our method can be applied to a \emph{single} image.
Thus, for a fair comparison, we collected a total set of 23 objects from existing personalization works \cite{TI22, treeVinker23, dreambooth23, customdiffusion2022}, where a small set of images is provided for each object.
We collected 20 style image references from \cite{styleHertz23,sohn2023styledrop}, along with 5 additional style images of our own.
From these sets, we randomly sampled 50 pairs of style and content images to compose our final evaluation set.

In terms of runtime, StyleAligned is zero-shot only for consistent style generation, while for content preservation it relies on LoRA to adapt the model to the desired concepts. Similarly, StyleDrop and ZipLoRA require LoRA training for content and style. Thus, our runtime is comparable to theirs. In contrast, ZipLoRA entails an additional training phase to merge the two LoRAs, which makes it more time-consuming than our approach.

\vspace{-0.2cm}
\paragraph{Qualitative Evaluation} We show representative comparison results in \Cref{fig:qual_comparison}, where on the left we show the style and content reference images. On the first four rows, we show the results of alternative approaches when applied with multiple content images, whereas our method uses a single image.
As can be seen, our method effectively preserves the subject from the content image while transferring the desired style. In contrast, other methods either overfit the content subject, thereby failing to alter its style (e.g., cat and sloth in ZipLoRA and StyleDrop), or they suffer from style image \ap{leakage}. For instance, in the cat example of StyleAligned (first row), the model generates two cats, matching the number of people in the style reference image. 
We also include an example of alternative methods applied to a single content image, where StyleDrop and ZipLoRA exhibit increased overfitting.
\vspace{-0.2cm}
\paragraph{Quantitative Evaluation} We measure content and style preservation by computing the cosine similarity between the embeddings of the input content and style references and the output image, utilizing the DINO ViT-B/8 embeddings \cite{DINO2021}. The average scores are presented in \Cref{tab:dino_scores}. Our method achieves the highest style alignment score, indicating its superior ability to adapt styles effectively. However, we observe lower object similarity scores, possibly due to content overfitting issues observed in alternative approaches.

To further support this observation, we conducted the same experiment using a single content image as a reference (scores shown in the \ap{Single} row). The results indicate a decrease in style consistency scores across all methods, accompanied by an increase in content preservation scores, suggesting overfitting.

\vspace{-0.2cm}
\paragraph{User Study} We conducted a user study to further validate the findings presented above.
Using $30$ random images from our evaluation set, we compared our results with the three alternative approaches. 
The participants were presented with the reference style and content images along with two combined results, one produced by our method and the other by an alternative method (with the results presented in random order). Participants were asked to choose the result that ``better transfers the style from the style image while preserving the content of the content image''.
We collected responses from $34$ participants for the survey, which contained a total of $1020$ answers. The results demonstrate a strong preference for our method, with 94\% of participants favoring our method over StyleAligned, 91\% over ZipLoRA, and 88\% over StyleDrop.

\section{Conclusions, Limitations and Future work}
\begin{figure}%
    \centering~\includegraphics[width=1\linewidth]{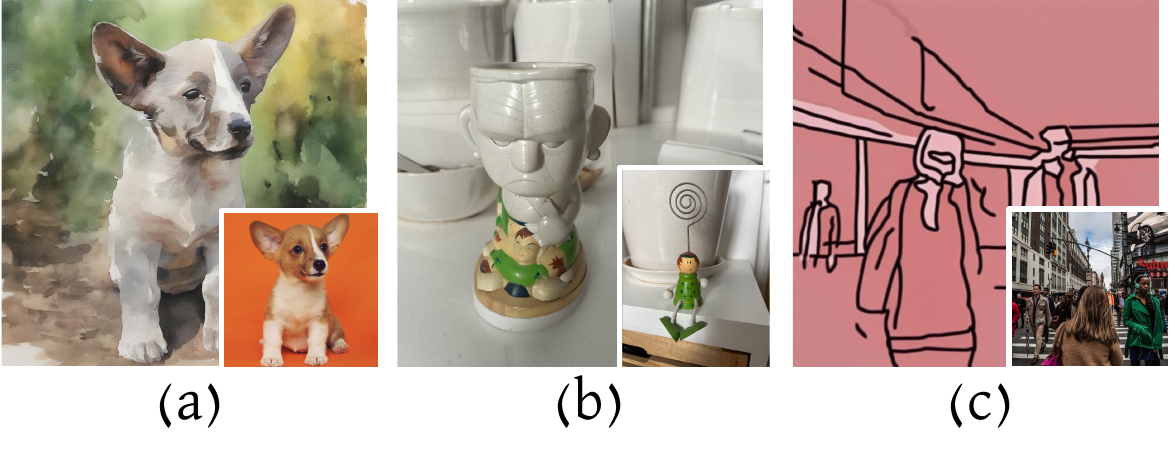}
    \vspace{-0.7cm}
    \caption{Method limitations. (a) Sub-optimal identity preservation due to color separation. (b) Style leakage from background objects. (c) Inability to adequately capture content in complex scenes.}
    \vspace{-0.2cm}
    \label{fig:limitation}
\end{figure}

We have presented a simple yet effective method to disentangle the style and content of a single input image. The style and content components are encoded separately with two B-LoRAs, providing high flexibility for independent use in various image stylization tasks. 
In contrast to existing methods that focus on style extraction, we employ a compound style-content learning approach that enables a better separation of style and content, enhancing stylization fidelity. 
While our work enables robust image stylization across various complex input images, it does have limitations. First, in our style-content separation procedure, the object's color is often included in the style component.
However, in some cases, color plays a crucial role in preserving identity. Therefore, when stylizing the content component, the results may not properly preserve the object's identity, as illustrated in \Cref{fig:limitation}(a).
Second, since we use a single reference image, our learned style component may encompass background elements rather than focusing solely on the central object, as demonstrated in \Cref{fig:limitation}(b).
Lastly, while our method effectively stylizes scene images, it may encounter challenges with complex scenes containing numerous elements. Consequently, it may struggle to accurately capture the scene structure, potentially compromising content preservation, as depicted in \Cref{fig:limitation}(c).

As for future research, one possible avenue is to further explore separation techniques within LoRA fine-tuning, to achieve more concrete separation into sub-components such as structure, shape, color, texture, etc. This could provide users with more control over the desired output.
Another direction for future work is to leverage the robustness of our approach and extend it to combine LoRA weights from multiple distinct objects or combine a few styles.

\section{Acknowledgements}
We would like to thank Amir Hertz and Yuval Alaluf for their insightful feedback. Additionally, some of the artistic paintings presented in this paper were created by the artist Judith Kondor Mochary. We thank the artist's family for granting us the privilege to use Judith's drawings.
This work was supported by the Israel Science Foundation under Grant No. 2492/20, 3441/21 and 1390/19, and Joint NSFC-ISF Research Grant Research Grant no. 3077/23.

{
    \small
    \bibliographystyle{ieeenat_fullname}
    \bibliography{egbib}
}

\clearpage
\appendix
\newpage
\twocolumn[
\centering
\Large
\textbf{\thetitle}\\
\vspace{5em}
Supplementary Material \\
\vspace{1.0em}
] %
\part{}
\vspace{-40pt}
\parttoc

\section{Comparisons} 
\begin{figure}[!ht]
  \centering\includegraphics[width=0.9\linewidth]{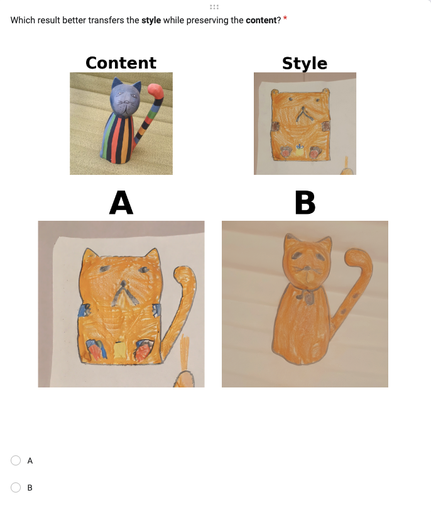}
  \caption{Screenshot from the user study. Each of the two images, labeled A and B, represents a result obtained from a different method. Participants were tasked with selecting the image they believe is better in terms of both style adaptation and content preservation.}
  \label{fig:user_study}
\end{figure}

\paragraph{User Study}
As described in the main paper, we conducted a user study to further validate our findings. We constructed an evaluation set comprising 50 unique pairs of style and content images, randomly sampled from a diverse pool of 23 objects and 25 style references. From this evaluation set, we selected 10 representative pairs for each of the competing methods: ZipLoRA, StyleDrop, and StyleAligned.
For each pair, we generated images using both the respective method and our approach, presenting them alongside the original style and content references, as illustrated in \Cref{fig:user_study} The generated images were displayed in a randomized order to avoid bias. Participants were asked to choose the result that \ap{better transfers the style from the style image while preserving the content of the content image.}
In total, we gathered 1020 responses from 34 participants, ensuring a comprehensive evaluation of our method against alternative approaches.

 \vspace{-10pt}

\begin{figure*}[t]
    \centering
    \setlength{\tabcolsep}{1.5pt}
    {\small
    \begin{tabular}{c c @{\hspace{0.17cm}} | @{\hspace{0.17cm}}c c c c @{\hspace{0.17cm}} | @{\hspace{0.17cm}}c}
        
     Content & Style & DB-LoRA & ZipLoRA 
         & \begin{tabular}[c]{@{}c@{}} StyleDrop \\ SDXL \end{tabular}
         & \begin{tabular}[c]{@{}c@{}} Style- \\ Aligned \end{tabular}
          & Ours \\
          
        \includegraphics[width=0.132\textwidth]{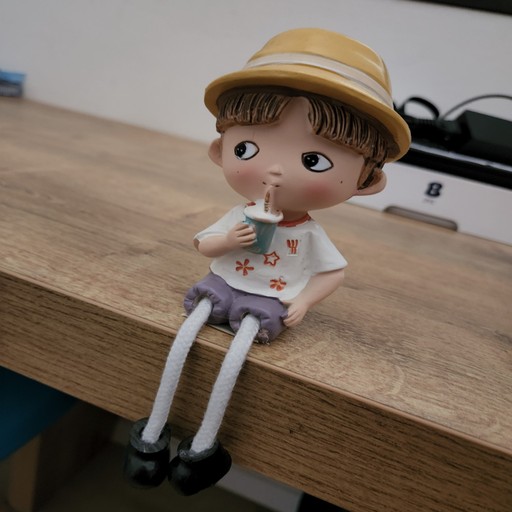} &
        \includegraphics[width=0.132\textwidth]{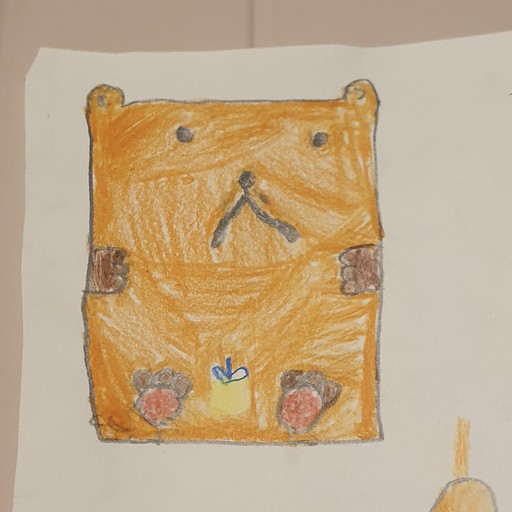} &
        
        \includegraphics[width=0.132\textwidth]{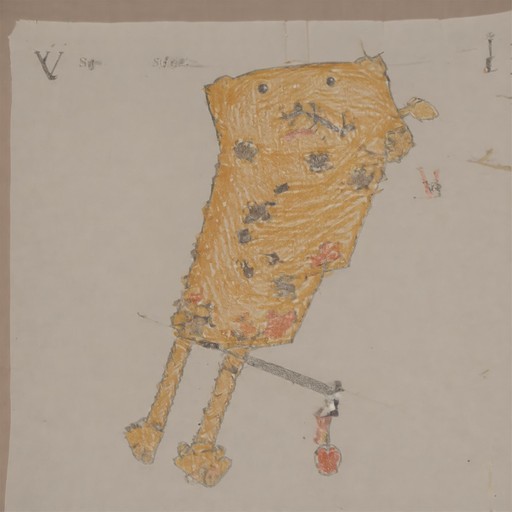} &
        \includegraphics[width=0.132\textwidth]{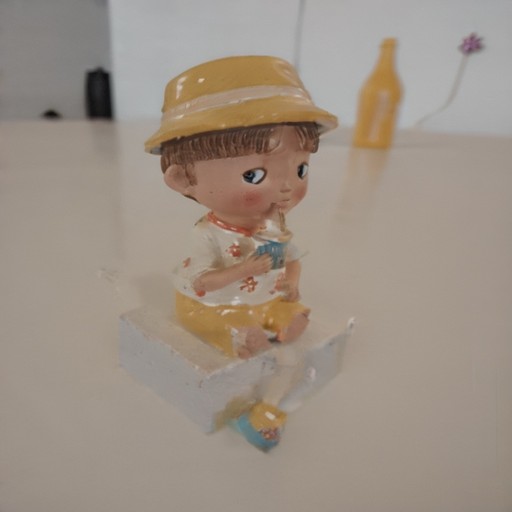} &
        \includegraphics[width=0.132\textwidth]{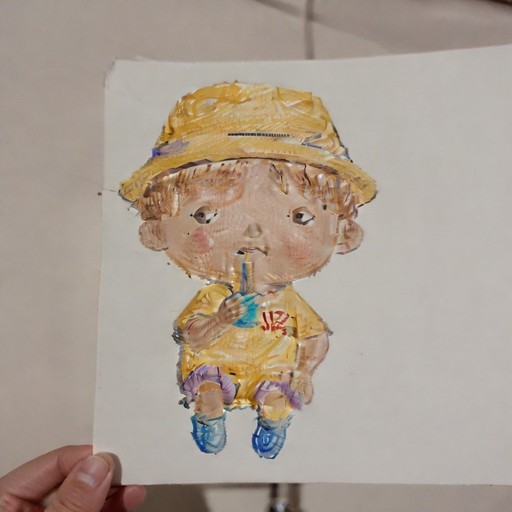} &
        \includegraphics[width=0.132\textwidth]{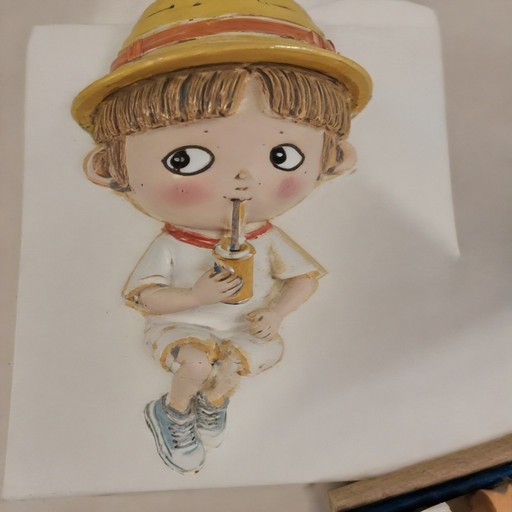} &
        \includegraphics[width=0.132\textwidth]{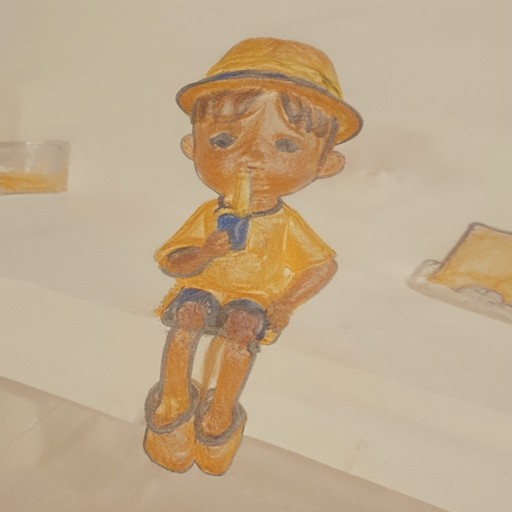} \\
        
        \includegraphics[width=0.132\textwidth]{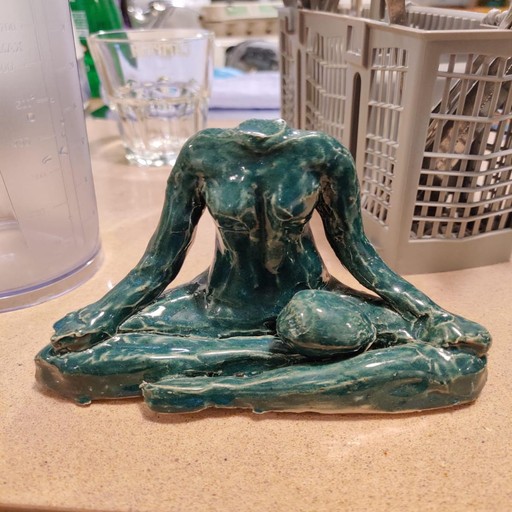} &
        \includegraphics[width=0.132\textwidth]{temp_figs/style_images/drawing3.png} &
        
        \includegraphics[width=0.132\textwidth]{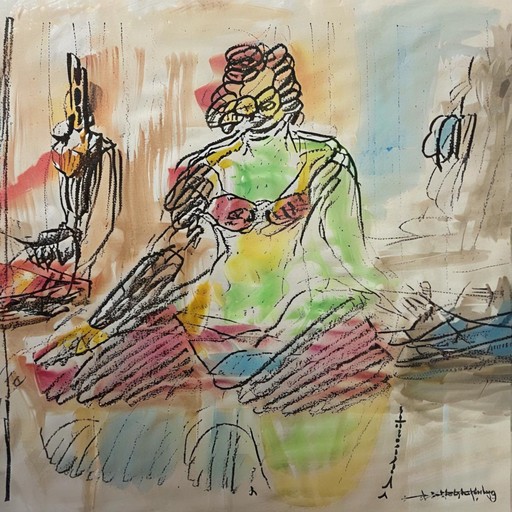} &
        \includegraphics[width=0.132\textwidth]{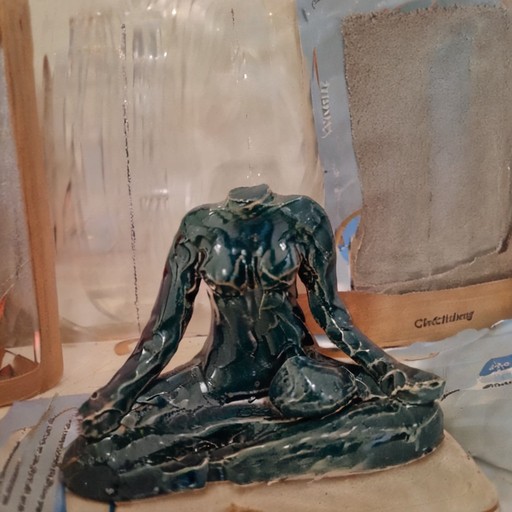} &
        \includegraphics[width=0.132\textwidth]{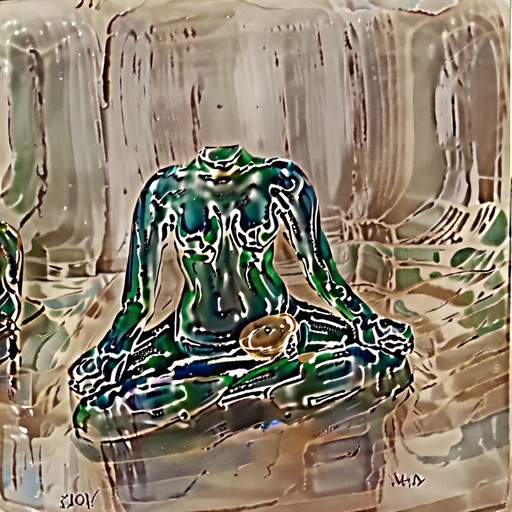} &
        \includegraphics[width=0.132\textwidth]{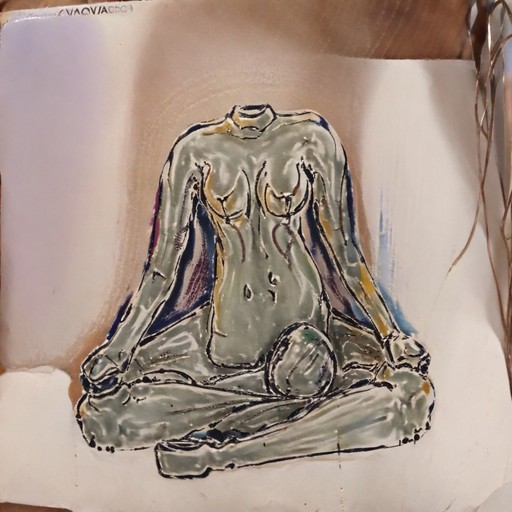} &
        \includegraphics[width=0.132\textwidth]{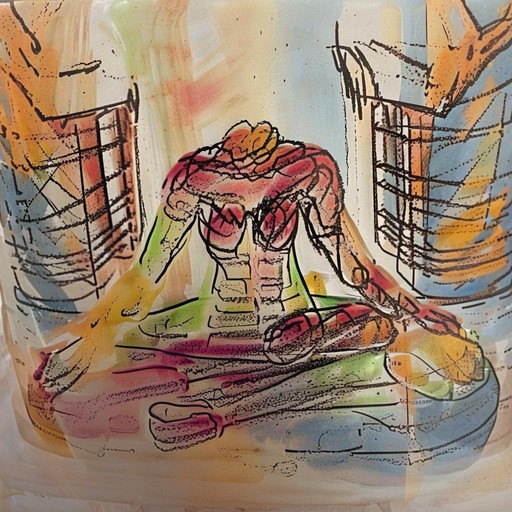} \\

        \includegraphics[width=0.132\textwidth]{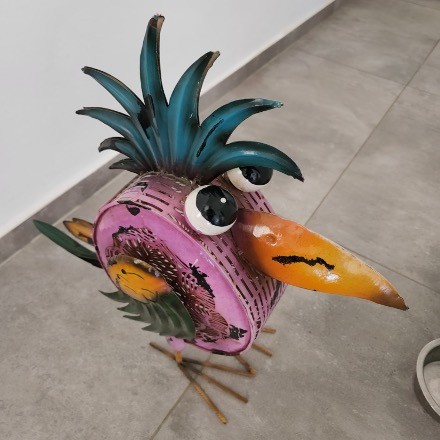} &
        \includegraphics[width=0.132\textwidth]{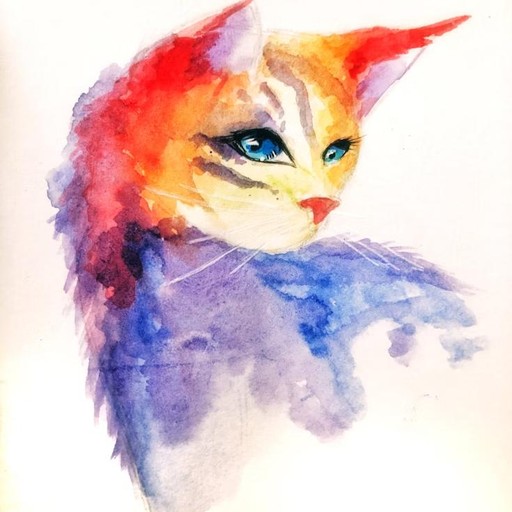} &
        
        \includegraphics[width=0.132\textwidth]{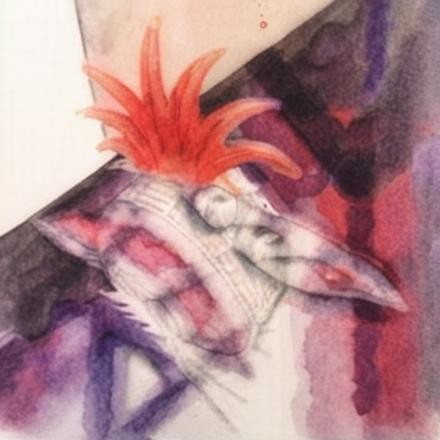} &
        \includegraphics[width=0.132\textwidth]{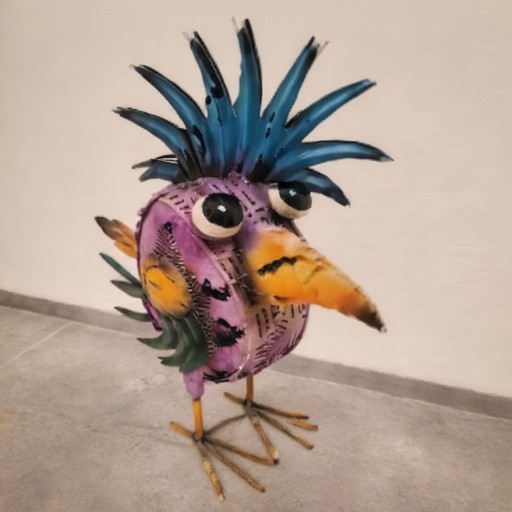} &
        \includegraphics[width=0.132\textwidth]{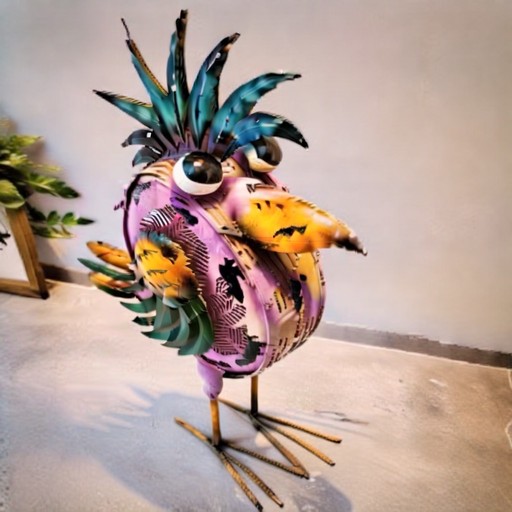} &
        \includegraphics[width=0.132\textwidth]{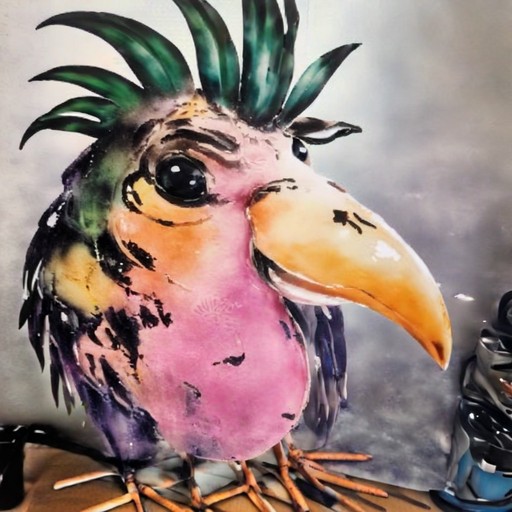} &
        \includegraphics[width=0.132\textwidth]{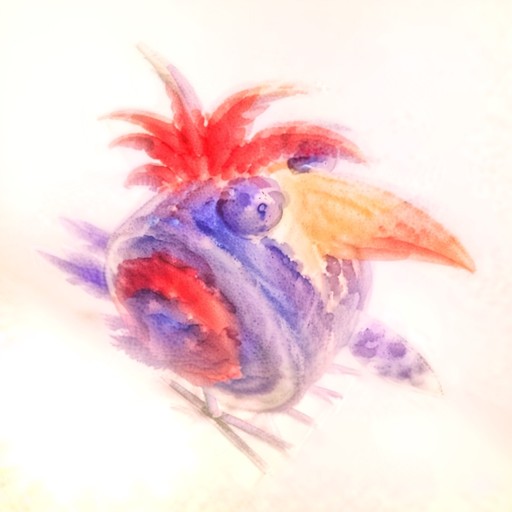} \\

        \includegraphics[width=0.132\textwidth]{temp_figs/content_images/bull.jpg} &
        \includegraphics[width=0.132\textwidth]{temp_figs/style_images/kiss.png} &
        
        \includegraphics[width=0.132\textwidth]{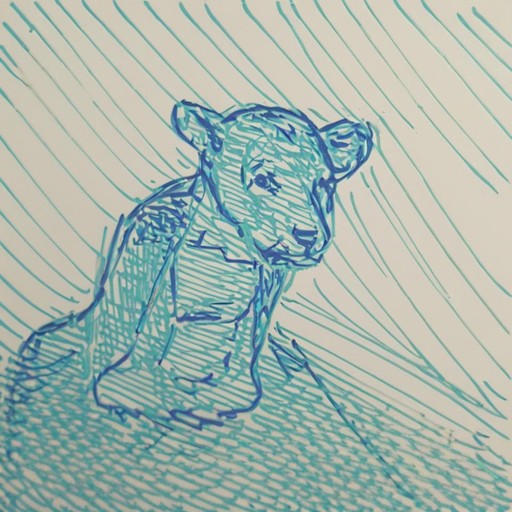} &
        \includegraphics[width=0.132\textwidth]{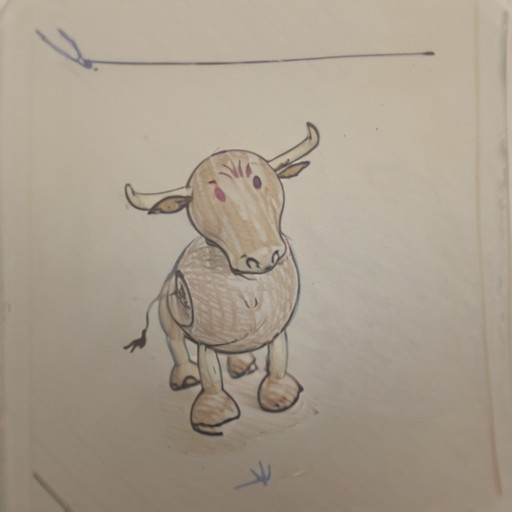} &
        \includegraphics[width=0.132\textwidth]{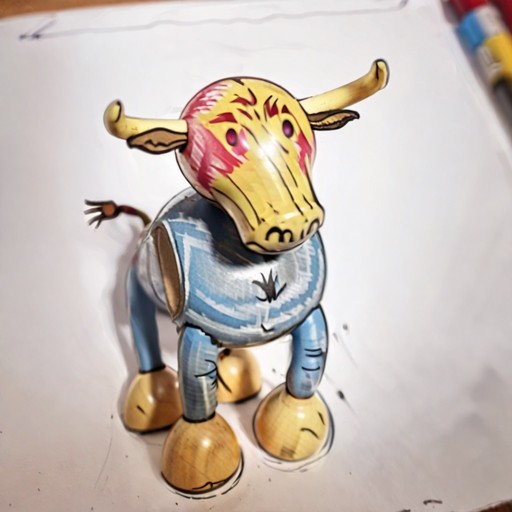} &
        \includegraphics[width=0.132\textwidth]{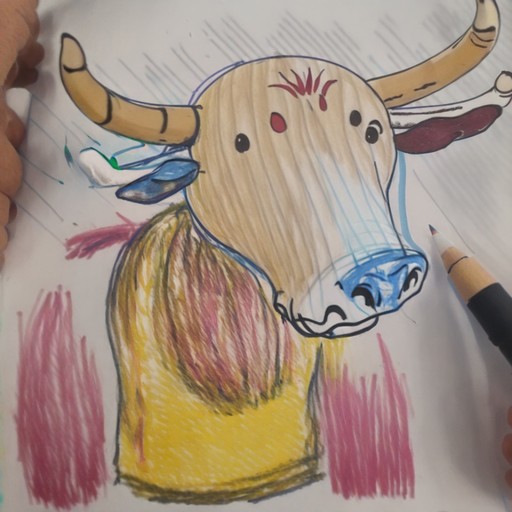} &
        \includegraphics[width=0.132\textwidth]{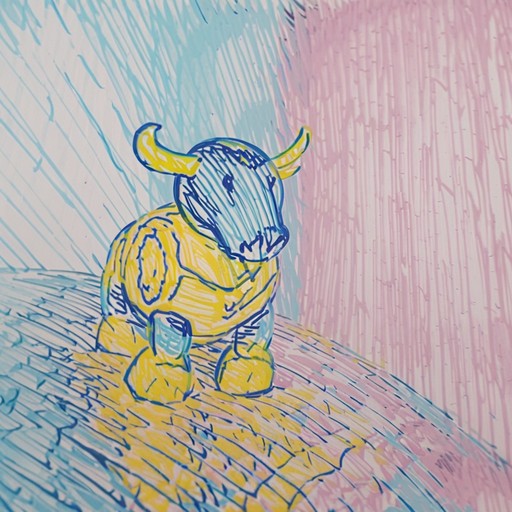} \\

        \includegraphics[width=0.132\textwidth]{temp_figs/content_images/fat_bird.jpg} &
        \includegraphics[width=0.132\textwidth]{temp_figs/style_images/pen_sketch.jpeg} &
        
        \includegraphics[width=0.132\textwidth]{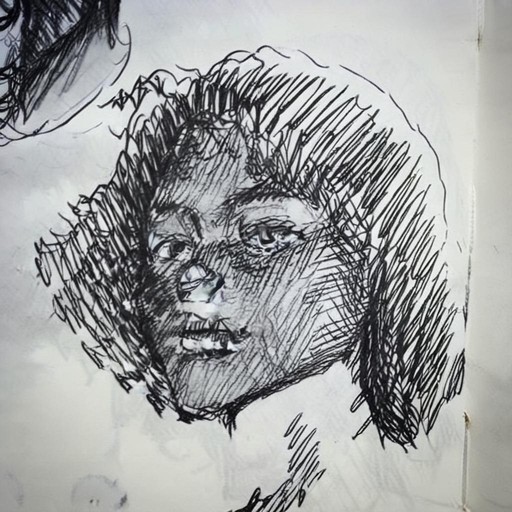} &
        \includegraphics[width=0.132\textwidth]{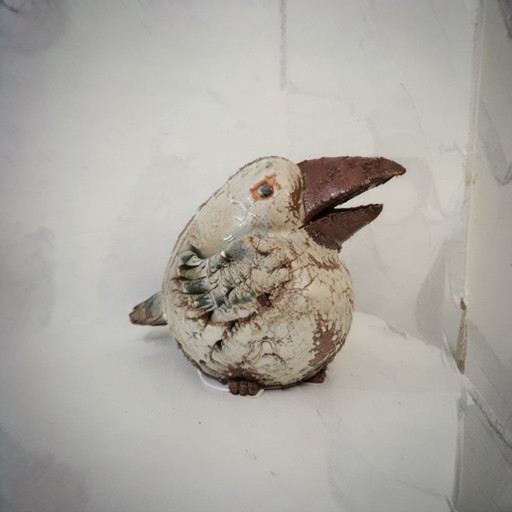} &
        \includegraphics[width=0.132\textwidth]{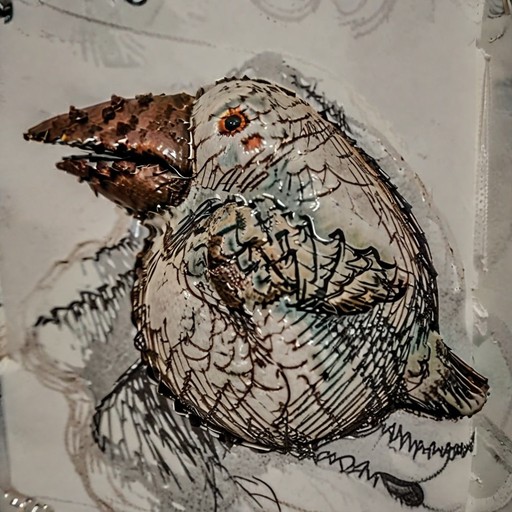} &
        \includegraphics[width=0.132\textwidth]{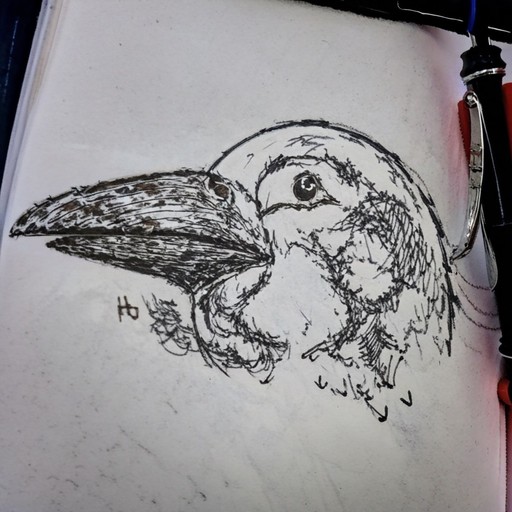} &
        \includegraphics[width=0.132\textwidth]{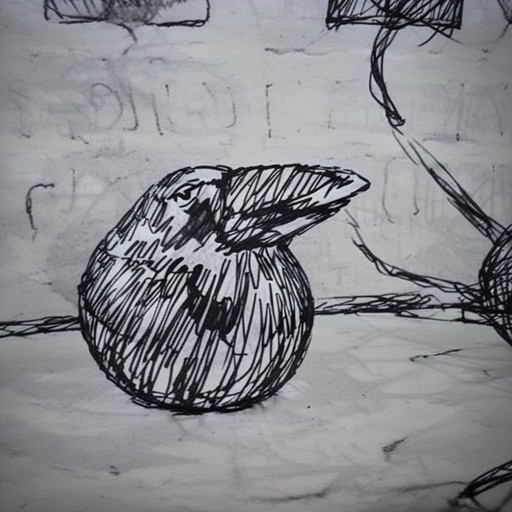} \\

        \includegraphics[width=0.132\textwidth]{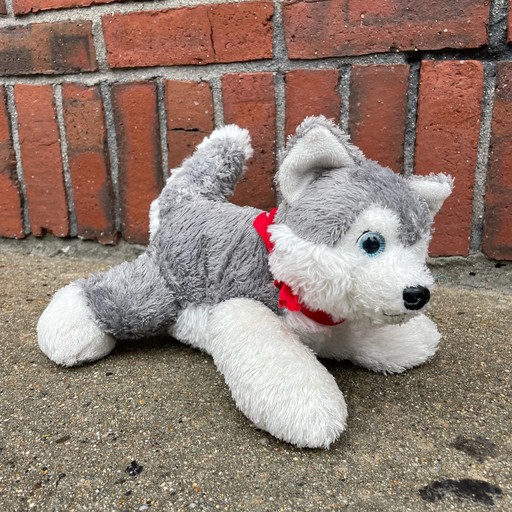} &
        \includegraphics[width=0.132\textwidth]{temp_figs/style_images/painting.jpg} &
        
        \includegraphics[width=0.132\textwidth]{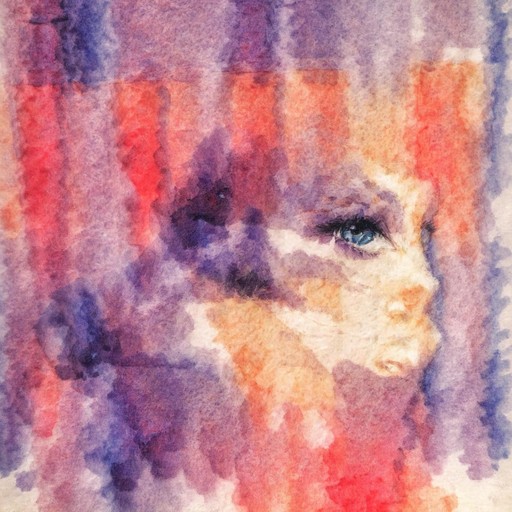} &
        \includegraphics[width=0.132\textwidth]{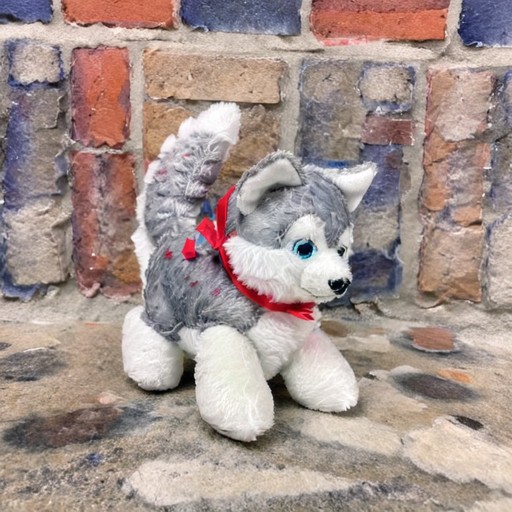} &
        \includegraphics[width=0.132\textwidth]{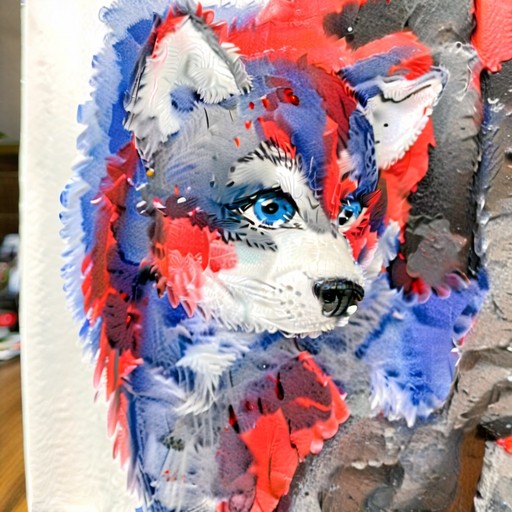} &
        \includegraphics[width=0.132\textwidth]{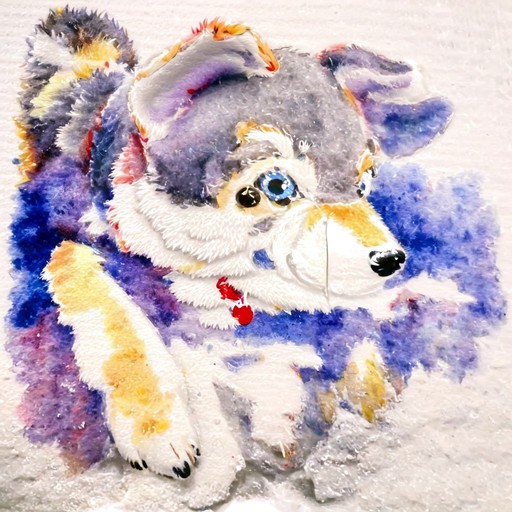} &
        \includegraphics[width=0.132\textwidth]{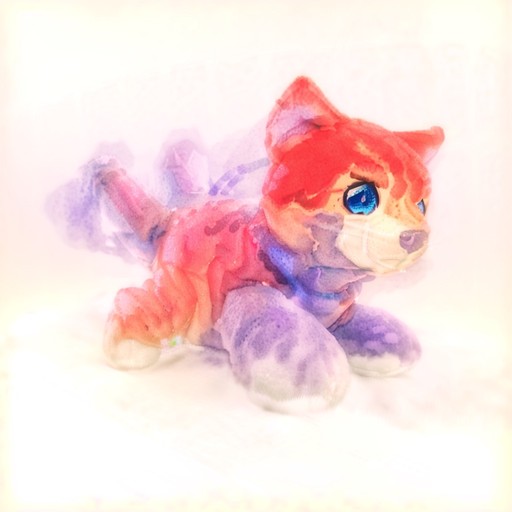} \\

        \includegraphics[width=0.132\textwidth]{temp_figs/content_images/dog2.jpg} &
        \includegraphics[width=0.132\textwidth]{temp_figs/style_images/kiss.png} &
        
        \includegraphics[width=0.132\textwidth]{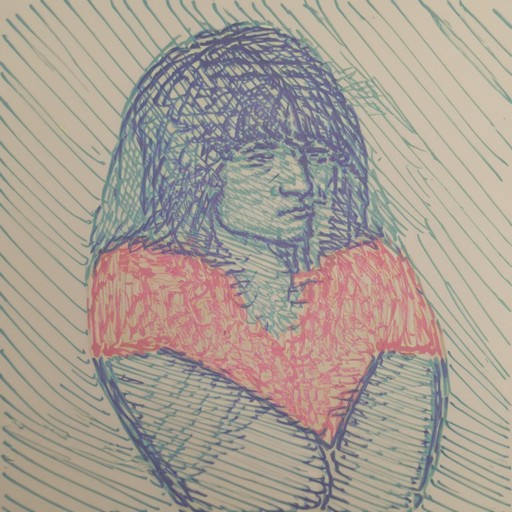} &
        \includegraphics[width=0.132\textwidth]{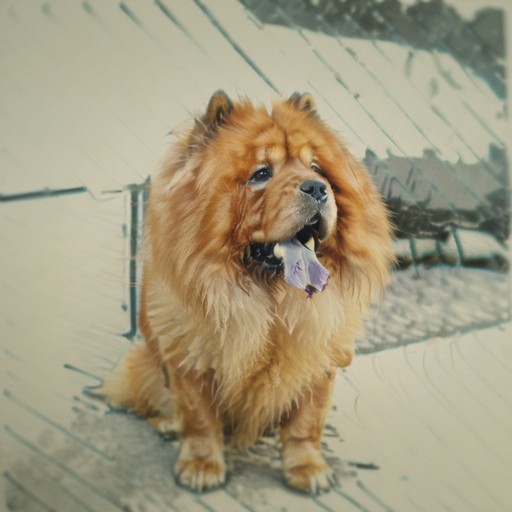} &
        \includegraphics[width=0.132\textwidth]{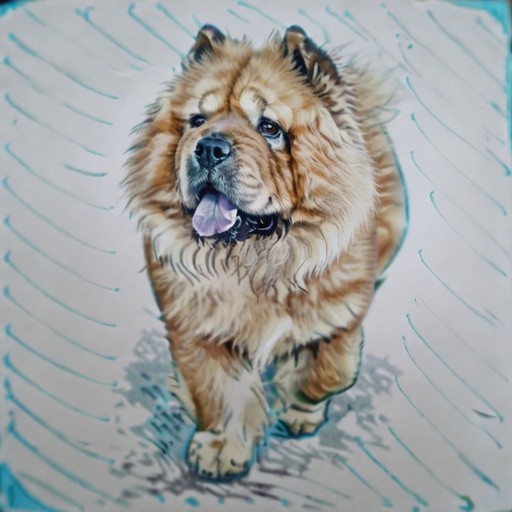} &
        \includegraphics[width=0.132\textwidth]{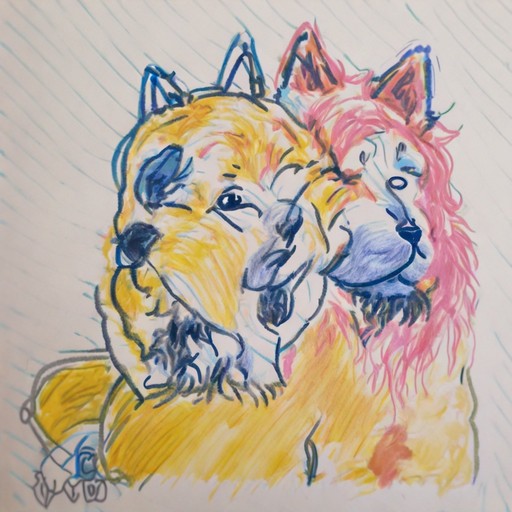} &
        \includegraphics[width=0.132\textwidth]{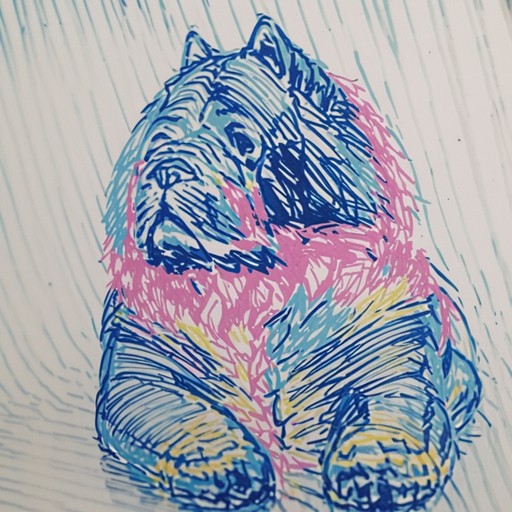} \\
        
    \end{tabular}
    }   
    \vspace{-0.11cm}
    \caption{Additional comparisons for image stylization based on reference image.}
    \label{fig:additional_comparison}
\end{figure*}

\begin{figure*}[t]
    \centering
    \setlength{\tabcolsep}{1.5pt}
    {\small
    \begin{tabular}{c c @{\hspace{0.17cm}} | @{\hspace{0.17cm}}c c c c @{\hspace{0.17cm}} | @{\hspace{0.17cm}}c}
        
     Content & Style & DB-LoRA & ZipLoRA 
         & \begin{tabular}[c]{@{}c@{}} StyleDrop \\ SDXL \end{tabular}
         & \begin{tabular}[c]{@{}c@{}} Style- \\ Aligned \end{tabular}
          & Ours \\
          
        \includegraphics[width=0.132\textwidth]{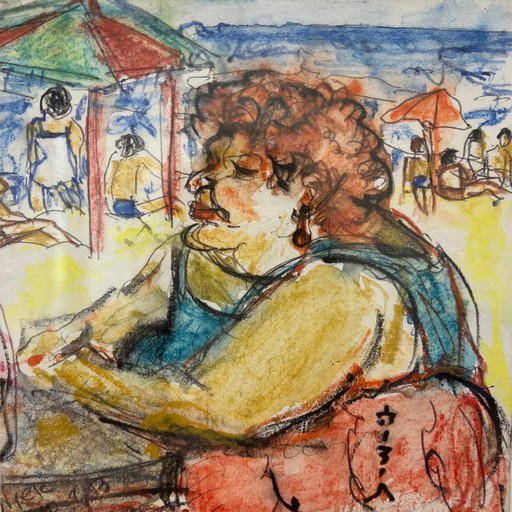} &
        \includegraphics[width=0.132\textwidth]{temp_figs/style_images/cartoon_line.png} &
        
        \includegraphics[width=0.132\textwidth]{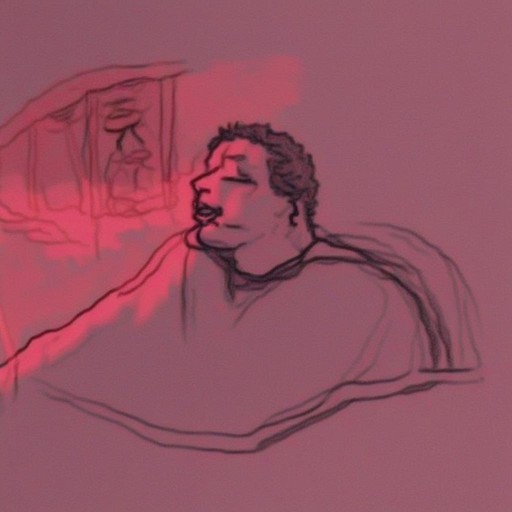} &
        \includegraphics[width=0.132\textwidth]{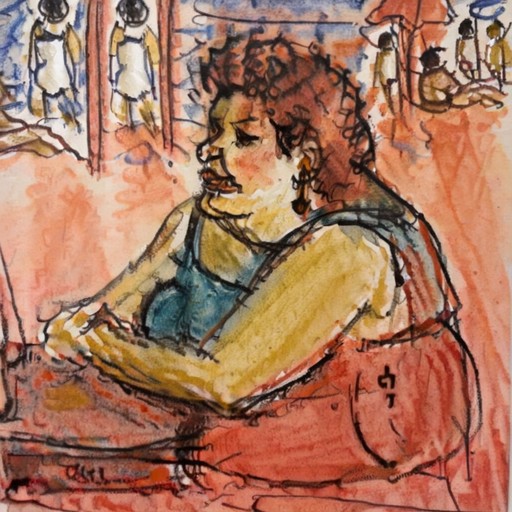} &
        \includegraphics[width=0.132\textwidth]{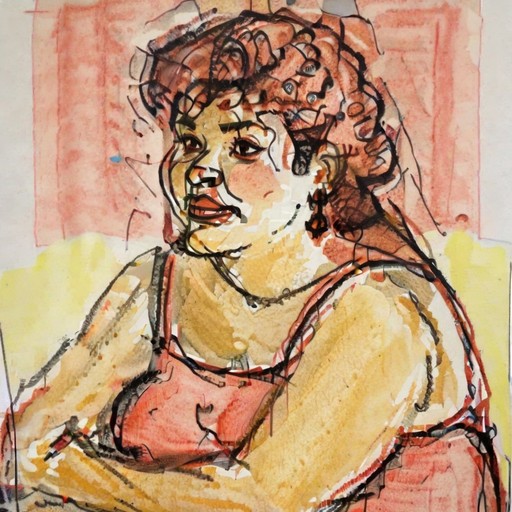} &
        \includegraphics[width=0.132\textwidth]{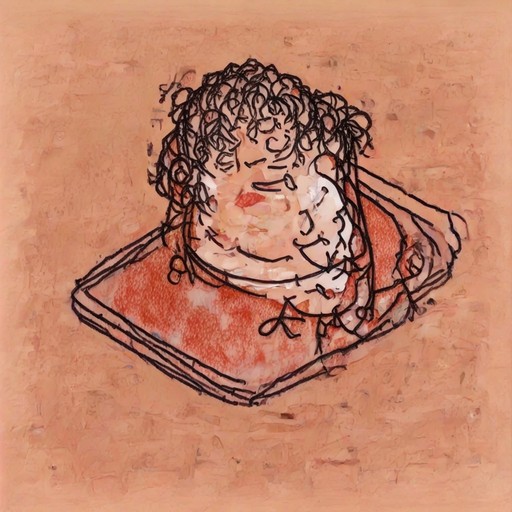} &
        \includegraphics[width=0.132\textwidth]{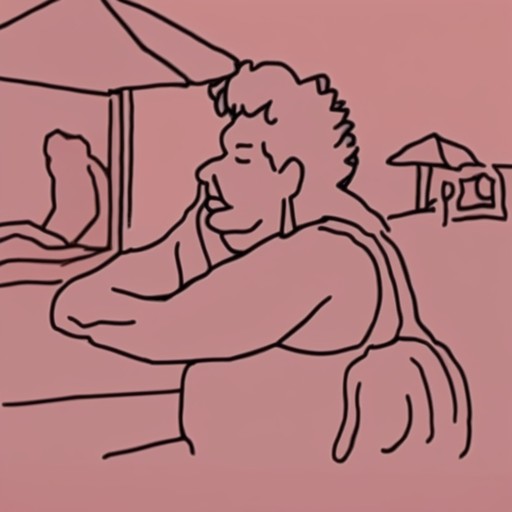} \\

        \includegraphics[width=0.132\textwidth]{temp_figs/style_images/working_cartoon.jpg} &
        \includegraphics[width=0.132\textwidth]{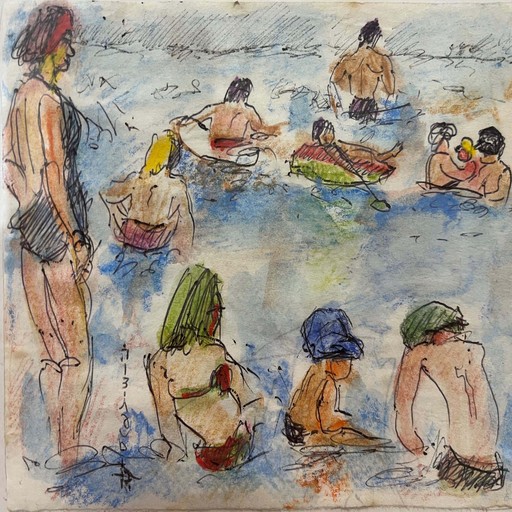} &
        \includegraphics[width=0.132\textwidth]{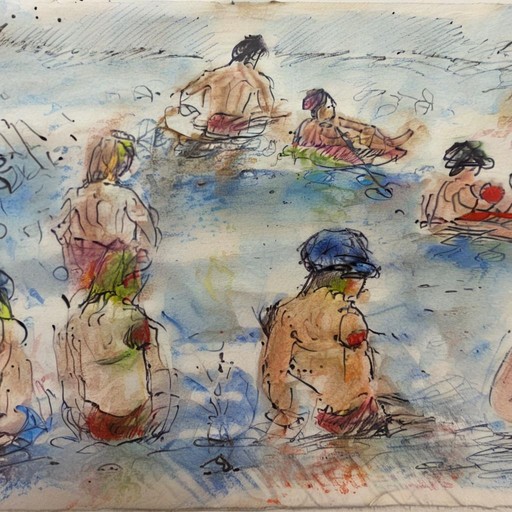} &
        \includegraphics[width=0.132\textwidth]{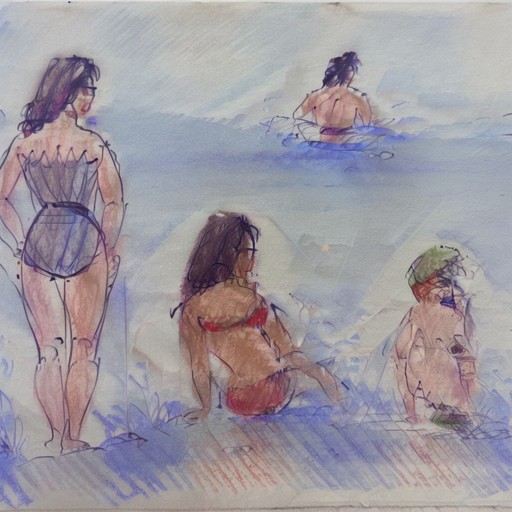} &
        \includegraphics[width=0.132\textwidth]{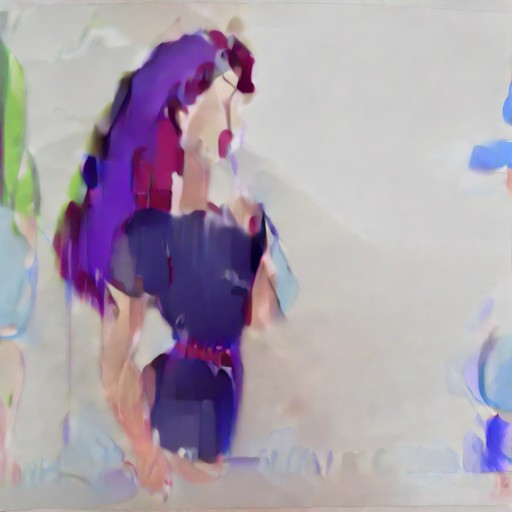} &
        \includegraphics[width=0.132\textwidth]{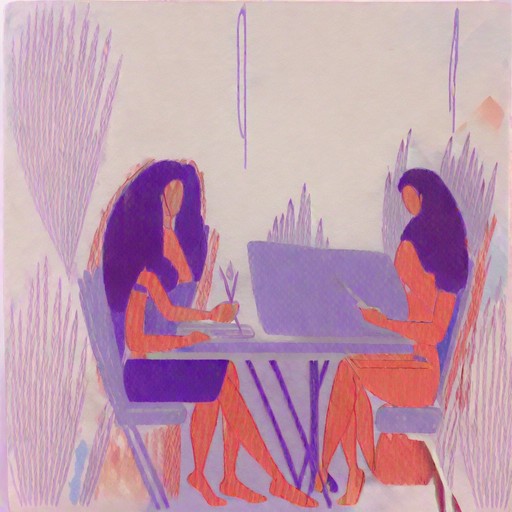} &
        \includegraphics[width=0.132\textwidth]{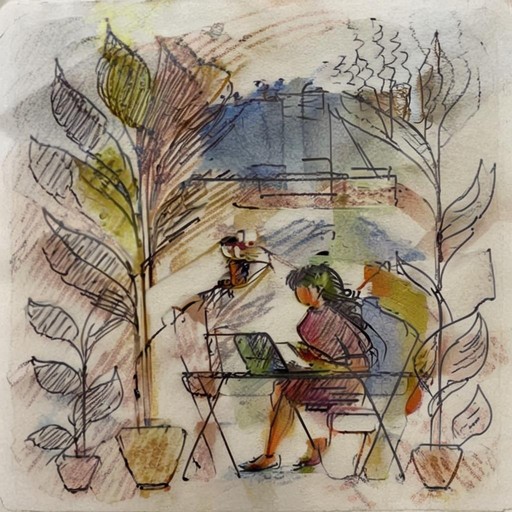} \\

        \includegraphics[width=0.132\textwidth]{temp_figs/style_images/painting.jpg} &
        \includegraphics[width=0.132\textwidth]{temp_figs/style_images/drawing3.png} &
        \includegraphics[width=0.132\textwidth]{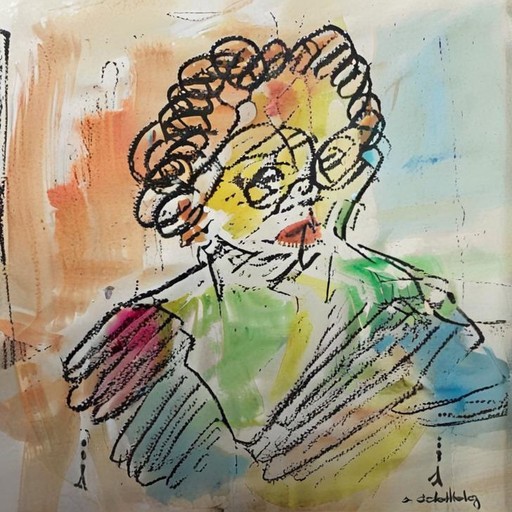} &
        \includegraphics[width=0.132\textwidth]{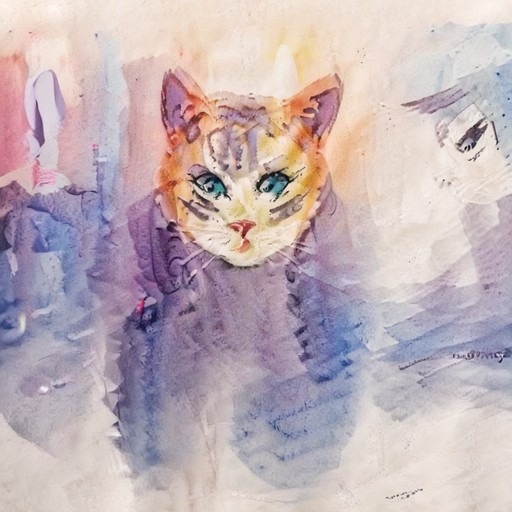} &
        \includegraphics[width=0.132\textwidth]{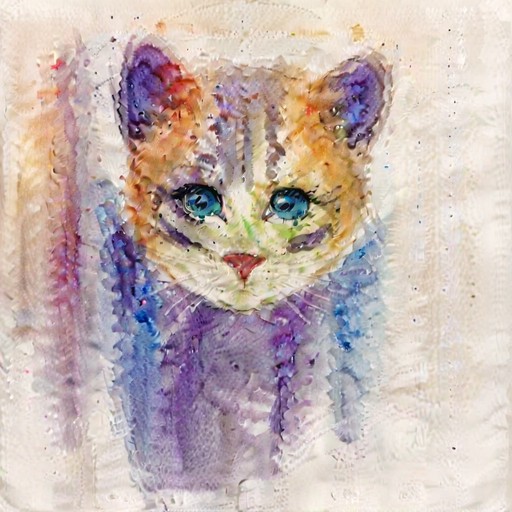} &
        \includegraphics[width=0.132\textwidth]{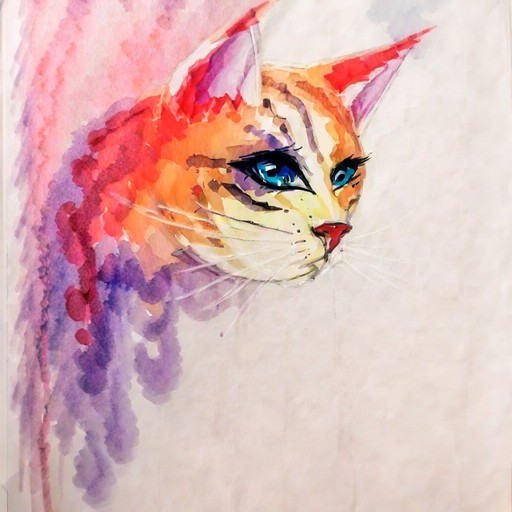} &
        \includegraphics[width=0.132\textwidth]{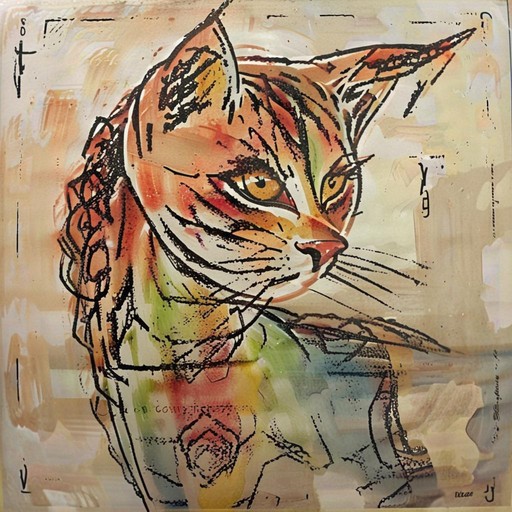} \\
        
        \includegraphics[width=0.132\textwidth]{temp_figs/style_images/crayon_drawing.jpg} &
        \includegraphics[width=0.132\textwidth]{temp_figs/style_images/pen_sketch.jpeg} &
        \includegraphics[width=0.132\textwidth]{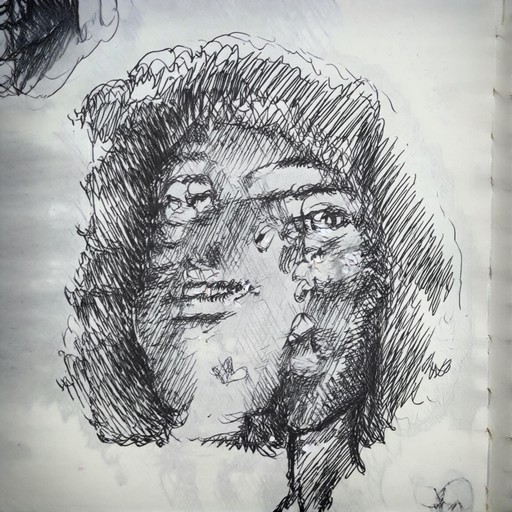} &
        \includegraphics[width=0.132\textwidth]{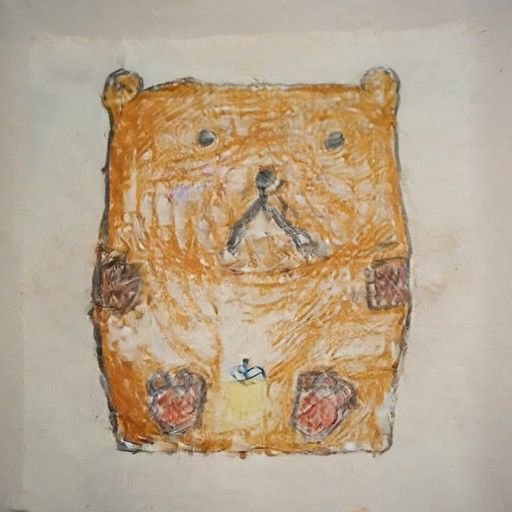} &
        \includegraphics[width=0.132\textwidth]{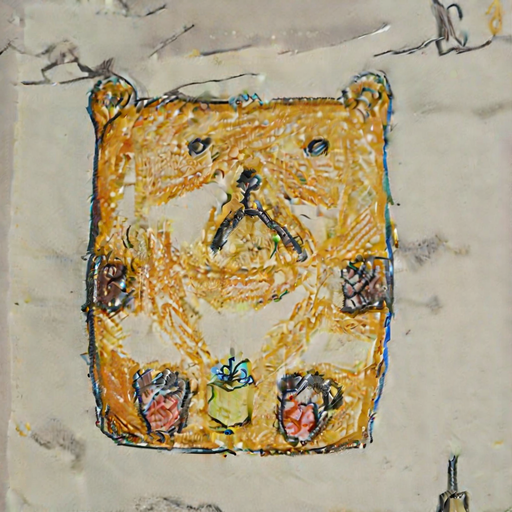} &
        \includegraphics[width=0.132\textwidth]{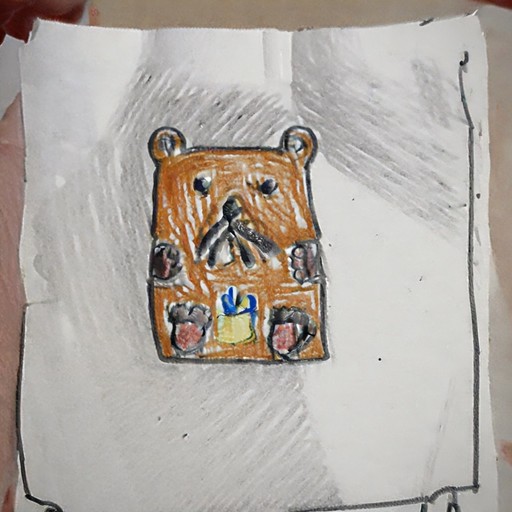} &
        \includegraphics[width=0.132\textwidth]{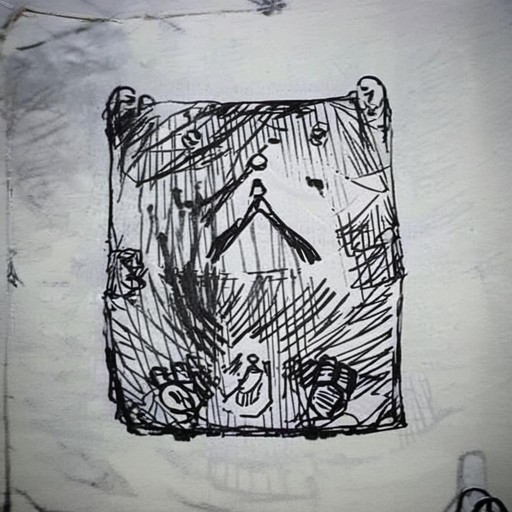} \\

        \includegraphics[width=0.132\textwidth]{temp_figs/style_images/kiss.png} &
        \includegraphics[width=0.132\textwidth]{temp_figs/style_images/painting.jpg} &
        \includegraphics[width=0.132\textwidth]{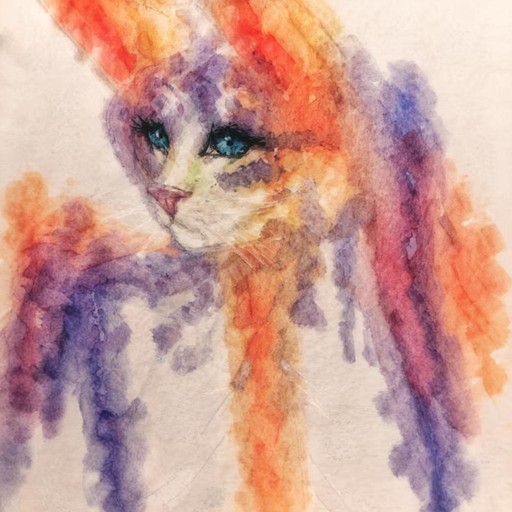} &
        \includegraphics[width=0.132\textwidth]{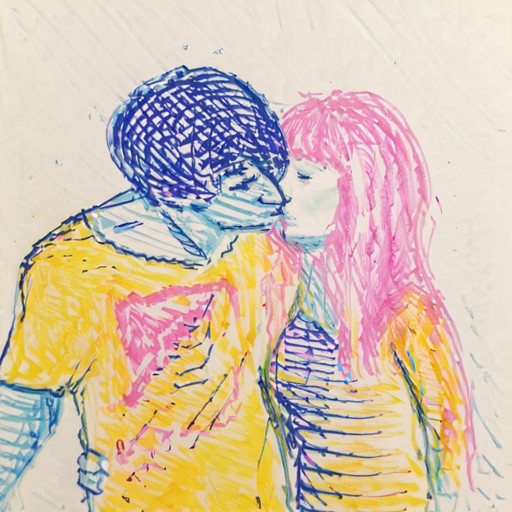} &
        \includegraphics[width=0.132\textwidth]{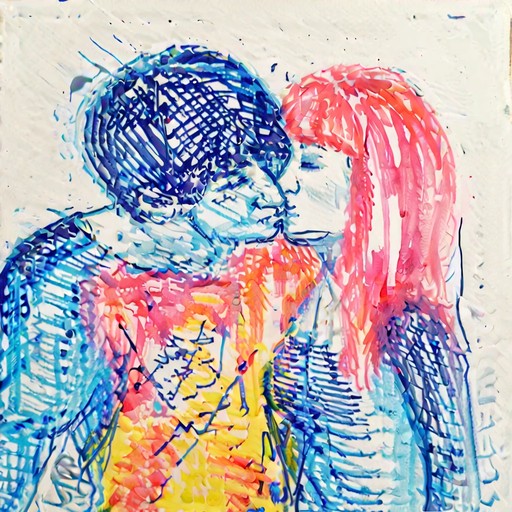} &
        \includegraphics[width=0.132\textwidth]{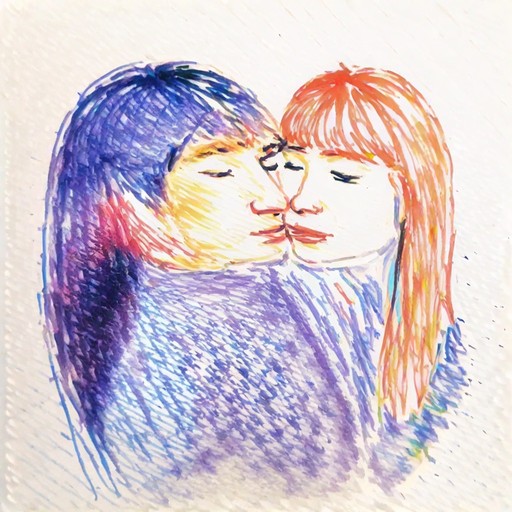} &
        \includegraphics[width=0.132\textwidth]{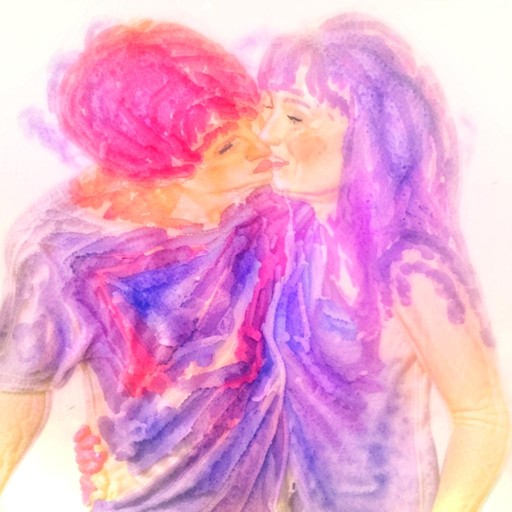} \\

    \end{tabular}
    }   
    \vspace{-0.11cm}
    \caption{Additional comparisons using challenging stylized images as content input. As can be seen, other methods encounter difficulties in disentangling the style and content from these images, consequently struggling to effectively transfer the style from one stylized image to another.  \copyright The paintings in the first three rows are by Judith Kondor Mochary}
    \label{fig:style_to_style_compare}
\end{figure*}

\begin{figure*}[t]
    \centering
    \setlength{\tabcolsep}{1.5pt}
    {\small
    \begin{tabular}{c c @{\hspace{0.17cm}} | @{\hspace{0.17cm}}c c c @{\hspace{0.17cm}} | @{\hspace{0.17cm}}c}
        
     Content & Style & DB-LoRA & ZipLoRA 
         & \begin{tabular}[c]{@{}c@{}} Style- \\ Aligned \end{tabular}
          & Ours \\
          
        \includegraphics[width=0.155\textwidth]{temp_figs/content_images/dog2.jpg} &
        \includegraphics[width=0.155\textwidth]{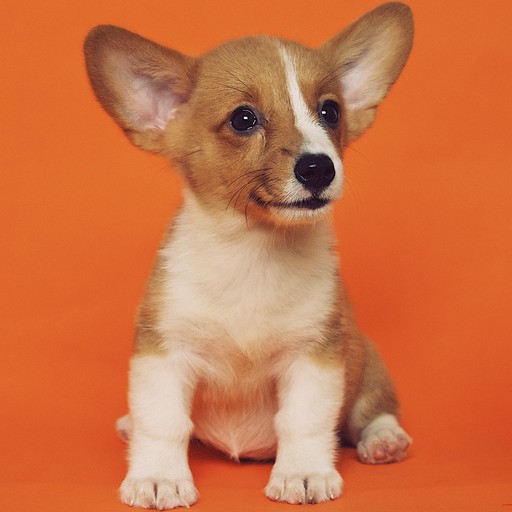} &
        
        \includegraphics[width=0.155\textwidth]{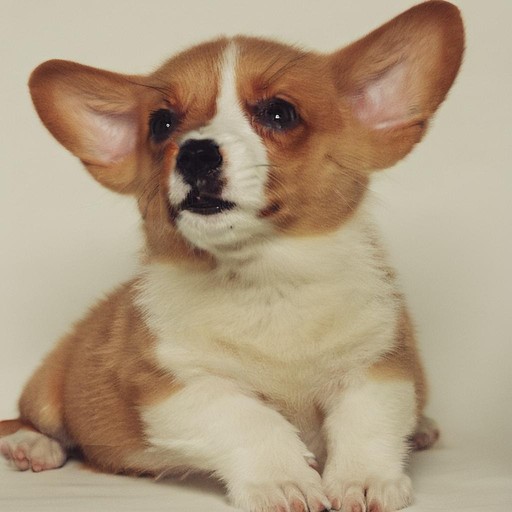} &
        \includegraphics[width=0.155\textwidth]{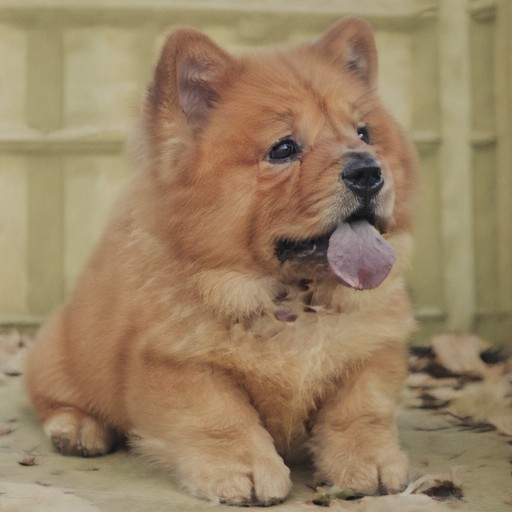} &
        \includegraphics[width=0.155\textwidth]{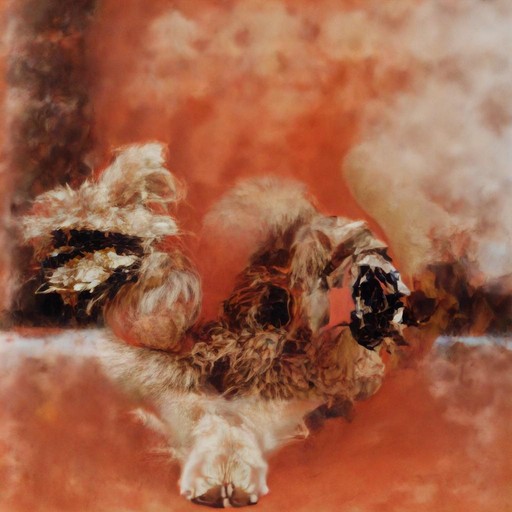} &
        \includegraphics[width=0.155\textwidth]{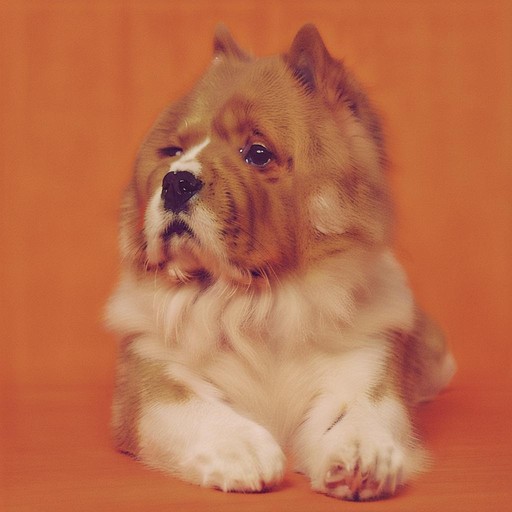} \\

        \includegraphics[width=0.155\textwidth]{temp_figs/content_images/bull.jpg} &
        \includegraphics[width=0.155\textwidth]{temp_figs/content_images/wolf_plushie.jpg} &
        \includegraphics[width=0.155\textwidth]{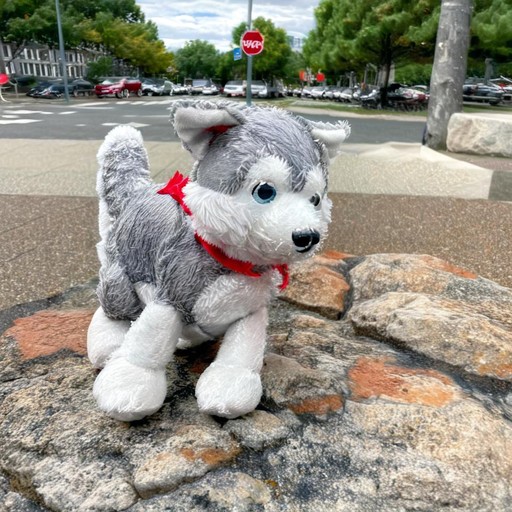} &
        \includegraphics[width=0.155\textwidth]{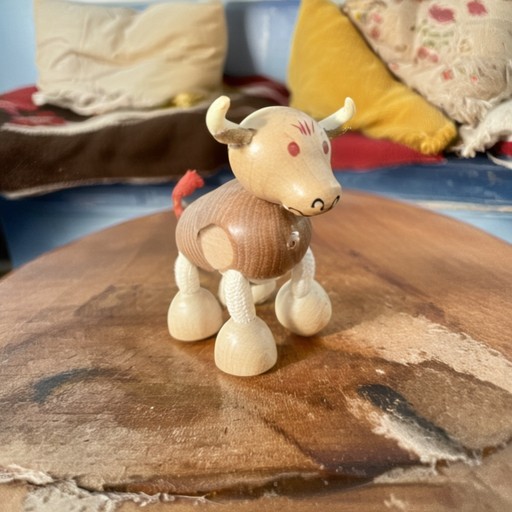} &
        \includegraphics[width=0.155\textwidth]{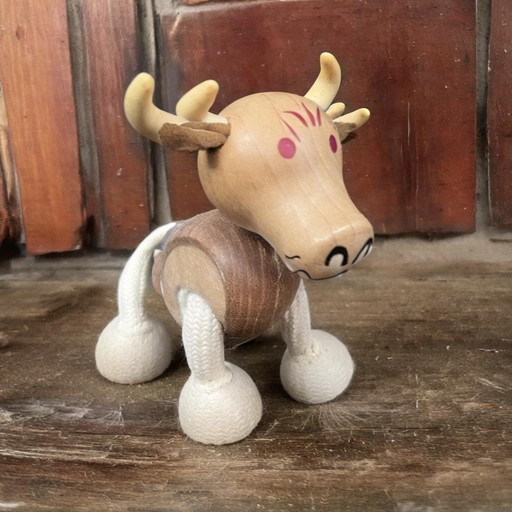} &
        \includegraphics[width=0.155\textwidth]{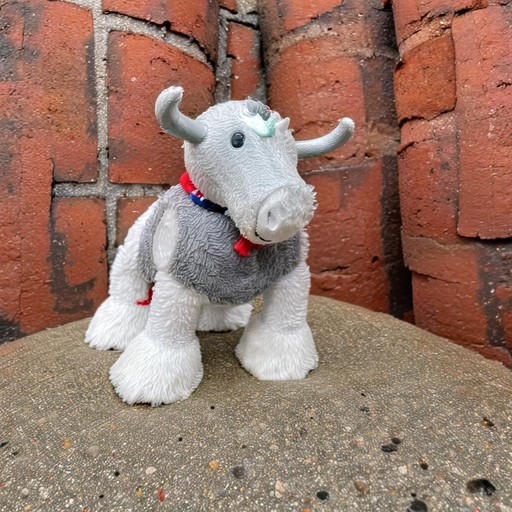} \\

        \includegraphics[width=0.155\textwidth]{temp_figs/content_images/statue.jpg} &
        \includegraphics[width=0.155\textwidth]{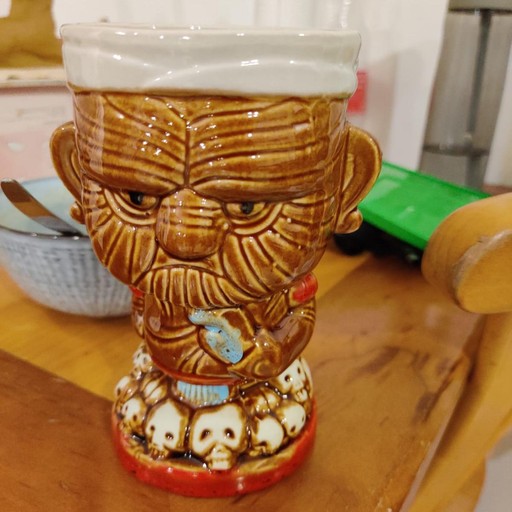} &
        \includegraphics[width=0.155\textwidth]{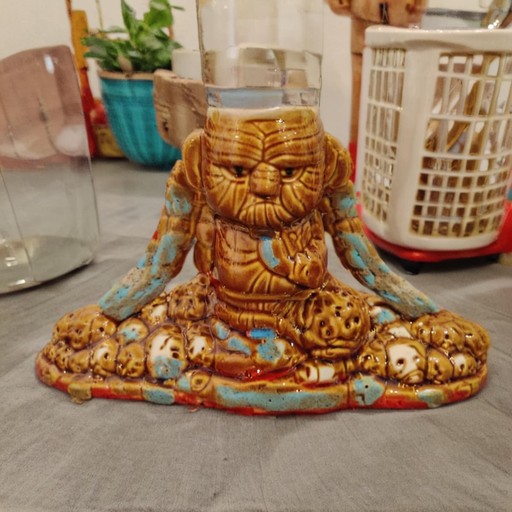} &
        \includegraphics[width=0.155\textwidth]{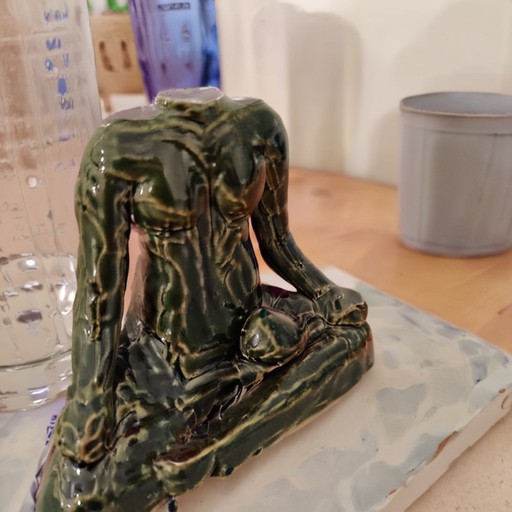} &
        \includegraphics[width=0.155\textwidth]{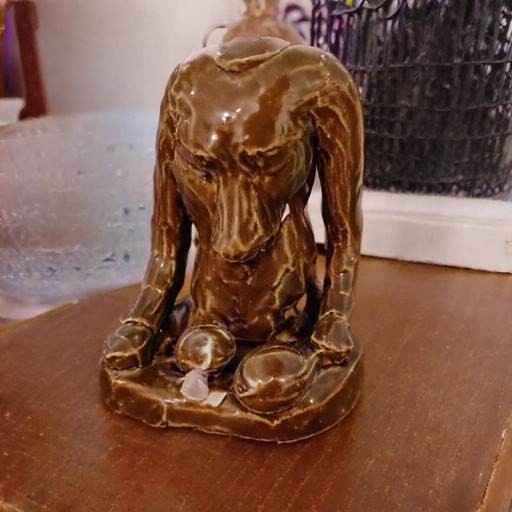} &
        \includegraphics[width=0.155\textwidth]{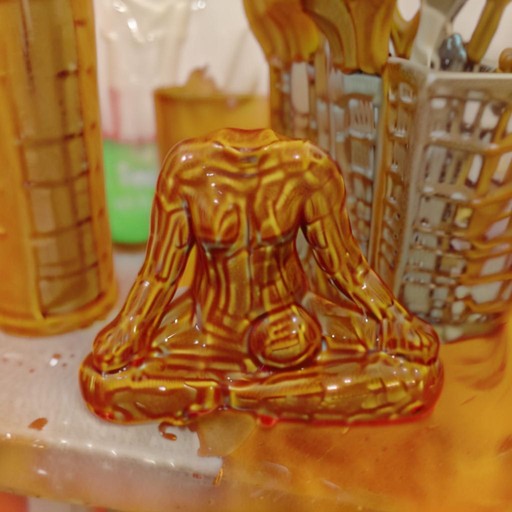} \\

        \includegraphics[width=0.155\textwidth]{temp_figs/content_images/fat_bird.jpg} &
        \includegraphics[width=0.155\textwidth]{temp_figs/content_images/dog2.jpg} &
        \includegraphics[width=0.155\textwidth]{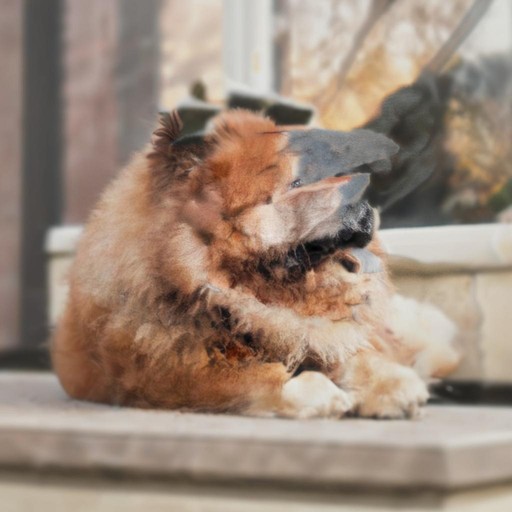} &
        \includegraphics[width=0.155\textwidth]{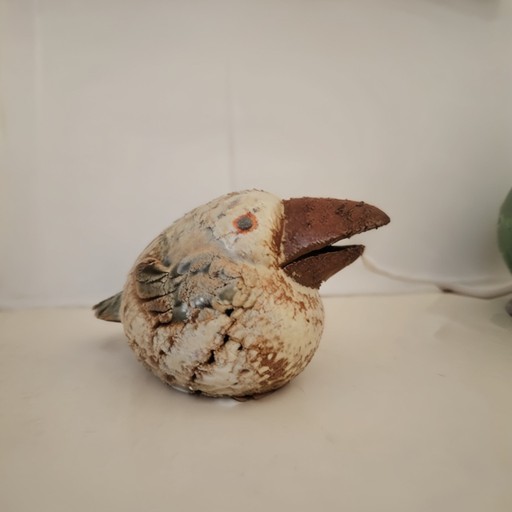} &
        \includegraphics[width=0.155\textwidth]{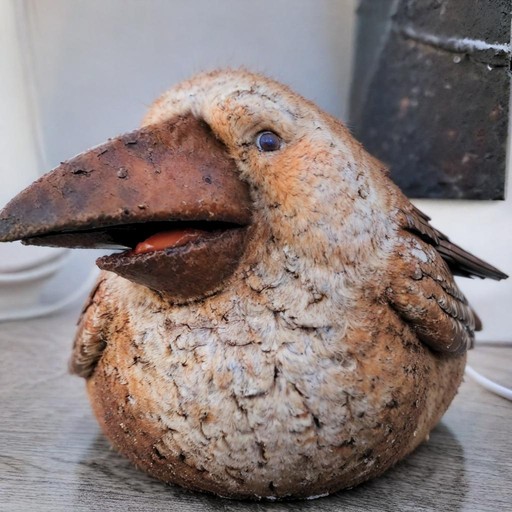} &
        \includegraphics[width=0.155\textwidth]{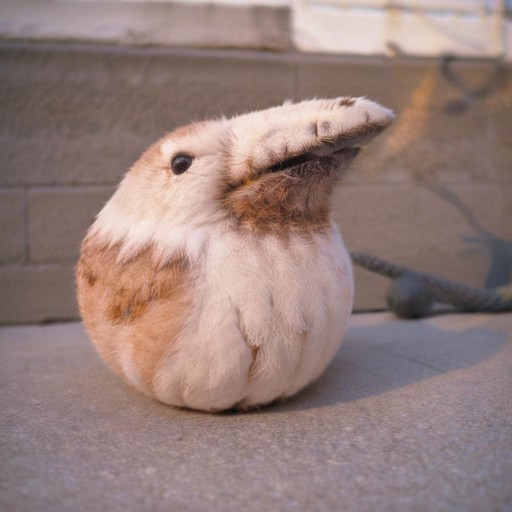} \\
        
        \includegraphics[width=0.155\textwidth]{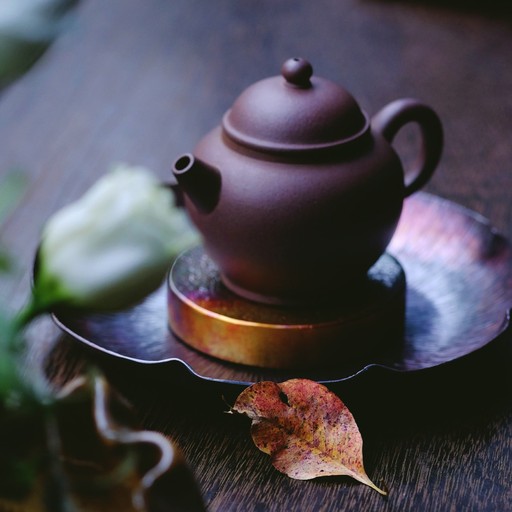} &
        \includegraphics[width=0.155\textwidth]{temp_figs/content_images/sloth.jpg} &
        \includegraphics[width=0.155\textwidth]{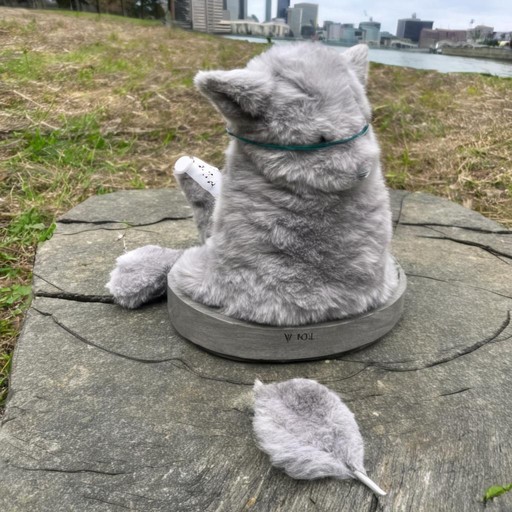} &
        \includegraphics[width=0.155\textwidth]{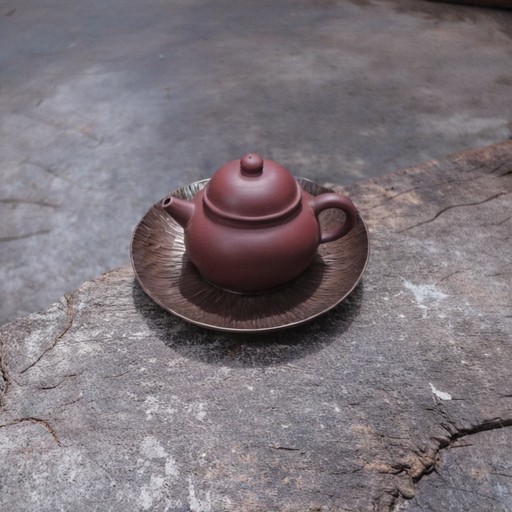} &
        \includegraphics[width=0.155\textwidth]{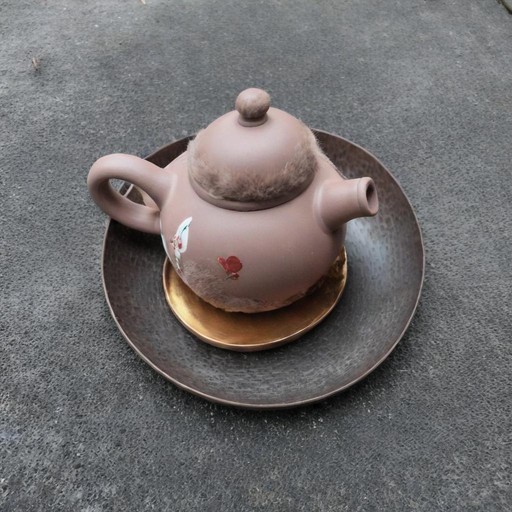} &
        \includegraphics[width=0.155\textwidth]{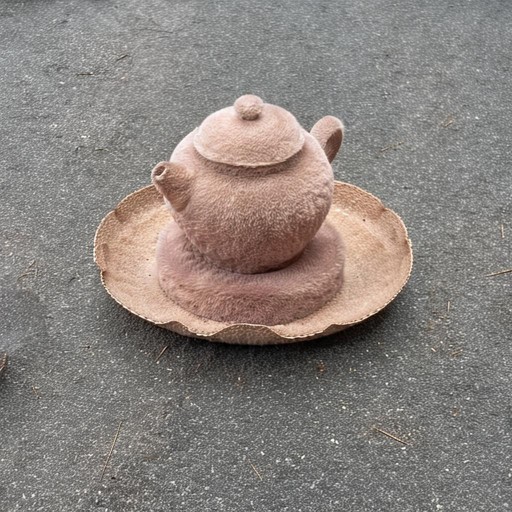} \\

    \end{tabular}
    }   
    \vspace{-0.11cm}
    \caption{Additional comparisons using challenging subject images as style reference. As can be seen, other methods encounter difficulties in disentangling the style and content from these images, consequently struggling to effectively transfer the style from one object to another.}
    \label{fig:object_to_object_compare}
\end{figure*}

\paragraph{Qualitative Comparisons}
In Section 5 of the main paper, we conducted a comparison of our B-LoRA method against four state-of-the-art baselines for image stylization incorporating personalization \cite{cnetSDXL,ziplora23,sohn2023styledrop,styleHertz23}. In this section, we delve deeper into the implementation details and present additional qualitative results.
To begin, we employed DreamBooth-LoRA \cite{DBlora} fine-tuning to obtain both style and content LoRAs, utilizing the same parameter configuration as ZipLoRA \cite{ziplora23}. For content images, we conducted fine-tuning across a set of images of the same object, except for the experiment involving a single image. However, for style LoRAs, we conducted fine-tuning using a single style image. We utilized the prompts provided in DreamBooth \cite{dreambooth23} and StyleDrop \cite{sohn2023styledrop}, specifically \ap{A [v] {\small\textless}object{\small\textgreater}} or \ap{A {\small\textless}object{\small\textgreater} in [s] style} for content and style, respectively. 
Subsequently, for ControlNet combined with DreamBooth-LoRA, we leveraged the publicly available implementation of ControlNet on SDXL from huggingface \cite{cnetSDXL}. this approach involved utilizing the style LoRAs we trained for style transfer while employing CannyEdge with thresholds of 100 and 200 for content guidance in ControlNet.
For StyleDrop \cite{sohn2023styledrop}, we followed the methodology outlined in StyleAligned \cite{styleHertz23} for fine-tuning the model over the style images, followed by fusing the content LoRAs with the SDXL weights. Similarly, for StyleAligned \cite{styleHertz23}, we utilized the authors' implementation for subject-driven generation alongside our content LoRAs.
Lastly, for ZipLoRA \cite{ziplora23}, we use the unofficial implementation \cite{unofficialZipLoRA} with default parameters.
We provide additional comparisons of our B-LoRA method against the aforementioned approaches using the same evaluation set presented in Section 5 of the main paper. These additional comparisons are illustrated in \Cref{fig:additional_comparison}.
Furthermore, we provide comparisons with challenging content inputs, such as stylized images, presented in \Cref{fig:style_to_style_compare}. We also showcase comparisons with challenging style inputs, such as object images, in \Cref{fig:object_to_object_compare}. These examples demonstrate the robustness of our method in handling diverse and complex content and style references.


\paragraph{Comparisons to Baselines Beyond SDXL-Based Approaches}
We provide additional comparisons of our method with three other image stylization techniques that do not rely on SDXL: StyTr2 \cite{StyTr22021}, AdaAttn \cite{AdaAttN2021}, and SWAG \cite{SWAG2021}. We evaluated the results using the same quantitative metrics described in the main paper. \Cref{fig:qual_no_sdxl_comparison} presents a qualitative comparison of the same set shown in the main paper, and \Cref{tab:dino_scores_sup} contains the quantitative results.

\begin{table}[!ht]
\caption{Quantitative comparison: We measure the average cosine similarity between the DINO features of the output image and the reference style and content. In this experiment, we use a single input image for evaluation.}
\vspace{-0.2cm}
\setlength{\tabcolsep}{3pt}
\renewcommand{\arraystretch}{1}
    \centering
    \begin{tabular}{c|cccc}
    \toprule
        ~ & StyTr2 & AdaAttn & SWAG & Ours \\
        \midrule
        \begin{tabular}[c]{@{}c@{}} Style \\  Transfer \end{tabular}  &  $0.83$ & $0.818$ & $0.883$ & $0.881$ \\
        \midrule
        Content &  $0.854$ & $0.828$& $0.788$ & $0.790$ \\
        \bottomrule
    \end{tabular}
    \label{tab:dino_scores_sup}
\end{table}

\begin{figure*}[ht]
    \centering
    \setlength{\tabcolsep}{1.5pt}
    {\small
    \begin{tabular}{c c @{\hspace{0.17cm}} | @{\hspace{0.17cm}}c c c @{\hspace{0.17cm}} | @{\hspace{0.17cm}}c}
        
     Content & Style & StyTr2 & AdaAttn 
         & SWAG & Ours \\
        \includegraphics[width=0.15\textwidth]{temp_figs/content_images/cat.jpeg} &
        \includegraphics[width=0.15\textwidth]{temp_figs/style_images/kiss.png} &
        
        \includegraphics[width=0.15\textwidth]{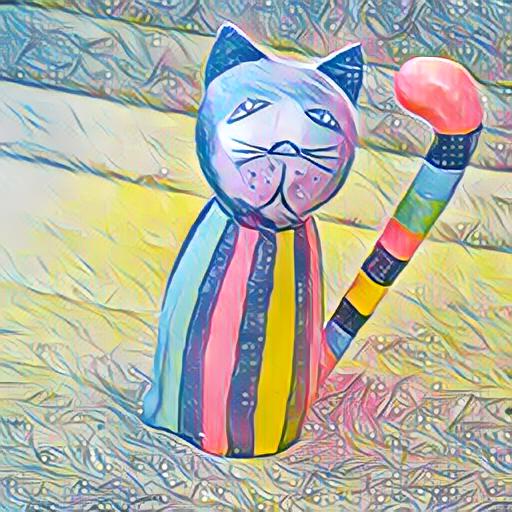} &
        \includegraphics[width=0.15\textwidth]{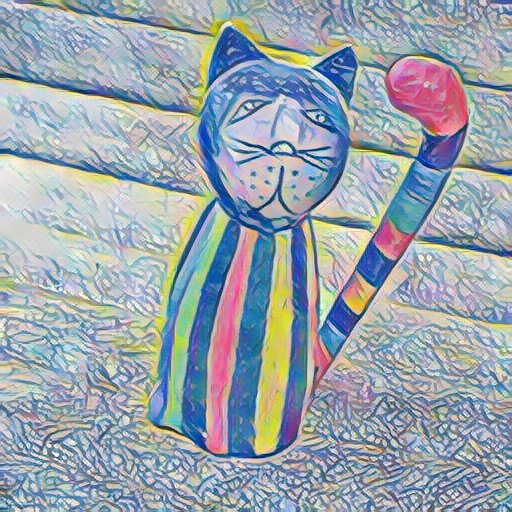} &
        \includegraphics[width=0.15\textwidth]{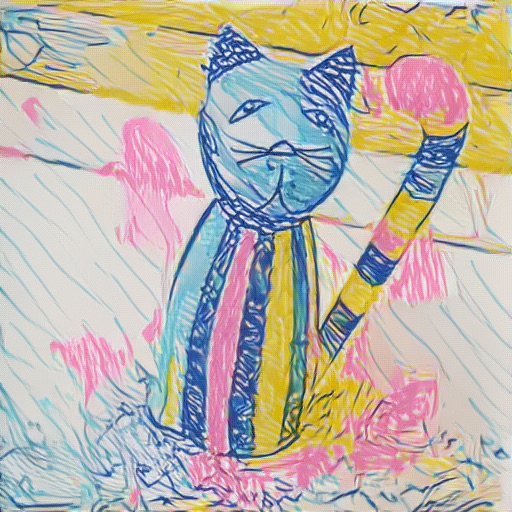} &
        \includegraphics[width=0.15\textwidth]{temp_figs/Fig_7/ours_cat.jpg} \\

        \includegraphics[width=0.15\textwidth]{temp_figs/content_images/sloth.jpg} &
        \includegraphics[width=0.15\textwidth]{temp_figs/style_images/drawing3.png} &
        \includegraphics[width=0.15\textwidth]{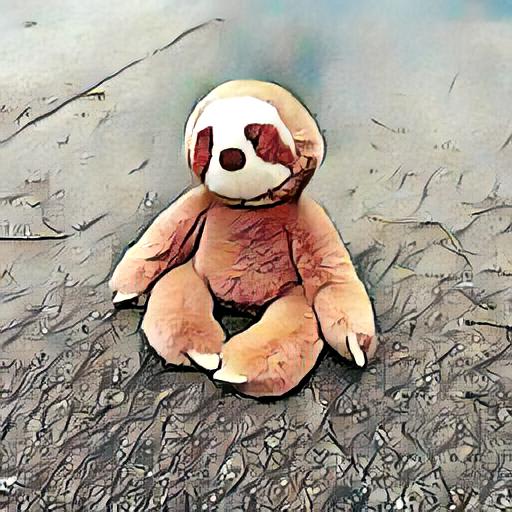} &
        \includegraphics[width=0.15\textwidth]{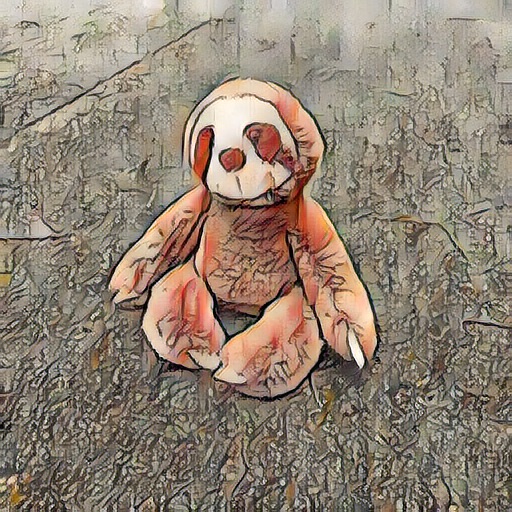} &
        \includegraphics[width=0.15\textwidth]{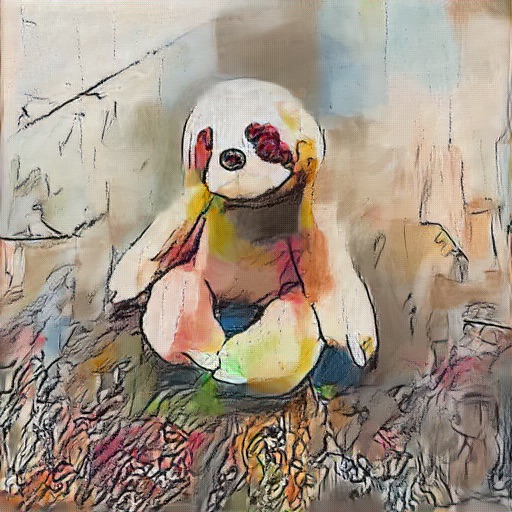} &
        \includegraphics[width=0.15\textwidth]{temp_figs/Fig_7/ours_sloth.jpeg} \\

        \includegraphics[width=0.15\textwidth]{temp_figs/content_images/teddybear.jpg} &
        \includegraphics[width=0.15\textwidth]{temp_figs/style_images/pen_sketch.jpeg} &
        \includegraphics[width=0.15\textwidth]{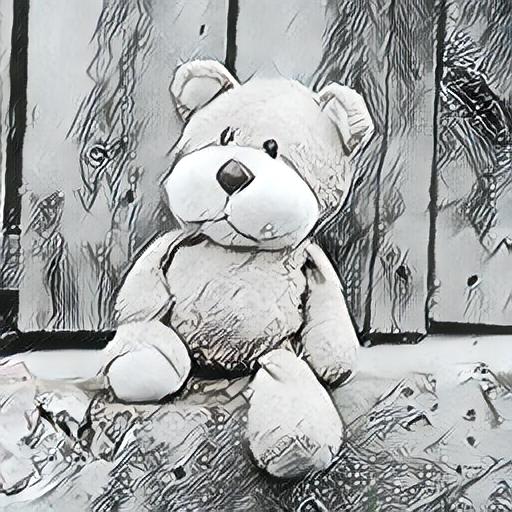} &
        \includegraphics[width=0.15\textwidth]{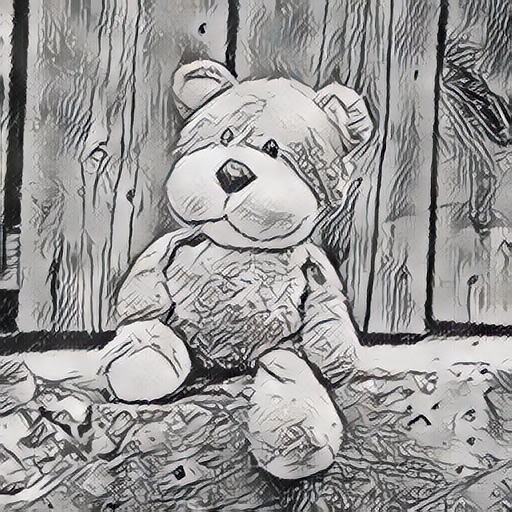} &
        \includegraphics[width=0.15\textwidth]{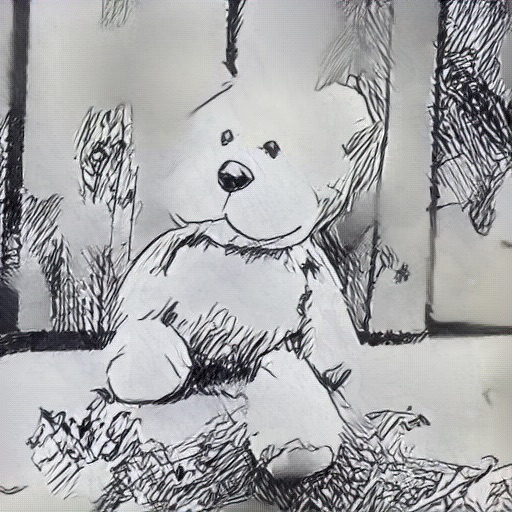} &
        \includegraphics[width=0.15\textwidth]{temp_figs/Fig_7/ours_teddybear.jpeg} \\
        
        \includegraphics[width=0.15\textwidth]{temp_figs/content_images/dog2.jpg} &
        \includegraphics[width=0.15\textwidth]{temp_figs/style_images/ink_sketch.jpeg} &
        \includegraphics[width=0.15\textwidth]{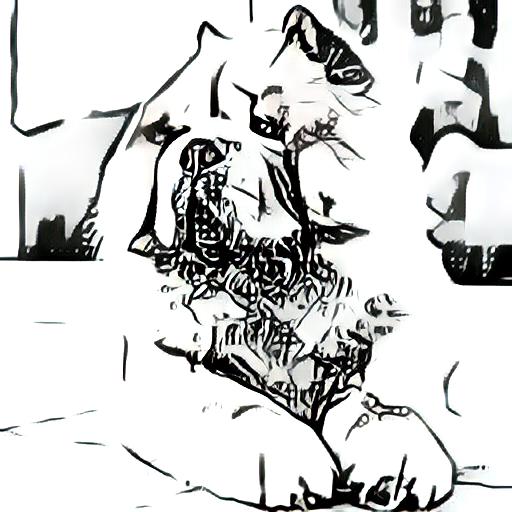} &
        \includegraphics[width=0.15\textwidth]{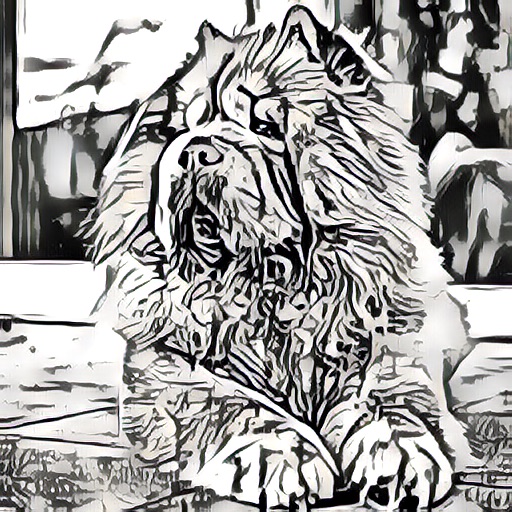} &
        \includegraphics[width=0.15\textwidth]{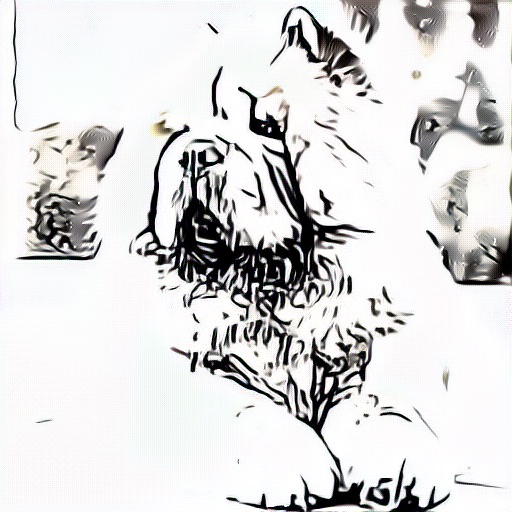} &
        \includegraphics[width=0.15\textwidth]{temp_figs/Fig_7/ours_dog2.jpeg} \\
    \end{tabular}
    }   
    \vspace{-0.11cm}
    \caption{Comparison with alternative approaches that do not rely on SDXL. The input style and content references are shown on the left.}
    \vspace{-0.4cm}
    \label{fig:qual_no_sdxl_comparison}

\end{figure*}

\begin{figure*}[ht]
    \centering
    \begin{tabular}{c @{\hspace{0.2cm}} c @{\hspace{0.2cm}} c c c c }
        Input Style & Method & \ap{A dog} & \ap{A moose} & \ap{A giraffe} & \ap{A girl} \\ 
        \multirow{2}{*}[0.06\textwidth]{\includegraphics[width=0.18\textwidth]{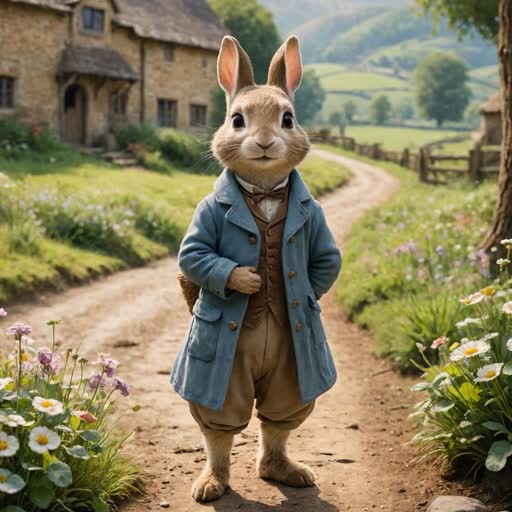}}
        & \raisebox{0.075\textwidth}{Ours} \hspace{0.05cm} & \includegraphics[width=0.15\textwidth]{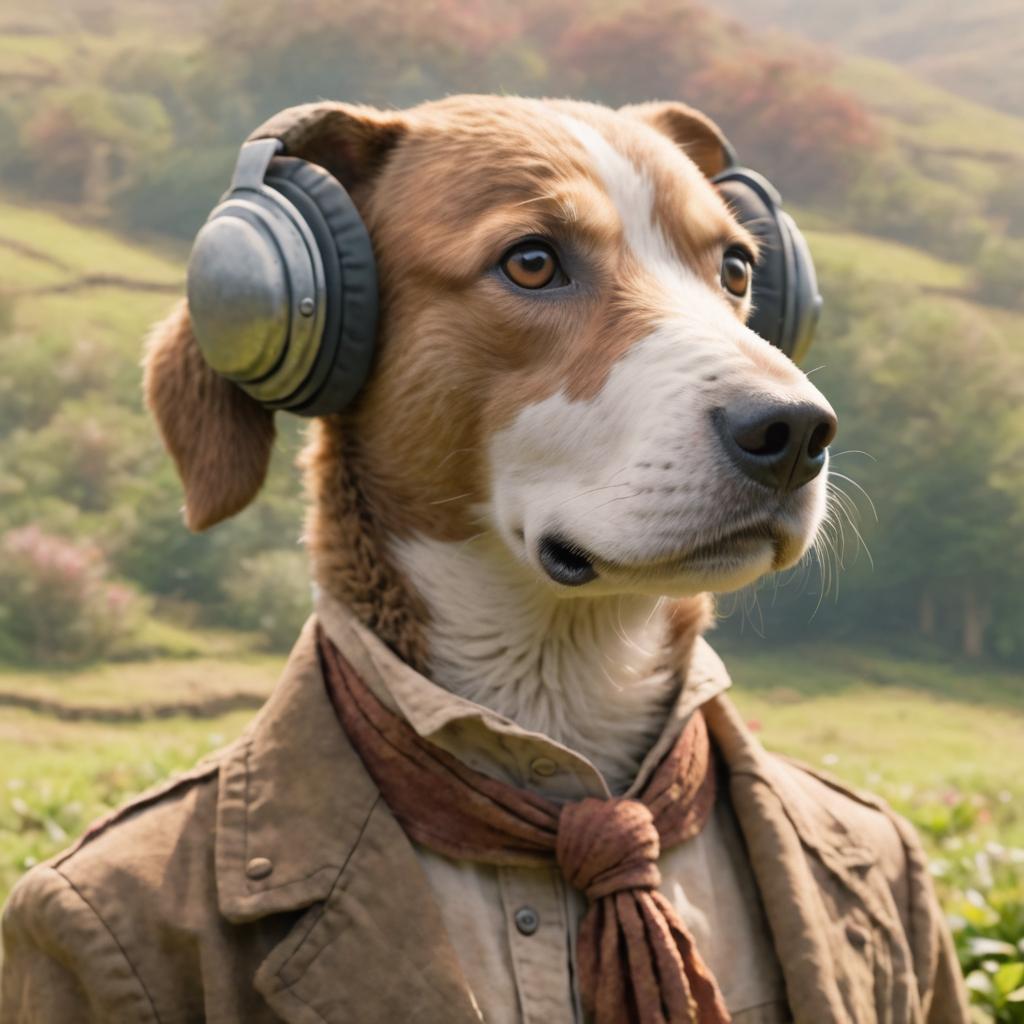} 
        & \includegraphics[width=0.15\textwidth]{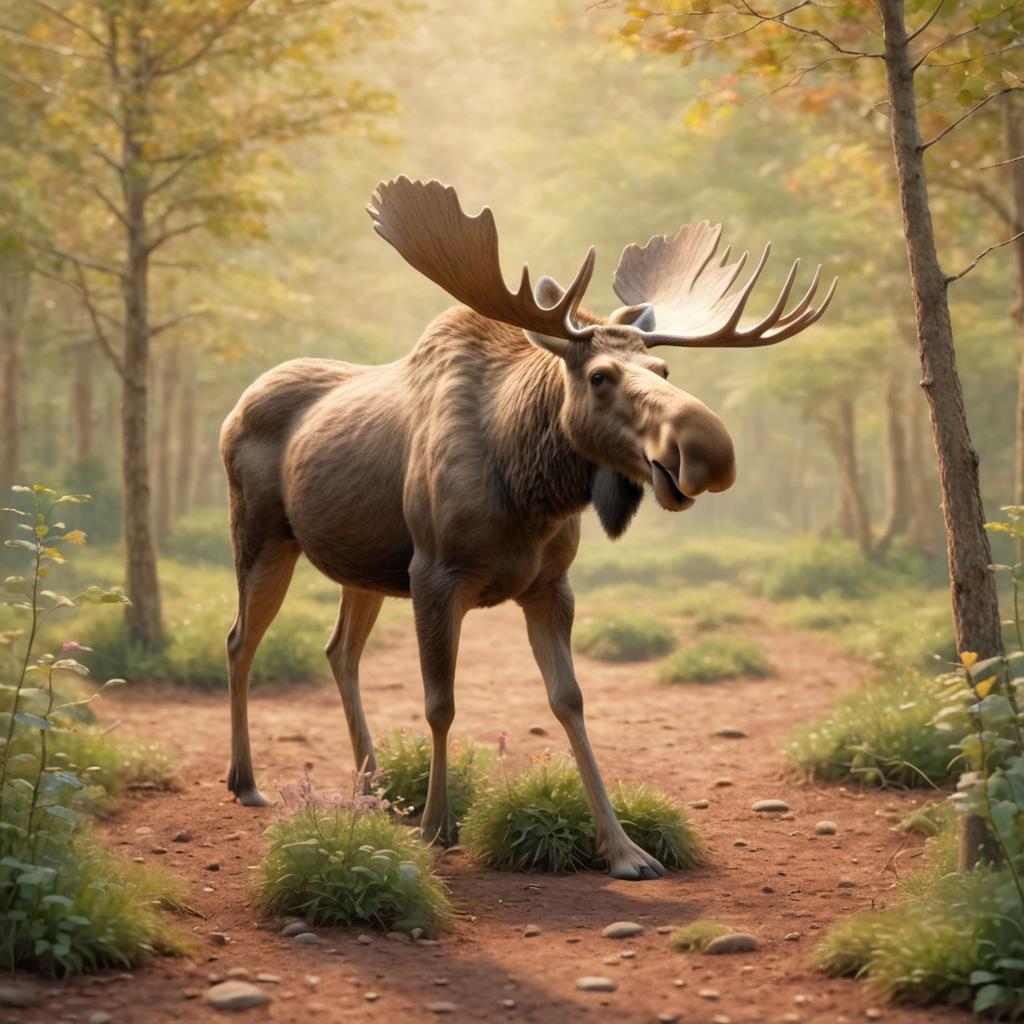} 
        & \includegraphics[width=0.15\textwidth]{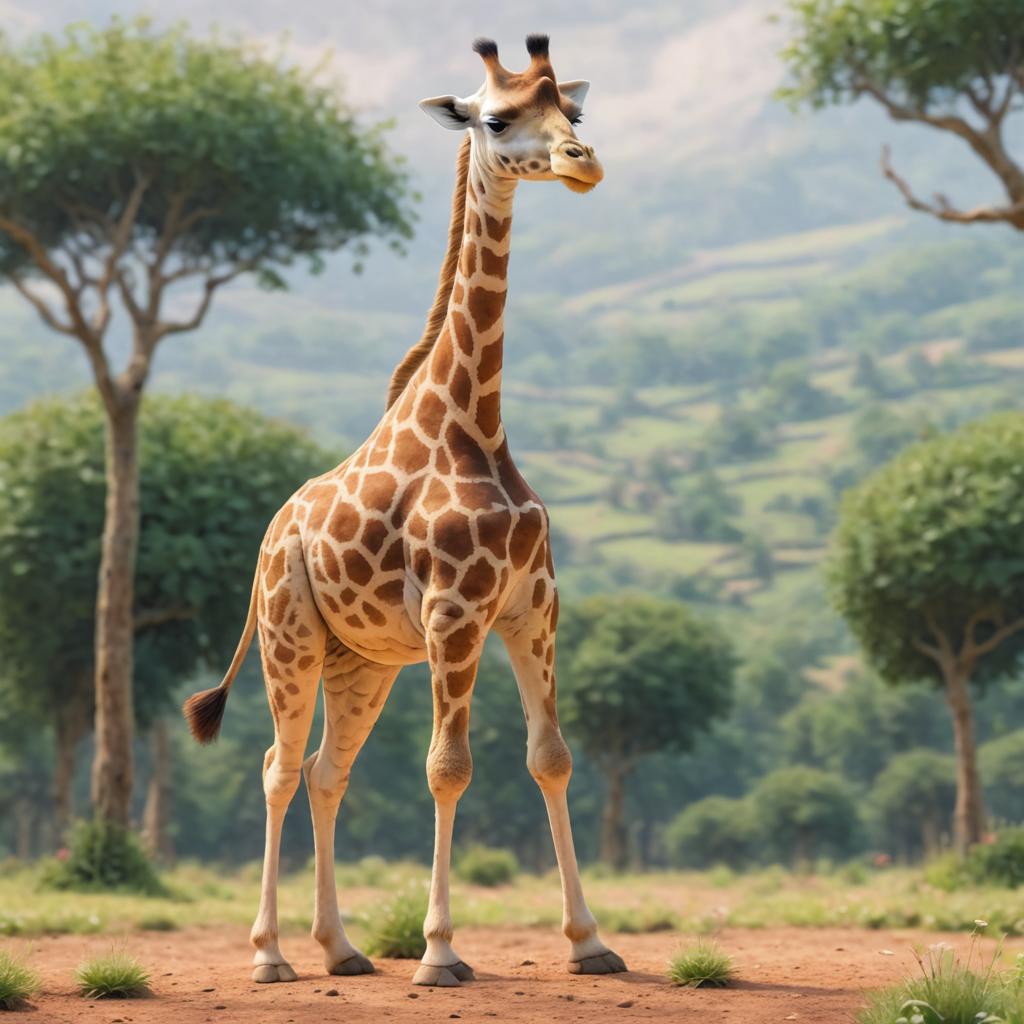} 
        & \includegraphics[width=0.15\textwidth]{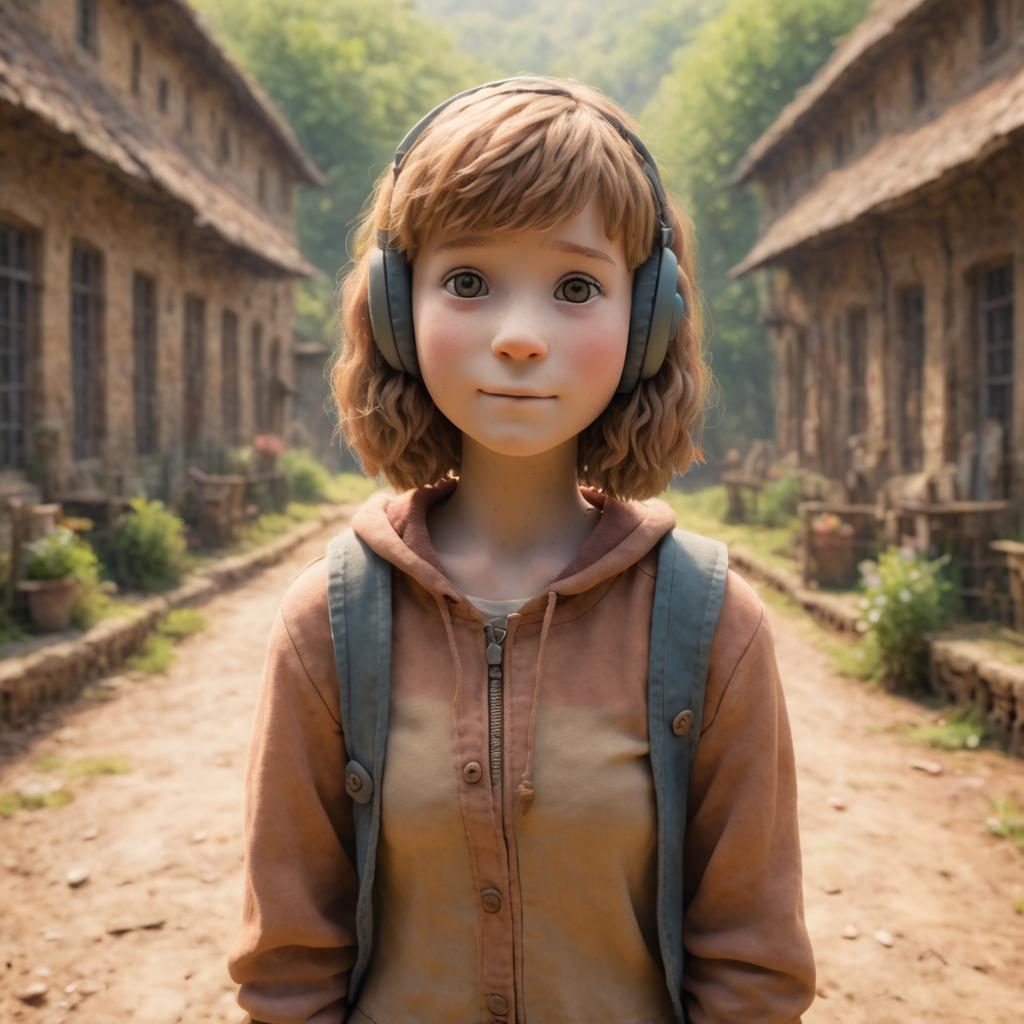} \\
        & \raisebox{0.075\textwidth}{InstantStyle} & \includegraphics[width=0.15\textwidth]{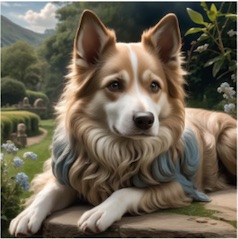} 
        & \includegraphics[width=0.15\textwidth]{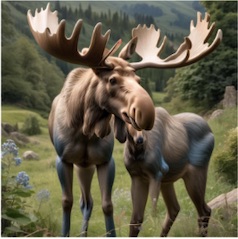} 
        & \includegraphics[width=0.15\textwidth]{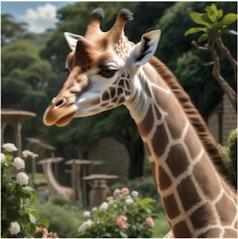} 
        & \includegraphics[width=0.15\textwidth]{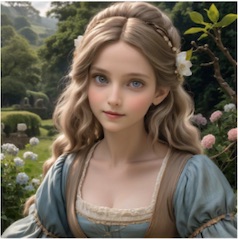} \\
        \multirow{2}{*}[0.06\textwidth]{\includegraphics[width=0.18\textwidth]{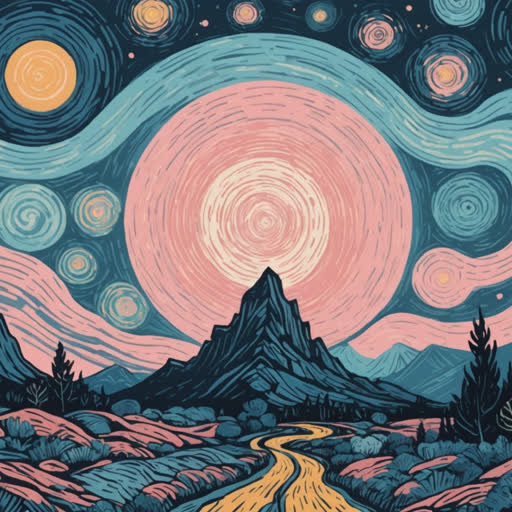}}
        & \raisebox{0.075\textwidth}{Ours} & \includegraphics[width=0.15\textwidth]{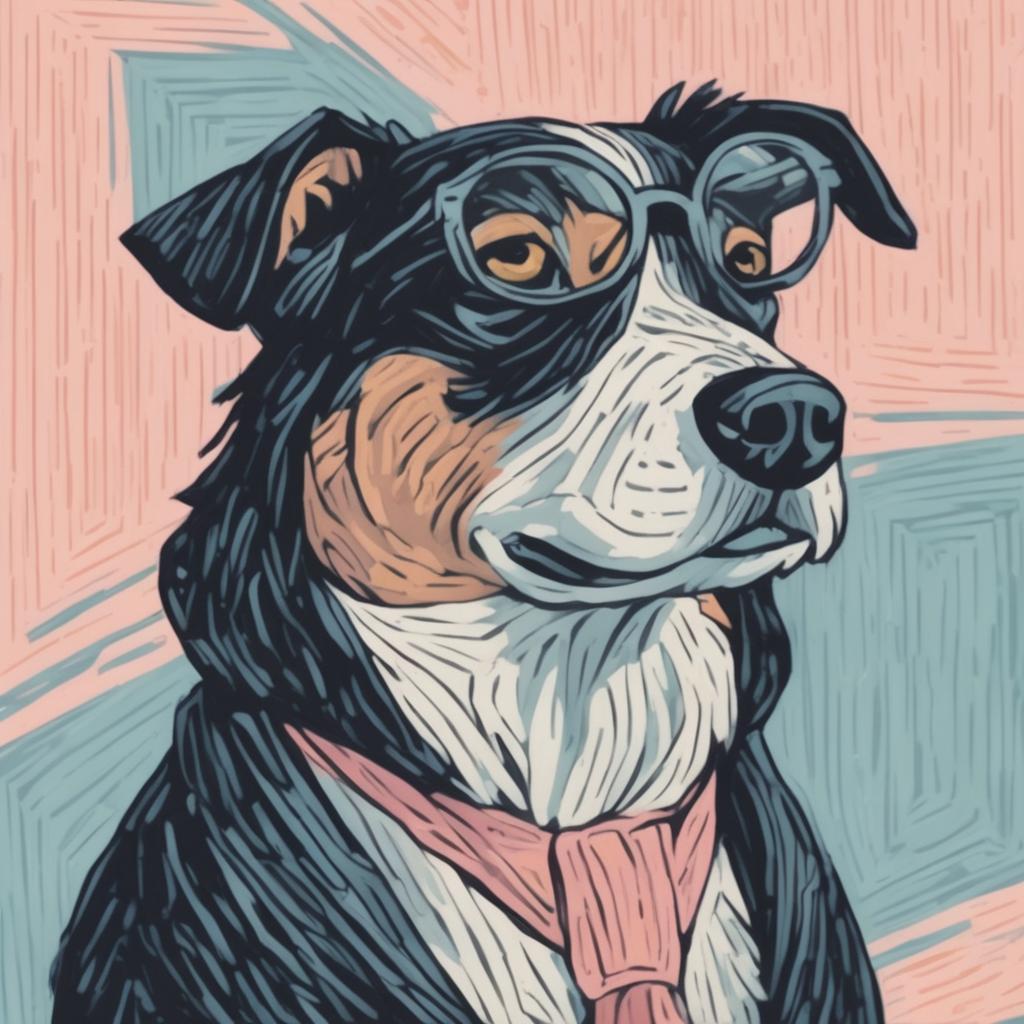} 
        & \includegraphics[width=0.15\textwidth]{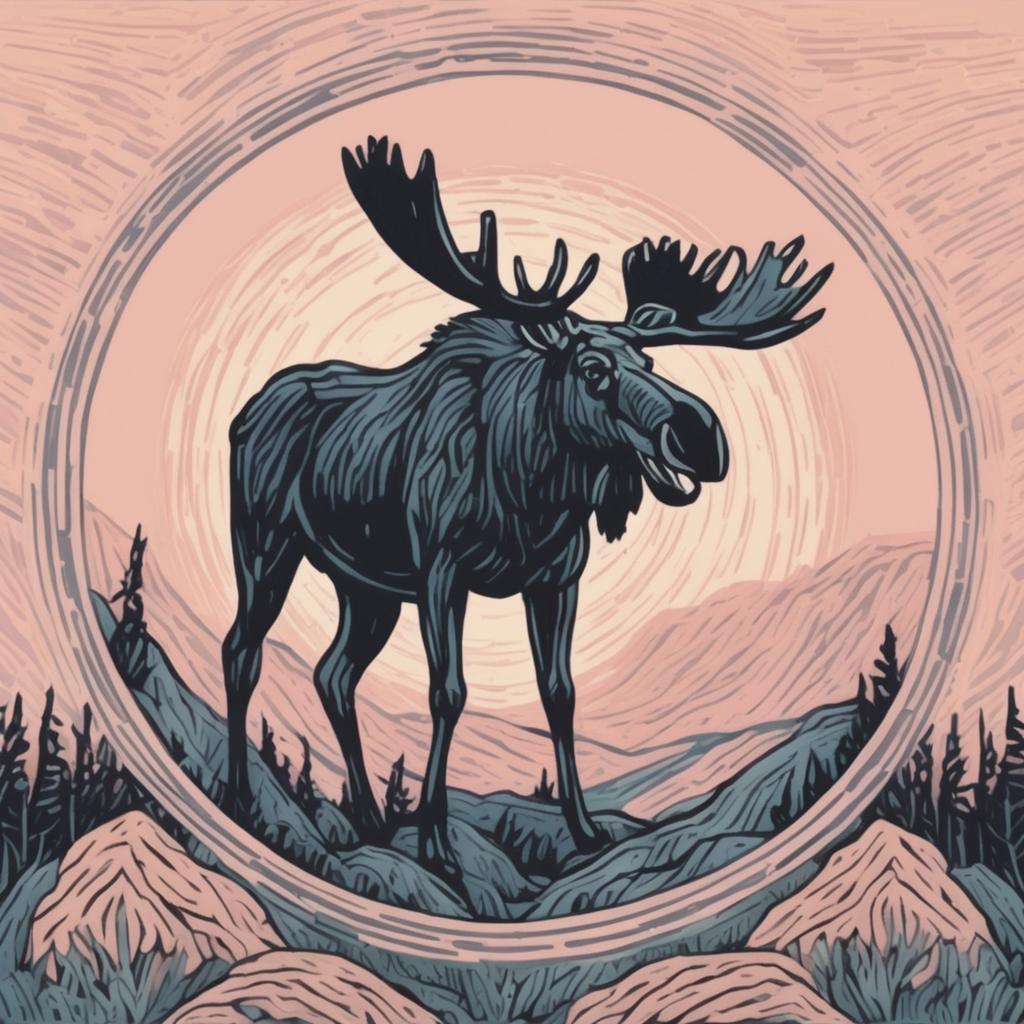} 
        & \includegraphics[width=0.15\textwidth]{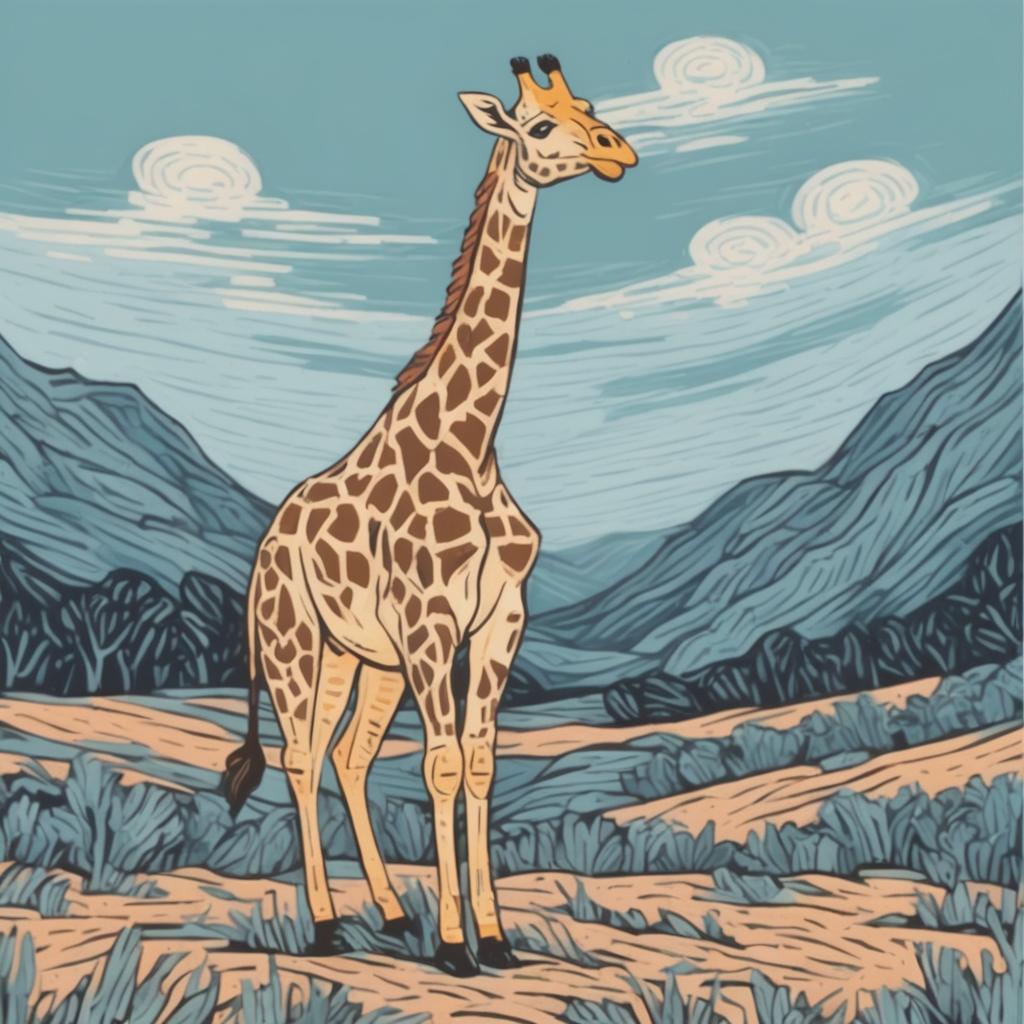} 
        & \includegraphics[width=0.15\textwidth]{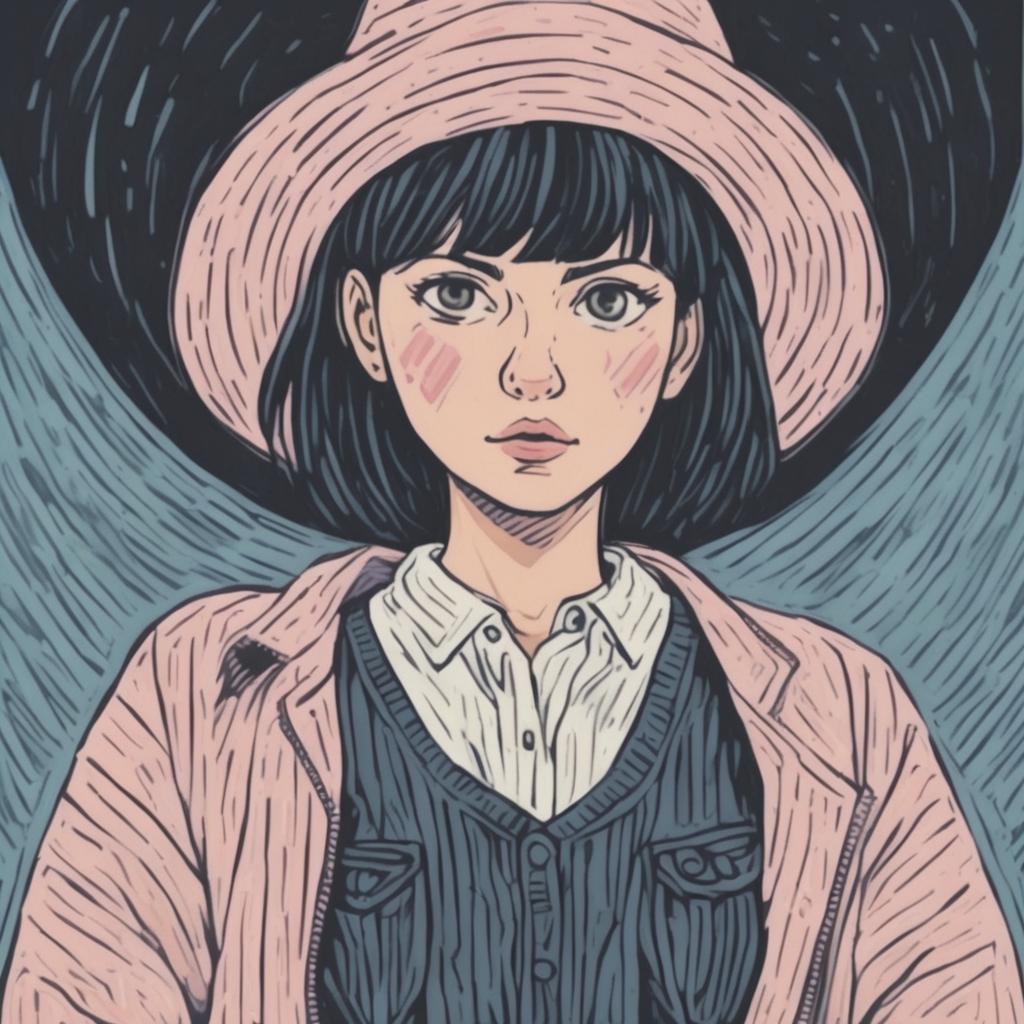} \\
        & \raisebox{0.075\textwidth}{InstantStyle} & \includegraphics[width=0.15\textwidth]{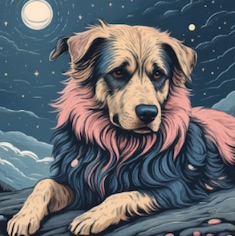} 
        & \includegraphics[width=0.15\textwidth]{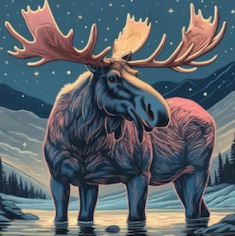} 
        & \includegraphics[width=0.15\textwidth]{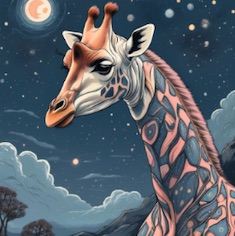} 
        & \includegraphics[width=0.15\textwidth]{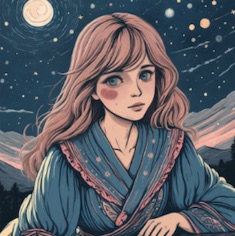} \\
        \multirow{2}{*}[0.06\textwidth]{\includegraphics[width=0.18\textwidth]{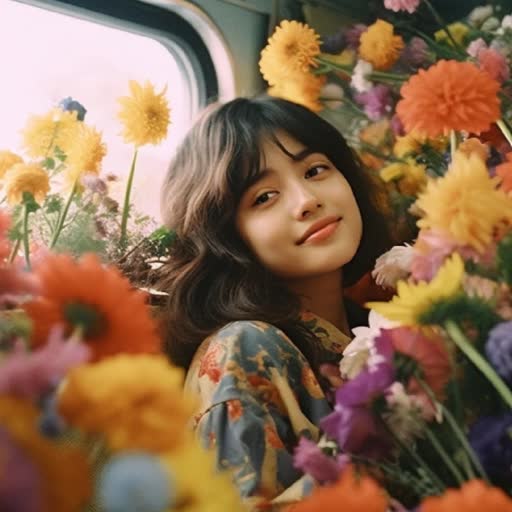}}
        & \raisebox{0.075\textwidth}{Ours} & \includegraphics[width=0.15\textwidth]{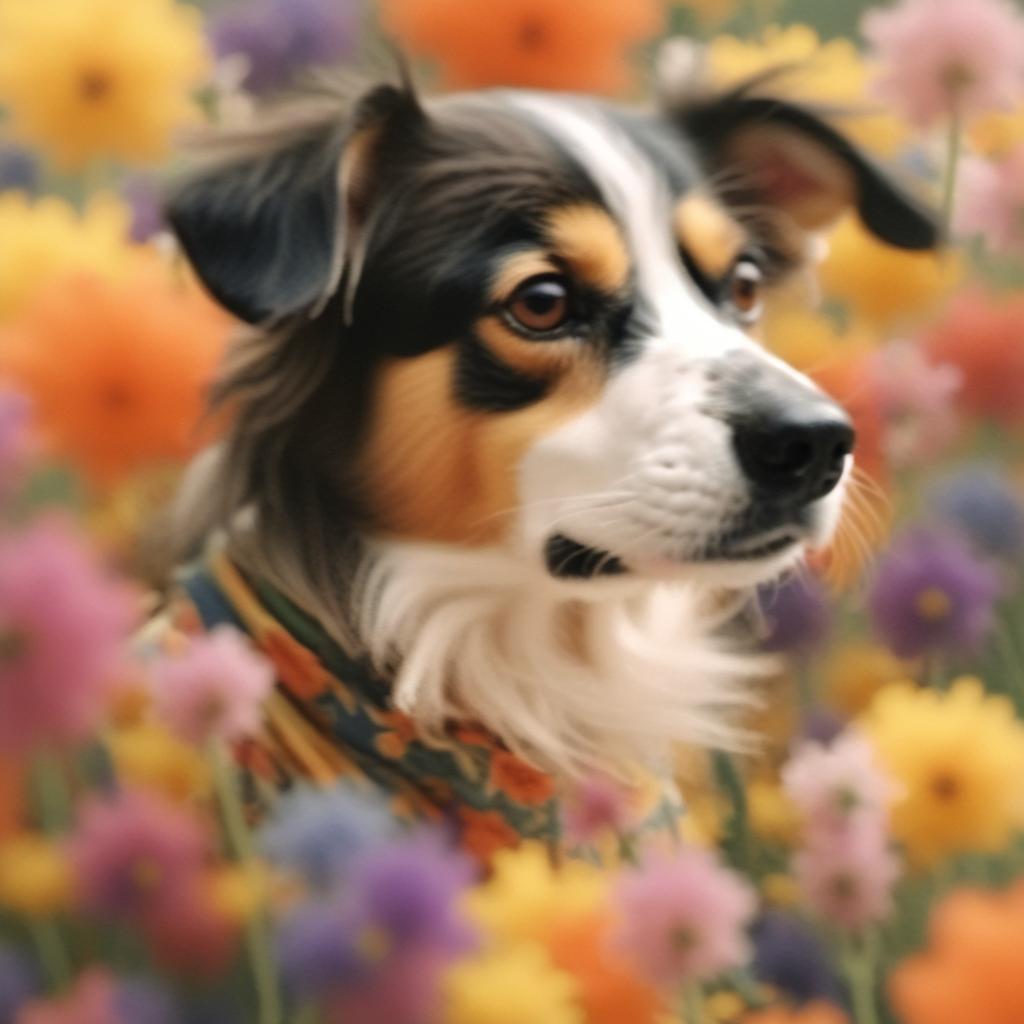} 
        & \includegraphics[width=0.15\textwidth]{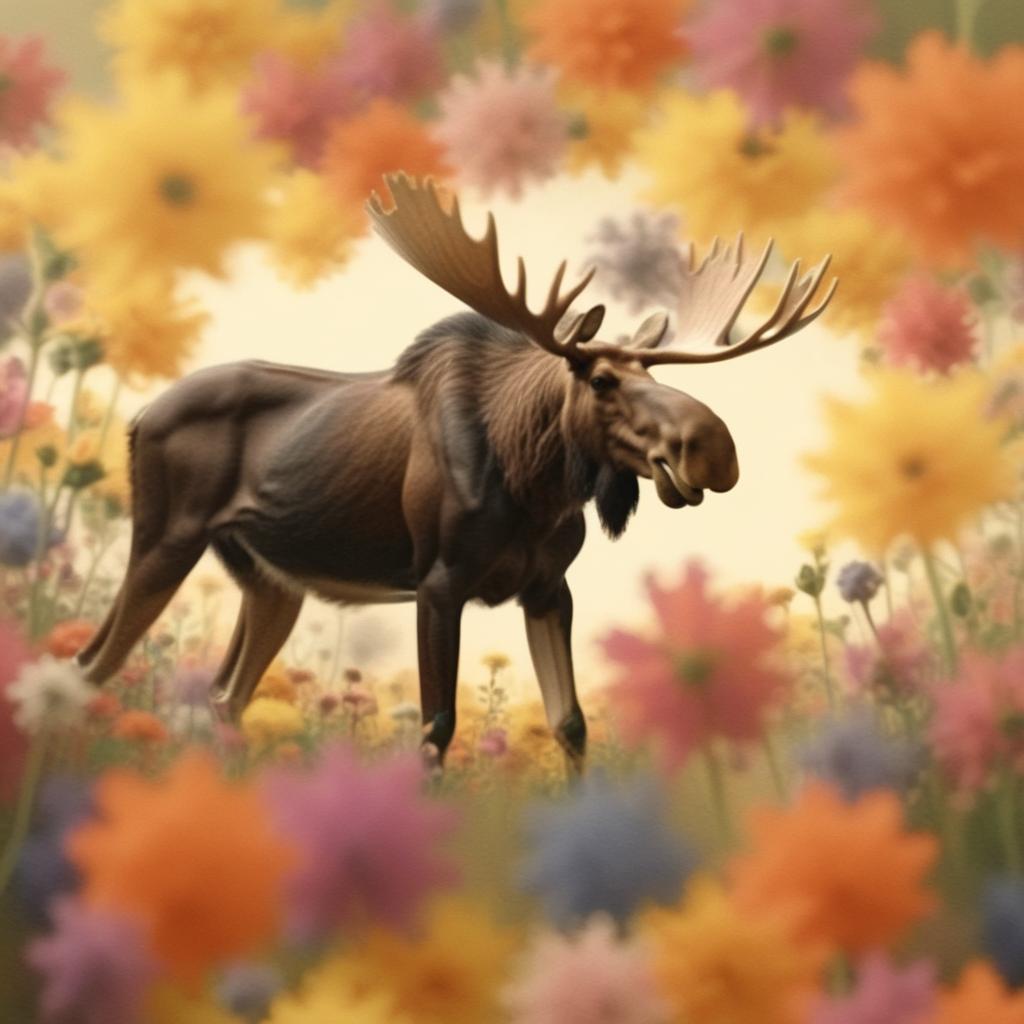} 
        & \includegraphics[width=0.15\textwidth]{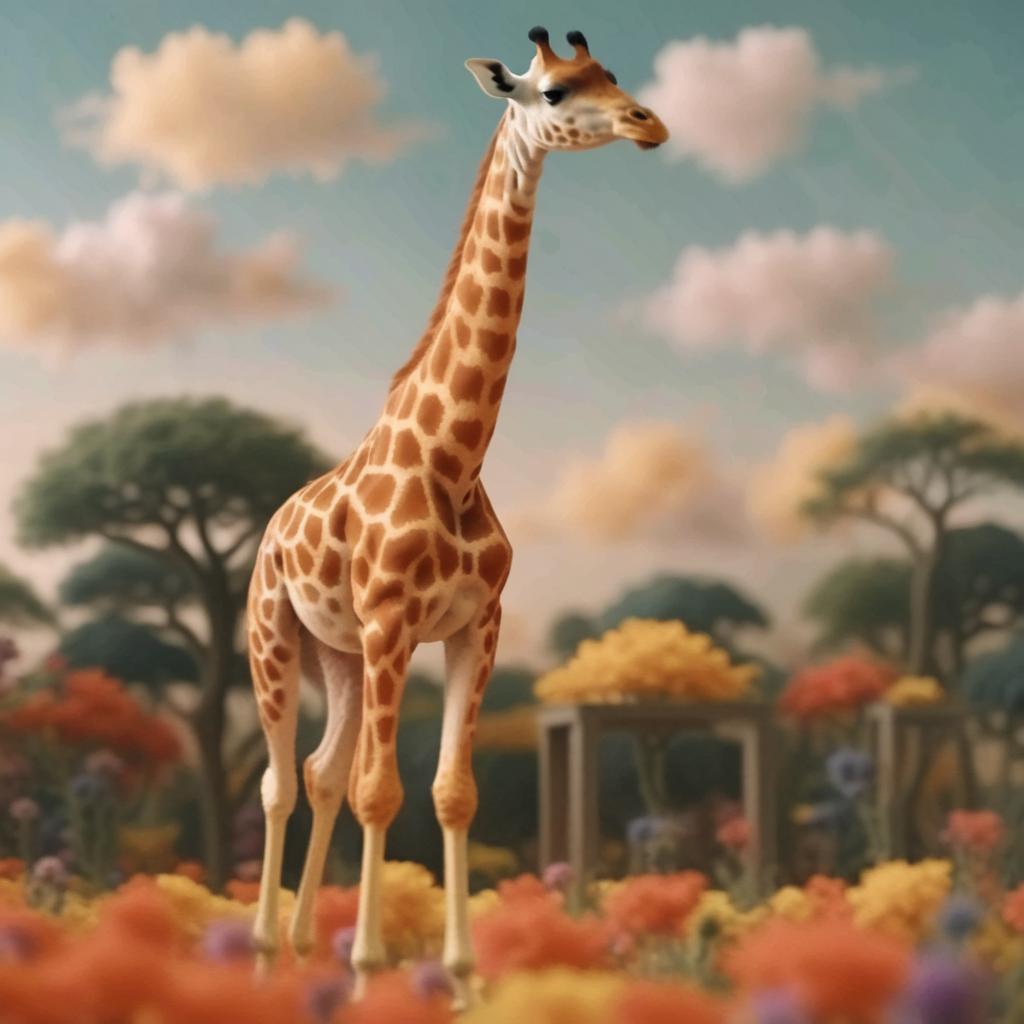} 
        & \includegraphics[width=0.15\textwidth]{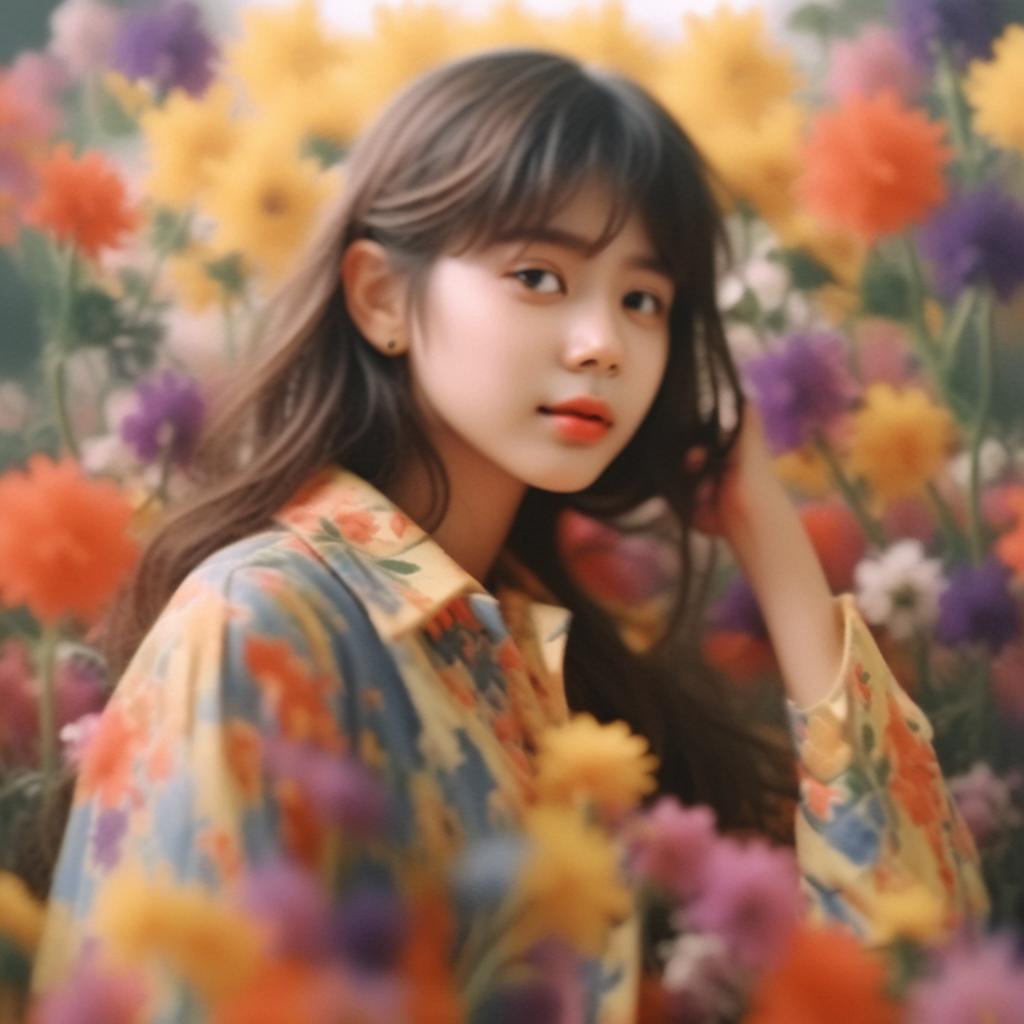} \\
        & \raisebox{0.075\textwidth}{InstantStyle} & \includegraphics[width=0.15\textwidth]{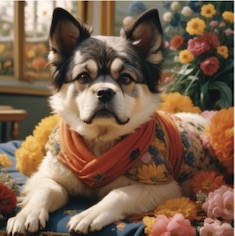} 
        & \includegraphics[width=0.15\textwidth]{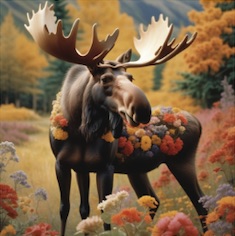} 
        & \includegraphics[width=0.15\textwidth]{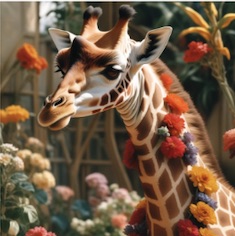} 
        & \includegraphics[width=0.15\textwidth]{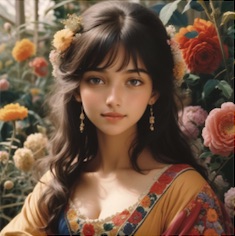} \\
    \end{tabular}
    \caption{Comparison of stylization results between our method and InstantStyle. The input style image is shown in the first column, followed by results generated by our method and InstantStyle for different prompts: \ap{A dog}, \ap{A moose}, \ap{A giraffe}, and \ap{A girl}. The images of InstantStyle are taken from the original paper. Both approaches achieve consistent style generation, demonstrating the effectiveness of style transfer.}
    \label{fig:comparison_style_instantstyle}
\end{figure*}

\begin{figure*}[ht]
    \centering
    {\small
    \begin{tabular}{c c @{\hspace{0.17cm}} | @{\hspace{0.17cm}}c c c @{\hspace{0.17cm}} | @{\hspace{0.17cm}}c}
        
     Content & Style & InstantStyle &  Ours \\
     
        \includegraphics[width=0.155\textwidth]{temp_figs/content_images/cat.jpeg} &
        \includegraphics[width=0.155\textwidth]{temp_figs/style_images/kiss.png} &
        
        \includegraphics[width=0.155\textwidth]{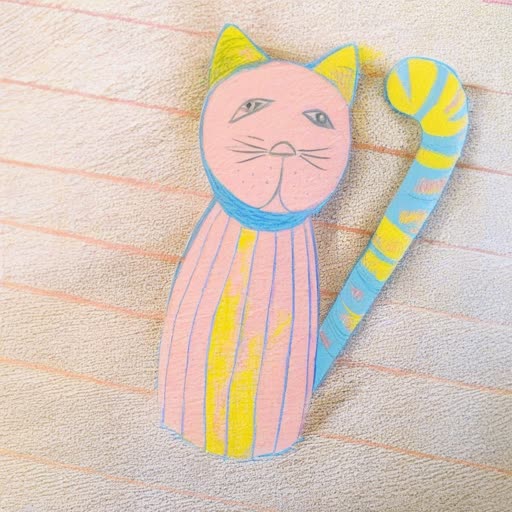} &
        \includegraphics[width=0.155\textwidth]{temp_figs/Fig_7/ours_cat.jpg} \\   

        \includegraphics[width=0.155\textwidth]{temp_figs/content_images/sloth.jpg} &
        \includegraphics[width=0.155\textwidth]{temp_figs/style_images/drawing3.png} &
        \includegraphics[width=0.155\textwidth]{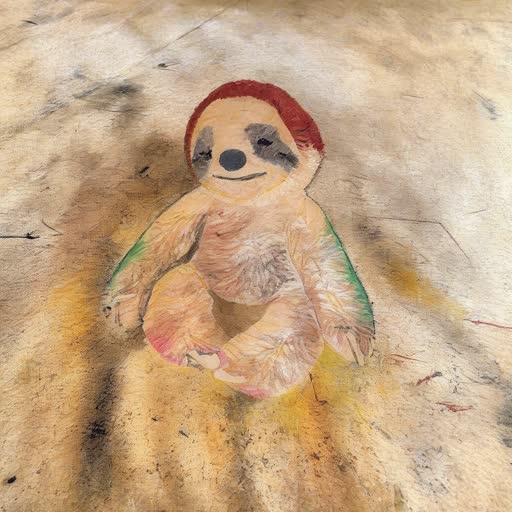} &
        \includegraphics[width=0.155\textwidth]{temp_figs/Fig_7/ours_sloth.jpeg} \\

        \includegraphics[width=0.155\textwidth]{temp_figs/content_images/teddybear.jpg} &
        \includegraphics[width=0.155\textwidth]{temp_figs/style_images/pen_sketch.jpeg} &
        \includegraphics[width=0.155\textwidth]{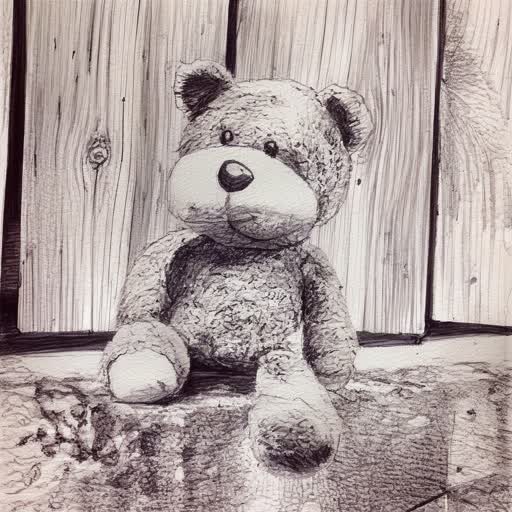} &
        \includegraphics[width=0.155\textwidth]{temp_figs/Fig_7/ours_teddybear.jpeg} \\
        
        \includegraphics[width=0.155\textwidth]{temp_figs/content_images/dog2.jpg} &
        \includegraphics[width=0.155\textwidth]{temp_figs/style_images/ink_sketch.jpeg} &
        \includegraphics[width=0.155\textwidth]{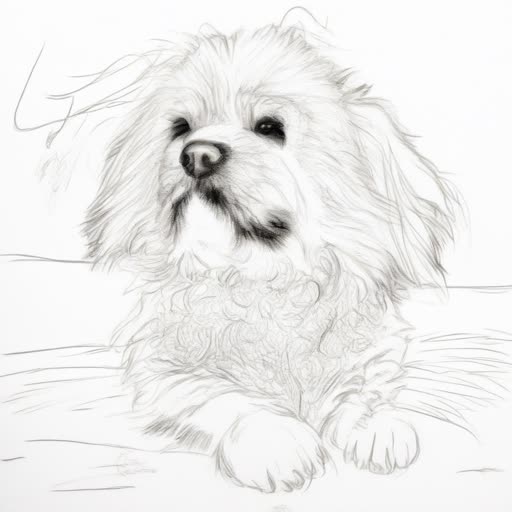} &
        \includegraphics[width=0.155\textwidth]{temp_figs/Fig_7/ours_dog2.jpeg} \\

        \includegraphics[width=0.155\textwidth]{temp_figs/content_images/bull.jpg} &
        \includegraphics[width=0.155\textwidth]{rebuttal_files/instantStyle_compare/references/landscape.jpg} &
        \includegraphics[width=0.155\textwidth]{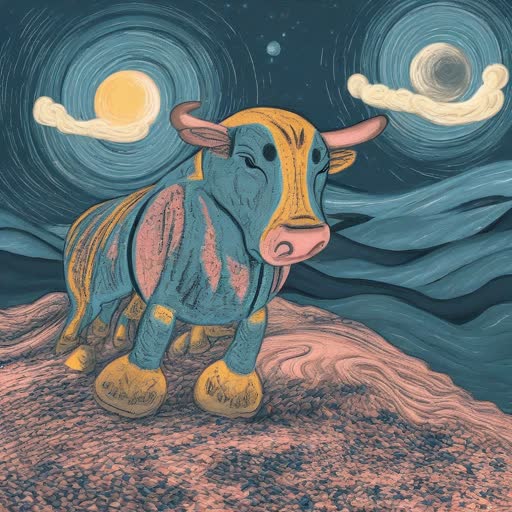} &
        \includegraphics[width=0.155\textwidth]{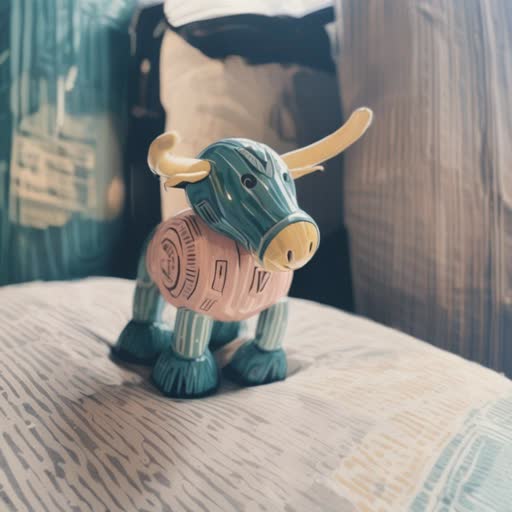} \\

        \includegraphics[width=0.155\textwidth]{temp_figs/content_images/fat_bird.jpg} &
        \includegraphics[width=0.155\textwidth]{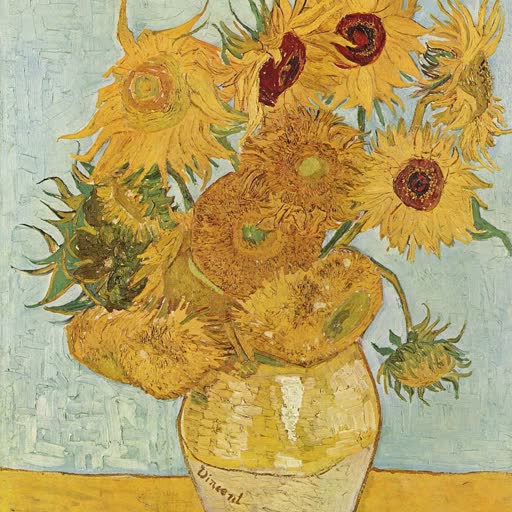} &
        \includegraphics[width=0.155\textwidth]{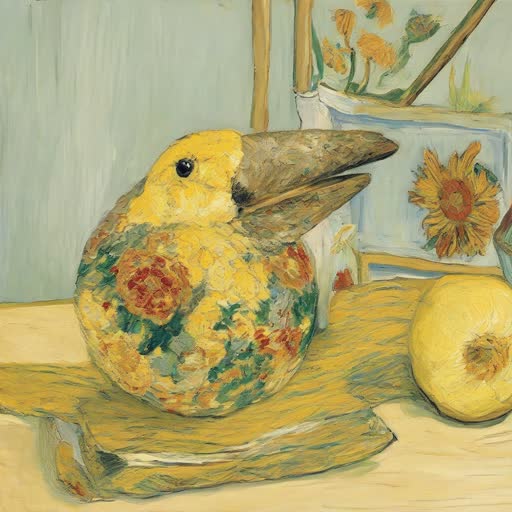} &
        \includegraphics[width=0.155\textwidth]{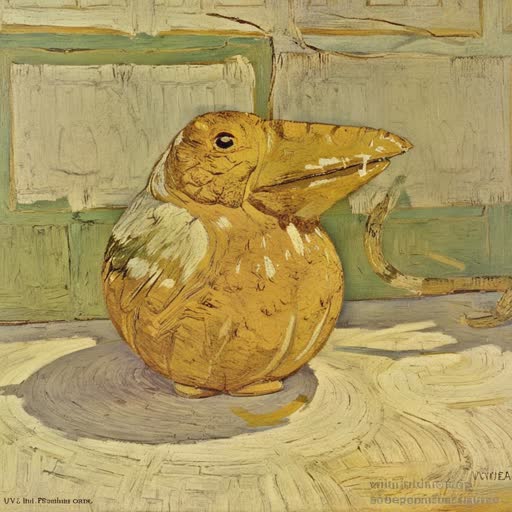} \\

        \includegraphics[width=0.155\textwidth]{temp_figs/content_images/dog2.jpg} &
        \includegraphics[width=0.155\textwidth]{rebuttal_files/instantStyle_compare/references/bunny.jpg} &
        \includegraphics[width=0.155\textwidth]{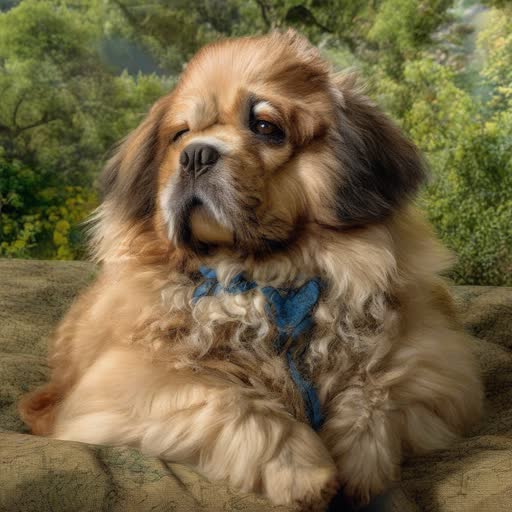} &
        \includegraphics[width=0.155\textwidth]{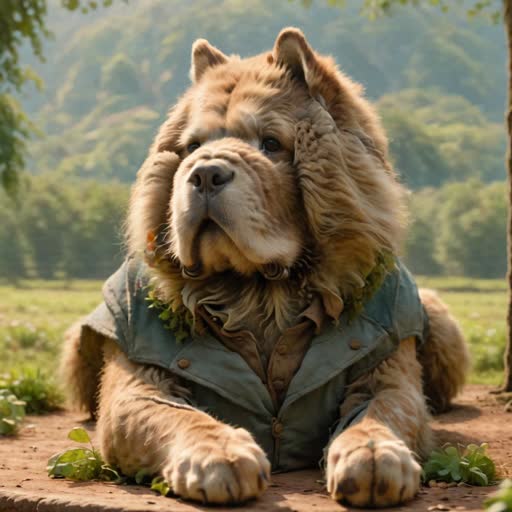} \\

    \end{tabular}
    }   
    \vspace{-0.11cm}
    \caption{Comparison of style and content mixing between our method and InstantStyle. The results illustrate cases where ControlNet, used by InstantStyle, may fail to adequately capture the content or may override the style. For example, in the fourth row, we can see that ControlNet failed to extract the shape of the dog, leading to unsatisfactory results, While our method demonstrates better content preservation. The images showcase the stylization applied to various content images, highlighting differences in how each approach handles content and style integration.}
    \vspace{-0.4cm}
    \label{fig:qual_instantstyle_mix_comparison}

\end{figure*}

\paragraph{Comparisons to InstantStyle}
InstantStyle \cite{InstantStyle2024} is a concurrent work to ours. Aimed at performing image stylization tasks based on a style image reference. InstantStyle achieves this by injecting the CLIP embedding of the style image into style-specific blocks within SDXL, similar to our method, where the fifth block is selected for the style condition. Notably, InstantStyle uses a trained IP-Adapter model and does not require fine-tuning, which is its main advantage over our method. Both approaches provide compelling results in consistent style generation, as presented in \Cref{fig:comparison_style_instantstyle}. For content conditioning, InstantStyle utilizes ControlNet, while our method separates content from style and extracts both. This allows for better content preservation in scenarios where ControlNet may not capture the content well enough or may override the style, as shown in \Cref{fig:qual_instantstyle_mix_comparison}. Additionally, InstantStyle requires the content component to be explicitly defined to subtract its CLIP embedding from the style embedding, whereas our approach learns the content and style implicitly. For a fair comparison, we trained our method using the style images from InstantStyle.

\section{Limitations}
\begin{figure*}[ht]
    \centering
    \setlength{\tabcolsep}{1.5pt}
    {\small
    \begin{tabular}{ c c c c c}
        
        Content &  ~ & ~ & ~ \\
        
        \includegraphics[width=0.193\textwidth]{temp_figs/content_images/dog6.jpg} &
        \includegraphics[width=0.193\textwidth]{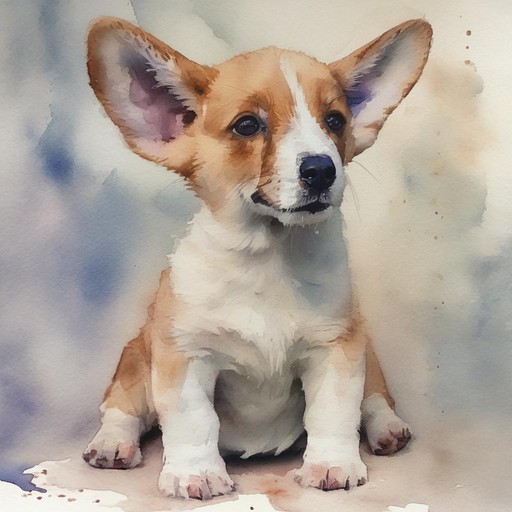} &
        \includegraphics[width=0.193\textwidth]{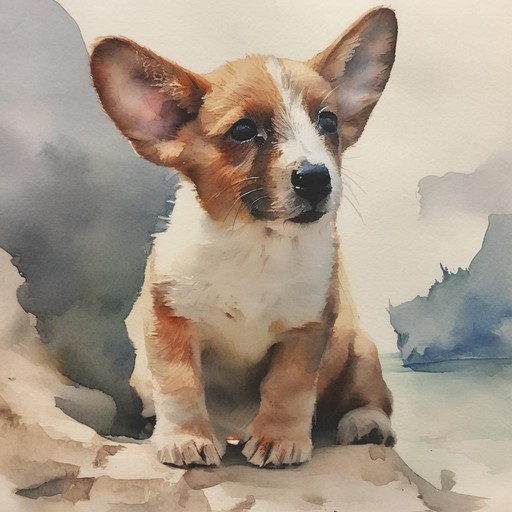} &
        \includegraphics[width=0.193\textwidth]{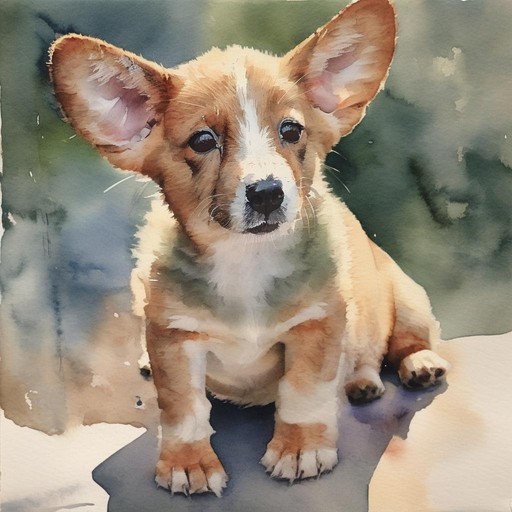} &
        \includegraphics[width=0.193\textwidth]{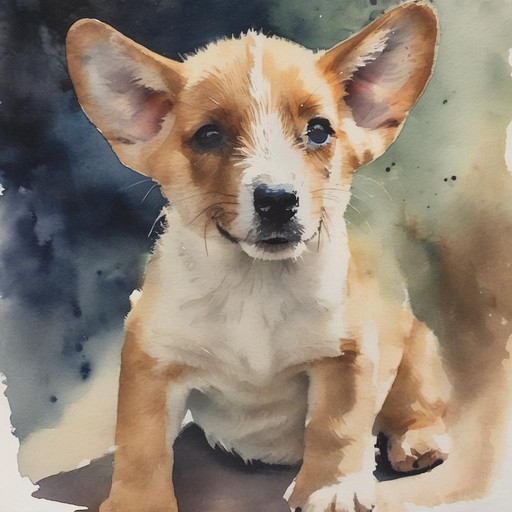} \\
    \end{tabular}
    
    }   
    \caption{To mitigate the limitation of sub-optimal identity preservation due to color separation, we propose combining adapters $\{\Delta W^4, \Delta W^5\}$, with $\Delta W^5$ assigned a coefficient $\alpha$ within the range of [0.4, 0.5]. This method preserves the original colors of the subject while allowing stylizations using text prompts. The generated contents depicted in the figure are based on the prompt \ap{Watercolor painting of [c]}.}
    \label{fig:limitations_color}
\end{figure*}

In Section 6 of the main paper, we discussed the limitations of our method. Here, we expand upon this section and propose potential approaches to mitigate these limitations.
The first limitation we aim to address is the sub-optimal identity preservation due to color separation. To overcome this issue, we propose applying a scaling factor of alpha between 0.4-0.5 to the style adapter $\Delta W^5$. This adjustment allows for preserving the original colors of the subject while minimizing interference with other style B-LoRA injections, as illustrated in \Cref{fig:limitations_color}.

To mitigate style leakage from background objects in the style reference image, we suggest preprocessing the training data by center cropping the desired style reference image. This approach increases precision by focusing on the central object during the B-LoRA training process.

Addressing the final limitation of adequately capturing content in complex scenes, we conducted an ablation study to explore the effect of injecting different prompts into different blocks of the network. Specifically, we conducted five experiments:

\textbf{(1)} Injecting our method's prompt \ap{A [c] in [s] style}, into all transformer blocks of the UNet.
\textbf{(2)} Injecting \ap{A [c]} into the content block $W^4$ while injecting \ap{A [s]} into all other blocks.
\textbf{(3)} The complement of (2), injecting \ap{A [s]} into the style block $W^5$ and \ap{A [c]} into all other blocks.
\textbf{(4)} Similar to (2), but injecting \ap{A [c]} into $W^4$ while other blocks receive \ap{A [c] in [s] style}.
\textbf{(5)} Similar to (3), but injecting \ap{A [s]} into $W^5$ while other blocks receive our method's prompt \ap{A [c] in [s] style}.

\begin{figure*}[ht]
    \centering
    \setlength{\tabcolsep}{1.5pt}
    {\small
    \begin{tabular}{ c c @{\hspace{0.2cm}} c c c c c}
        
        Content &  Style & (1) & (2) & (3) & (4) & (5) \\
        
        \includegraphics[width=0.135\textwidth]{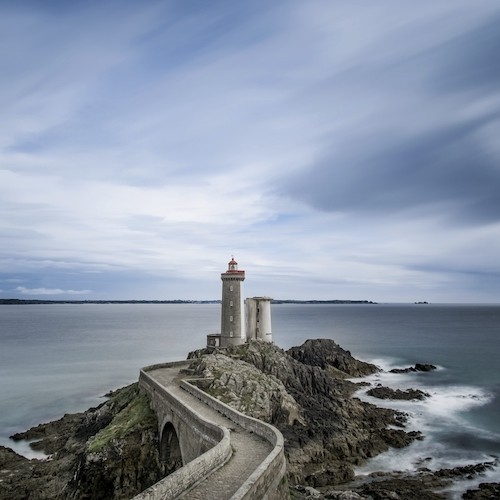} &
        \includegraphics[width=0.135\textwidth]{temp_figs/style_images/pen_sketch.jpeg} &
        \includegraphics[width=0.135\textwidth]{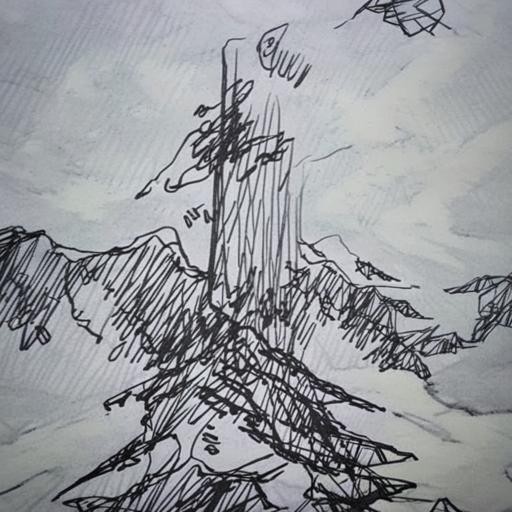} &
        \includegraphics[width=0.135\textwidth]{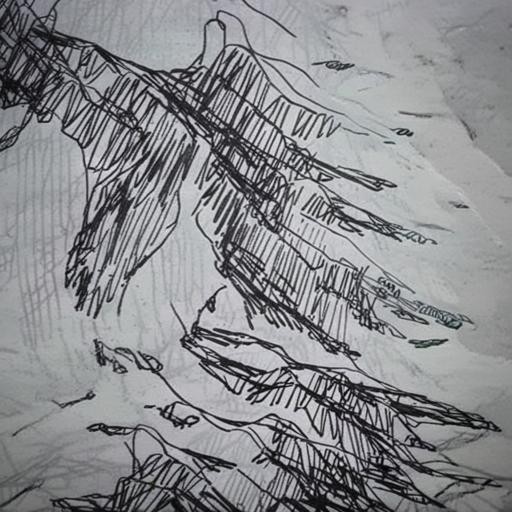} &
        \includegraphics[width=0.135\textwidth]{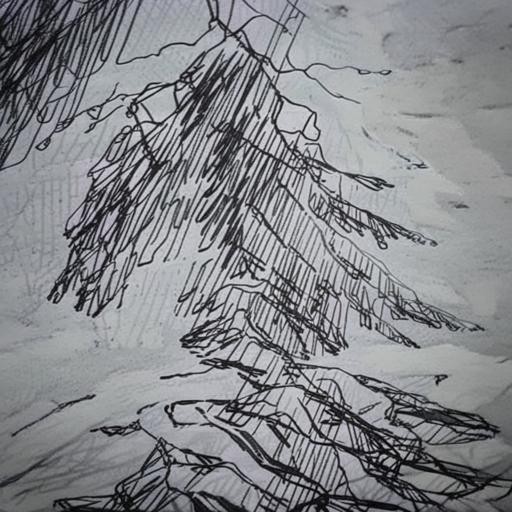} &
        \includegraphics[width=0.135\textwidth]{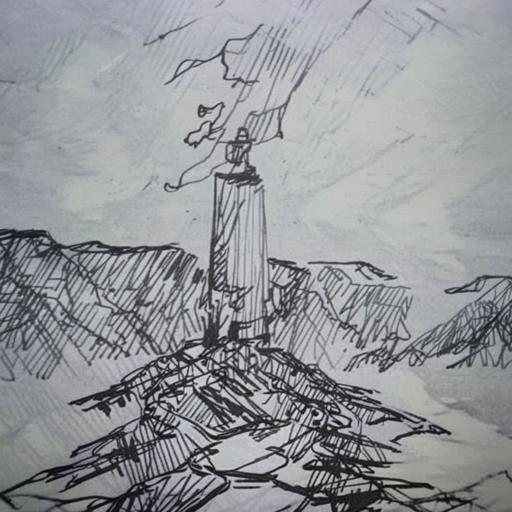} &
        \includegraphics[width=0.135\textwidth]{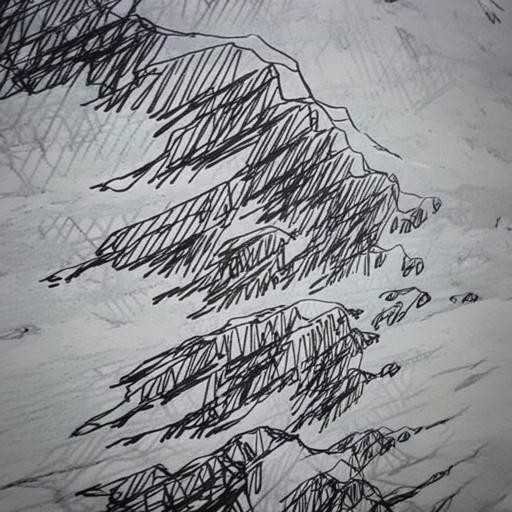} \\

        \includegraphics[width=0.135\textwidth]{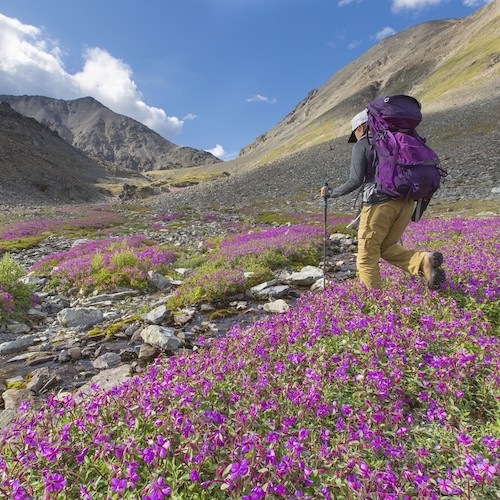} &
        \includegraphics[width=0.135\textwidth]{temp_figs/style_images/drawing3.png} &
        \includegraphics[width=0.135\textwidth]{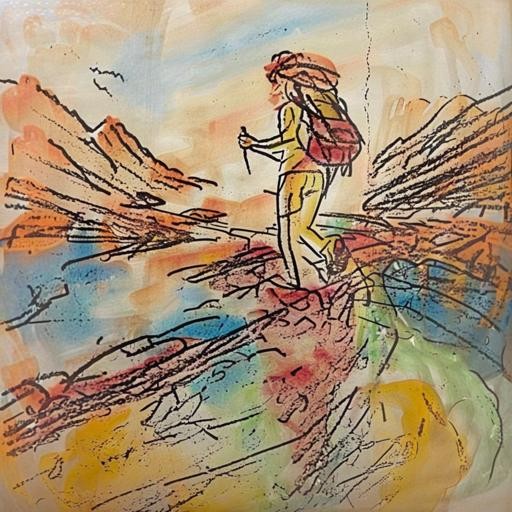} &
        \includegraphics[width=0.135\textwidth]{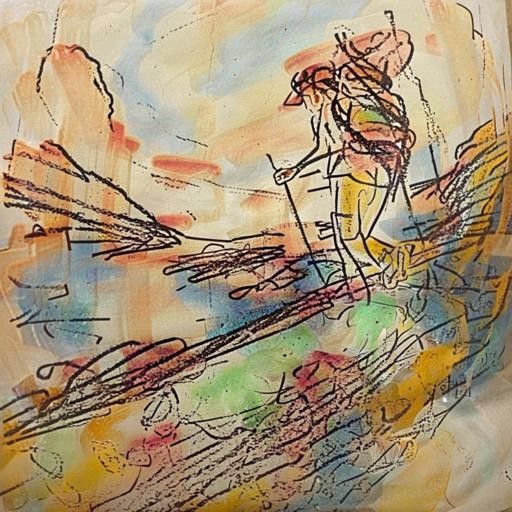} &
        \includegraphics[width=0.135\textwidth]{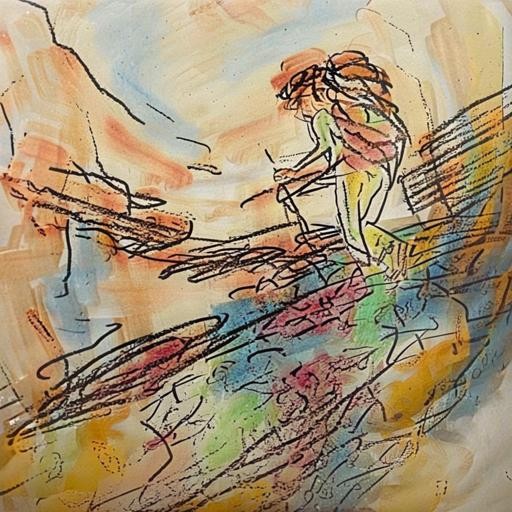} &
        \includegraphics[width=0.135\textwidth]{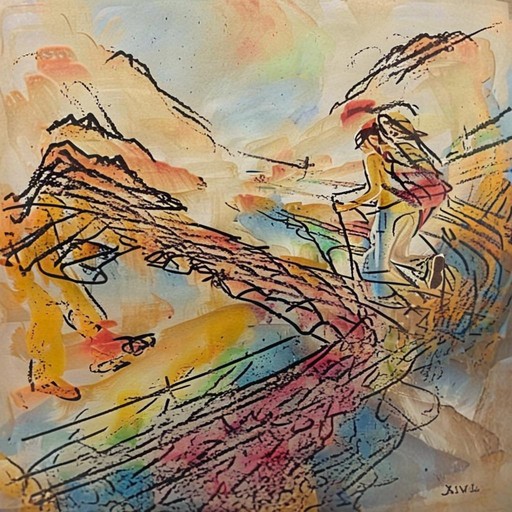} &
        \includegraphics[width=0.135\textwidth]{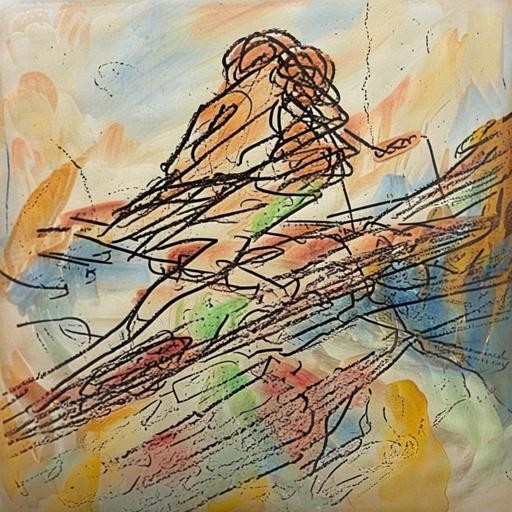} \\

        \includegraphics[width=0.135\textwidth]{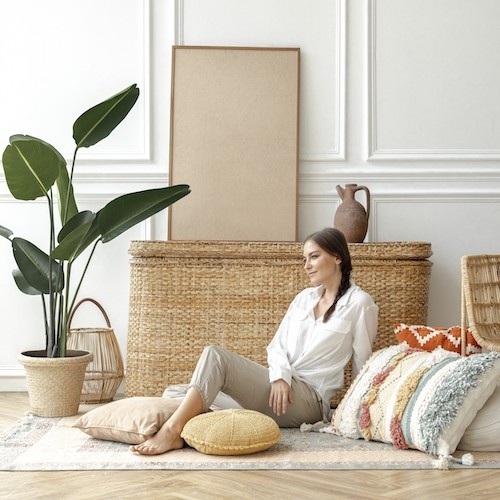} &
        \includegraphics[width=0.135\textwidth]{temp_figs/style_images/pen_sketch.jpeg} &
        \includegraphics[width=0.135\textwidth]{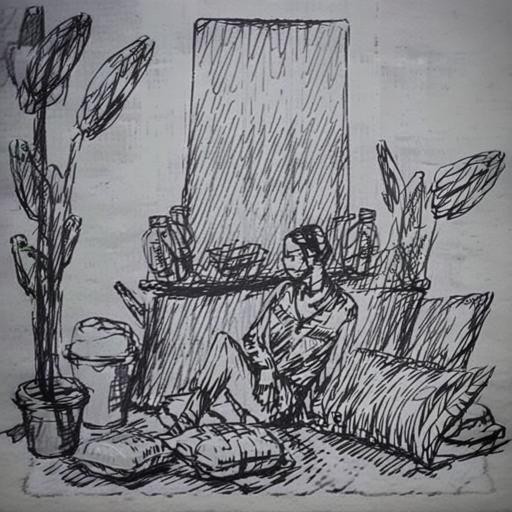} &
        \includegraphics[width=0.135\textwidth]{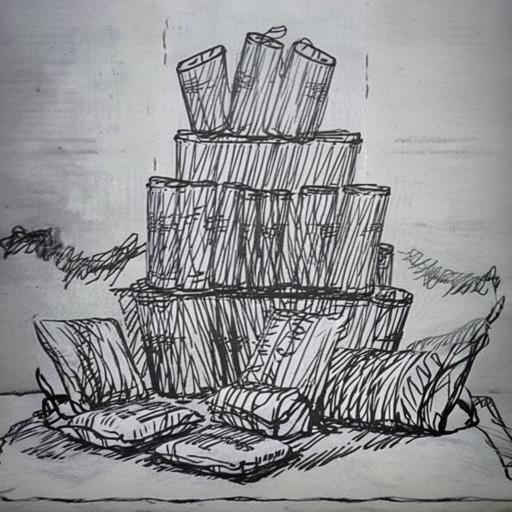} &
        \includegraphics[width=0.135\textwidth]{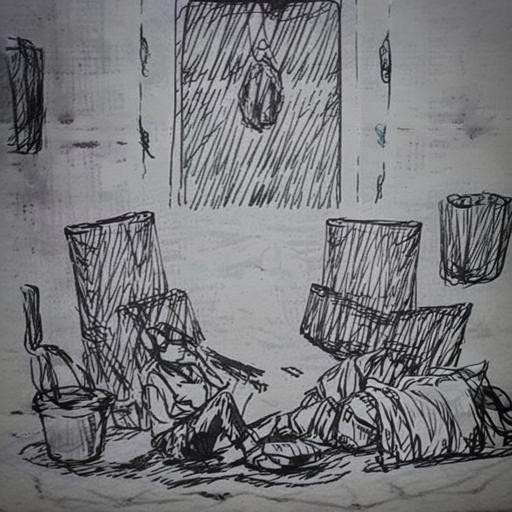} &
        \includegraphics[width=0.135\textwidth]{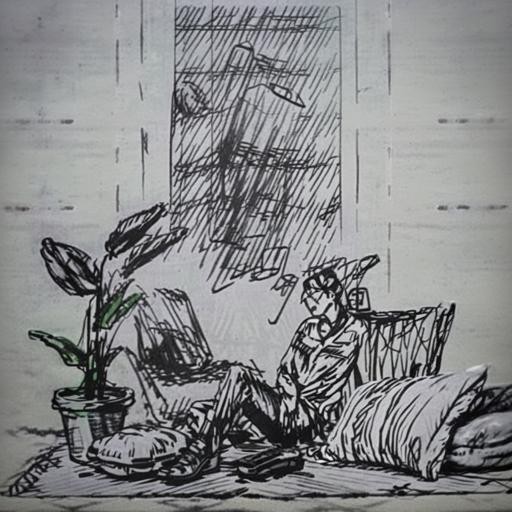} &
        \includegraphics[width=0.135\textwidth]{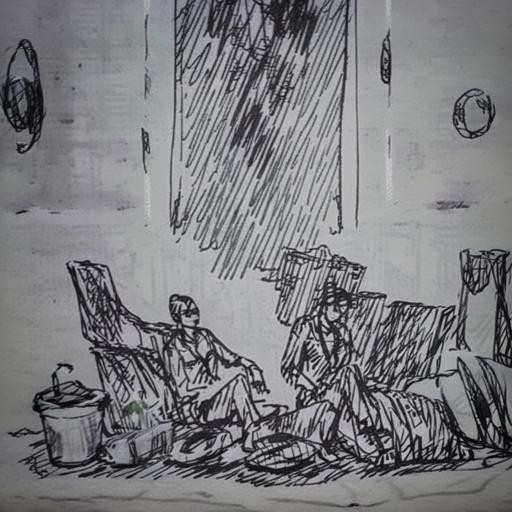} \\

    \end{tabular}
    
    }   
    \caption{Qualitative results of an ablation study investigating the effect of injecting different prompts into different blocks of the network to address the limitation of capturing content in complex scenes. Five experiments were conducted presented in the five columns [1-5]: \textbf{(1)} Injecting our method prompt, denoted as $p_1$ = \ap{A [c] in [s] style}, into the entire Unet.
\textbf{(2)} Injecting \ap{A [c]} into the content block $W^4$ while all other blocks receive \ap{A [s]}.
\textbf{(3)} The complement, injecting \ap{A [s]} into the style block $W^5$ and \ap{A [c]} into all other blocks.
\textbf{(4)} Similar to 2, but injecting \ap{A [c]} into $W^4$ while other blocks receive \ap{A [c] in [s] style}.
\textbf{(5)} Similar to 3, but injecting \ap{A [s]} into $W^5$ while other blocks receive our method's prompt \ap{A [c] in [s] style}.
 As can be seen the (4) columns contains the best results.}
    \label{fig:limitations_prompt_ablation}
\end{figure*}

We present the results of these experiments in \Cref{fig:limitations_prompt_ablation}. Our findings indicate that injecting the prompt \ap{A [c]} into $W^4$ while other blocks receive the prompt \ap{A [c] in [s] style} often leads to improved generation results, particularly for complex scenes containing numerous elements.

\section{Analysis and Ablation}
\begin{figure*}[h]
    \centering
    \setlength{\tabcolsep}{1.5pt}
    {\small
    \begin{tabular}{ c c c c c c c c c}
        
        $\Delta W^0$ &  $\Delta W^1$ & $\Delta W^2$ & $\Delta W^3$ & $\Delta W^4$ & $\Delta W^5$ & $\Delta W^6$ & $\Delta W^7$ & ~ \\
        
        \includegraphics[width=0.115\textwidth]{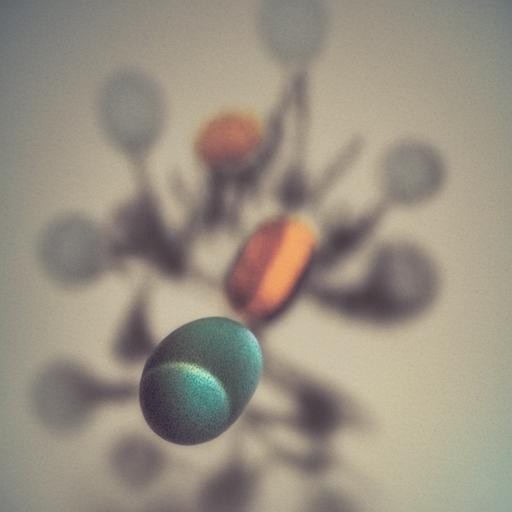} &
        \includegraphics[width=0.115\textwidth]{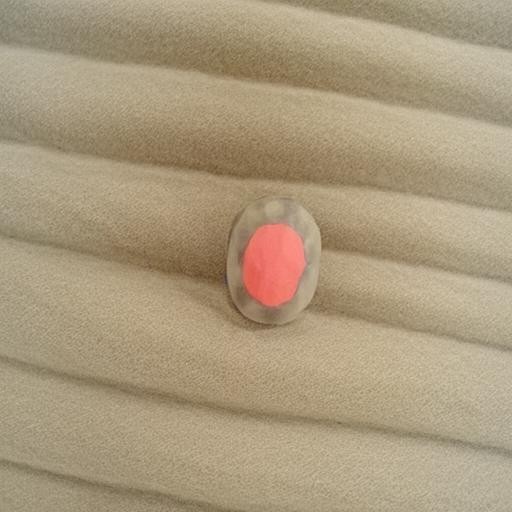} &
        \includegraphics[width=0.115\textwidth]{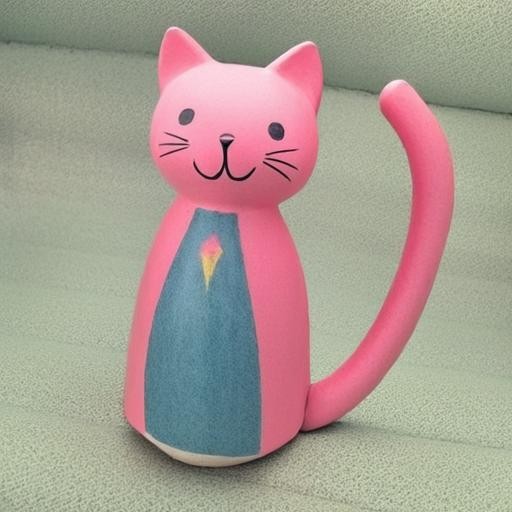} &
        \includegraphics[width=0.115\textwidth]{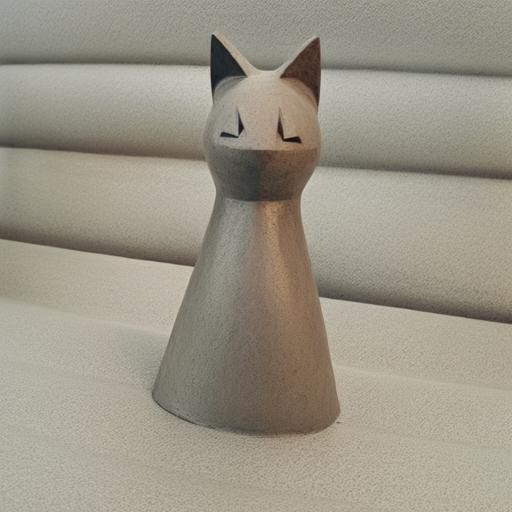} &
        \includegraphics[width=0.115\textwidth]{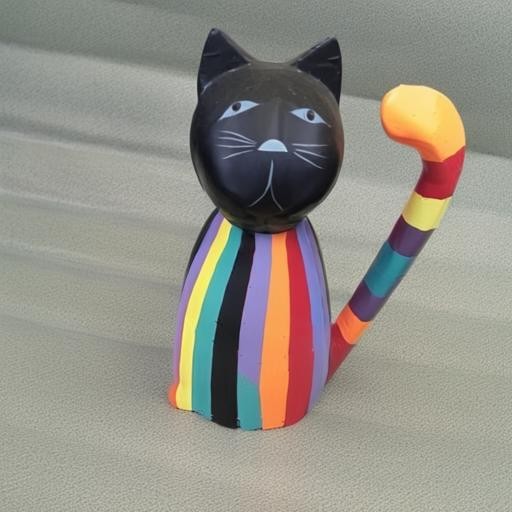} &
        \includegraphics[width=0.115\textwidth]{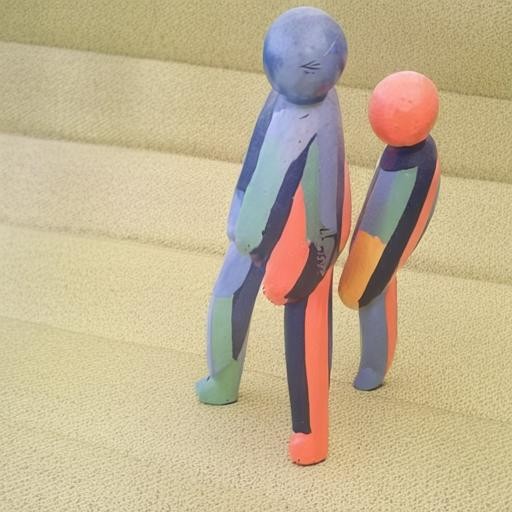} &
        \includegraphics[width=0.115\textwidth]{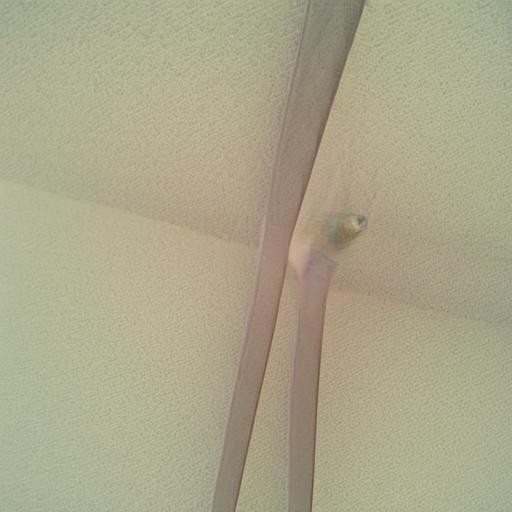} &
        \includegraphics[width=0.115\textwidth]{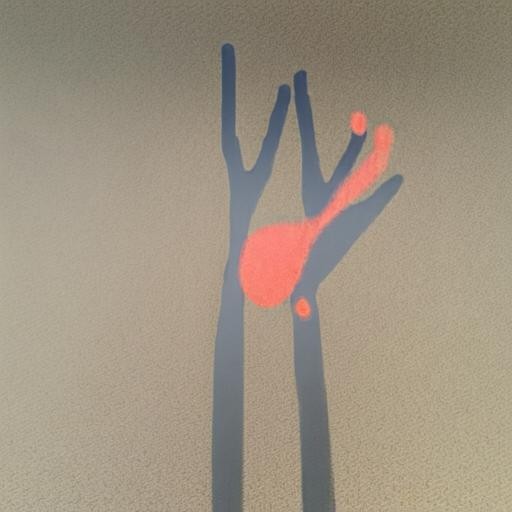} &  \raisebox{0.55cm}{\rotatebox[origin=l]{0}{$\Delta W^0$}} \\
        ~ &
        \includegraphics[width=0.115\textwidth]{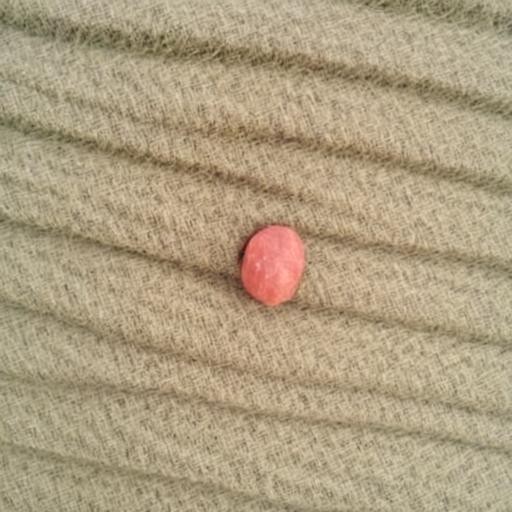} &
        \includegraphics[width=0.115\textwidth]{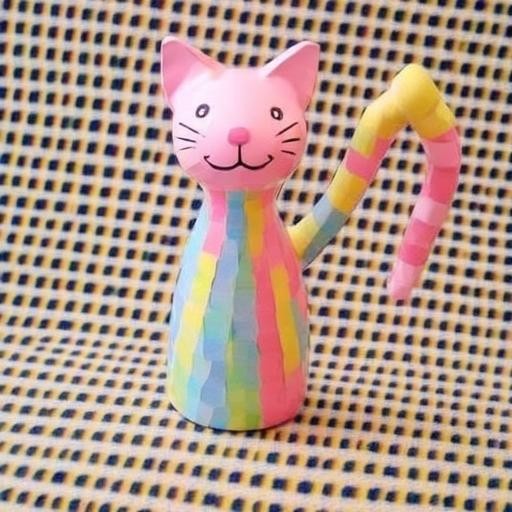} &
        \includegraphics[width=0.115\textwidth]{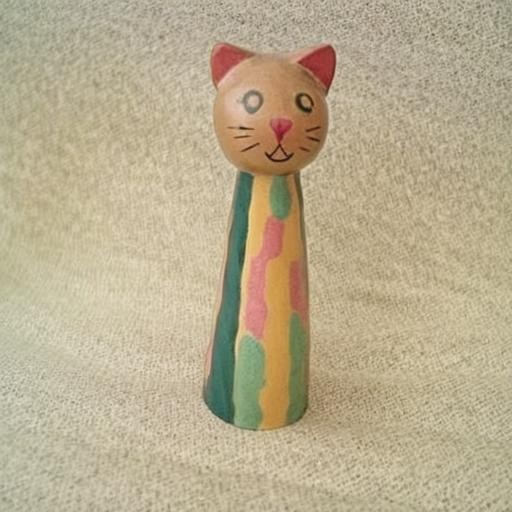} &
        \includegraphics[width=0.115\textwidth]{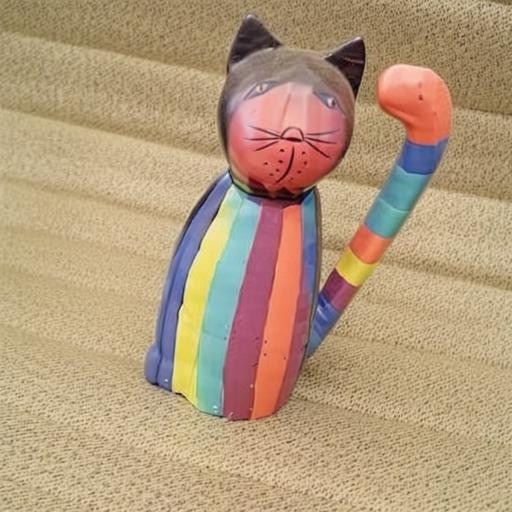} &
        \includegraphics[width=0.115\textwidth]{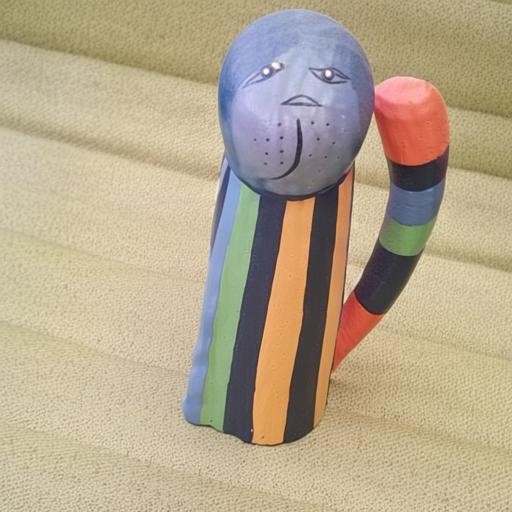} &
        \includegraphics[width=0.115\textwidth]{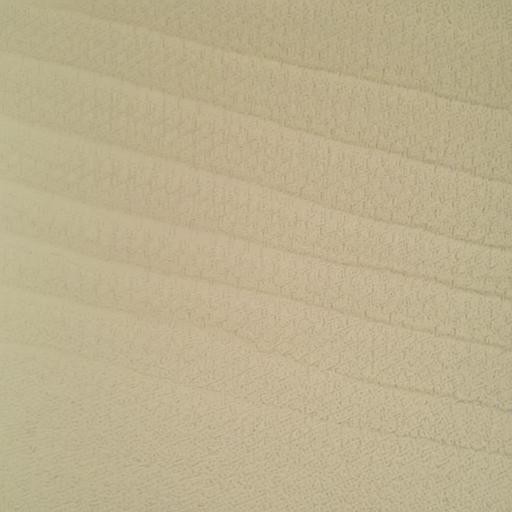} &
        \includegraphics[width=0.115\textwidth]{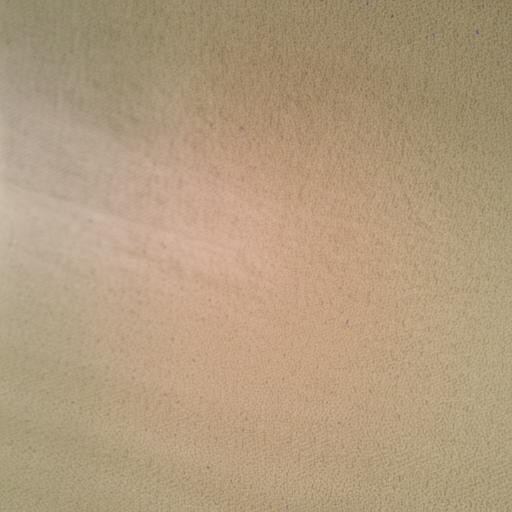} &  \raisebox{0.55cm}{\rotatebox[origin=l]{0}{$\Delta W^1$}} \\
        ~ & ~ &
        \includegraphics[width=0.115\textwidth]{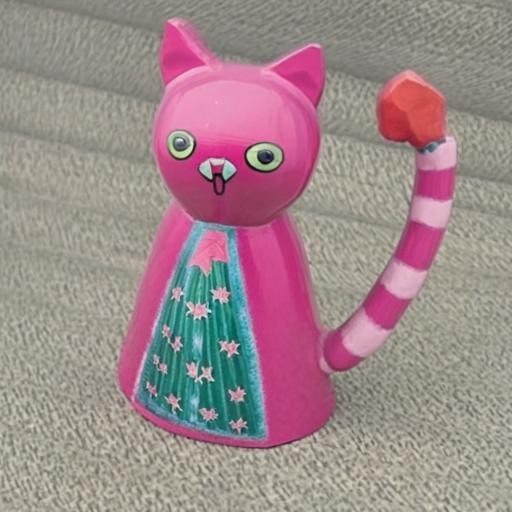} &
        \includegraphics[width=0.115\textwidth]{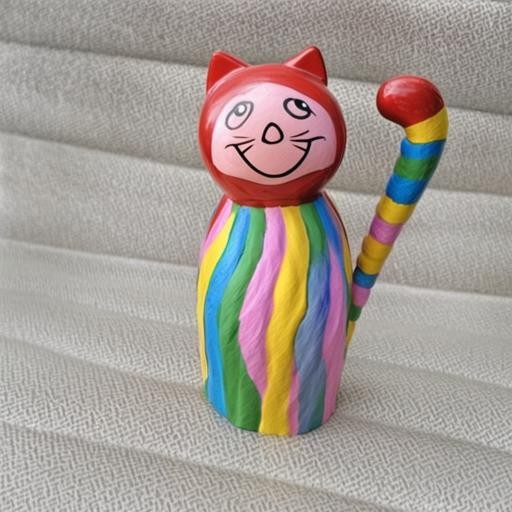} &
        \includegraphics[width=0.115\textwidth]{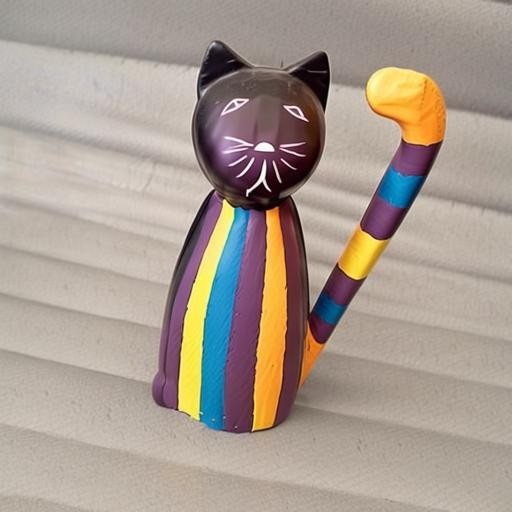} &
        \includegraphics[width=0.115\textwidth]{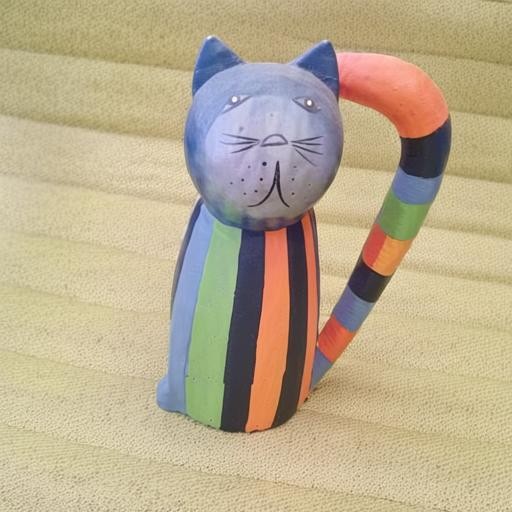} &
        \includegraphics[width=0.115\textwidth]{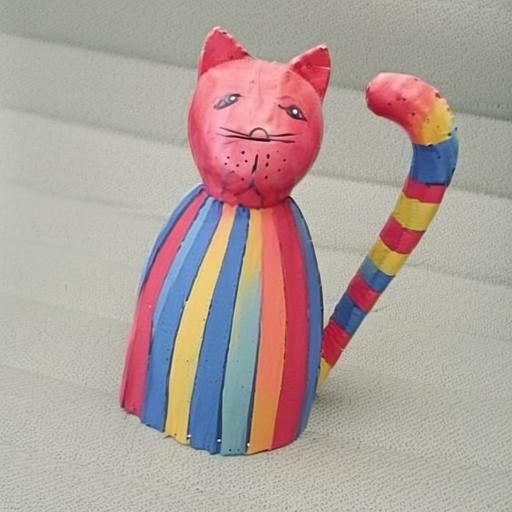} &
        \includegraphics[width=0.115\textwidth]{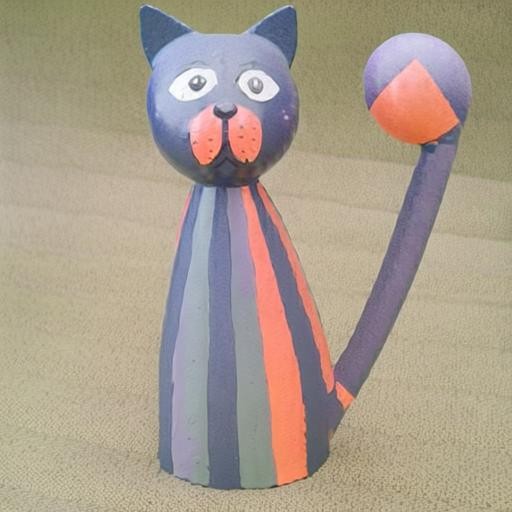}  & \raisebox{0.55cm}{\rotatebox[origin=l]{0}{$\Delta W^2$}} \\
        ~ & ~ & ~ &
        \includegraphics[width=0.115\textwidth]{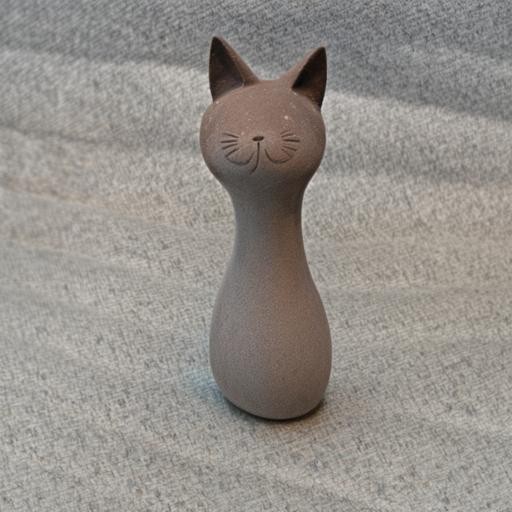} &
        \includegraphics[width=0.115\textwidth]{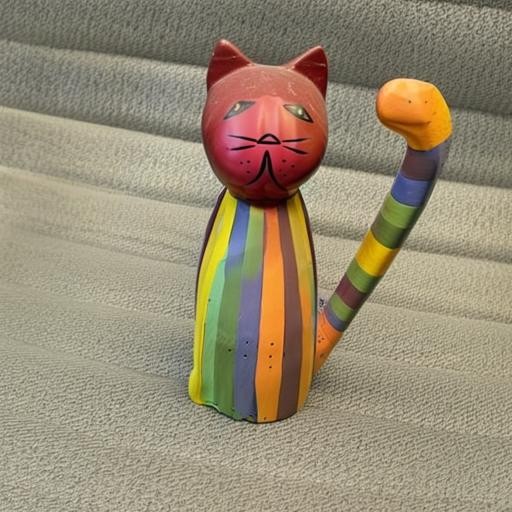} &
        \includegraphics[width=0.115\textwidth]{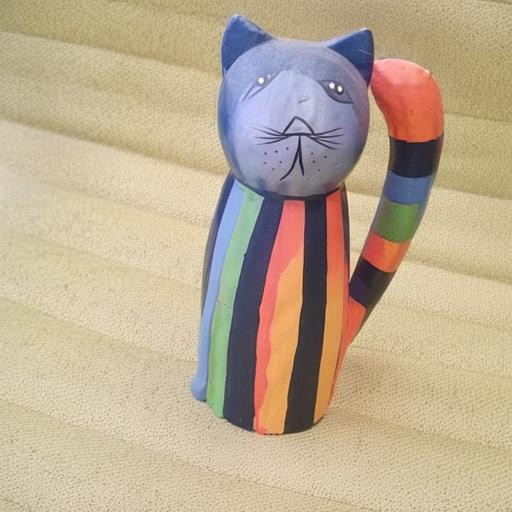} &
        \includegraphics[width=0.115\textwidth]{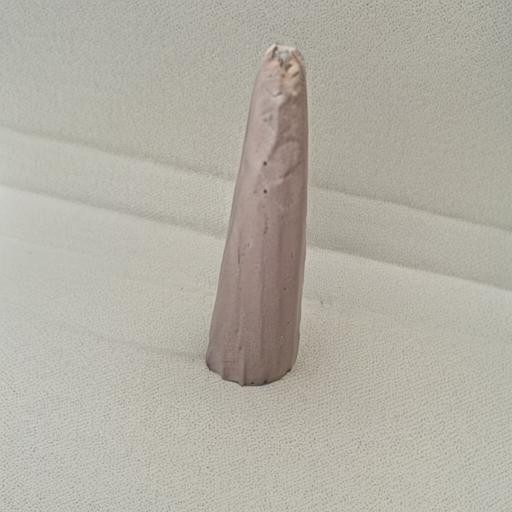} &
        \includegraphics[width=0.115\textwidth]{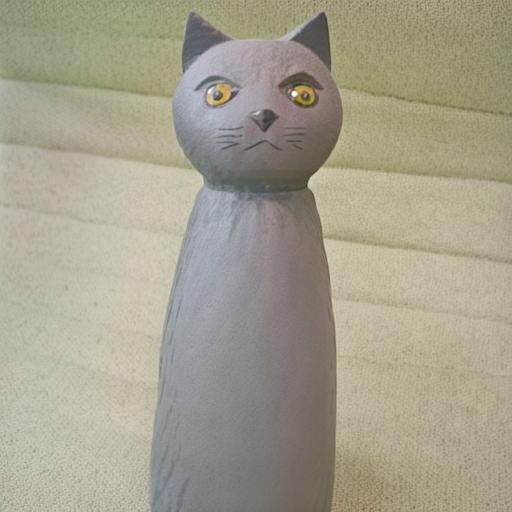} &  \raisebox{0.55cm}{\rotatebox[origin=l]{0}{$\Delta W^3$}} \\
        ~ & ~ & ~ & ~ &
        \includegraphics[width=0.115\textwidth]{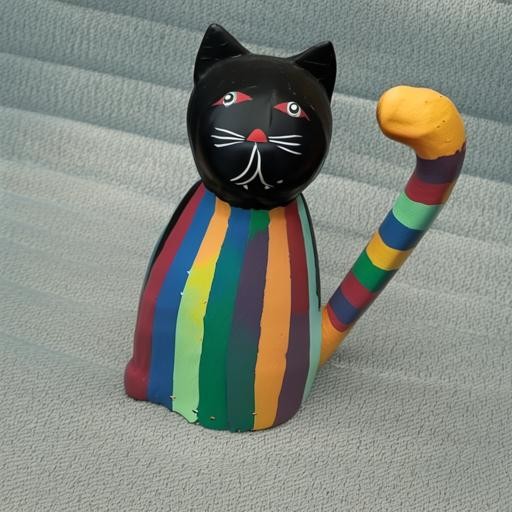} &
        \includegraphics[width=0.115\textwidth]{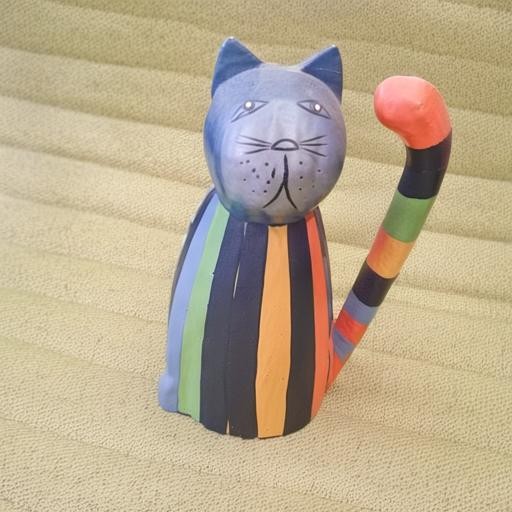} &
        \includegraphics[width=0.115\textwidth]{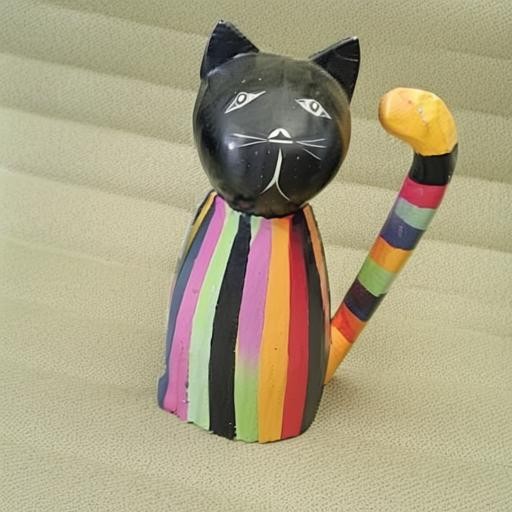} &
        \includegraphics[width=0.115\textwidth]{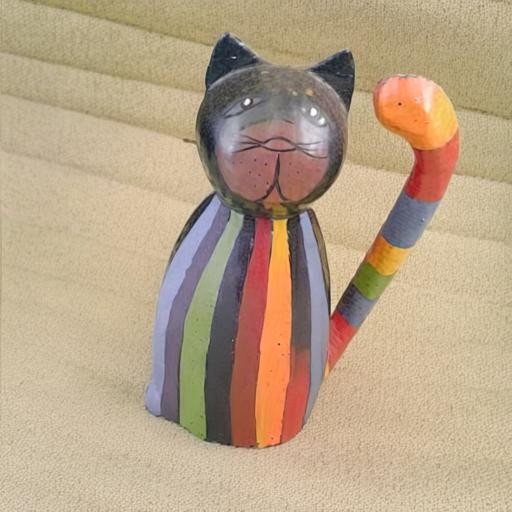} &  \raisebox{0.55cm}{\rotatebox[origin=l]{0}{$\Delta W^4$}}  \\
        ~ & ~ & ~ & ~ & ~ & 
        \includegraphics[width=0.115\textwidth]{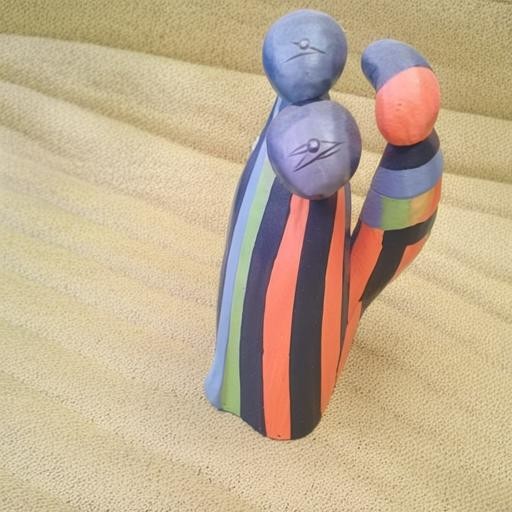} &
        \includegraphics[width=0.115\textwidth]{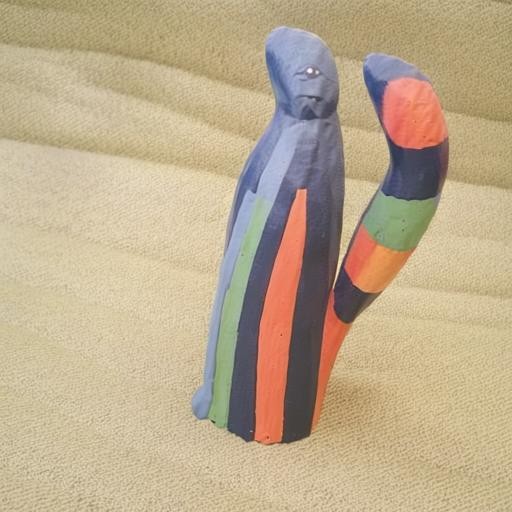} &
        \includegraphics[width=0.115\textwidth]{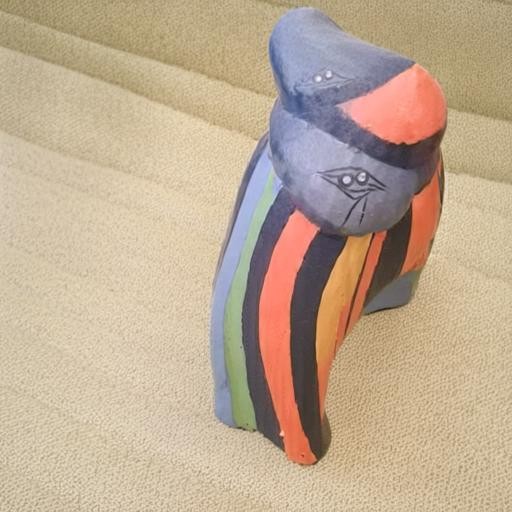} &  \raisebox{0.55cm}{\rotatebox[origin=l]{0}{$\Delta W^5$}} \\
        ~ & ~ & ~ & ~ & ~ & ~ &
        \includegraphics[width=0.115\textwidth]{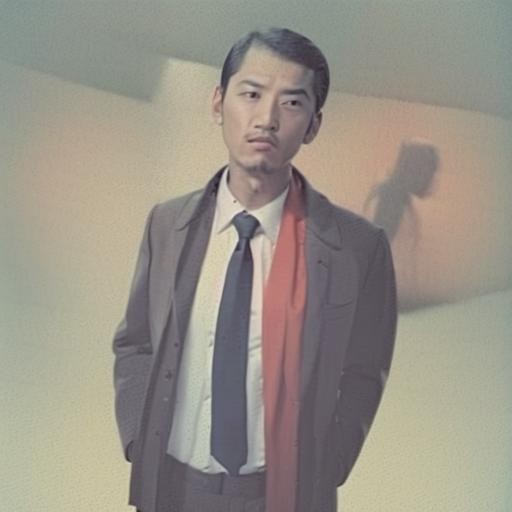} &
        \includegraphics[width=0.115\textwidth]{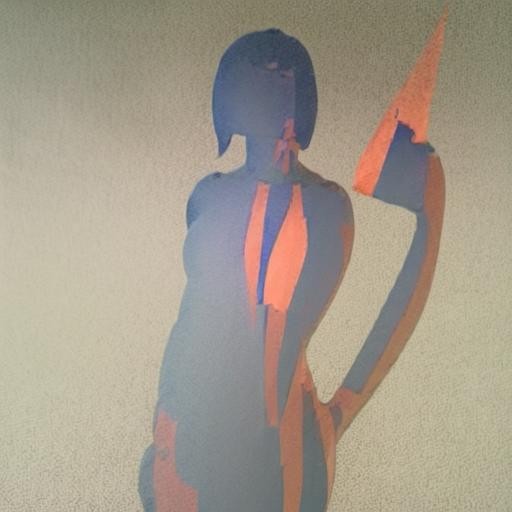} &  \raisebox{0.55cm}{\rotatebox[origin=l]{0}{$\Delta W^6$}}  \\
        ~ & ~ & ~ & ~ & ~ & ~ & ~ &
        \includegraphics[width=0.115\textwidth]{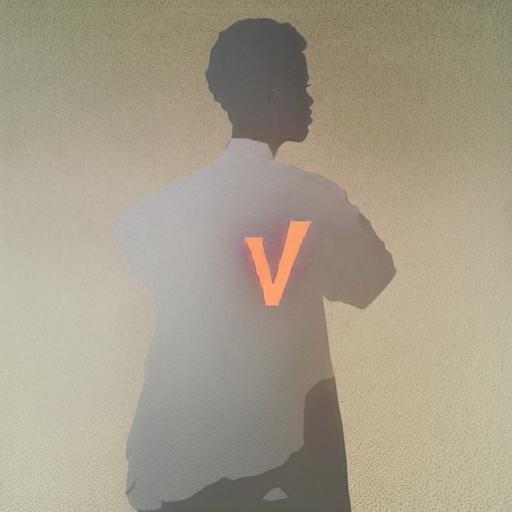} &  \raisebox{0.55cm}{\rotatebox[origin=l]{0}{$\Delta W^7$}}  \\

    \end{tabular}
    }   
    \caption{Qualitative results of the ablation study showcasing the reconstruction images for prompt \ap{A [v]} after training LoRAs for different block combinations of the SDXL Unet. Each cell (i, j) represents a specific block combination, with the diagonal representing output generated by training a single block. Notably, cells (4, 5) demonstrate the most consistent and optimal reconstruction for content and style, respectively}
    \label{fig:optimization_grid1}
    
\end{figure*}

\begin{figure*}[ht]
    \centering
    \setlength{\tabcolsep}{1.5pt}
    {\small
    \begin{tabular}{ c c c c c c c c c}
        
        $\Delta W^0$ &  $\Delta W^1$ & $\Delta W^2$ & $\Delta W^3$ & $\Delta W^4$ & $\Delta W^5$ & $\Delta W^6$ & $\Delta W^7$ & ~ \\
        
        \includegraphics[width=0.115\textwidth]{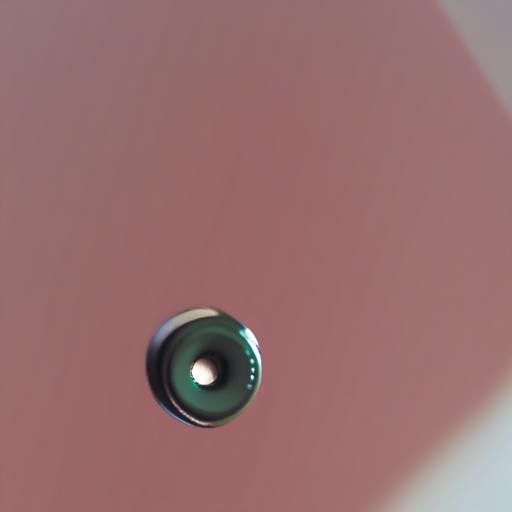} &
        \includegraphics[width=0.115\textwidth]{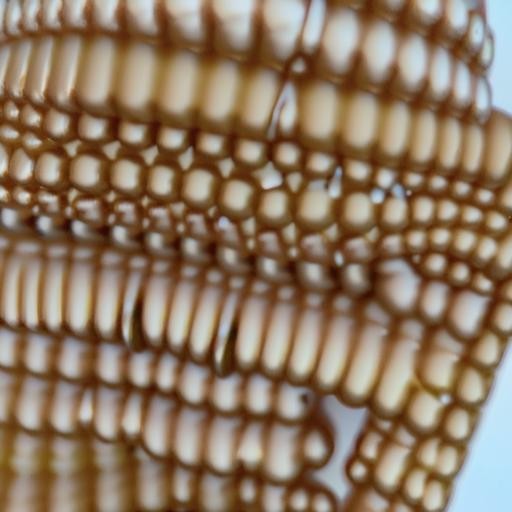} &
        \includegraphics[width=0.115\textwidth]{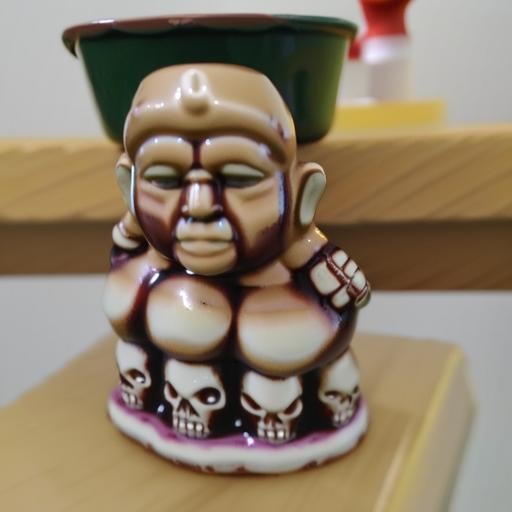} &
        \includegraphics[width=0.115\textwidth]{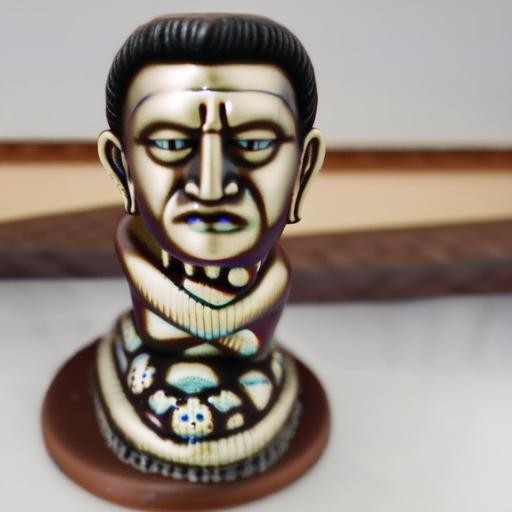} &
        \includegraphics[width=0.115\textwidth]{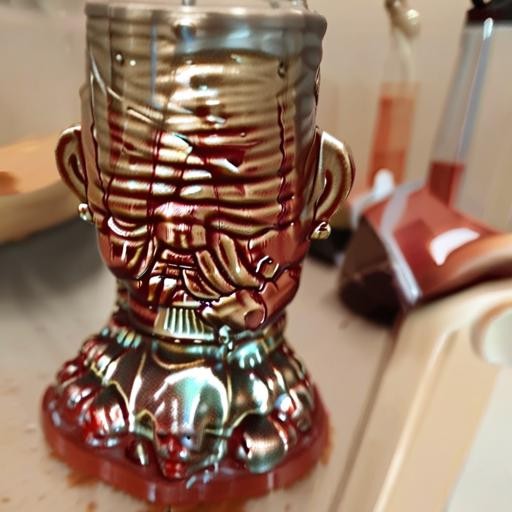} &
        \includegraphics[width=0.115\textwidth]{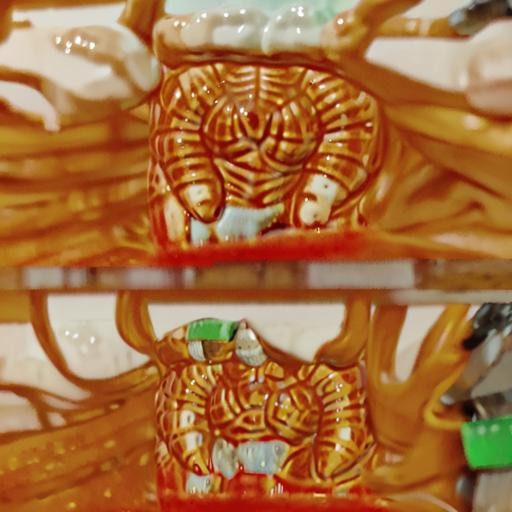} &
        \includegraphics[width=0.115\textwidth]{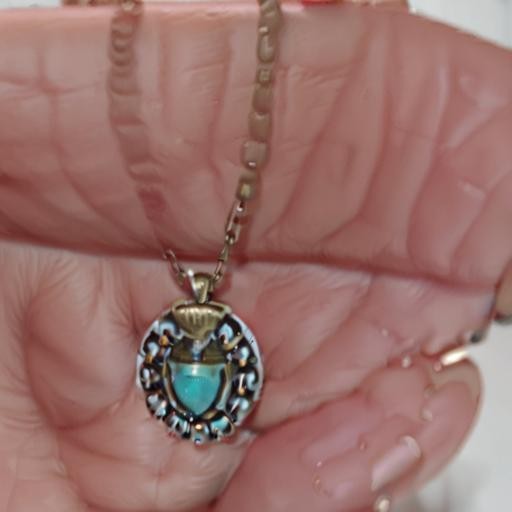} &
        \includegraphics[width=0.115\textwidth]{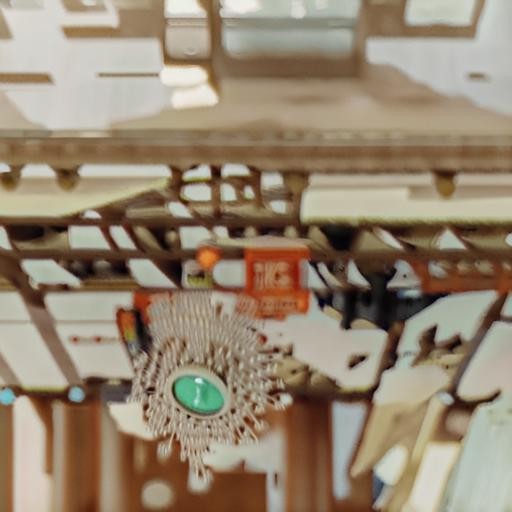} &  \raisebox{0.55cm}{\rotatebox[origin=l]{0}{$\Delta W^0$}} \\
        ~ &
        \includegraphics[width=0.115\textwidth]{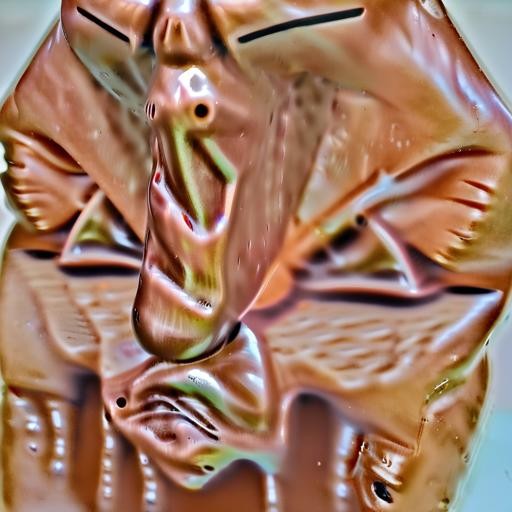} &
        \includegraphics[width=0.115\textwidth]{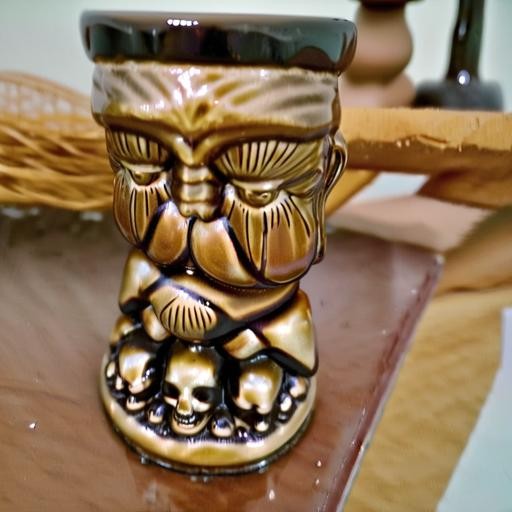} &
        \includegraphics[width=0.115\textwidth]{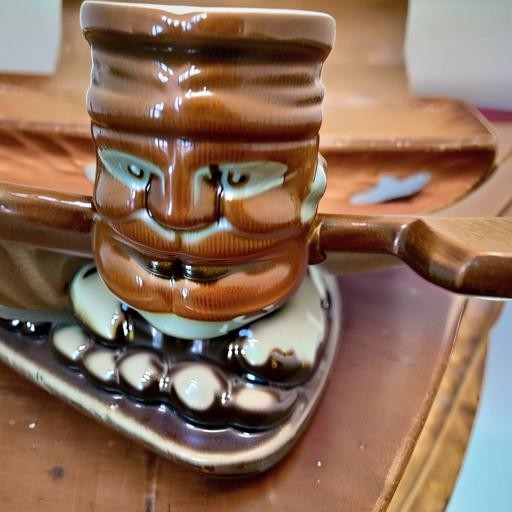} &
        \includegraphics[width=0.115\textwidth]{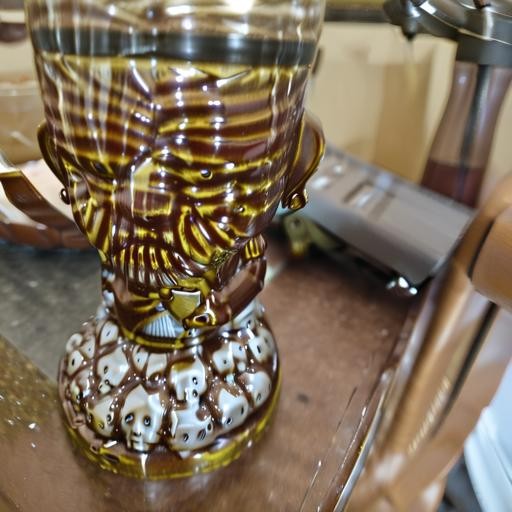} &
        \includegraphics[width=0.115\textwidth]{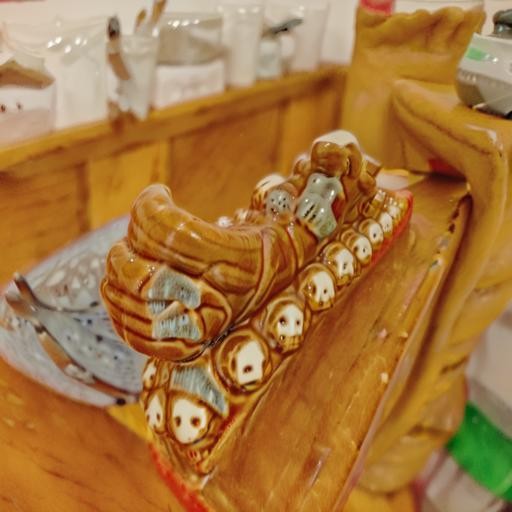} &
        \includegraphics[width=0.115\textwidth]{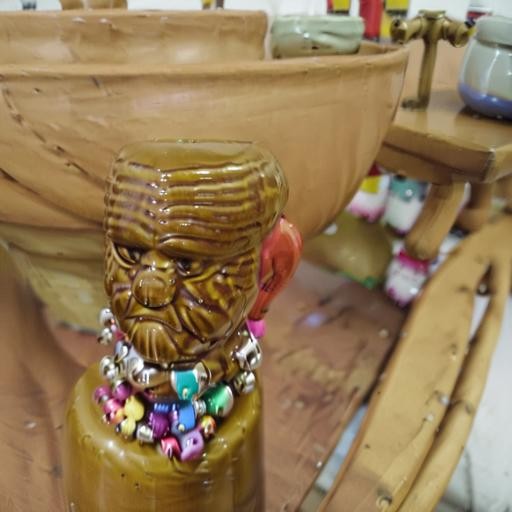} &
        \includegraphics[width=0.115\textwidth]{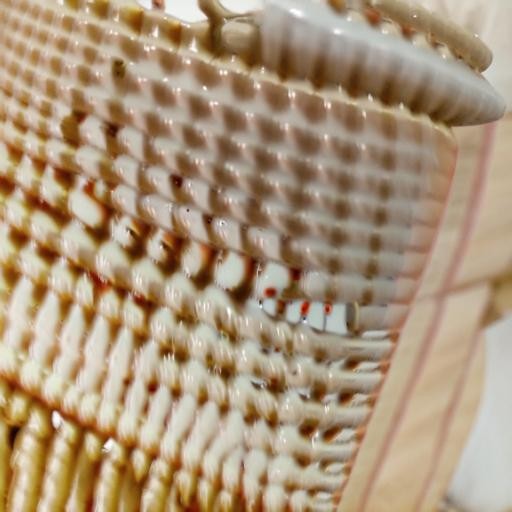} &  \raisebox{0.55cm}{\rotatebox[origin=l]{0}{$\Delta W^1$}}\\
        ~ & ~ &
        \includegraphics[width=0.115\textwidth]{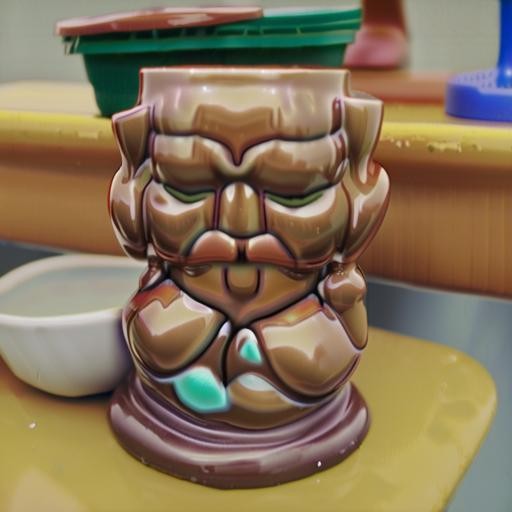} &
        \includegraphics[width=0.115\textwidth]{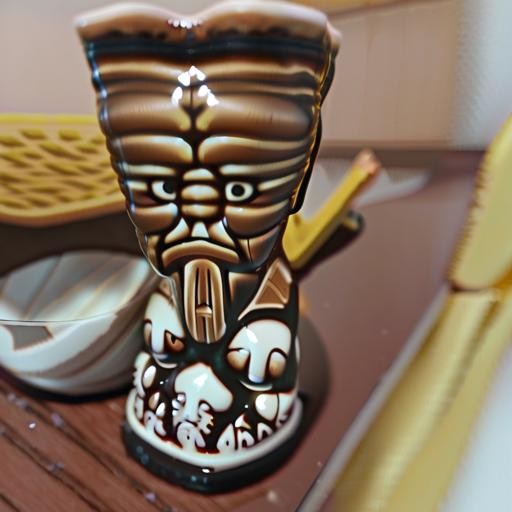} &
        \includegraphics[width=0.115\textwidth]{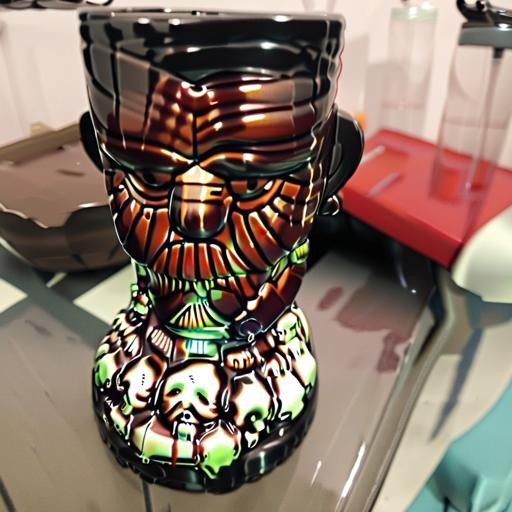} &
        \includegraphics[width=0.115\textwidth]{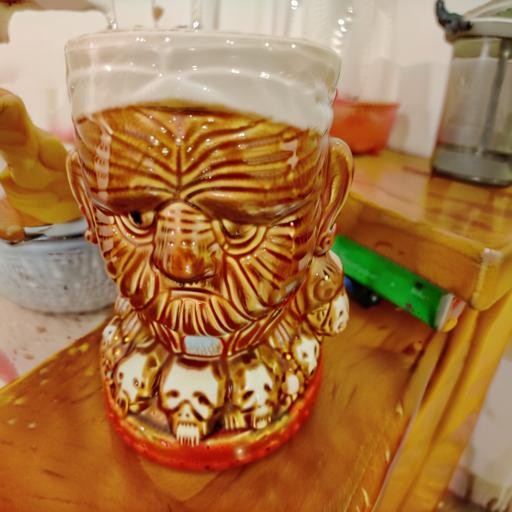} &
        \includegraphics[width=0.115\textwidth]{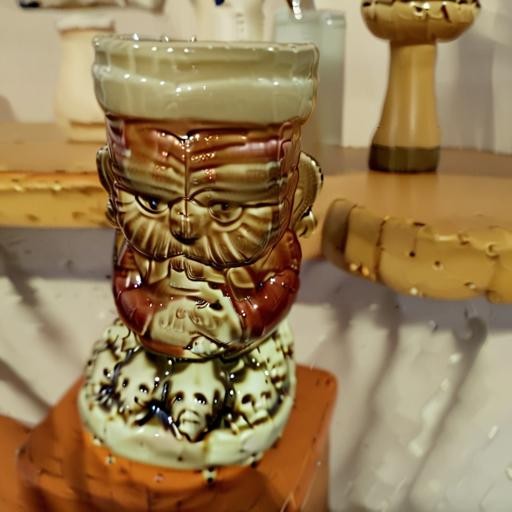} &
        \includegraphics[width=0.115\textwidth]{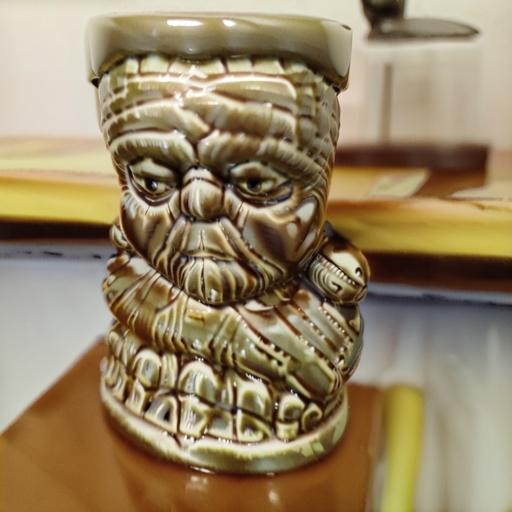} &  \raisebox{0.55cm}{\rotatebox[origin=l]{0}{$\Delta W^2$}}\\
        ~ & ~ & ~ &
        \includegraphics[width=0.115\textwidth]{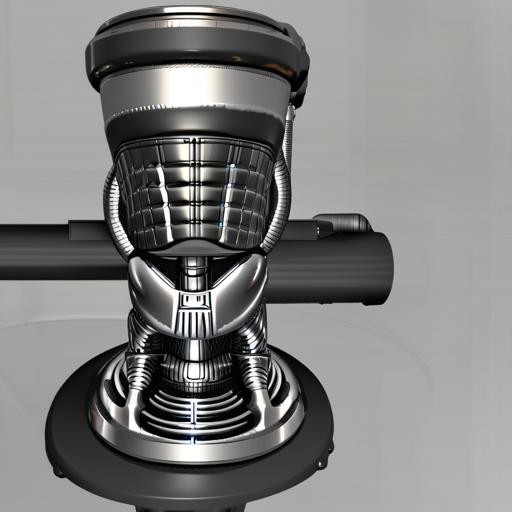} &
        \includegraphics[width=0.115\textwidth]{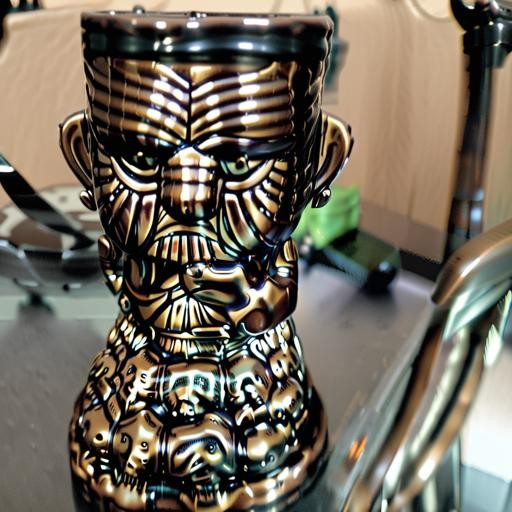} &
        \includegraphics[width=0.115\textwidth]{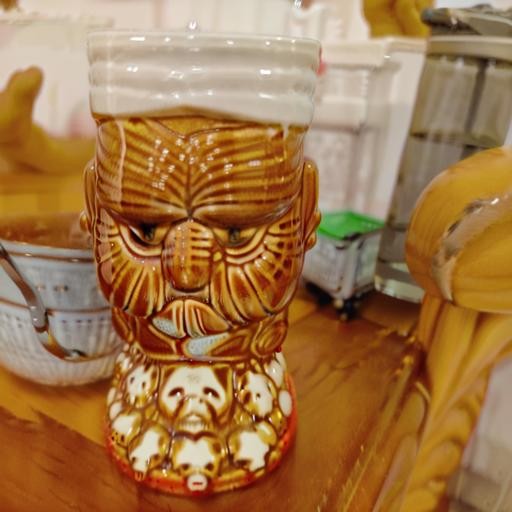} &
        \includegraphics[width=0.115\textwidth]{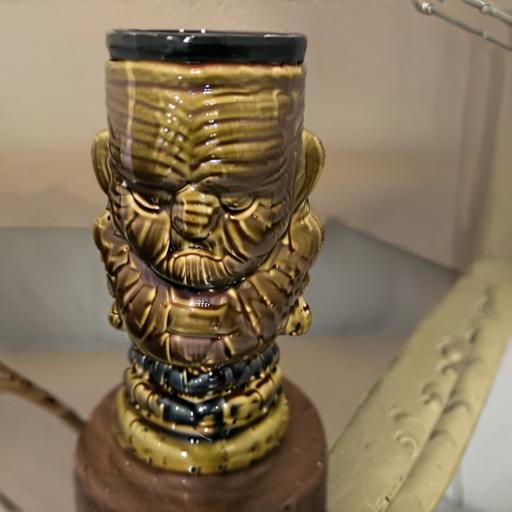} &
        \includegraphics[width=0.115\textwidth]{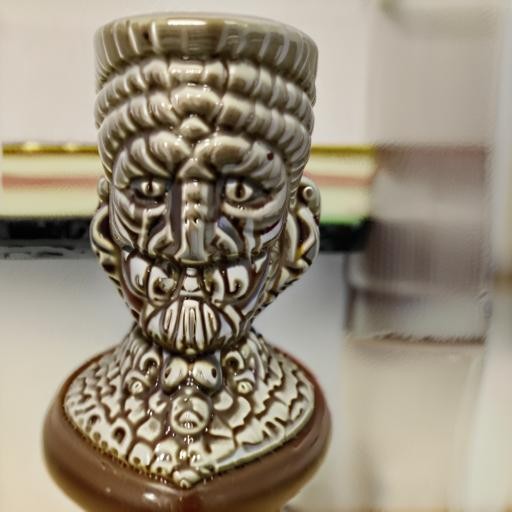} &  \raisebox{0.55cm}{\rotatebox[origin=l]{0}{$\Delta W^3$}}\\
        ~ & ~ & ~ & ~ &
        \includegraphics[width=0.115\textwidth]{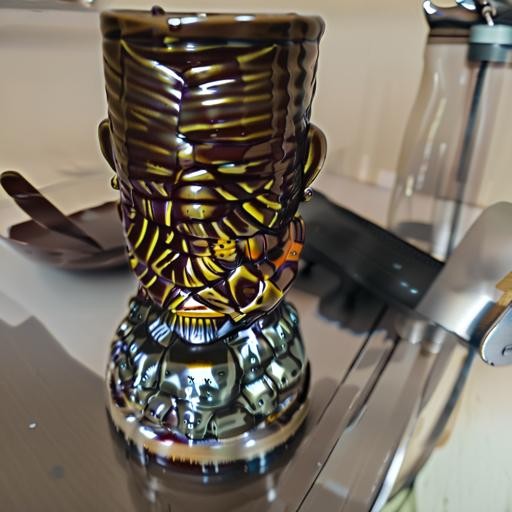} &
        \includegraphics[width=0.115\textwidth]{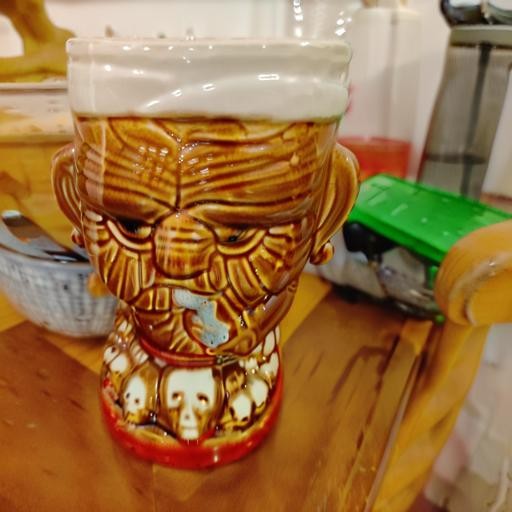} &
        \includegraphics[width=0.115\textwidth]{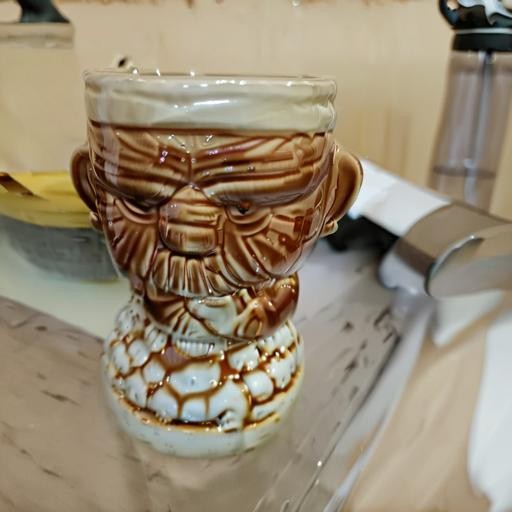} &
        \includegraphics[width=0.115\textwidth]{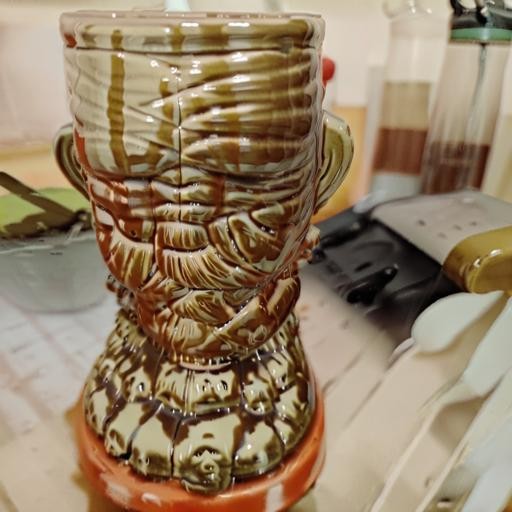} &  \raisebox{0.55cm}{\rotatebox[origin=l]{0}{$\Delta W^4$}}\\
        ~ & ~ & ~ & ~ & ~ & 
        \includegraphics[width=0.115\textwidth]{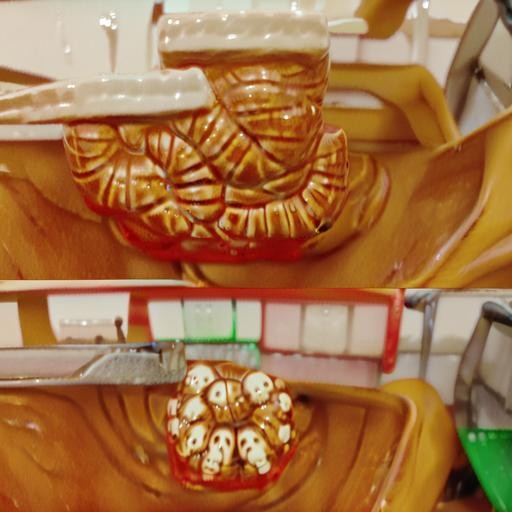} &
        \includegraphics[width=0.115\textwidth]{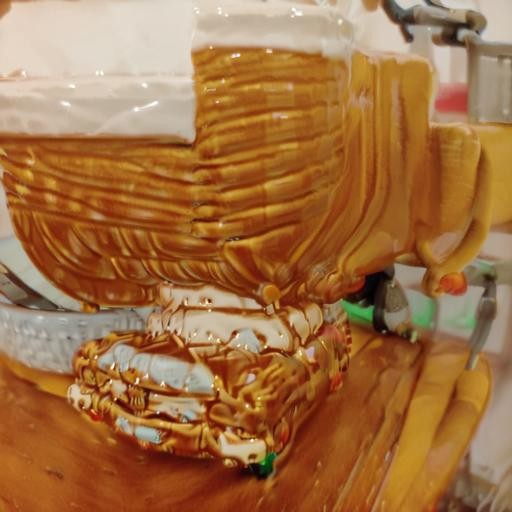} &
        \includegraphics[width=0.115\textwidth]{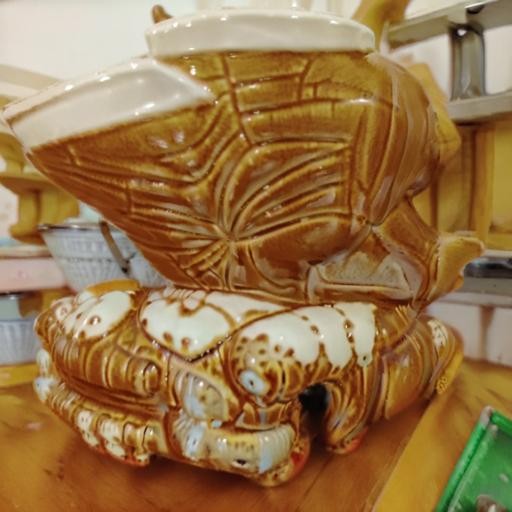} &  \raisebox{0.55cm}{\rotatebox[origin=l]{0}{$\Delta W^5$}}\\
        ~ & ~ & ~ & ~ & ~ & ~ &
        \includegraphics[width=0.115\textwidth]{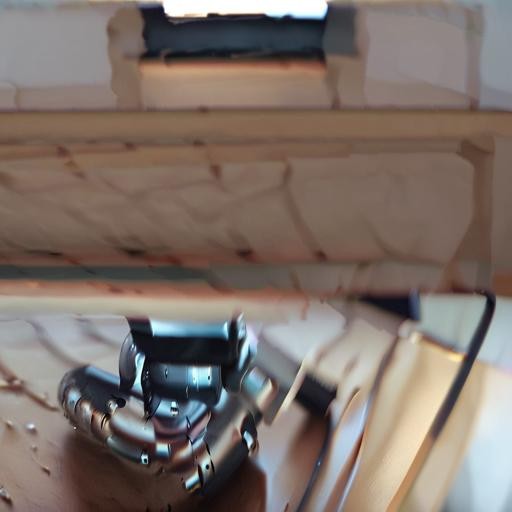} &
        \includegraphics[width=0.115\textwidth]{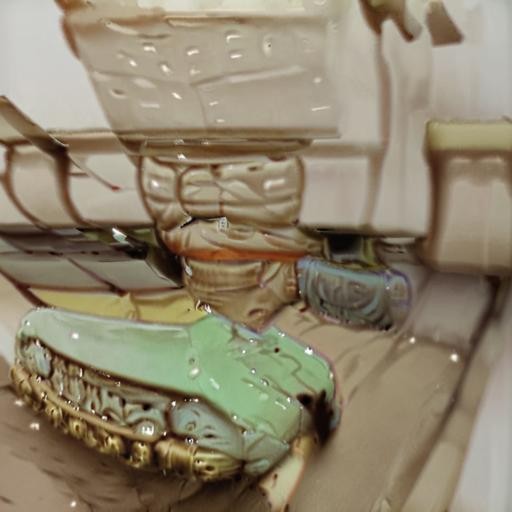} &  \raisebox{0.55cm}{\rotatebox[origin=l]{0}{$\Delta W^6$}}\\
        ~ & ~ & ~ & ~ & ~ & ~ & ~ &
        \includegraphics[width=0.115\textwidth]{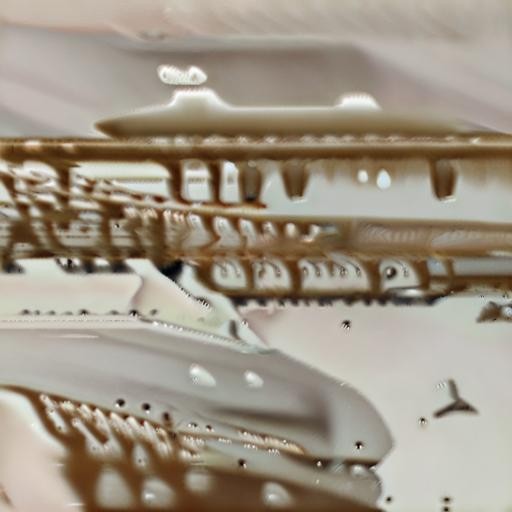} &  \raisebox{0.55cm}{\rotatebox[origin=l]{0}{$\Delta W^7$}}\\

    \end{tabular}
    }   
    \caption{Qualitative results of the ablation study showcasing the reconstruction images for prompt \ap{A [v]} after training LoRAs for different block combinations of the SDXL Unet. Each cell (i, j) represents a specific block combination, with the diagonal representing output generated by training a single block. Notably, cells (4, 5) demonstrate the most consistent and optimal reconstruction for content and style, respectively.}
    \label{fig:optimization_grid2}
\end{figure*}

\paragraph{Layers Optimization}
As detailed in Section 4 of the main paper, the SDXL UNet comprises 11 transformer blocks, with the high-resolution blocks containing 2 attention layers each and the middle 6 blocks containing 10 attention layers each (see Figure 3 in the main paper).
To explore the impact of different block combinations on the resulting image, we divided the UNet into 8 blocks $\{W^0_0\ldots W^7_0\}$, where $\{W^1_0\ldots W^6_0\}$  represent the bottleneck blocks, as discussed in Section 4, and designated $W^0_0$ and $W^7_0$ for the high-resolution blocks at the edges. We aimed to assess the effects of optimizing various block combinations $\{\Delta W^i, \Delta W^j\}$ by jointly training the LoRA weights of the corresponding blocks. Qualitative results are depicted in \Cref{fig:optimization_grid1,fig:optimization_grid2}, where each cell (i, j) represents the reconstruction image for the prompt \ap{A [v]} after training the LoRAs solely for the i-th and j-th blocks of the SDXL Unet. The diagonal entries represent output generated by training a single block.
Upon examination, we observed that optimizing $\{\Delta W^4, \Delta W^5\}$ consistently produced the most satisfactory results for content and style, respectively, outperforming other combinations. Notably, the reconstruction in cell (4, 5) yielded the best results achievable among all combinations, supporting our findings in the main paper.
Furthermore, we noted that the combination of blocks 2 and 5 also achieved satisfactory reconstruction. However, employing this combination may lead to less disentanglement of style from content, as $\Delta W^5$ needs to \ap{cover} $\Delta W^2$ by learning content details instead of focusing primarily on style, as intended. This observation further solidifies our choice of optimizing $\{\Delta W^4, \Delta W^5\}$ for effective style-content separation.

\vspace{-4pt}
\paragraph{Prompt Selection}
To validate our choice of the prompt \ap{A [v]} during optimization, we conducted an ablation study regarding the prompt used during training. As described in the DreamBooth \cite{dreambooth23} paper, the authors suggest that the most efficient way to conduct the fine-tuning process is by using the prompt \ap{A [v] {\small\textless}class-name{\small\textgreater}}, where [v] is the token dedicated for personalization, and {\small\textless}class-name{\small\textgreater} is the class of the object depicted in the input image. We compare our method of optimizing $\Delta W^4$ and $\Delta W^5$ with the prompt \ap{A [v]} against using the suggested \ap{A [v] {\small\textless}class-name{\small\textgreater}} prompt.

\begin{figure*}[ht]
    \centering
    \setlength{\tabcolsep}{1.5pt}
    {\small
    \begin{tabular}{c c @{\hspace{0.17cm}} | @{\hspace{0.17cm}}c c c @{\hspace{0.17cm}} | @{\hspace{0.17cm}}c}
        
     Content & Style & (1) & (2) & (3) & Ours \\

    \includegraphics[width=0.15\textwidth]{temp_figs/content_images/scary_mug.jpg} &
    \includegraphics[width=0.15\textwidth]{temp_figs/content_images/cat.jpeg} &
    
    \includegraphics[width=0.15\textwidth]{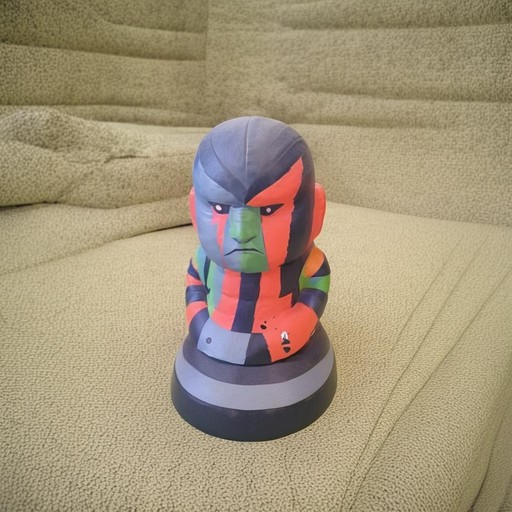} &
    \includegraphics[width=0.15\textwidth]{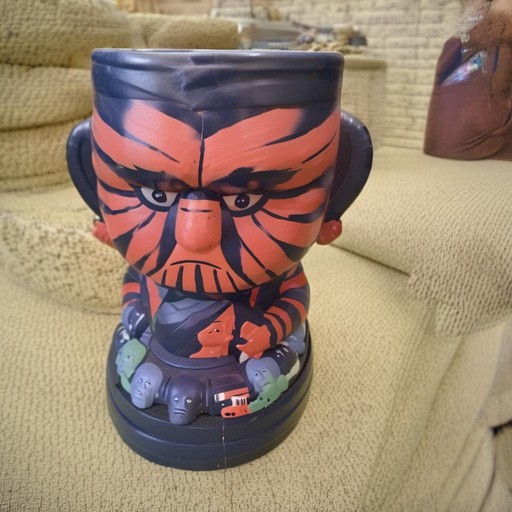} &
    \includegraphics[width=0.15\textwidth]{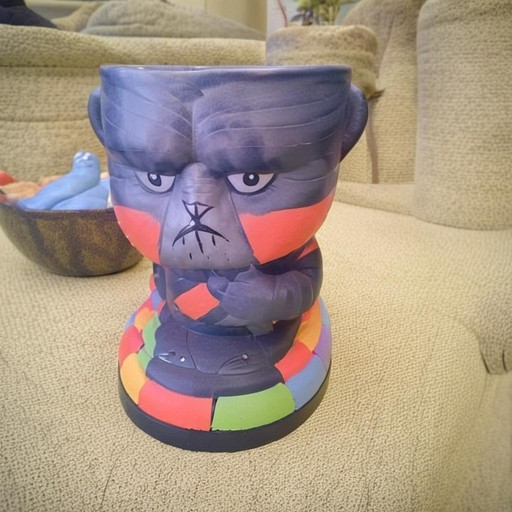} &
    \includegraphics[width=0.15\textwidth]{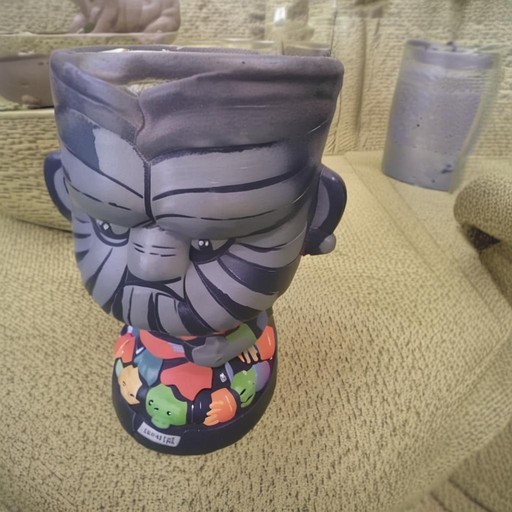} \\

    \includegraphics[width=0.15\textwidth]{temp_figs/content_images/cat.jpeg} &
    \includegraphics[width=0.15\textwidth]{temp_figs/content_images/scary_mug.jpg} &
    
    \includegraphics[width=0.15\textwidth]{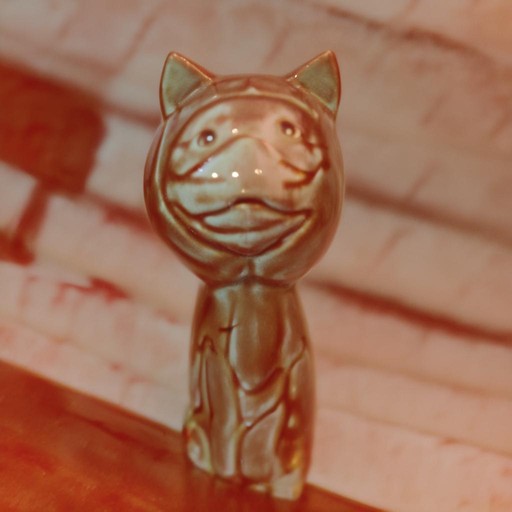} &
    \includegraphics[width=0.15\textwidth]{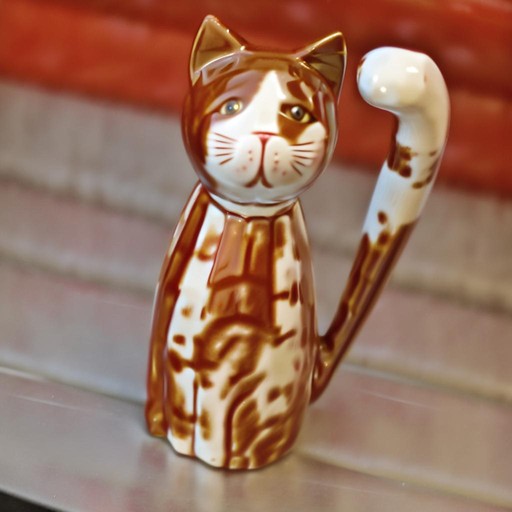} &
    \includegraphics[width=0.15\textwidth]{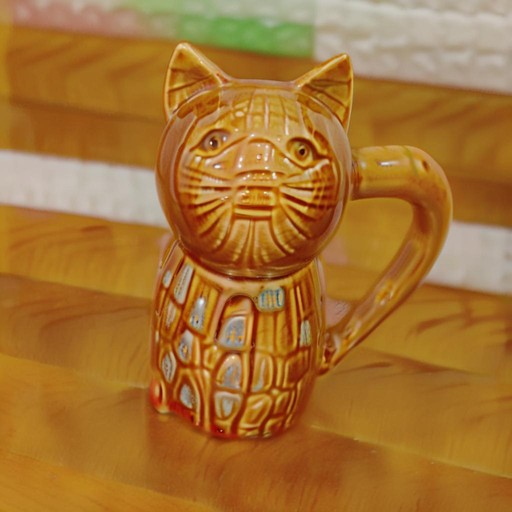} &
    \includegraphics[width=0.15\textwidth]{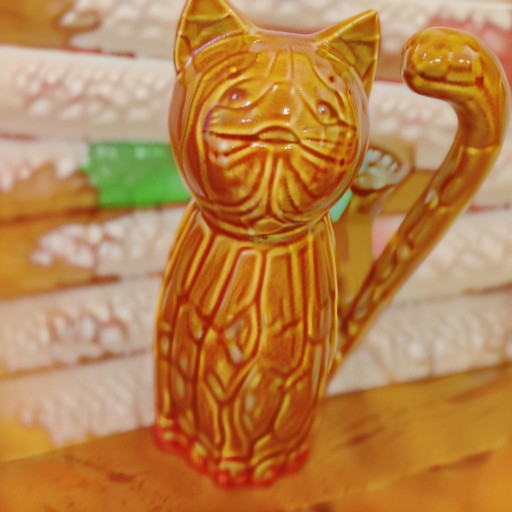} \\

    \includegraphics[width=0.15\textwidth]{temp_figs/content_images/dog2.jpg} &
    \includegraphics[width=0.15\textwidth]{temp_figs/content_images/wolf_plushie.jpg} &
    
    \includegraphics[width=0.15\textwidth]{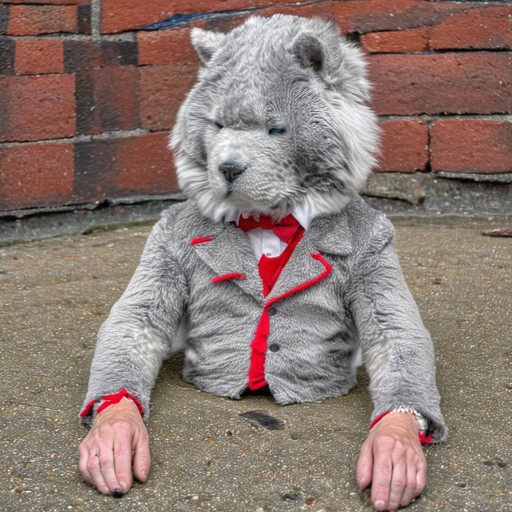} &
    \includegraphics[width=0.15\textwidth]{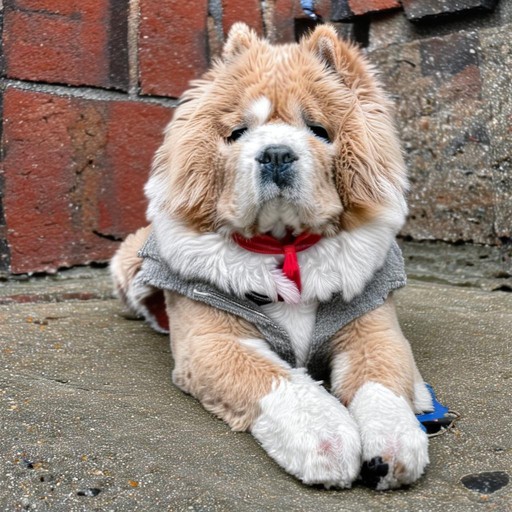} &
    \includegraphics[width=0.15\textwidth]{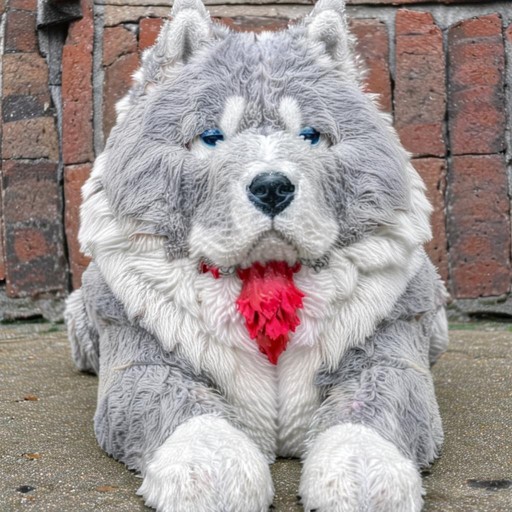} &
    \includegraphics[width=0.15\textwidth]{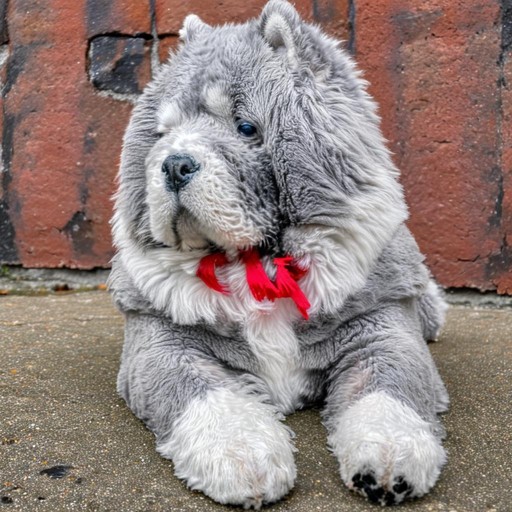} \\

    \includegraphics[width=0.15\textwidth]{temp_figs/content_images/scary_mug.jpg} &
    \includegraphics[width=0.15\textwidth]{temp_figs/content_images/dog2.jpg} &
    
    \includegraphics[width=0.15\textwidth]{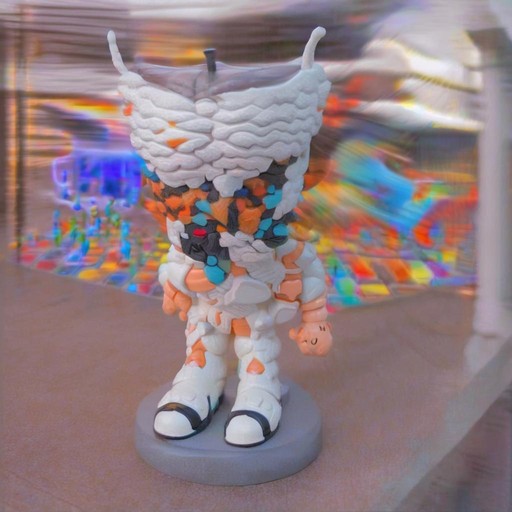} &
    \includegraphics[width=0.15\textwidth]{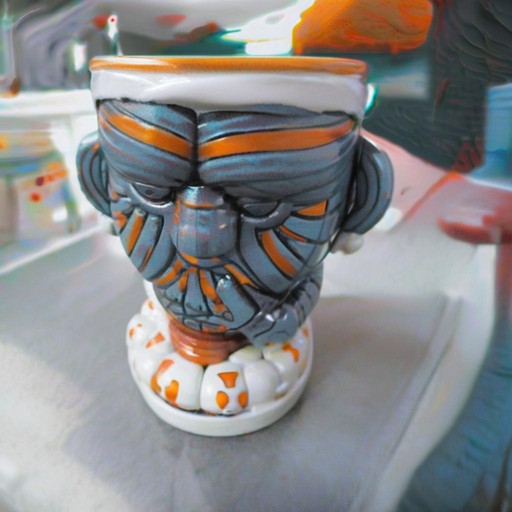} &
    \includegraphics[width=0.15\textwidth]{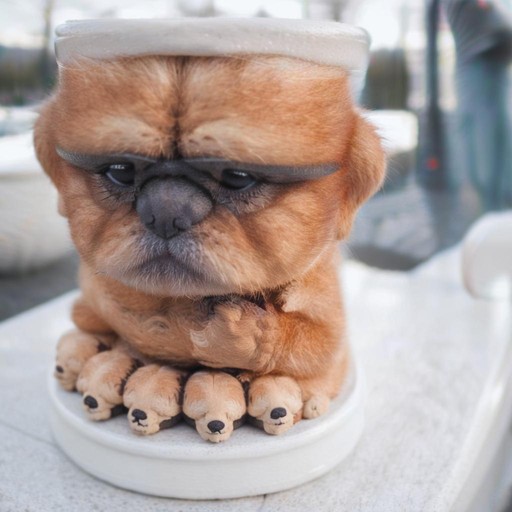} &
    \includegraphics[width=0.15\textwidth]{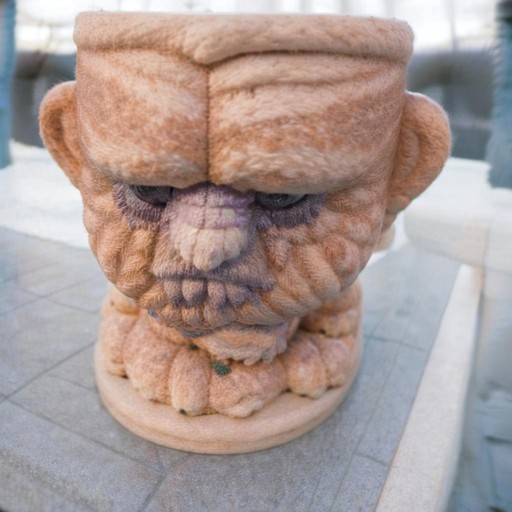} \\

    \includegraphics[width=0.15\textwidth]{temp_figs/content_images/wolf_plushie.jpg} &
    \includegraphics[width=0.15\textwidth]{temp_figs/content_images/cat.jpeg} &
    
    \includegraphics[width=0.15\textwidth]{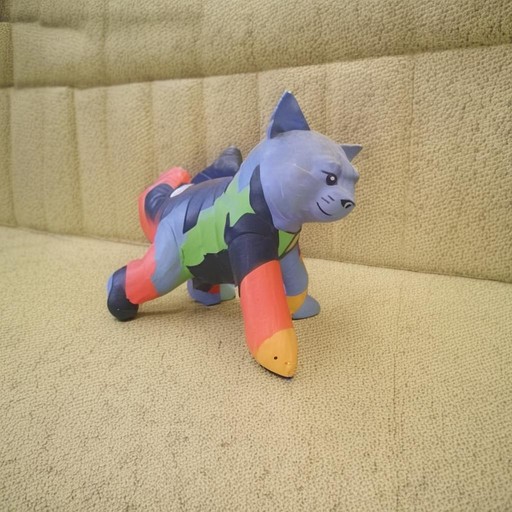} &
    \includegraphics[width=0.15\textwidth]{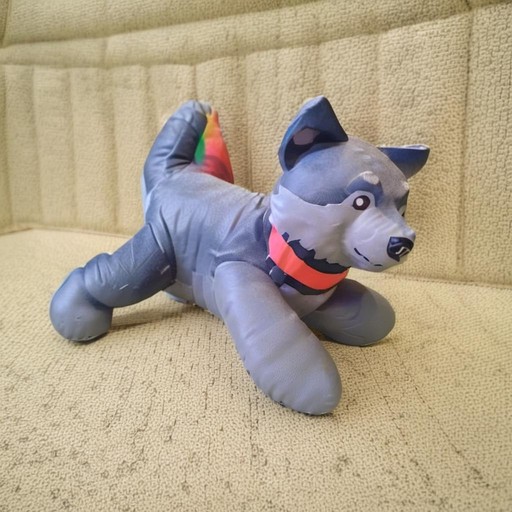} &
    \includegraphics[width=0.15\textwidth]{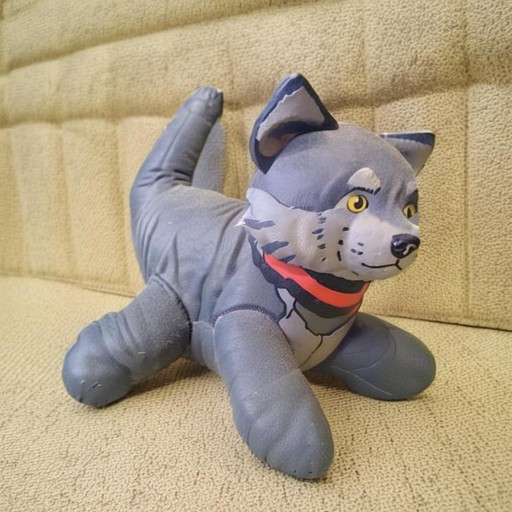} &
    \includegraphics[width=0.15\textwidth]{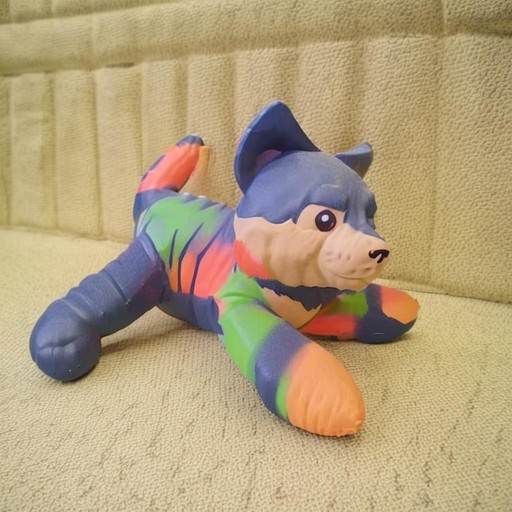} \\
    
    \end{tabular}
    }   
    \vspace{-0.11cm}
    \caption{Ablation study on the impact of different prompts for style transfer between objects. The first three columns use the prompts: (1) \ap{A [c1] in [c2] style}, (2) \ap{A [c1] {\small\textless}obj1{\small\textgreater} in [c2] style}, (3) \ap{A [c1] {\small\textless}obj1{\small\textgreater} in [c2] {\small\textless}obj2{\small\textgreater} style}, respectively. The fourth column shows our method using the prompt \ap{A [v]} without class names during optimization of $\Delta W^4$ and $\Delta W^5$. Our approach in the fourth column better preserves the content object's structure while effectively transferring the style from the other object.}
    \vspace{-0.4cm}
    \label{fig:prompt_ablation}
\end{figure*}

In \Cref{fig:prompt_ablation}, we demonstrate the impact of different prompts on style transfer between objects by fusing $\Delta W^4_{c1}$ and $\Delta W^5_{c2}$ to transfer the style of object1 to object2. We use four different prompts: \textbf{(1)} \ap{A [c1] in [c2] style} (our method), \textbf{(2)}  \ap{A [c1] {\small\textless}obj1{\small\textgreater} in [c2] style}, \textbf{(3)}  \ap{A [c1] {\small\textless}obj1{\small\textgreater} in [c2] {\small\textless}obj2{\small\textgreater} style}, and \textbf{(4)}  our method optimized without the class name.

As can be seen, the first column, using \ap{A [c1] in [c2] style}, fails to reconstruct the object's structure correctly. The second column, with \ap{A [c1] {\small\textless}obj1{\small\textgreater} in [c2] style}, successfully reconstructs the content but struggles to transfer the style. In the third column, using \ap{A [c1] {\small\textless}obj1{\small\textgreater} in [c2] {\small\textless}obj2{\small\textgreater} style}, the structure of the resulting image is affected by the obj2 class name.

In contrast, our method in the fourth column, optimized without the class name, is able to preserve the content image's structure and effectively transfer the style from the other object. This demonstrates the effectiveness of our approach using the prompt \ap{A [v]} during optimization.

\begin{figure*}[ht]
    \centering
    {\small
    \includegraphics[width=1\linewidth]{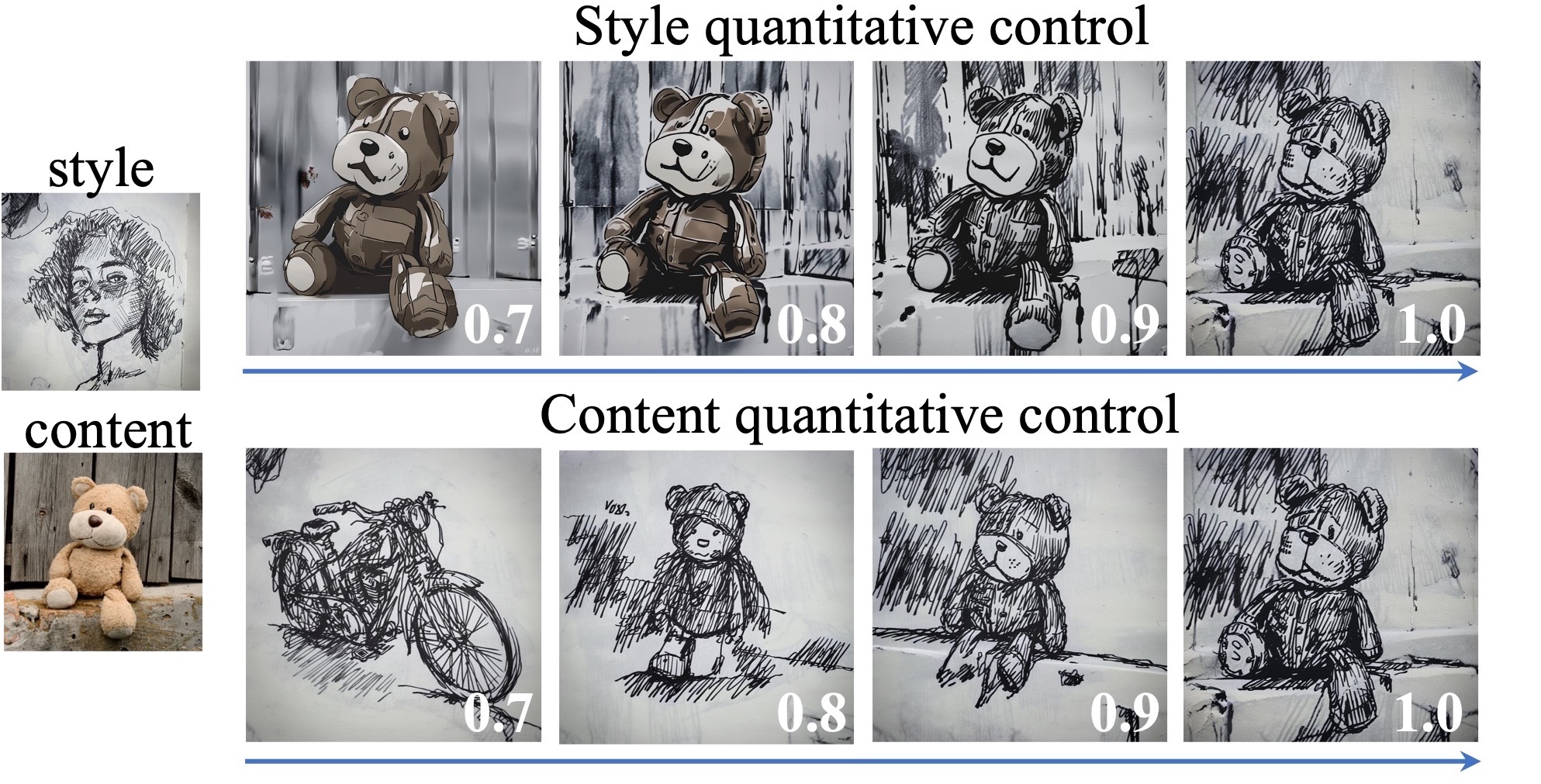}
    }   
    \caption{On the left is the style-content input pair. On the right is quantitative control over style and content by altering the $\alpha$ parameter, shown in white.}
    \label{fig:alpha_effect}
\end{figure*}

\paragraph{Alpha Effect}
As mentioned in the main paper, by the end of the training, we can obtain the tuned model weights using $W=W_0+\Delta W$, where $\Delta W$ is our trained B-LoRA update.
The strength of the fine-tuning merge equation can be adjusted and controlled by the alpha scalar: $W=W_0 +\alpha \Delta W$. (in our experiments $\alpha=1$). We demonstrate alpha's effect on style and content components in \Cref{fig:alpha_effect}. As can be seen, when the alpha value is small, both the style and the content may be lost.

\section{B-LoRA for Personalization}
\begin{figure*}[t]
    \centering
    \setlength{\tabcolsep}{1.5pt}
    {\small
    \begin{tabular}{ c c @{\hspace{0.2cm}} c c c c}
        
        Content & Style & \begin{tabular}[c]{@{}c@{}}``playing\\with a ball'' \end{tabular} & \begin{tabular}[c]{@{}c@{}}``catching \\a frisbie'' \end{tabular}  & \begin{tabular}[c]{@{}c@{}}``wearing\\a hat'' \end{tabular}  & \begin{tabular}[c]{@{}c@{}}``with a\\crown'' \end{tabular}\\
        \includegraphics[width=0.15\textwidth]{temp_figs/content_images/dog2.jpg} &
        \includegraphics[width=0.15\textwidth]{temp_figs/style_images/working_cartoon.jpg} &
        
        \includegraphics[width=0.15\textwidth]{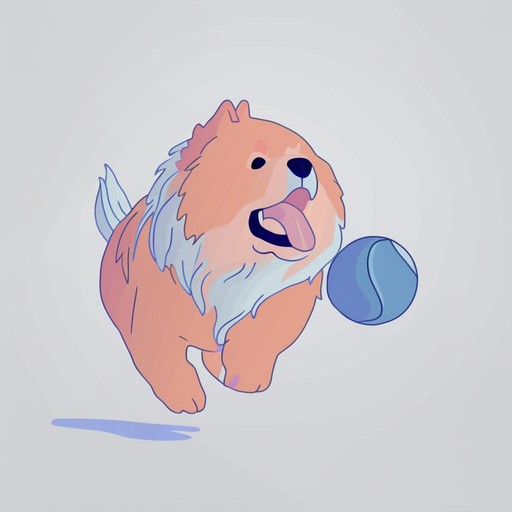} &
        \includegraphics[width=0.15\textwidth]{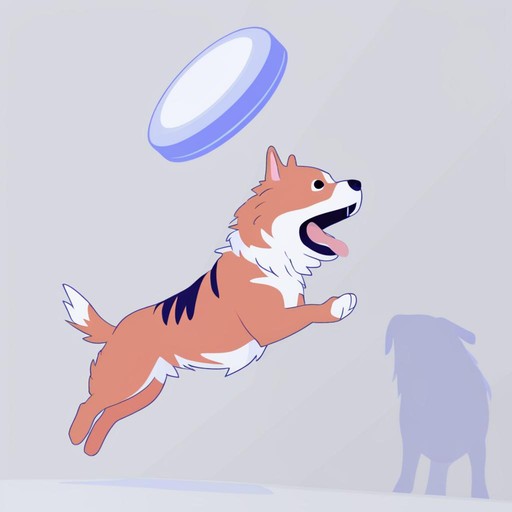} &
        \includegraphics[width=0.15\textwidth]{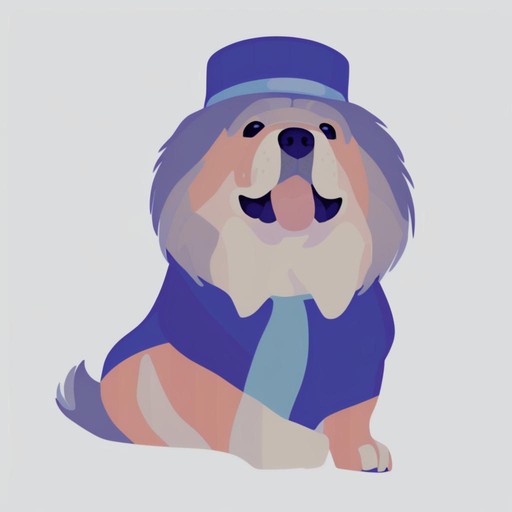} &
        \includegraphics[width=0.15\textwidth]{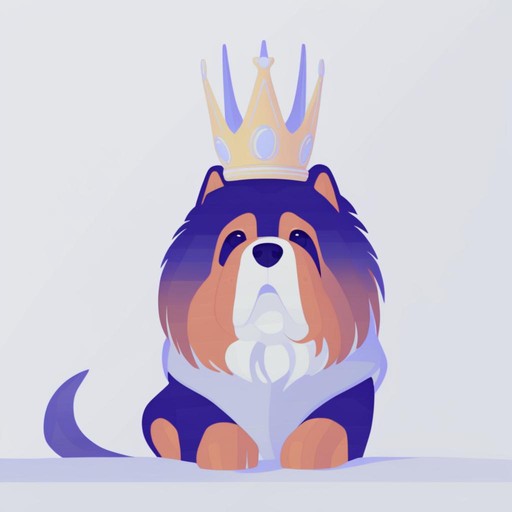} \\

        \includegraphics[width=0.15\textwidth]{temp_figs/content_images/dog2.jpg} &
        \includegraphics[width=0.15\textwidth]{temp_figs/content_images/wolf_plushie.jpg} &
        
        \includegraphics[width=0.15\textwidth]{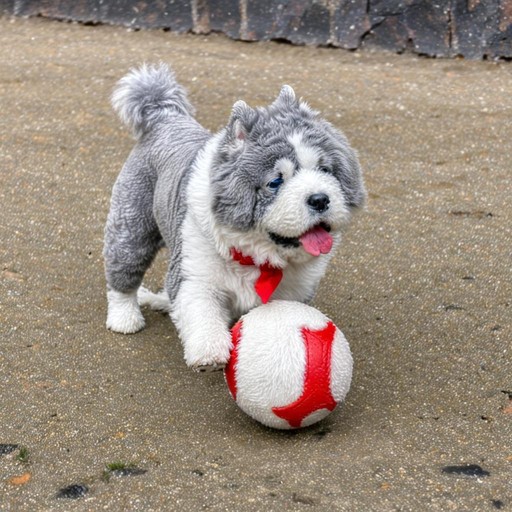} &
        \includegraphics[width=0.15\textwidth]{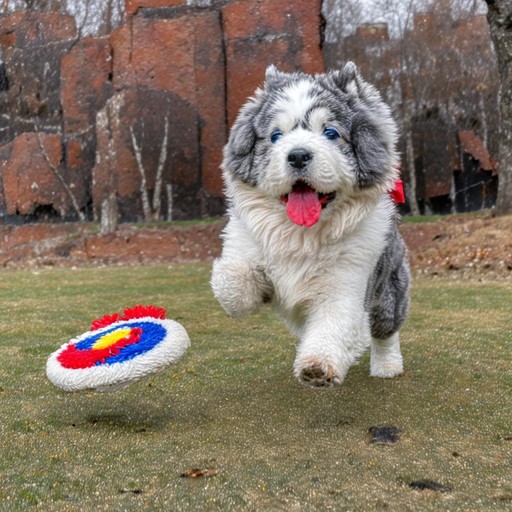} &
        \includegraphics[width=0.15\textwidth]{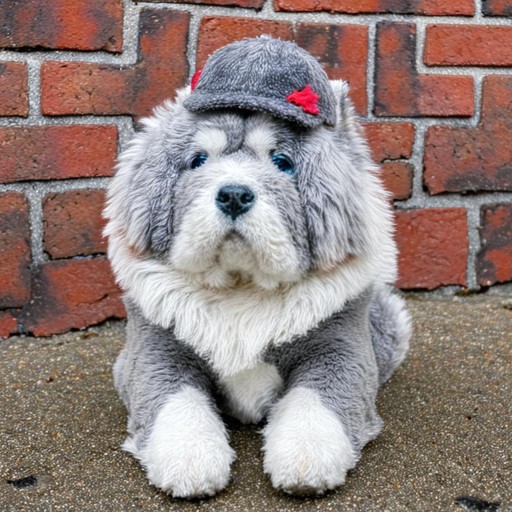} &
        \includegraphics[width=0.15\textwidth]{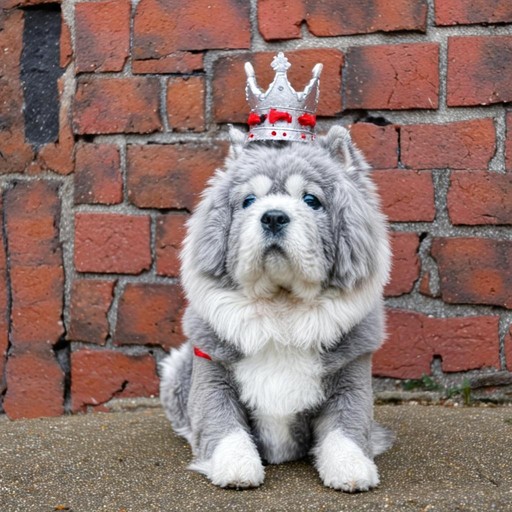} \\

        \includegraphics[width=0.15\textwidth]{temp_figs/content_images/cat.jpeg} &
        \includegraphics[width=0.15\textwidth]{temp_figs/style_images/cartoon_line.png} &
        
        \includegraphics[width=0.15\textwidth]{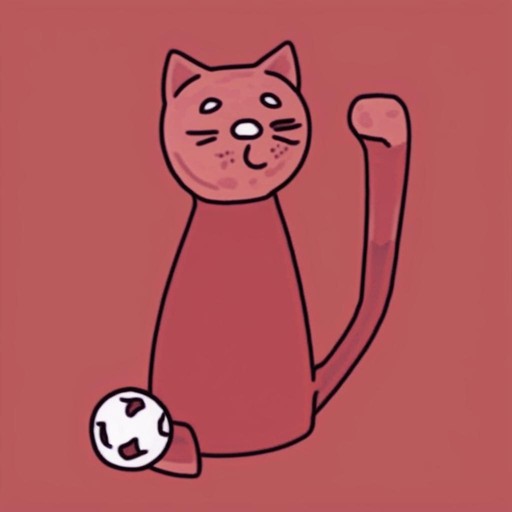} &
        \includegraphics[width=0.15\textwidth]{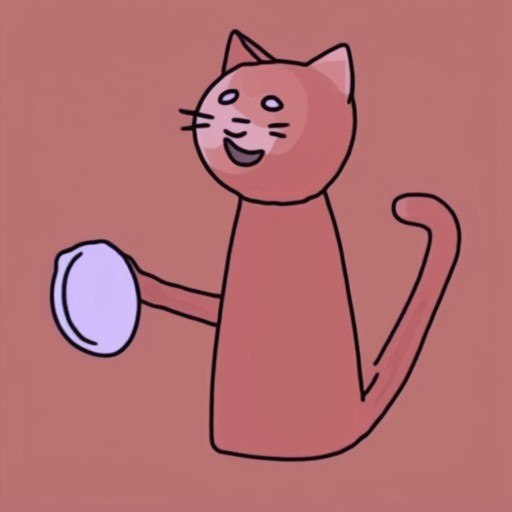} &
        \includegraphics[width=0.15\textwidth]{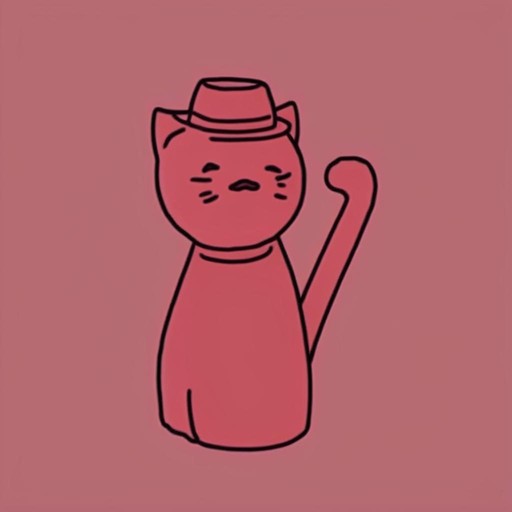} &
        \includegraphics[width=0.15\textwidth]{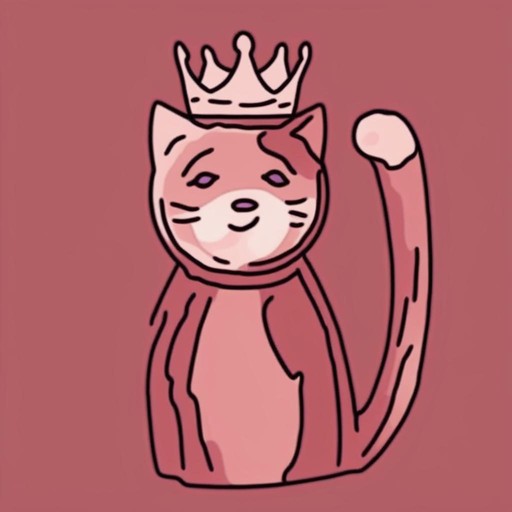} \\

        \includegraphics[width=0.15\textwidth]{temp_figs/content_images/child.jpg} &
        \includegraphics[width=0.15\textwidth]{temp_figs/style_images/watercolor.png} &
        
        \includegraphics[width=0.15\textwidth]{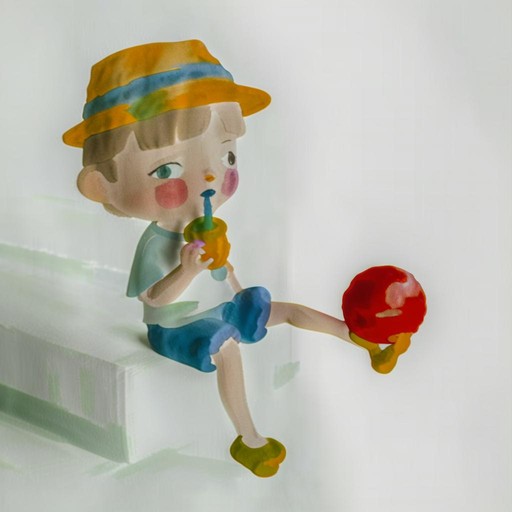} &
        \includegraphics[width=0.15\textwidth]{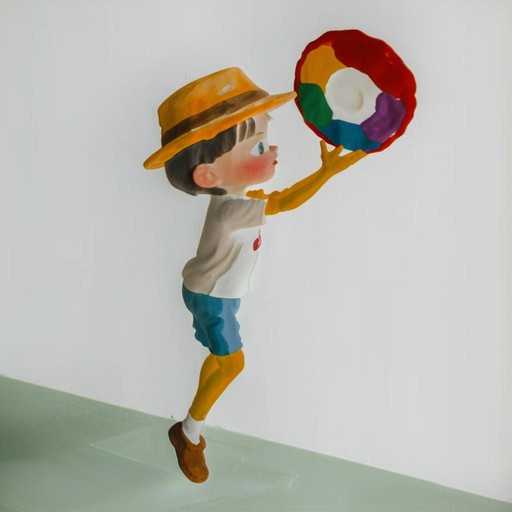} &
        \includegraphics[width=0.15\textwidth]{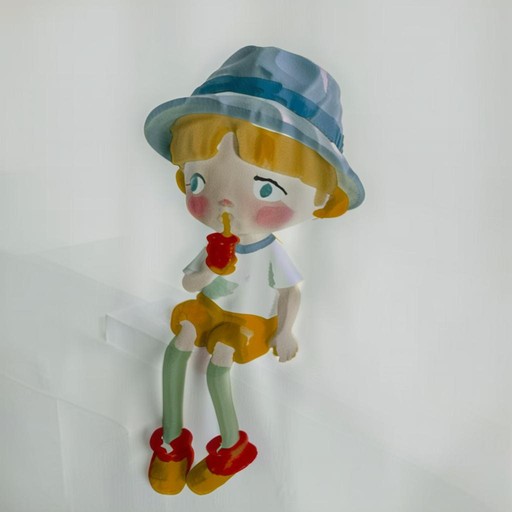} &
        \includegraphics[width=0.15\textwidth]{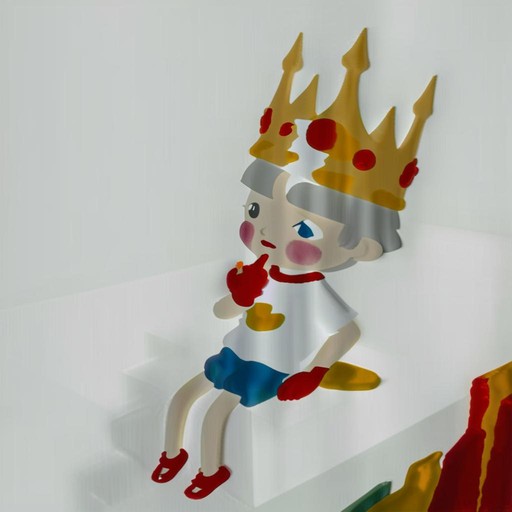} \\

        \includegraphics[width=0.15\textwidth]{temp_figs/content_images/bull.jpg} &
        \includegraphics[width=0.15\textwidth]{temp_figs/style_images/drawing3.png} &
        
        \includegraphics[width=0.15\textwidth]{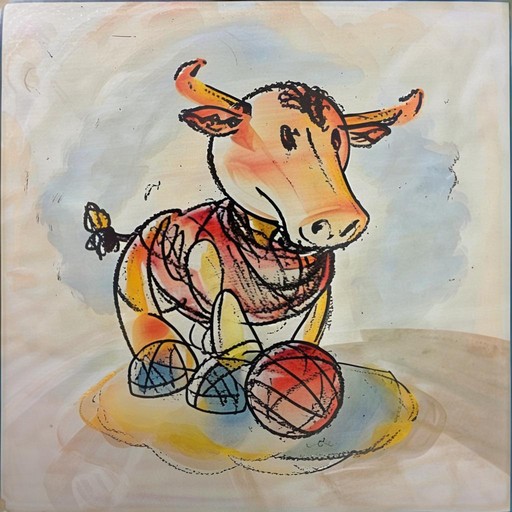} &
        \includegraphics[width=0.15\textwidth]{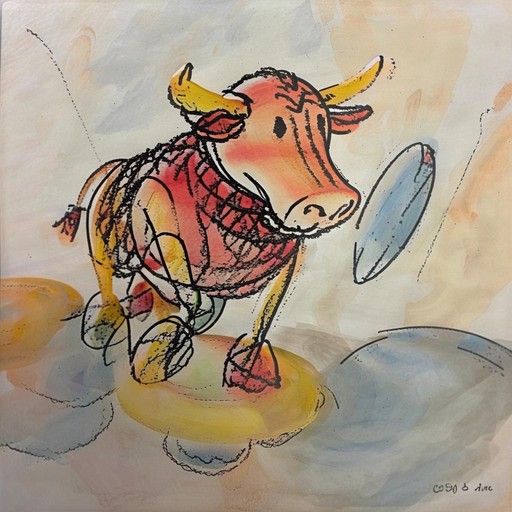} &
        \includegraphics[width=0.15\textwidth]{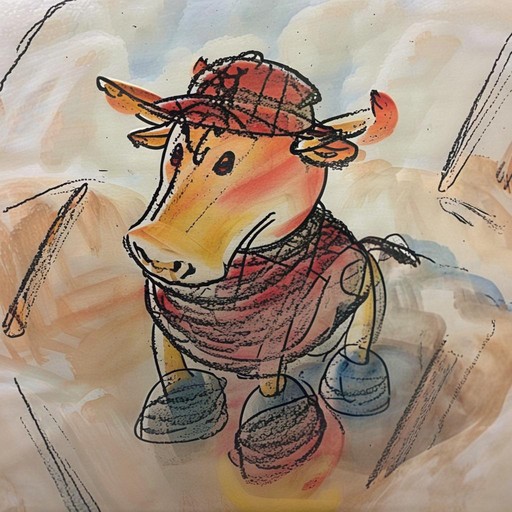} &
        \includegraphics[width=0.15\textwidth]{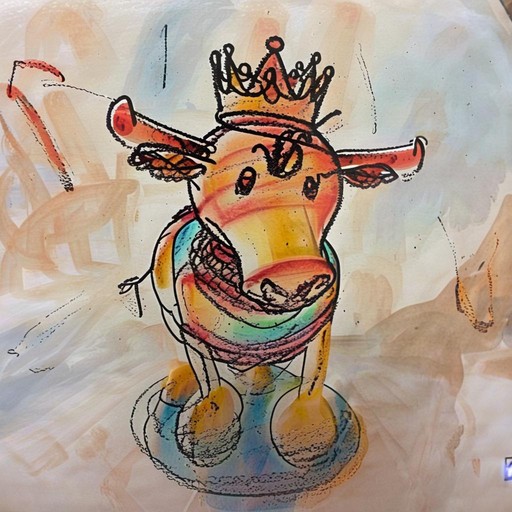} \\

        \includegraphics[width=0.15\textwidth]{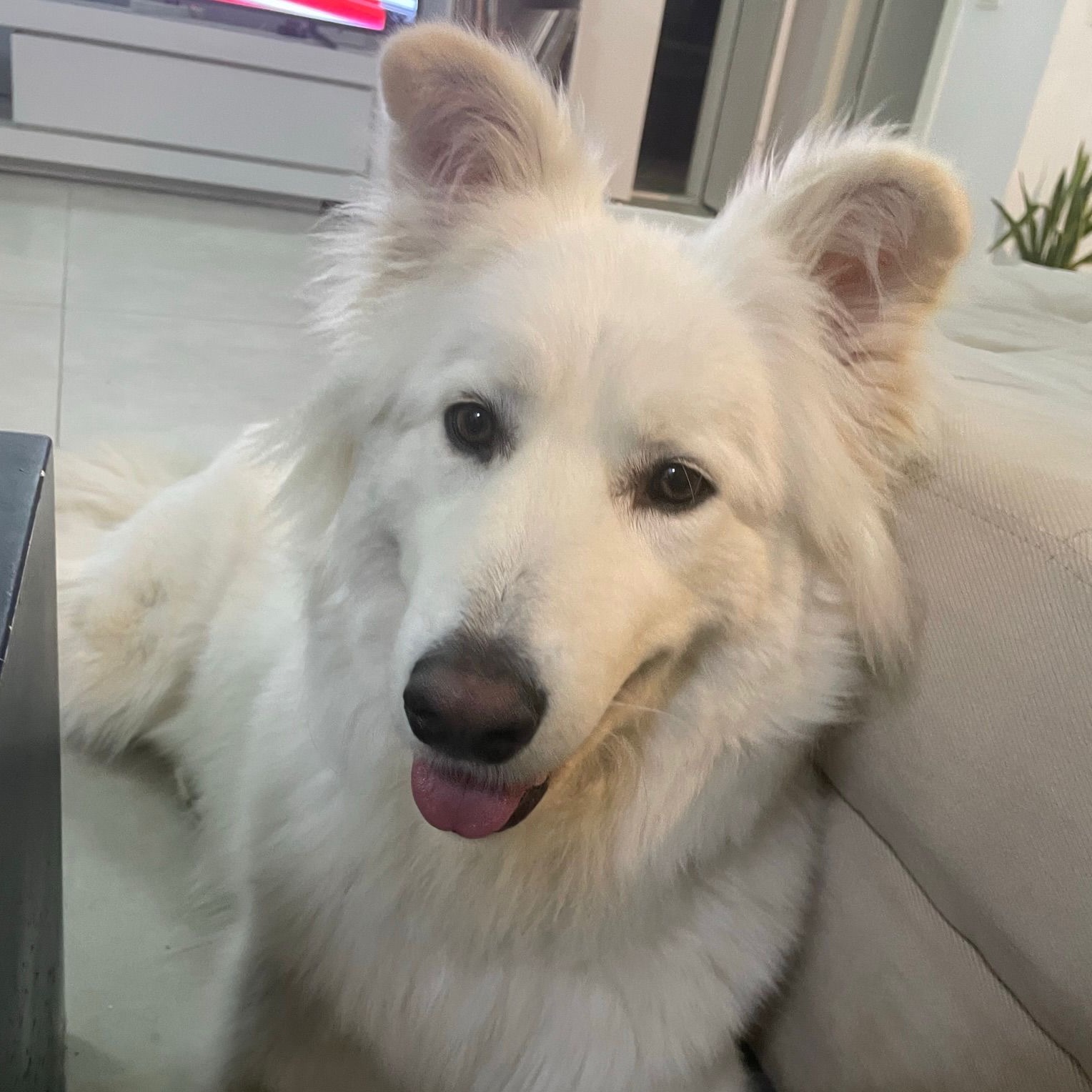} &
        \includegraphics[width=0.15\textwidth]{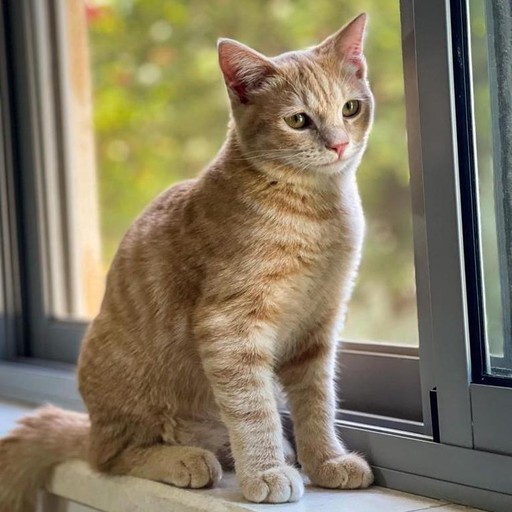} &
        \includegraphics[width=0.15\textwidth]{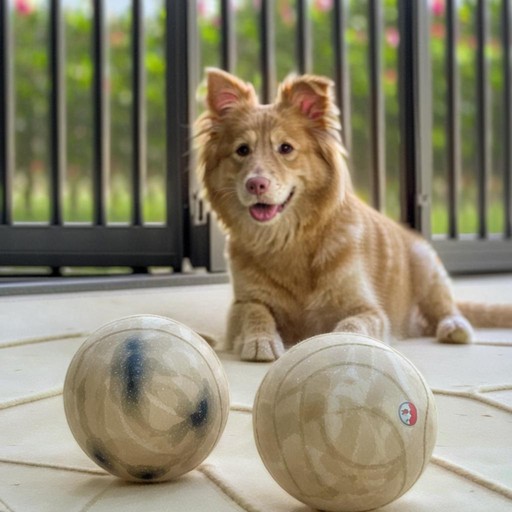} &
        \includegraphics[width=0.15\textwidth]{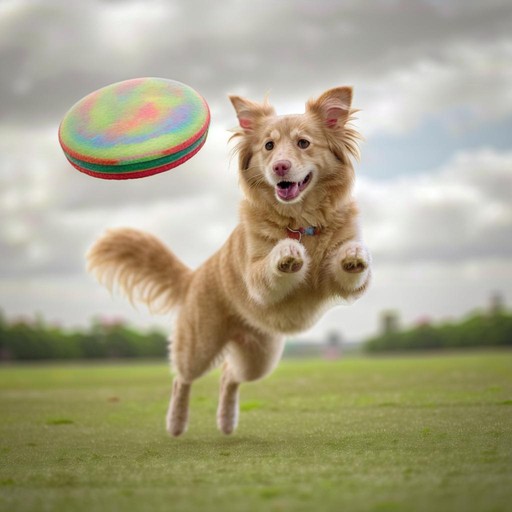} &
        \includegraphics[width=0.15\textwidth]{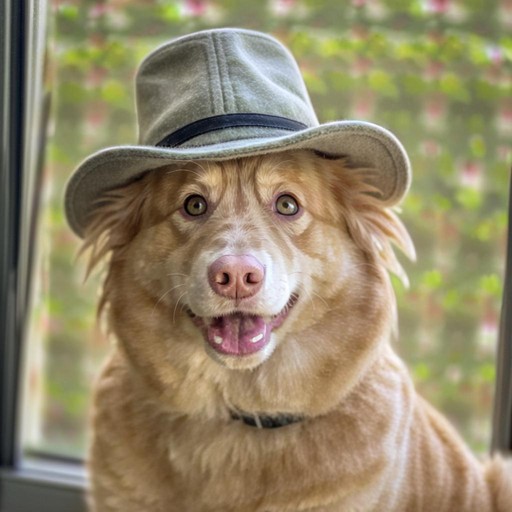} &
        \includegraphics[width=0.15\textwidth]{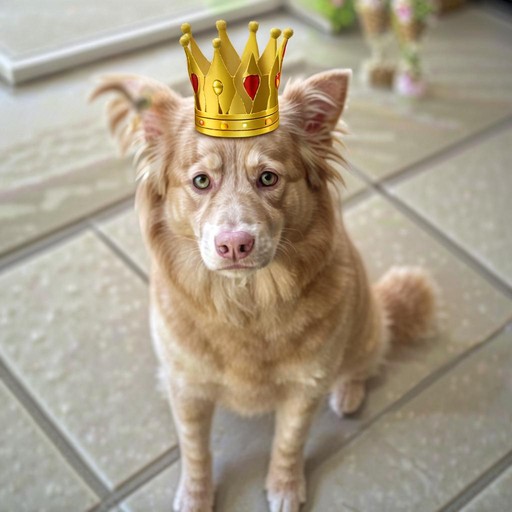}\\
    \end{tabular}
    
    }   
    \caption{While maintaining the stylistic characteristics of the style, our method effectively re-contextualizes the content object. Note that our approach is capable of transferring the style from either a style or object reference image.}
    \label{fig:personalization1}
\end{figure*}

\begin{figure*}[t]
    \centering
    \setlength{\tabcolsep}{1.5pt}
    {\small
    \begin{tabular}{ c c @{\hspace{0.2cm}} c c c c}
        
        Content & Style & \begin{tabular}[c]{@{}c@{}}``riding a\\bicycle'' \end{tabular}& \ap{sleeping} & \begin{tabular}[c]{@{}c@{}}``in a\\boat'' \end{tabular} & \begin{tabular}[c]{@{}c@{}}``driving\\a car'' \end{tabular}\\
        \includegraphics[width=0.15\textwidth]{temp_figs/content_images/dog2.jpg} &
        \includegraphics[width=0.15\textwidth]{temp_figs/style_images/working_cartoon.jpg} &
        
        \includegraphics[width=0.15\textwidth]{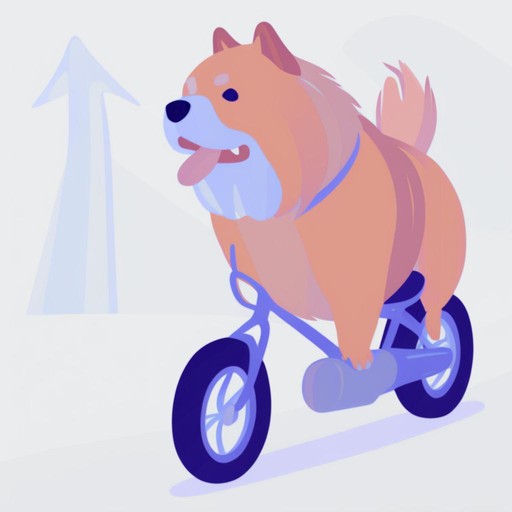} &
        \includegraphics[width=0.15\textwidth]{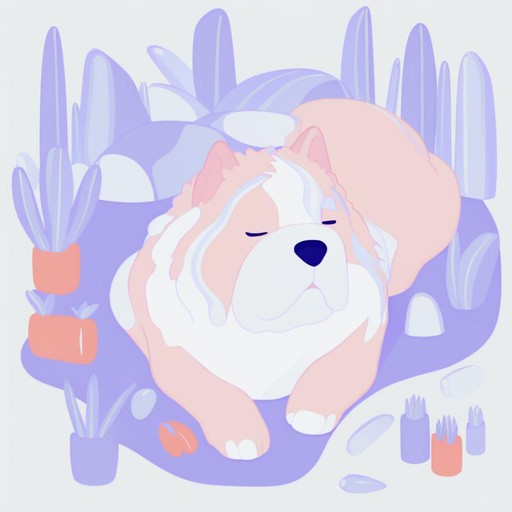} &
        \includegraphics[width=0.15\textwidth]{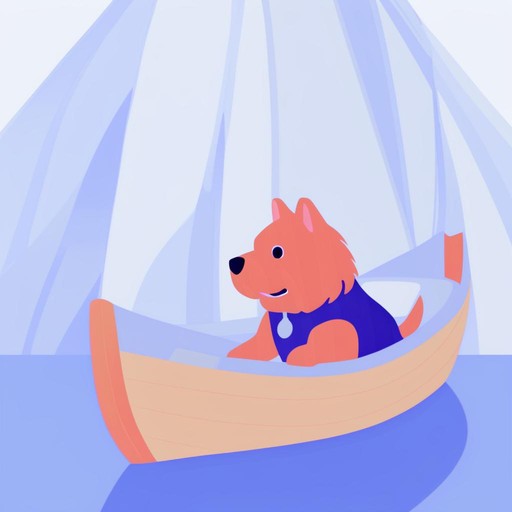} &
        \includegraphics[width=0.15\textwidth]{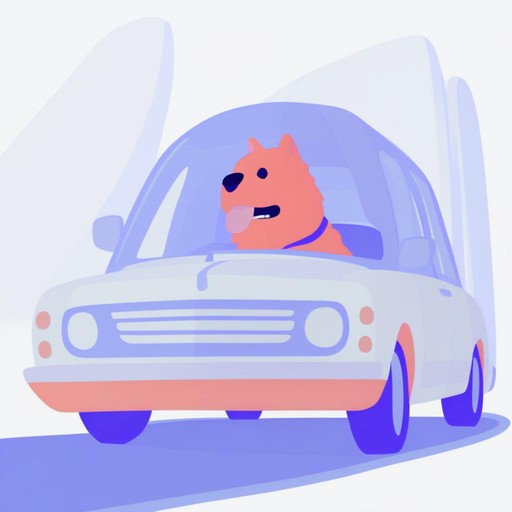} \\

        \includegraphics[width=0.15\textwidth]{temp_figs/content_images/dog2.jpg} &
        \includegraphics[width=0.15\textwidth]{temp_figs/content_images/wolf_plushie.jpg} &
        
        \includegraphics[width=0.15\textwidth]{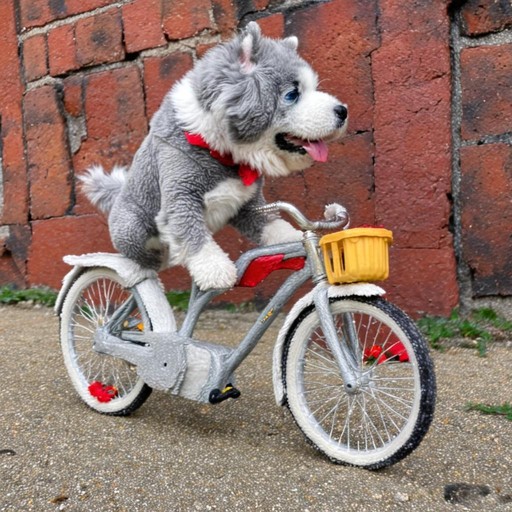} &
        \includegraphics[width=0.15\textwidth]{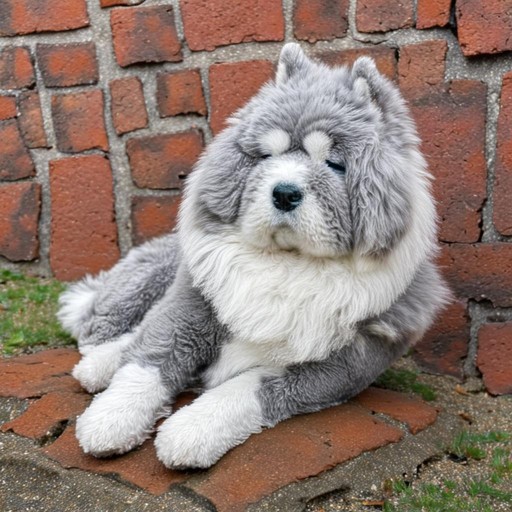} &
        \includegraphics[width=0.15\textwidth]{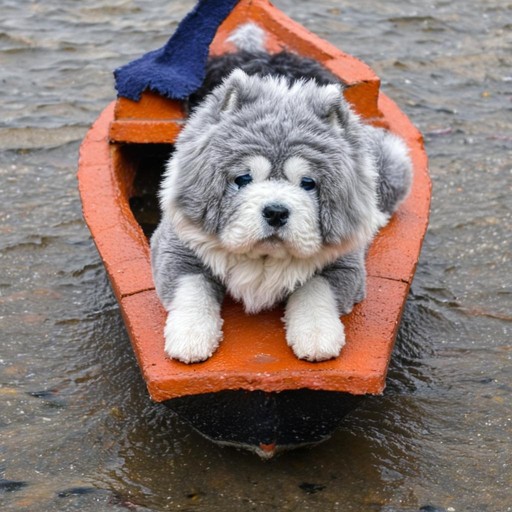} &
        \includegraphics[width=0.15\textwidth]{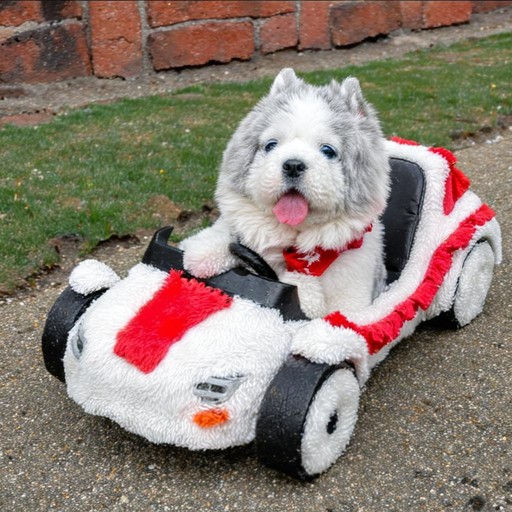} \\

        \includegraphics[width=0.15\textwidth]{temp_figs/content_images/cat.jpeg} &
        \includegraphics[width=0.15\textwidth]{temp_figs/style_images/cartoon_line.png} &
        
        \includegraphics[width=0.15\textwidth]{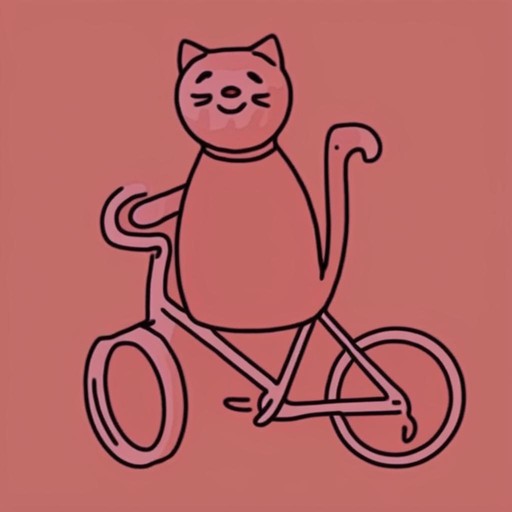} &
        \includegraphics[width=0.15\textwidth]{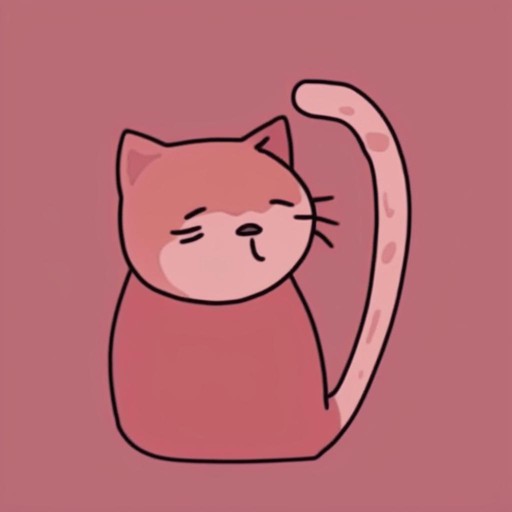} &
        \includegraphics[width=0.15\textwidth]{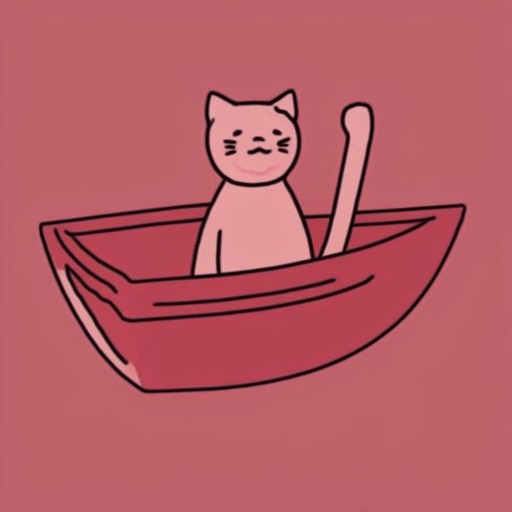} &
        \includegraphics[width=0.15\textwidth]{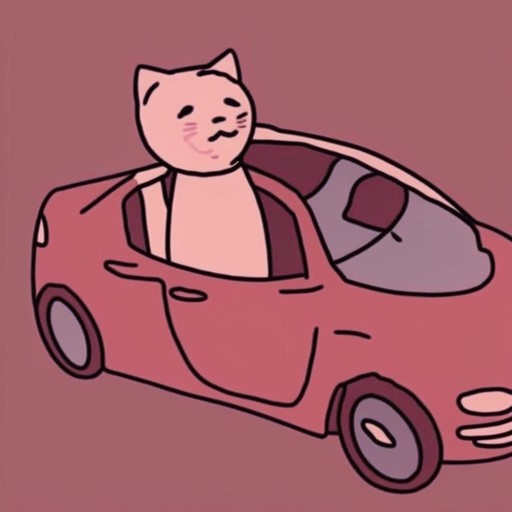} \\

        \includegraphics[width=0.15\textwidth]{temp_figs/content_images/child.jpg} &
        \includegraphics[width=0.15\textwidth]{temp_figs/style_images/watercolor.png} &
        
        \includegraphics[width=0.15\textwidth]{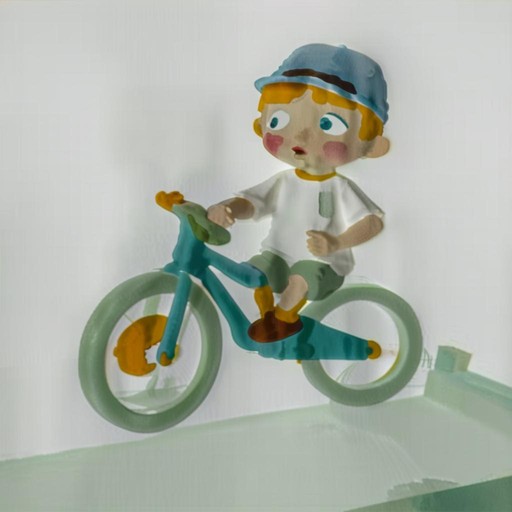} &
        \includegraphics[width=0.15\textwidth]{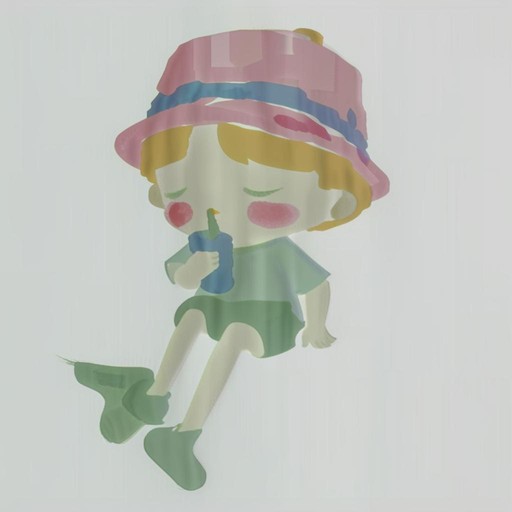} &
        \includegraphics[width=0.15\textwidth]{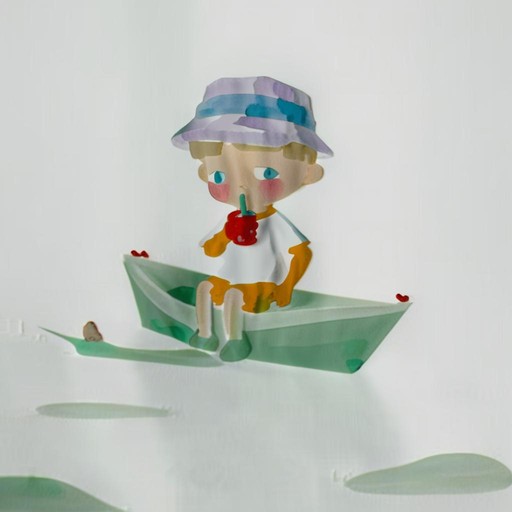} &
        \includegraphics[width=0.15\textwidth]{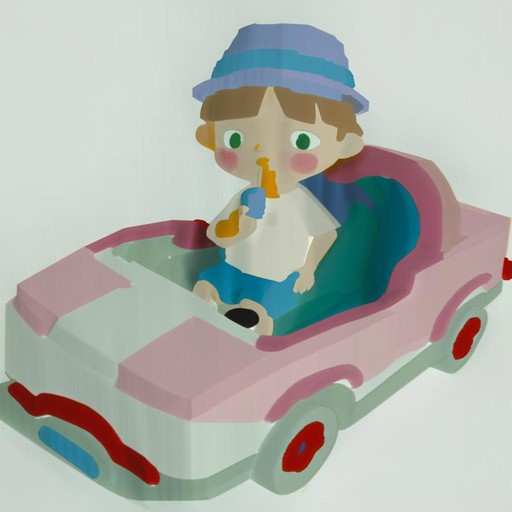} \\

        \includegraphics[width=0.15\textwidth]{temp_figs/content_images/bull.jpg} &
        \includegraphics[width=0.15\textwidth]{temp_figs/style_images/drawing3.png} &
        
        \includegraphics[width=0.15\textwidth]{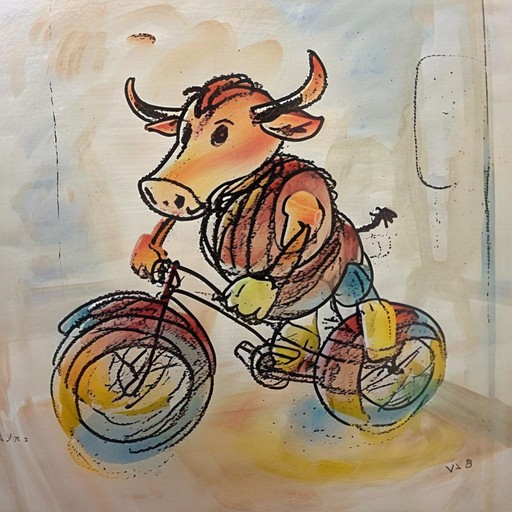} &
        \includegraphics[width=0.15\textwidth]{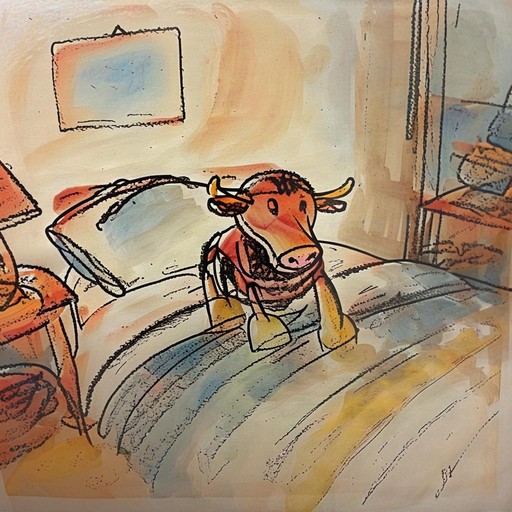} &
        \includegraphics[width=0.15\textwidth]{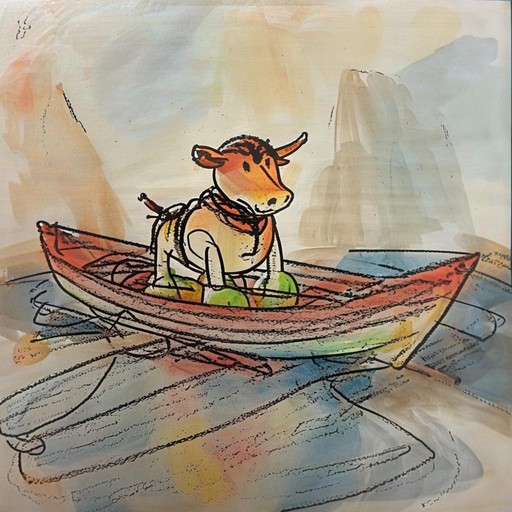} &
        \includegraphics[width=0.15\textwidth]{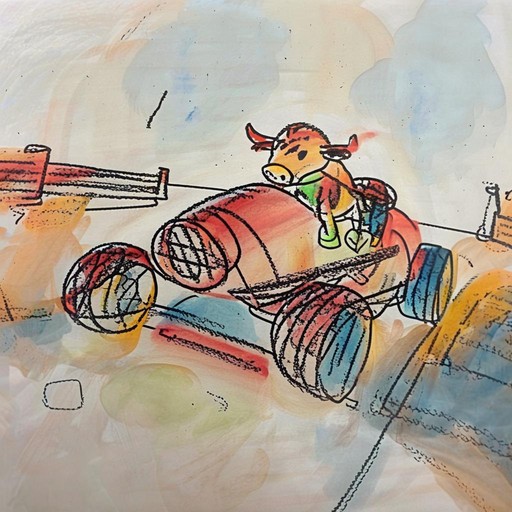} \\

        \includegraphics[width=0.15\textwidth]{temp_figs/content_images/olly.JPG} &
        \includegraphics[width=0.15\textwidth]{temp_figs/content_images/berli.jpeg} &
        
        \includegraphics[width=0.15\textwidth]{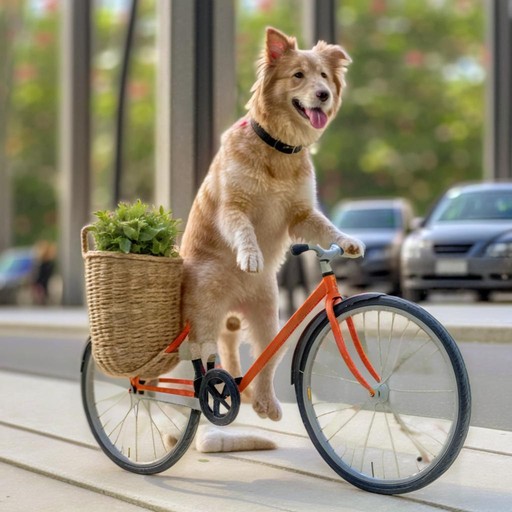} &
        \includegraphics[width=0.15\textwidth]{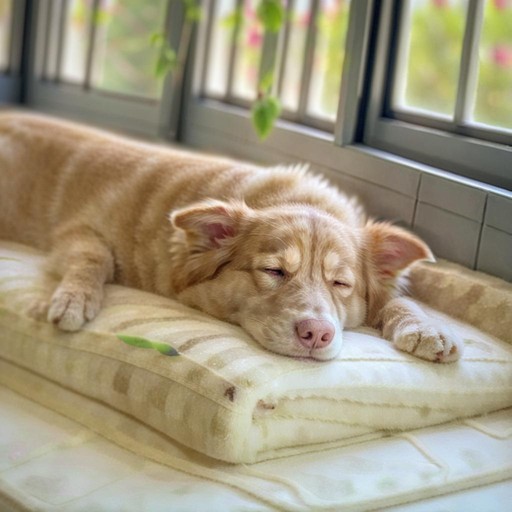} &
        \includegraphics[width=0.15\textwidth]{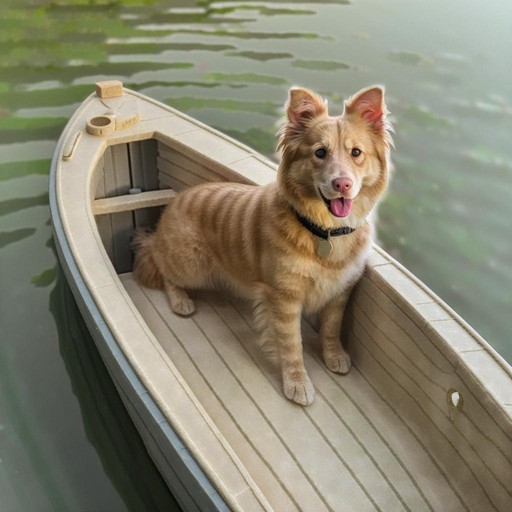} &
        \includegraphics[width=0.15\textwidth]{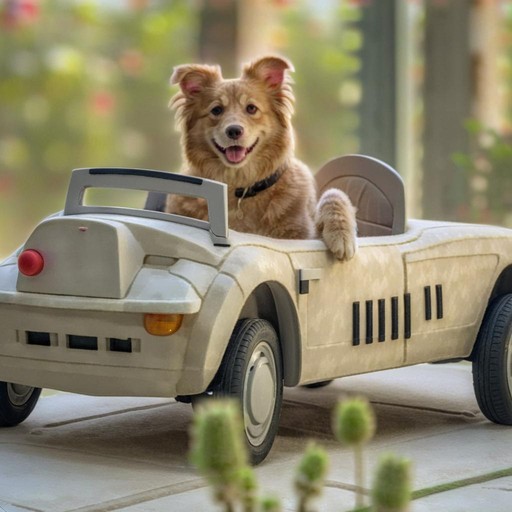} \\
    \end{tabular}
    
    }   
    \caption{While maintaining the stylistic characteristics of the style, our method effectively re-contextualizes the content object. Note that our approach is capable of transferring the style from either a style or object reference image.}
    \label{fig:personalization2}
\end{figure*}

Throughout the paper, our method has been implemented using a single image for decoupling style and content. However, by training our method using multiple images for content, we can recontextualize reference objects while preserving stylization quality. In \Cref{fig:personalization1,fig:personalization2}, we showcase the versatility of our method by combining various B-LoRAs for style and content with text prompts. Note that the style can be derived from the reference style or from other objects.

\section{Additional Results}
\begin{figure*}
    \centering
    \setlength{\tabcolsep}{1.5pt}
    {\small
    \begin{NiceTabular}{c @{\hspace{0.15cm}} c c c c c}

        \diagbox{Input \\ Content}{Style} &
        \hspace{0.11cm}
        \includegraphics[width=0.15\textwidth]{temp_figs/style_images/drawing2.jpg} &
        \includegraphics[width=0.15\textwidth]{temp_figs/style_images/drawing1.jpg} &
        \includegraphics[width=0.15\textwidth]{temp_figs/style_images/drawing3.png} &
        \includegraphics[width=0.15\textwidth]{temp_figs/style_images/kiss.png} &
        \includegraphics[width=0.15\textwidth]{temp_figs/style_images/pen_sketch.jpeg} \\

        \includegraphics[width=0.15\textwidth]{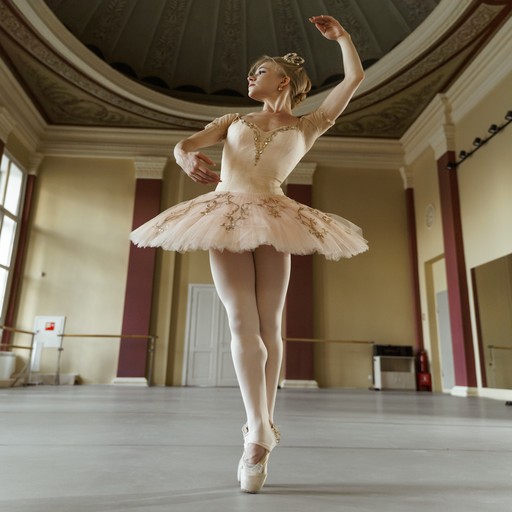} &
        \hspace{0.11cm}
        \includegraphics[width=0.15\textwidth]{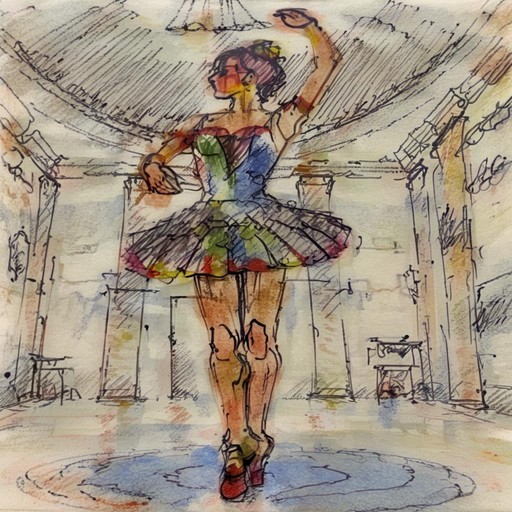} &
        \includegraphics[width=0.15\textwidth]{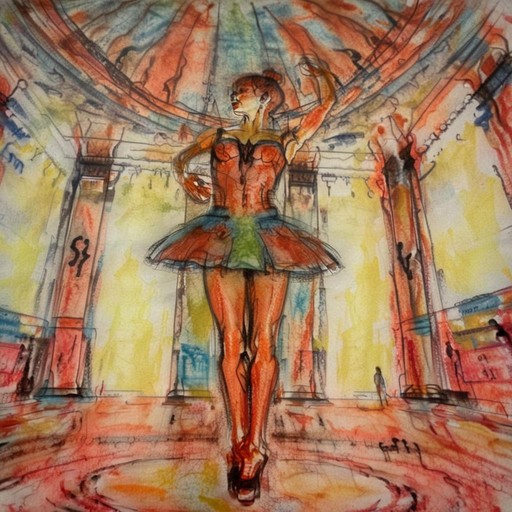} &
        \includegraphics[width=0.15\textwidth]{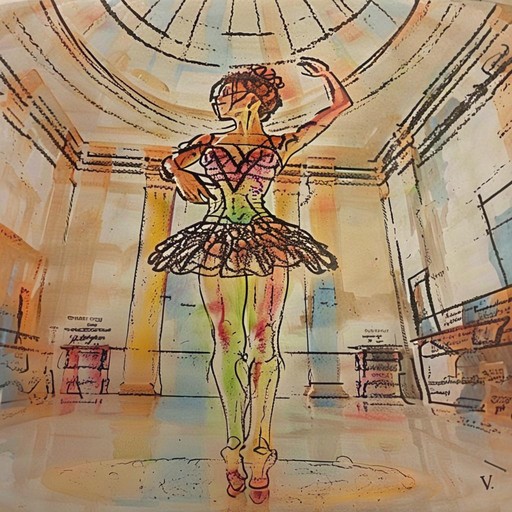} &
        \includegraphics[width=0.15\textwidth]{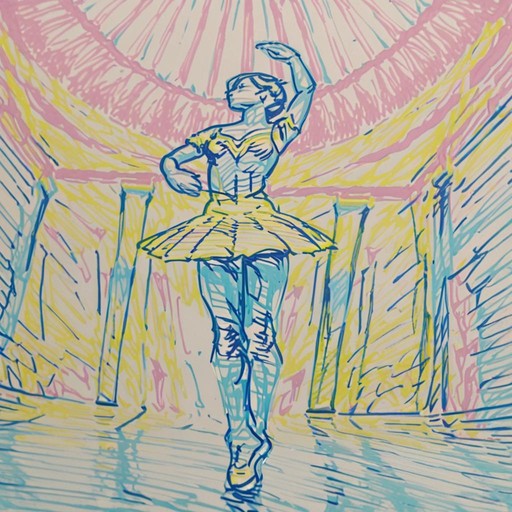} &
        \includegraphics[width=0.15\textwidth]{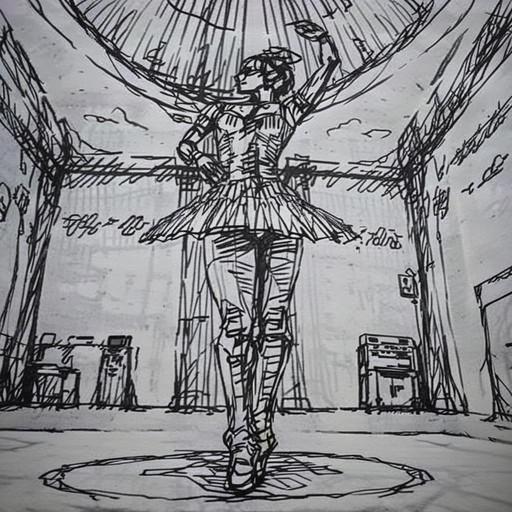} \\

        \includegraphics[width=0.15\textwidth]{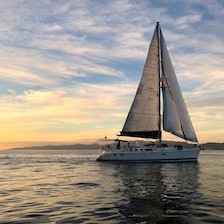} &
        \hspace{0.11cm}
        \includegraphics[width=0.15\textwidth]{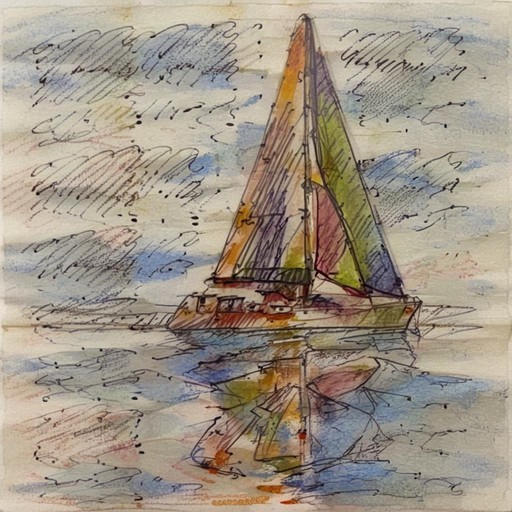} &
        \includegraphics[width=0.15\textwidth]{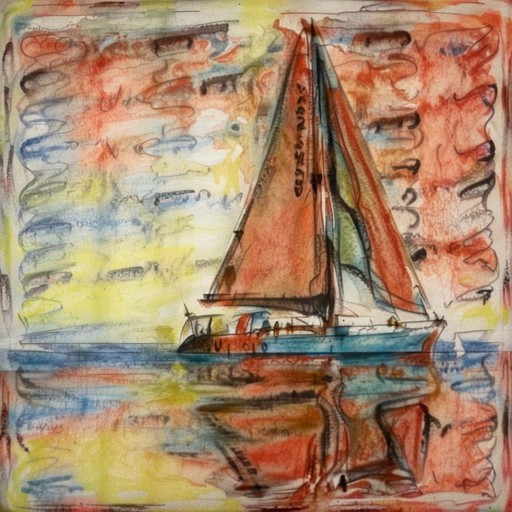} &
        \includegraphics[width=0.15\textwidth]{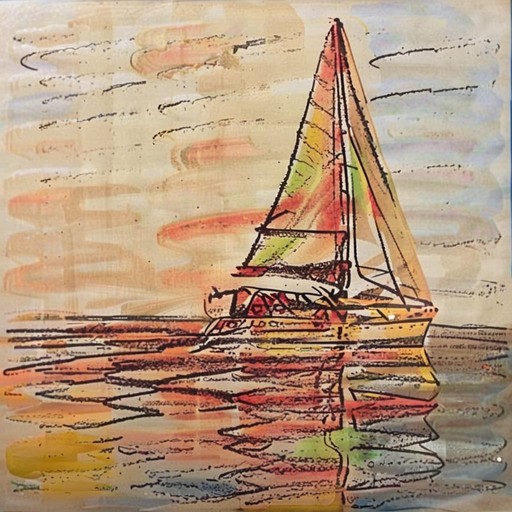} &
        \includegraphics[width=0.15\textwidth]{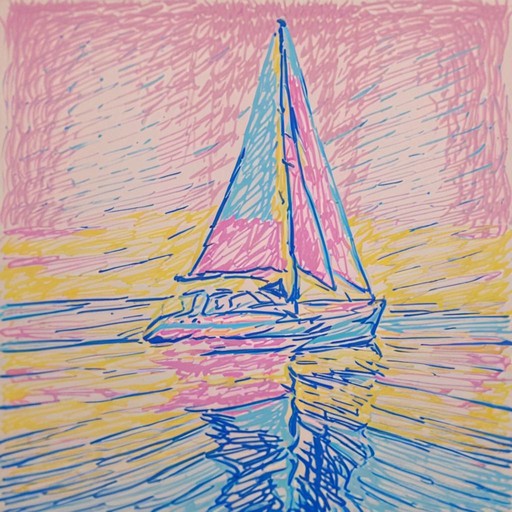} &
        \includegraphics[width=0.15\textwidth]{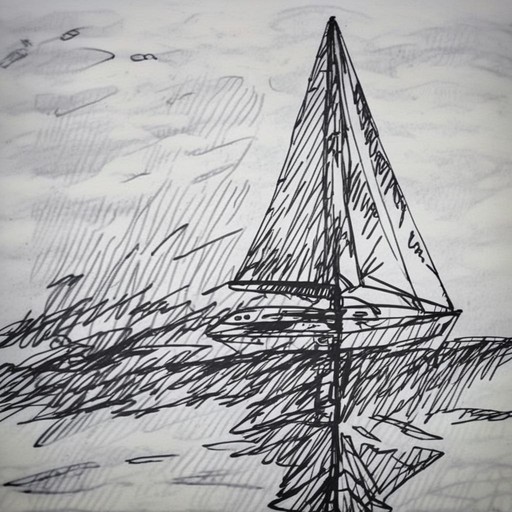} \\

        \includegraphics[width=0.15\textwidth]{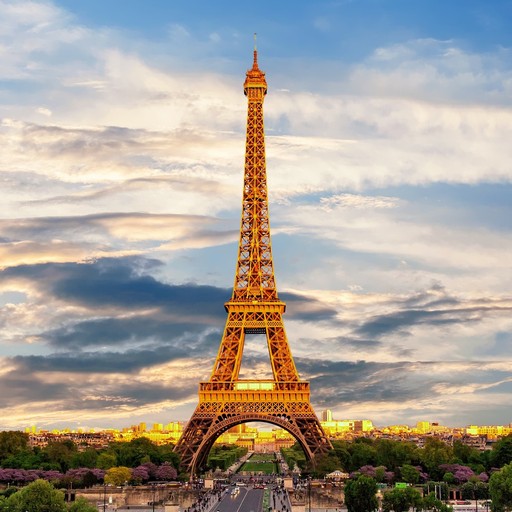} &
        \hspace{0.11cm}
        \includegraphics[width=0.15\textwidth]{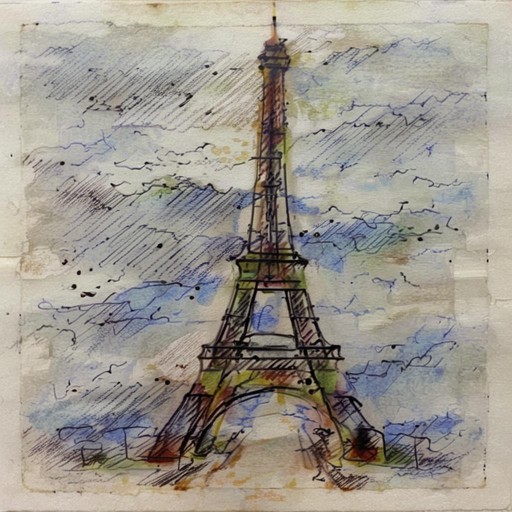} &
        \includegraphics[width=0.15\textwidth]{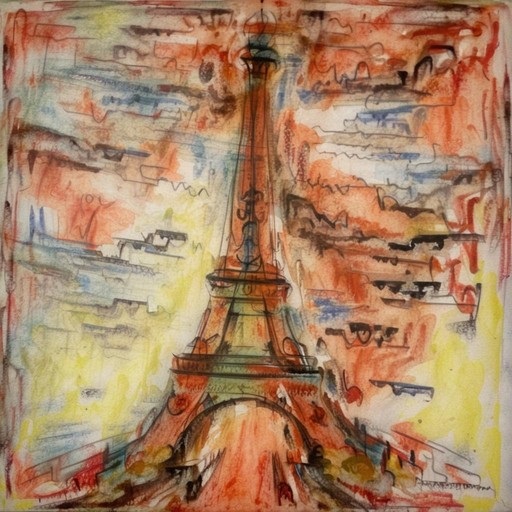} &
        \includegraphics[width=0.15\textwidth]{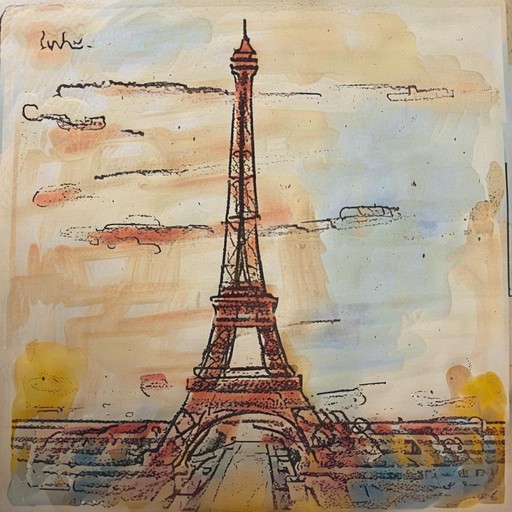} &
        \includegraphics[width=0.15\textwidth]{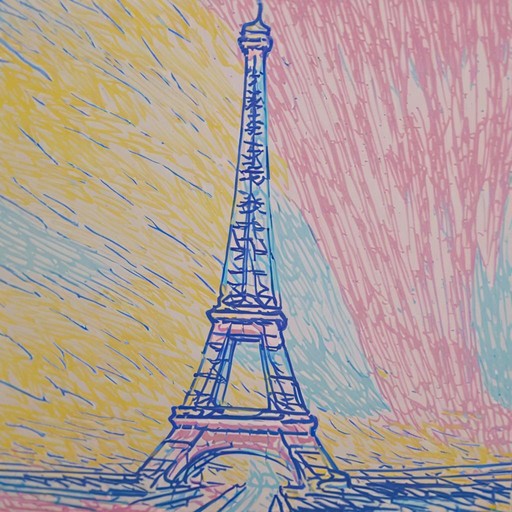} &
        \includegraphics[width=0.15\textwidth]{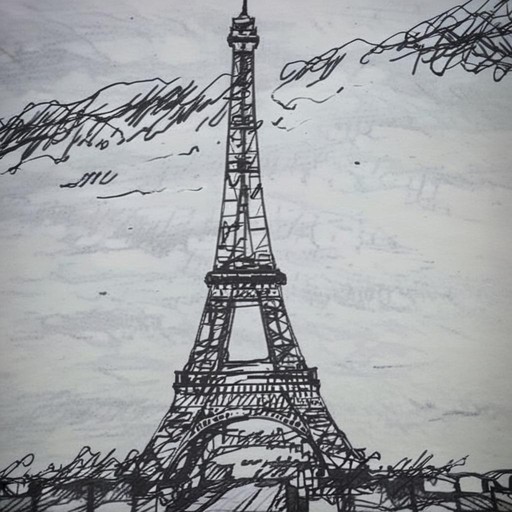} \\

         \includegraphics[width=0.15\textwidth]{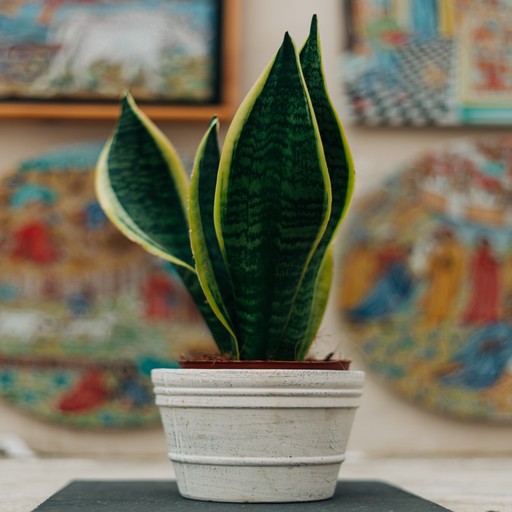} &
        \hspace{0.11cm}
        \includegraphics[width=0.15\textwidth]{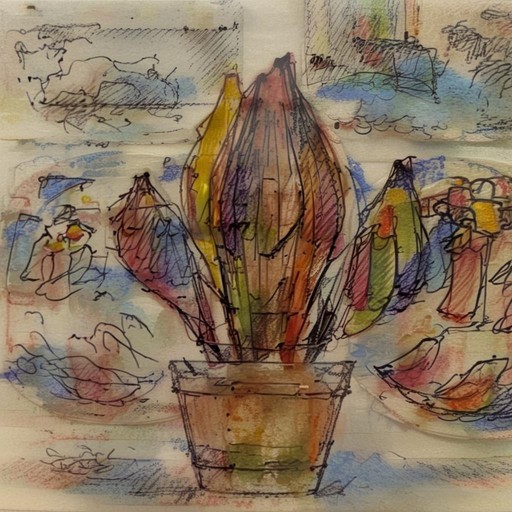} &
        \includegraphics[width=0.15\textwidth]{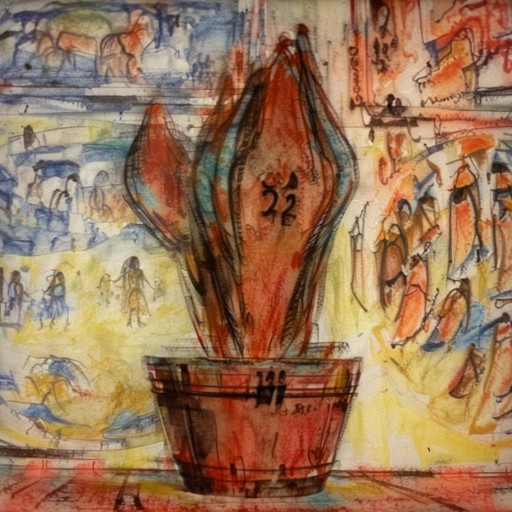} &
        \includegraphics[width=0.15\textwidth]{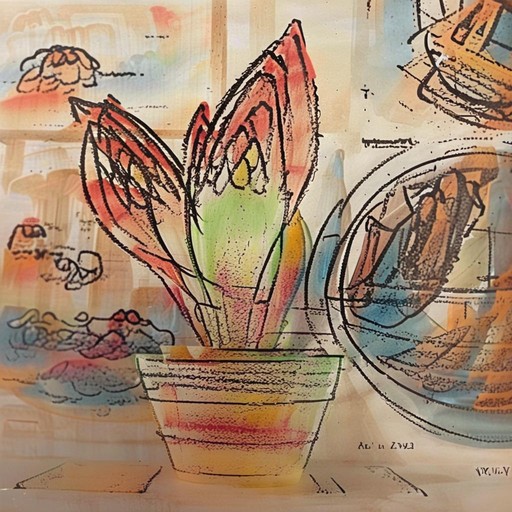} &
        \includegraphics[width=0.15\textwidth]{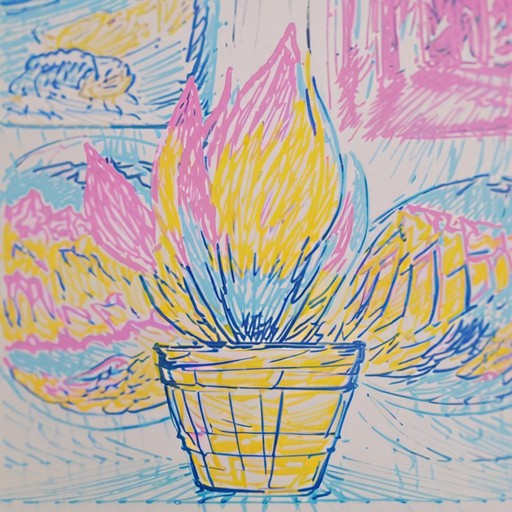} &
        \includegraphics[width=0.15\textwidth]{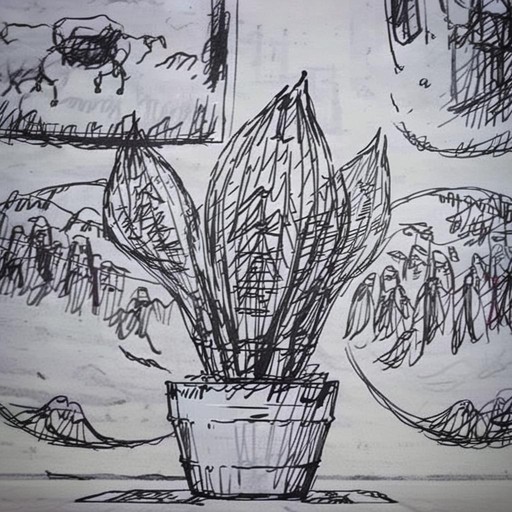} \\

        \includegraphics[width=0.15\textwidth]{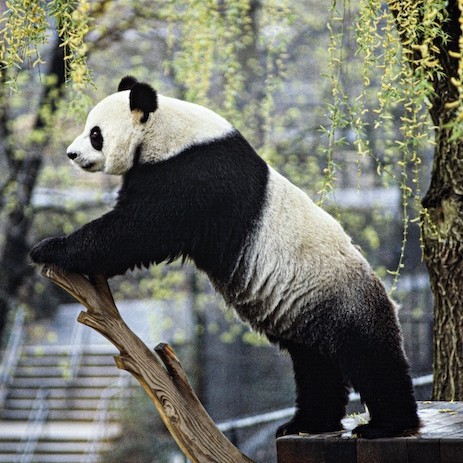} &
        \hspace{0.11cm}
        \includegraphics[width=0.15\textwidth]{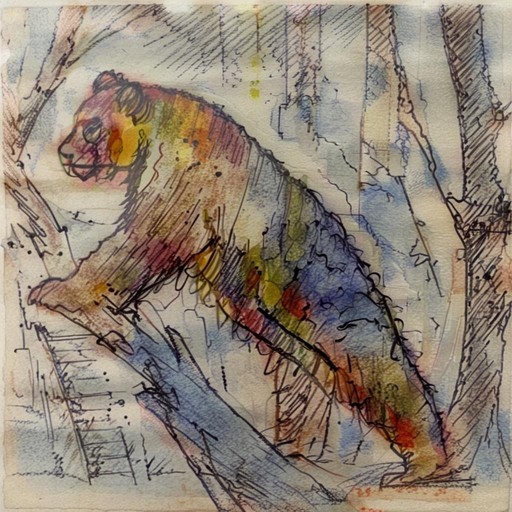} &
        \includegraphics[width=0.15\textwidth]{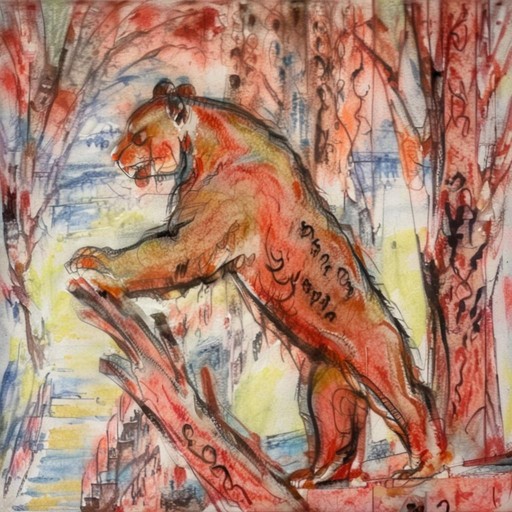} &
        \includegraphics[width=0.15\textwidth]{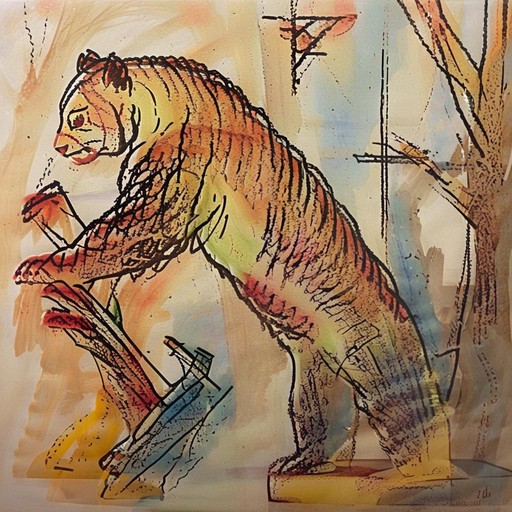} &
        \includegraphics[width=0.15\textwidth]{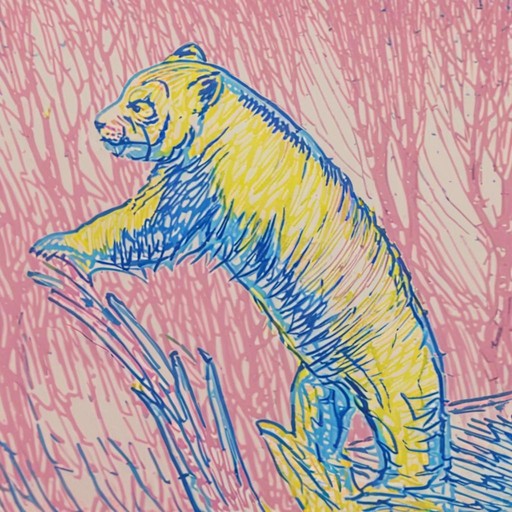} &
        \includegraphics[width=0.15\textwidth]{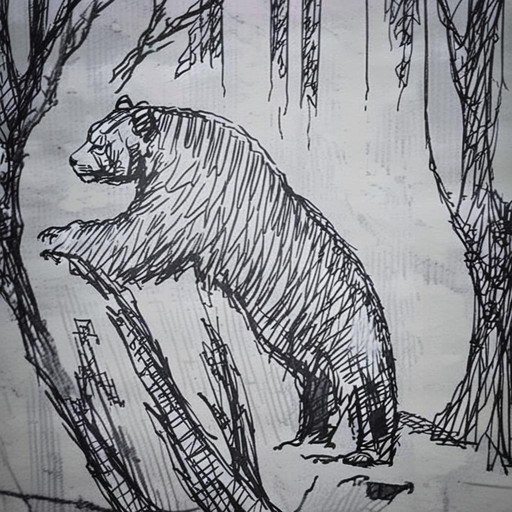} \\

        \includegraphics[width=0.15\textwidth]{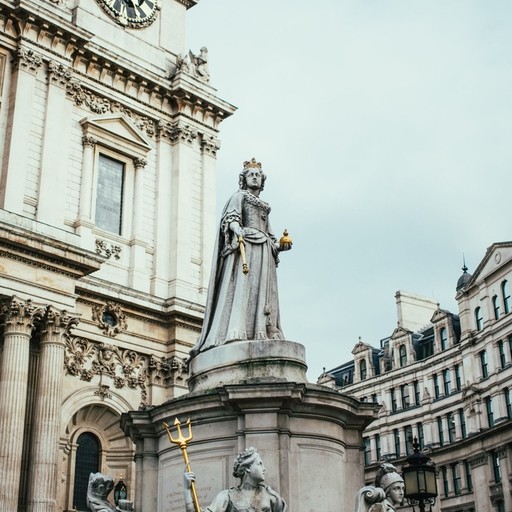} &
        \hspace{0.11cm}
        \includegraphics[width=0.15\textwidth]{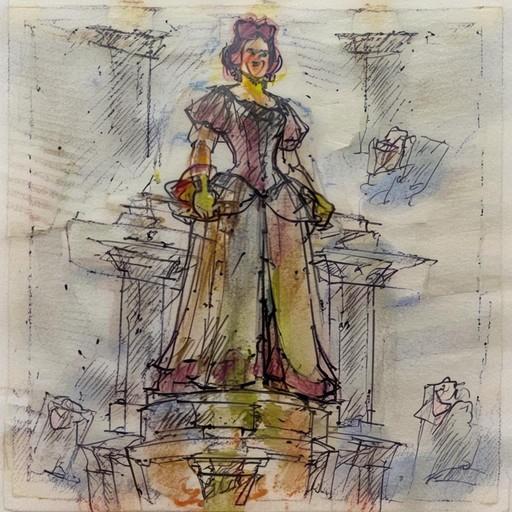} &
        \includegraphics[width=0.15\textwidth]{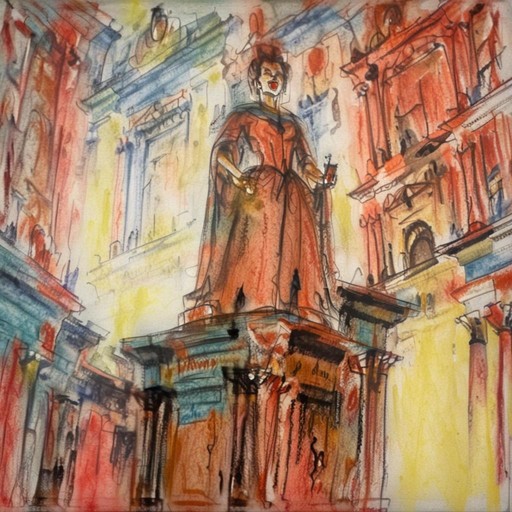} &
        \includegraphics[width=0.15\textwidth]{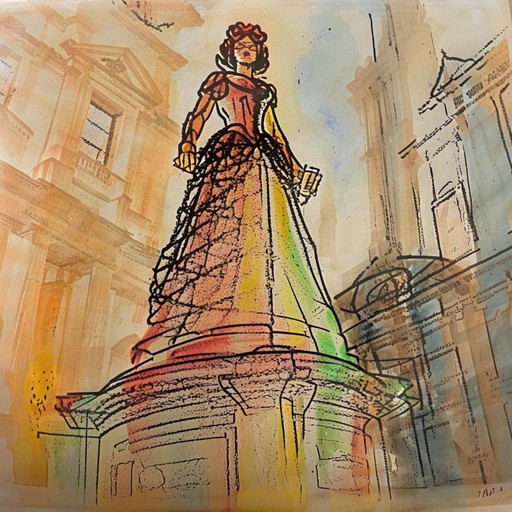} &
        \includegraphics[width=0.15\textwidth]{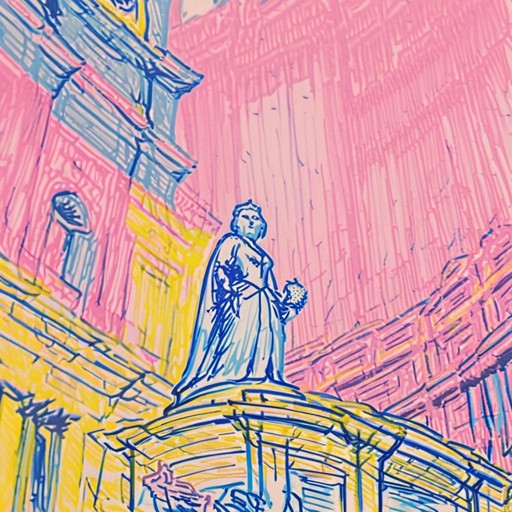} &
        \includegraphics[width=0.15\textwidth]{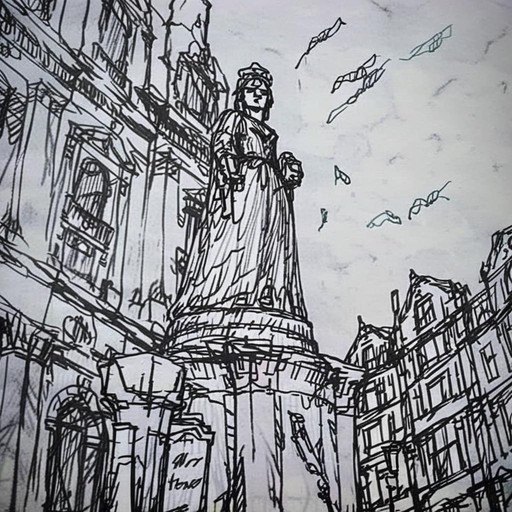} \\

        \includegraphics[width=0.15\textwidth]{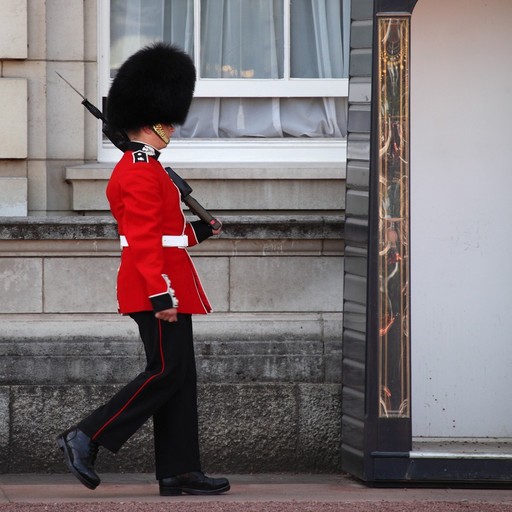} &
        \hspace{0.11cm}
        \includegraphics[width=0.15\textwidth]{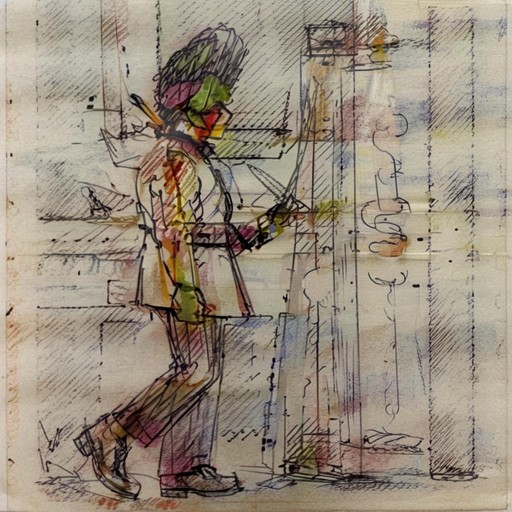} &
        \includegraphics[width=0.15\textwidth]{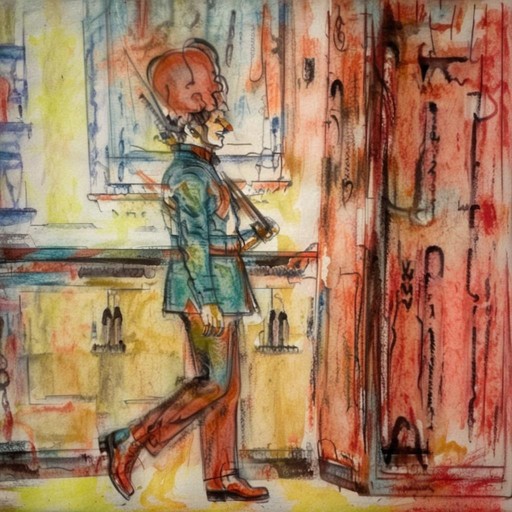} &
        \includegraphics[width=0.15\textwidth]{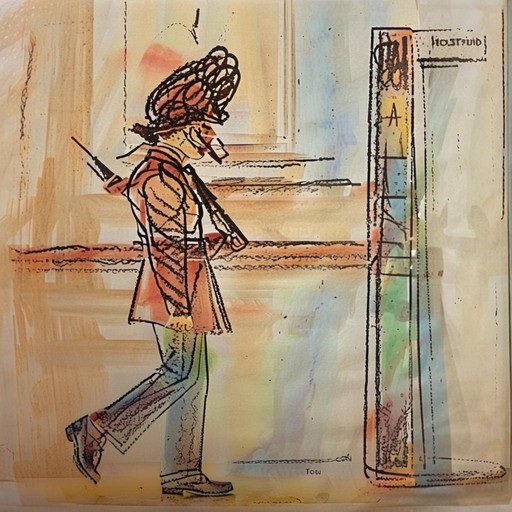} &
        \includegraphics[width=0.15\textwidth]{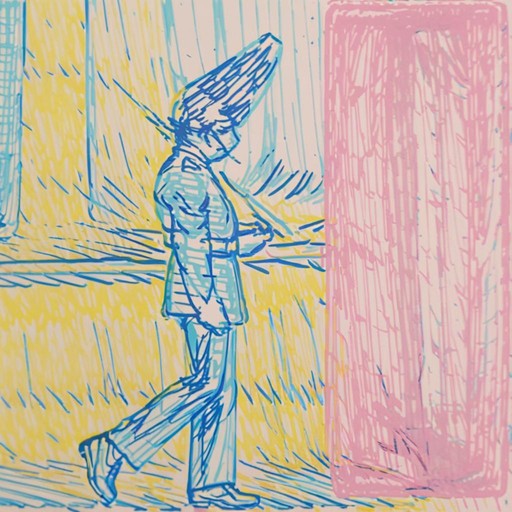} &
        \includegraphics[width=0.15\textwidth]{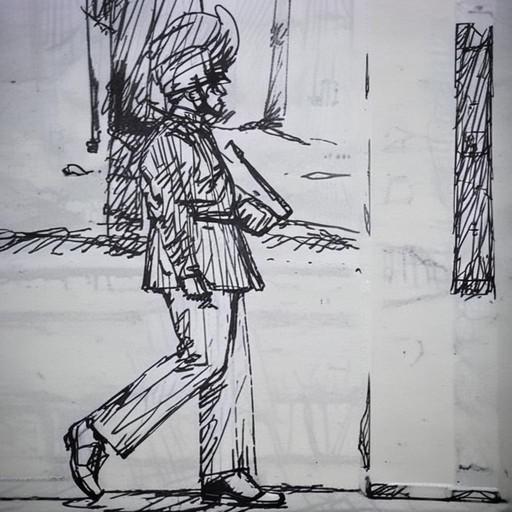} \\
        
    \end{NiceTabular}
    
    }
    \vspace{-7.5pt}
    \caption{Image stylization based on image style reference using B-LoRA, illustrating the performance on challenging content image references. \copyright The paintings in the first three columns are by Judith Kondor Mochary.}
    \vspace{-0.2cm}
    \label{fig:scene1}
\end{figure*}

\begin{figure*}
    \centering
    \setlength{\tabcolsep}{1.5pt}
    {\small
    \begin{NiceTabular}{c @{\hspace{0.15cm}} c c c c}

        \diagbox{Input \\ Content}{Style} &
        \hspace{0.11cm}
        \includegraphics[width=0.15\textwidth]{temp_figs/style_images/painting.jpg} &
        \includegraphics[width=0.15\textwidth]{temp_figs/style_images/watercolor.png} &
        \includegraphics[width=0.15\textwidth]{temp_figs/style_images/working_cartoon.jpg} &
        \includegraphics[width=0.15\textwidth]{temp_figs/style_images/cartoon_line.png} \\

        \includegraphics[width=0.15\textwidth]{temp_figs/scene_images/ballerina.jpg} &
        \hspace{0.11cm}
        \includegraphics[width=0.15\textwidth]{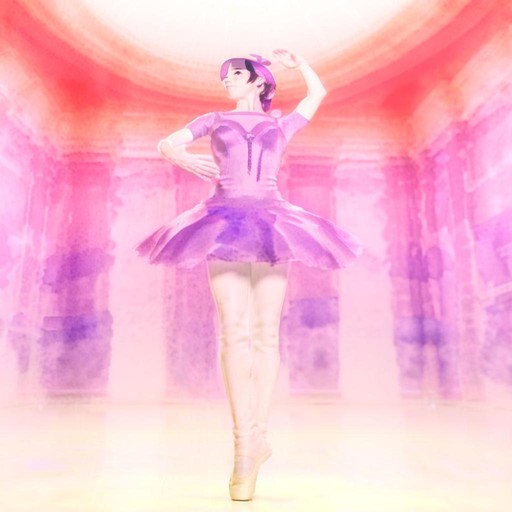} &
        \includegraphics[width=0.15\textwidth]{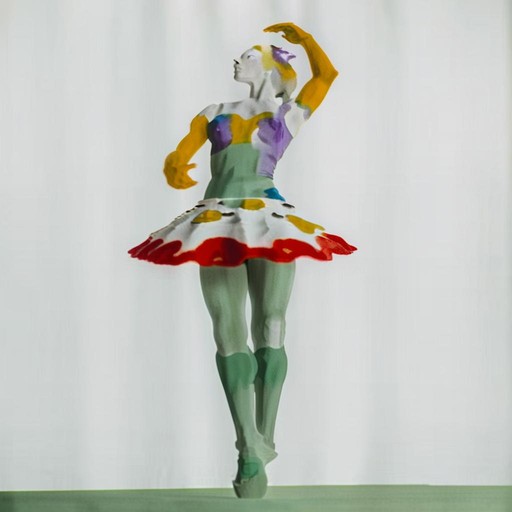} &
        \includegraphics[width=0.15\textwidth]{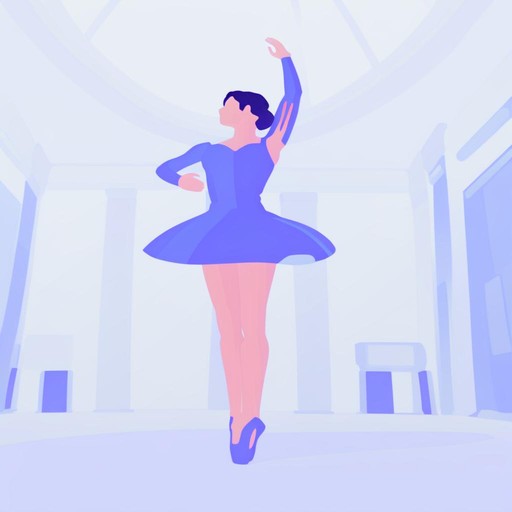} &
        \includegraphics[width=0.15\textwidth]{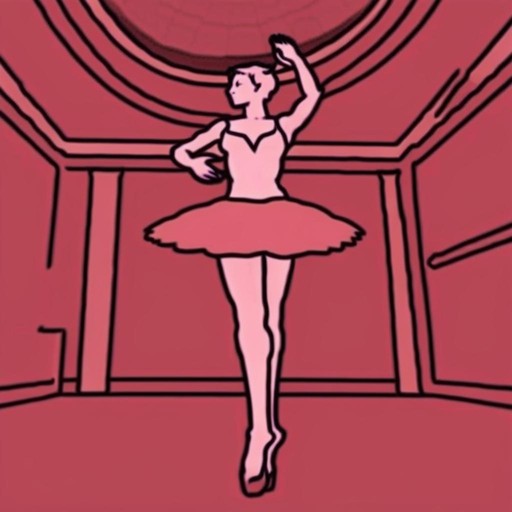} \\

        \includegraphics[width=0.15\textwidth]{temp_figs/scene_images/boat.jpg} &
        \hspace{0.11cm}
        \includegraphics[width=0.15\textwidth]{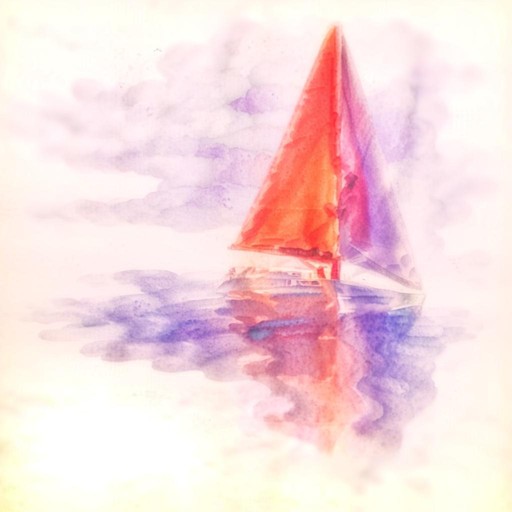} &
        \includegraphics[width=0.15\textwidth]{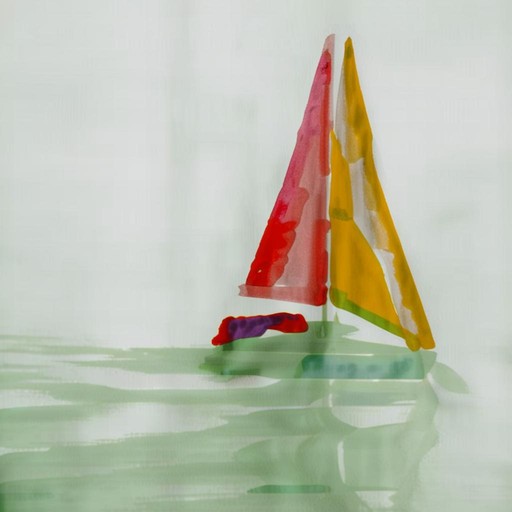} &
        \includegraphics[width=0.15\textwidth]{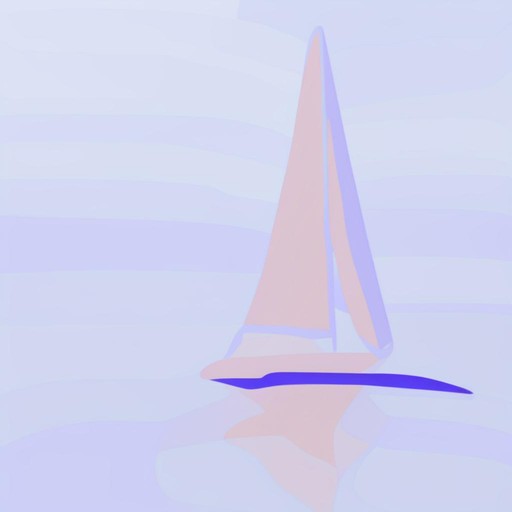} &
        \includegraphics[width=0.15\textwidth]{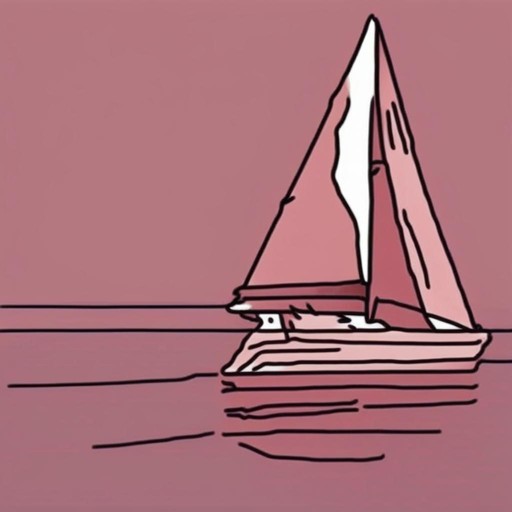} \\

        \includegraphics[width=0.15\textwidth]{temp_figs/scene_images/eiffel_tower.jpg} &
        \hspace{0.11cm}
        \includegraphics[width=0.15\textwidth]{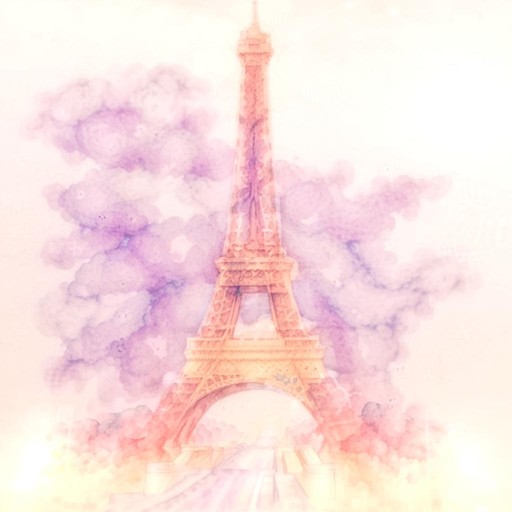} &
        \includegraphics[width=0.15\textwidth]{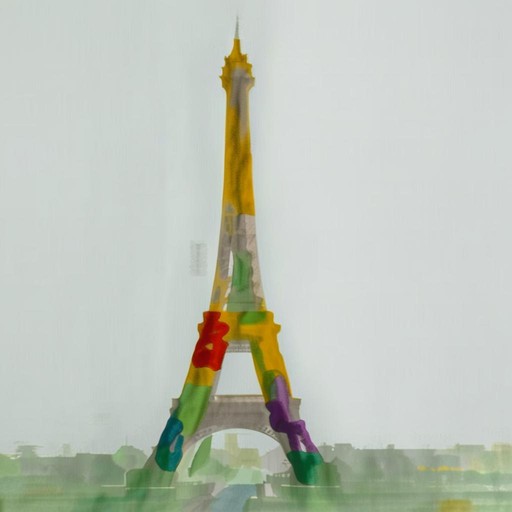} &
        \includegraphics[width=0.15\textwidth]{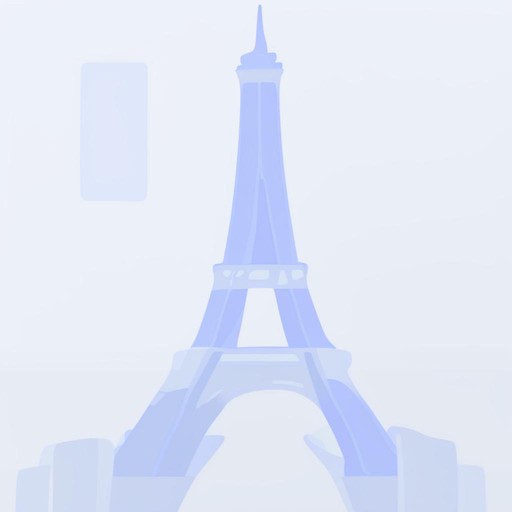} &
        \includegraphics[width=0.15\textwidth]{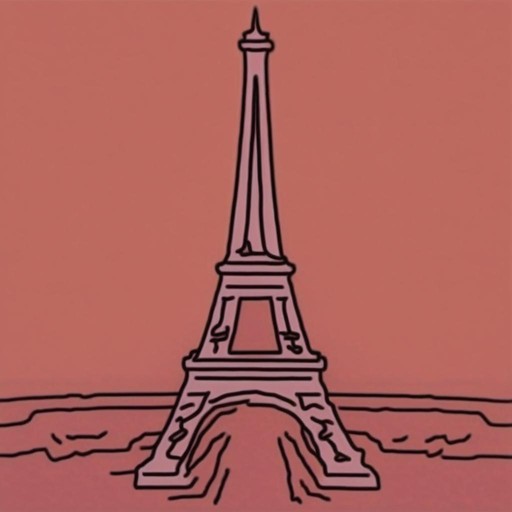} \\

         \includegraphics[width=0.15\textwidth]{temp_figs/scene_images/plant.jpg} &
        \hspace{0.11cm}
        \includegraphics[width=0.15\textwidth]{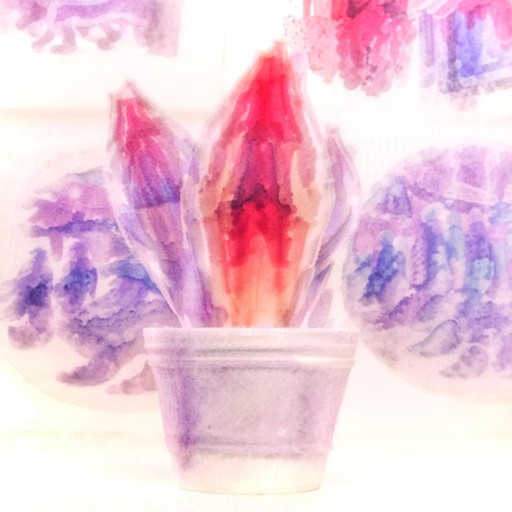} &
        \includegraphics[width=0.15\textwidth]{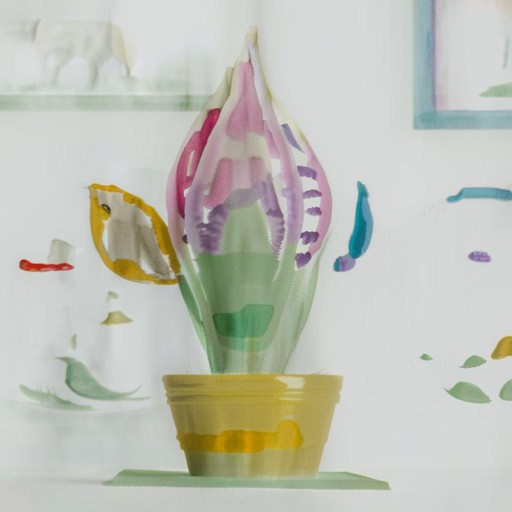} &
        \includegraphics[width=0.15\textwidth]{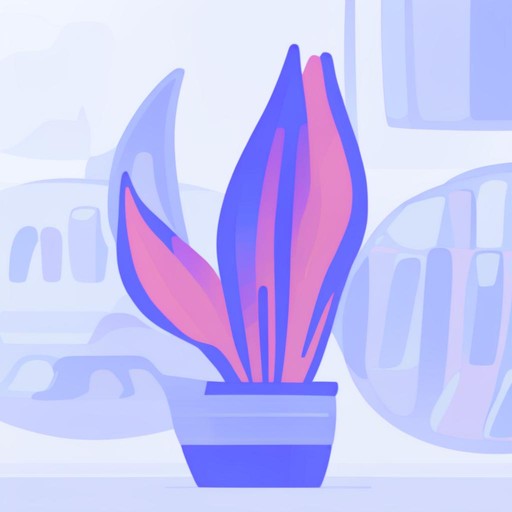} &
        \includegraphics[width=0.15\textwidth]{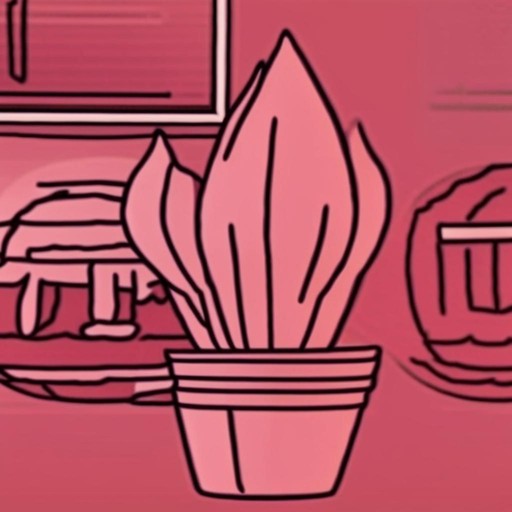} \\

        \includegraphics[width=0.15\textwidth]{temp_figs/scene_images/panda.jpg} &
        \hspace{0.11cm}
        \includegraphics[width=0.15\textwidth]{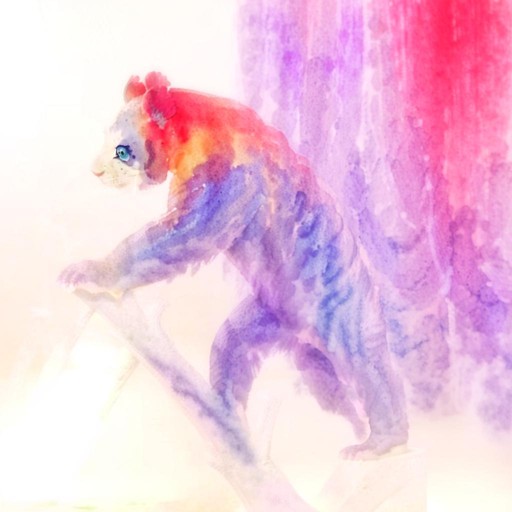} &
        \includegraphics[width=0.15\textwidth]{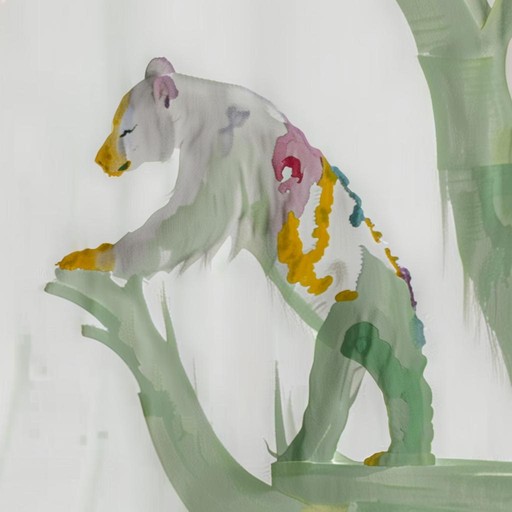} &
        \includegraphics[width=0.15\textwidth]{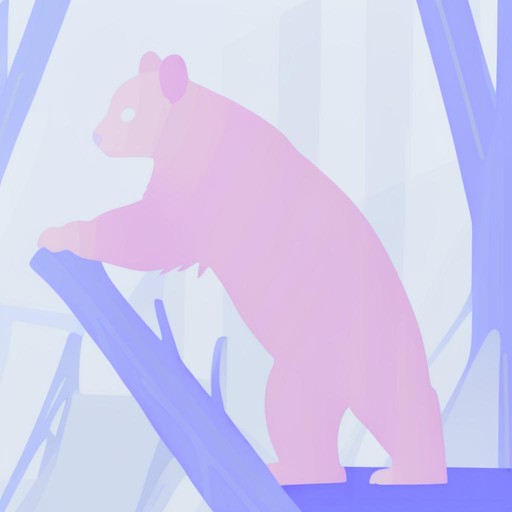} &
        \includegraphics[width=0.15\textwidth]{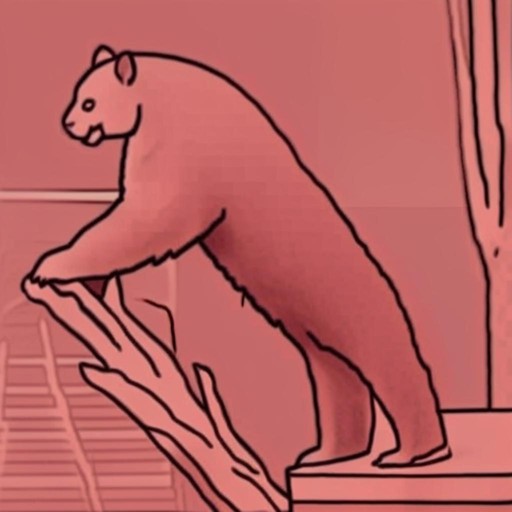}\\

        \includegraphics[width=0.15\textwidth]{temp_figs/scene_images/queen.jpg} &
        \hspace{0.11cm}
        \includegraphics[width=0.15\textwidth]{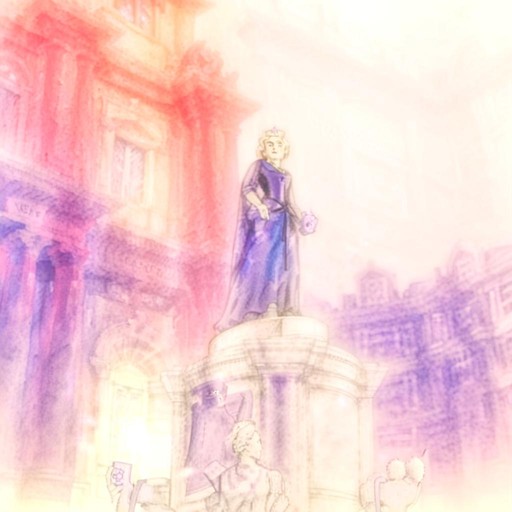} &
        \includegraphics[width=0.15\textwidth]{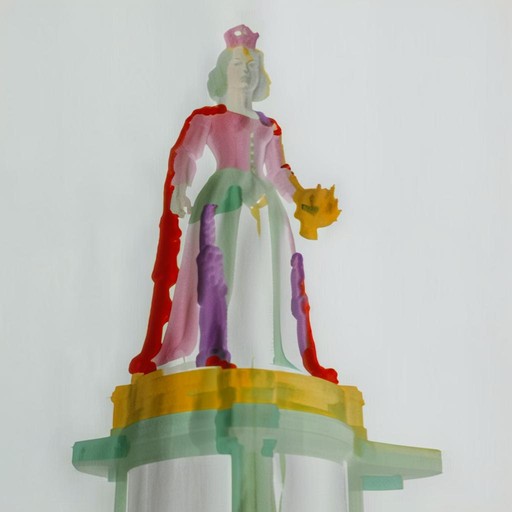} &
        \includegraphics[width=0.15\textwidth]{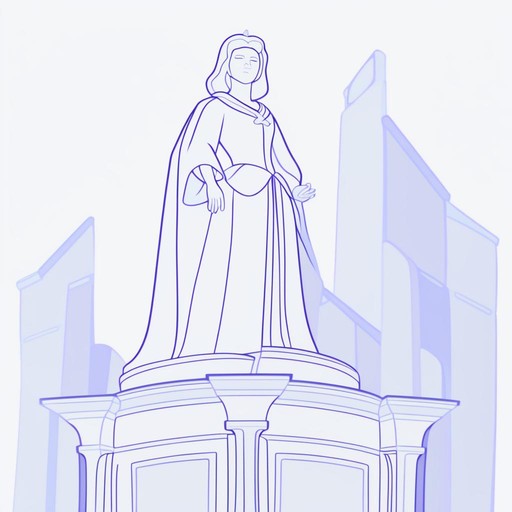} &
        \includegraphics[width=0.15\textwidth]{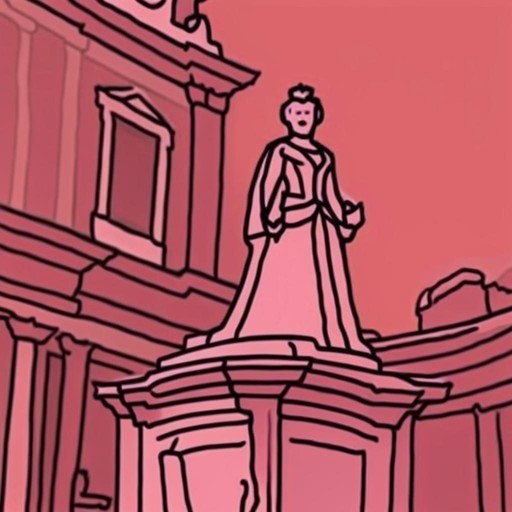} \\

        \includegraphics[width=0.15\textwidth]{temp_figs/scene_images/royal_guard.jpg} &
        \hspace{0.11cm}
        \includegraphics[width=0.15\textwidth]{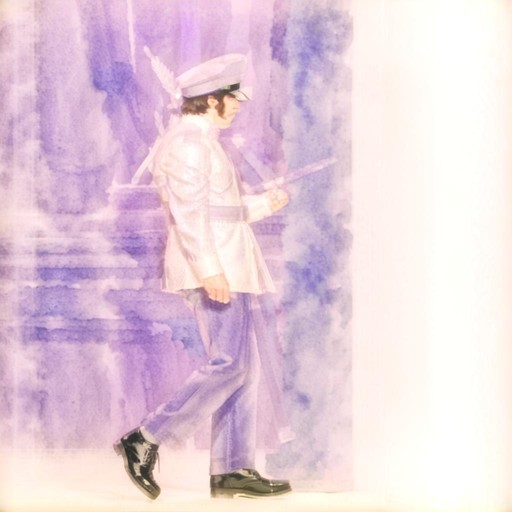} &
        \includegraphics[width=0.15\textwidth]{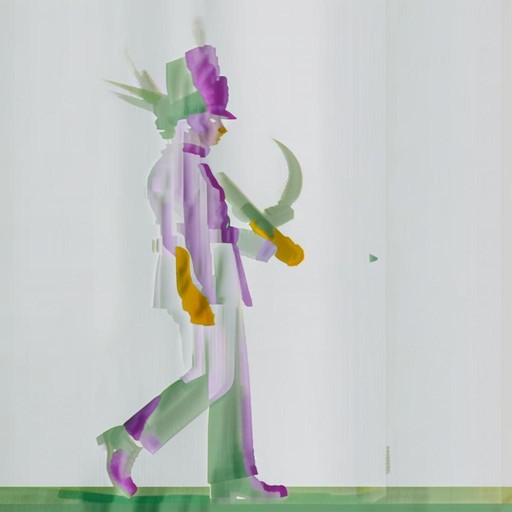} &
        \includegraphics[width=0.15\textwidth]{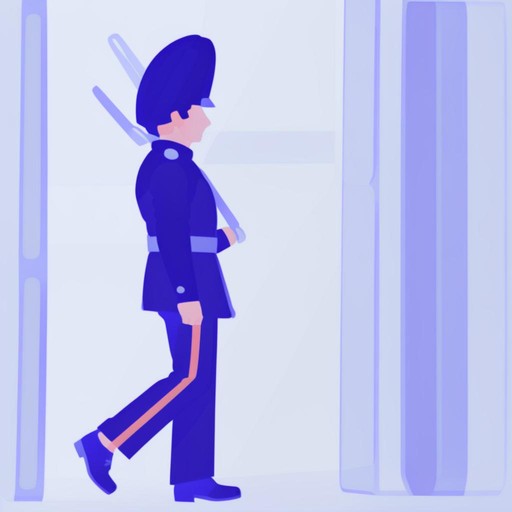} &
        \includegraphics[width=0.15\textwidth]{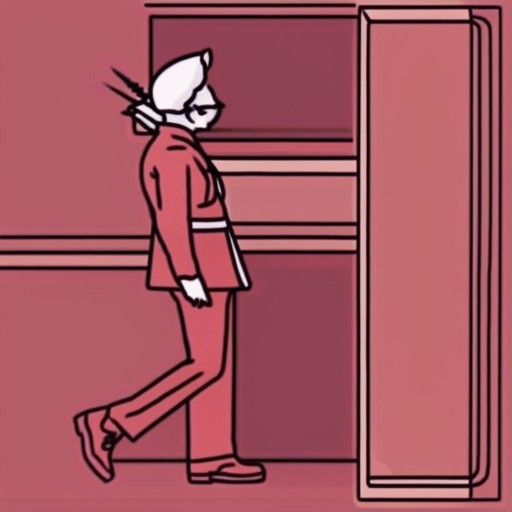} \\
        
    \end{NiceTabular}
    
    }
    \vspace{-7.5pt}
    \caption{Image stylization based on image style reference using B-LoRA, illustrating the performance on challenging content image references.}
    \vspace{-0.2cm}
    \label{fig:scene2}
\end{figure*}

\begin{figure*}
    \centering
    \setlength{\tabcolsep}{1.5pt}
    {\small
    \begin{NiceTabular}{c @{\hspace{0.2cm}} c c c c c}

        \diagbox{Input \\ Content}{Style} &
        \hspace{0.11cm}
        \includegraphics[width=0.15\textwidth]{temp_figs/style_images/pen_sketch.jpeg} &
        \includegraphics[width=0.15\textwidth]{temp_figs/style_images/drawing3.png} &
        \includegraphics[width=0.15\textwidth]{temp_figs/style_images/painting.jpg} &
        \includegraphics[width=0.15\textwidth]{temp_figs/style_images/drawing1.jpg} &
        \includegraphics[width=0.15\textwidth]{temp_figs/style_images/working_cartoon.jpg} \\

        \includegraphics[width=0.15\textwidth]{temp_figs/style_images/pen_sketch.jpeg} &
        \hspace{0.11cm}
        \includegraphics[width=0.15\textwidth]{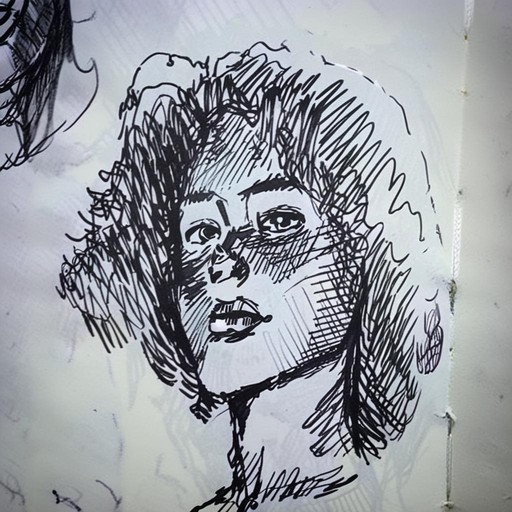} &
        \includegraphics[width=0.15\textwidth]{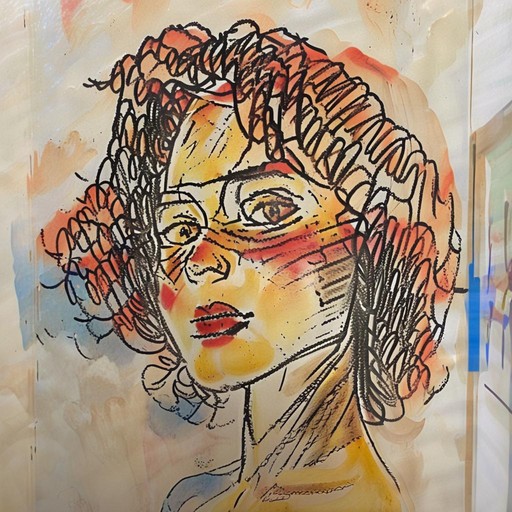} &
        \includegraphics[width=0.15\textwidth]{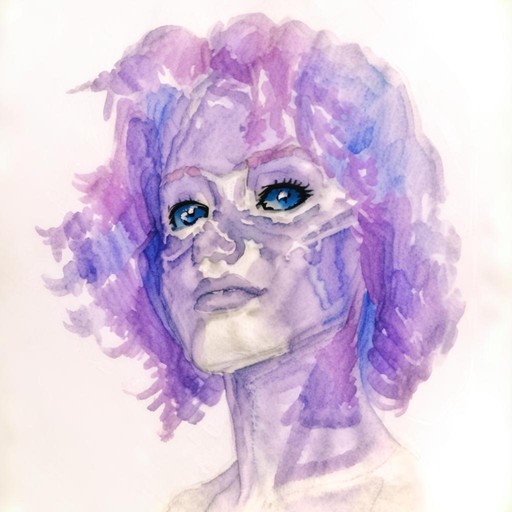} &
        \includegraphics[width=0.15\textwidth]{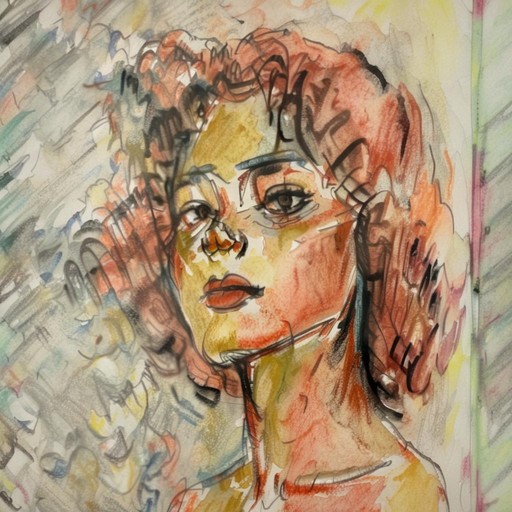} &
        \includegraphics[width=0.15\textwidth]{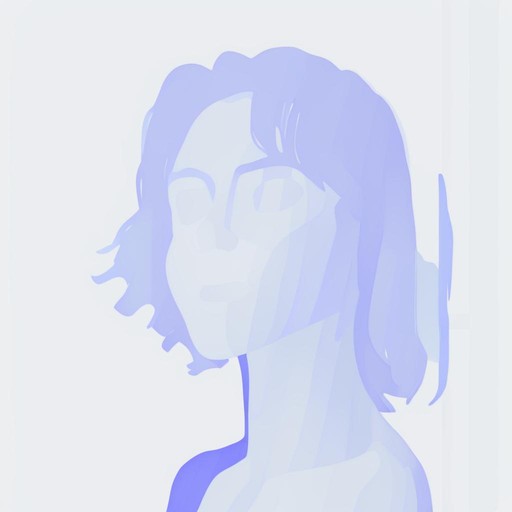} \\

        \includegraphics[width=0.15\textwidth]{temp_figs/style_images/drawing3.png} &
        \hspace{0.11cm}
        \includegraphics[width=0.15\textwidth]{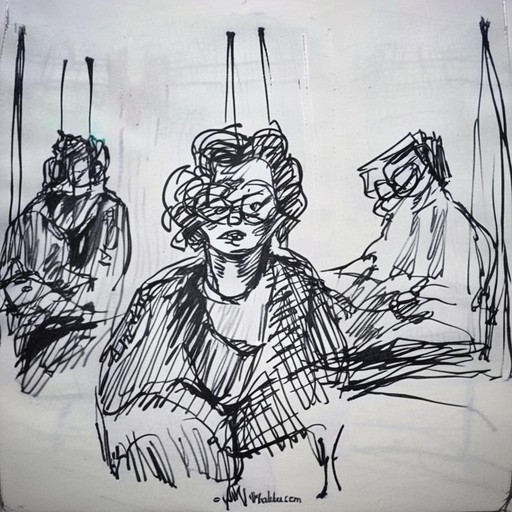} &
        \includegraphics[width=0.15\textwidth]{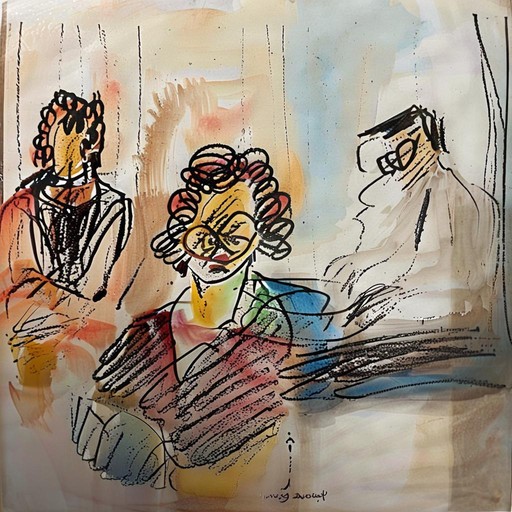} &
        \includegraphics[width=0.15\textwidth]{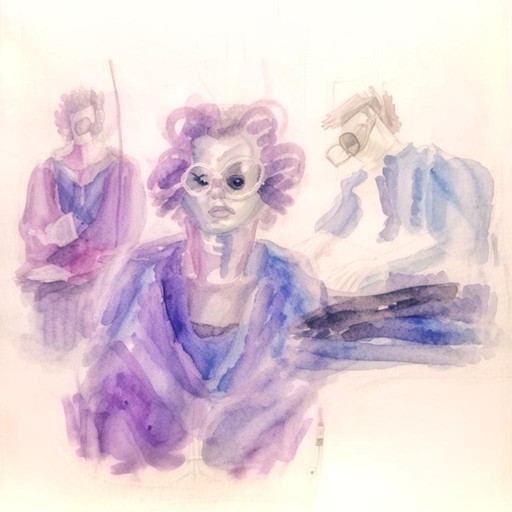} &
        \includegraphics[width=0.15\textwidth]{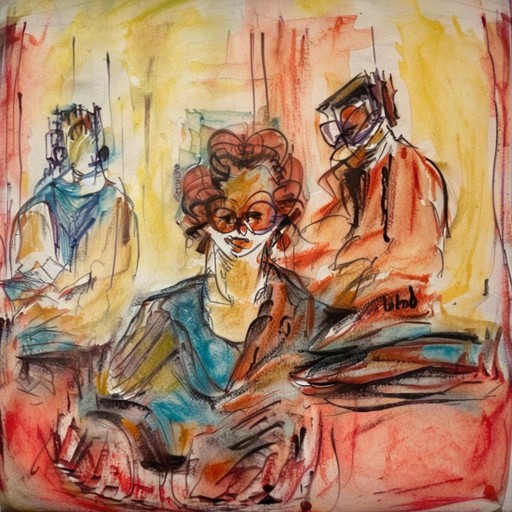} &
        \includegraphics[width=0.15\textwidth]{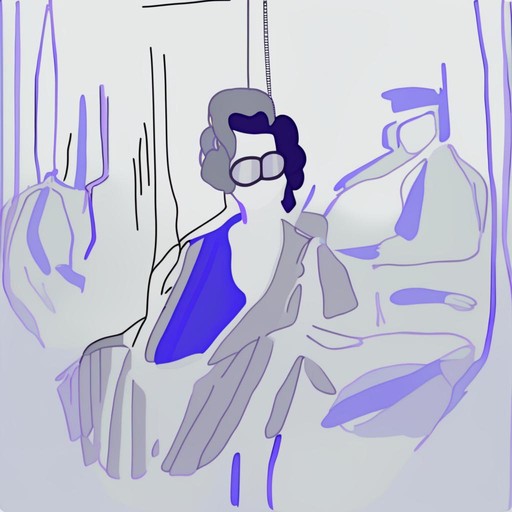} \\

        \includegraphics[width=0.15\textwidth]{temp_figs/style_images/painting.jpg} &
        \hspace{0.11cm}
        \includegraphics[width=0.15\textwidth]{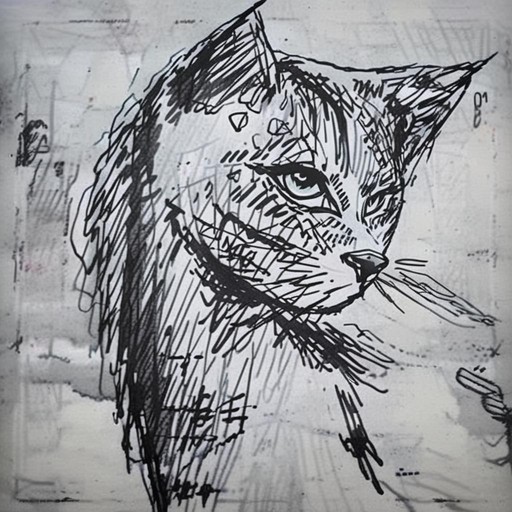} &
        \includegraphics[width=0.15\textwidth]{sup_figs/style_mixing/painting_drawing3.jpg} &
        \includegraphics[width=0.15\textwidth]{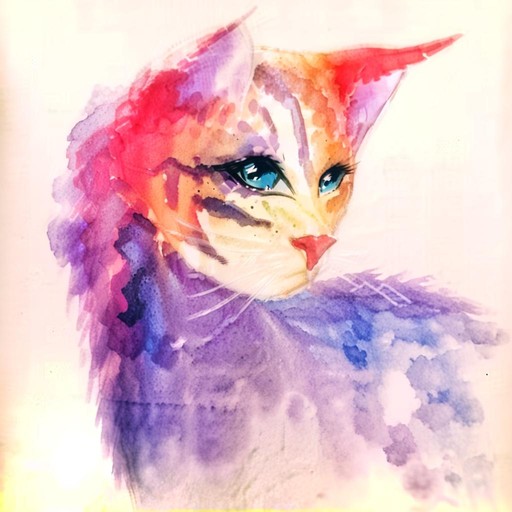} &
        \includegraphics[width=0.15\textwidth]{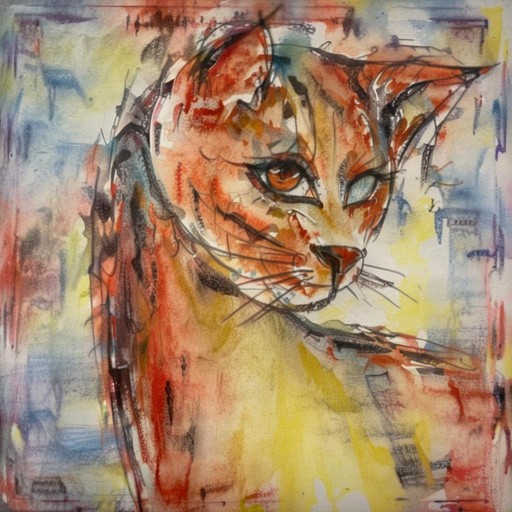} &
        \includegraphics[width=0.15\textwidth]{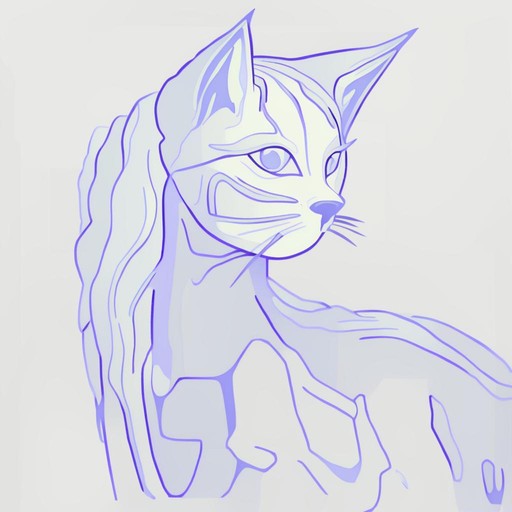} \\
        
        \includegraphics[width=0.15\textwidth]{temp_figs/style_images/drawing1.jpg} &
        \hspace{0.11cm}
        \includegraphics[width=0.15\textwidth]{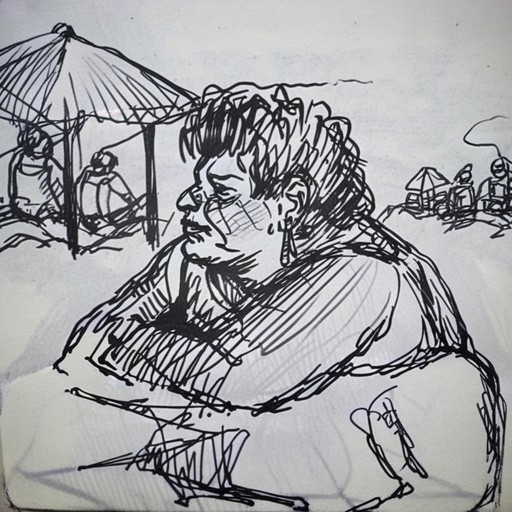} &
        \includegraphics[width=0.15\textwidth]{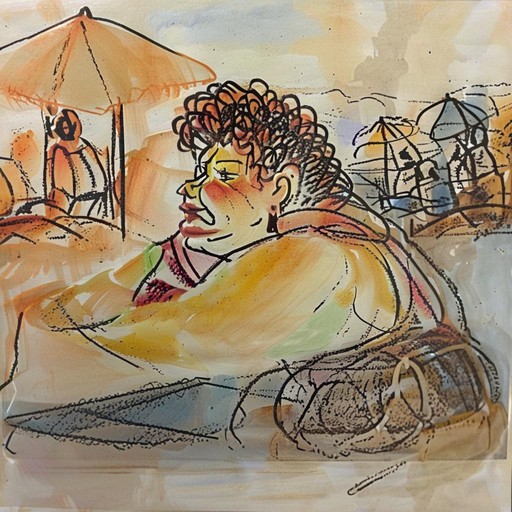} &
        \includegraphics[width=0.15\textwidth]{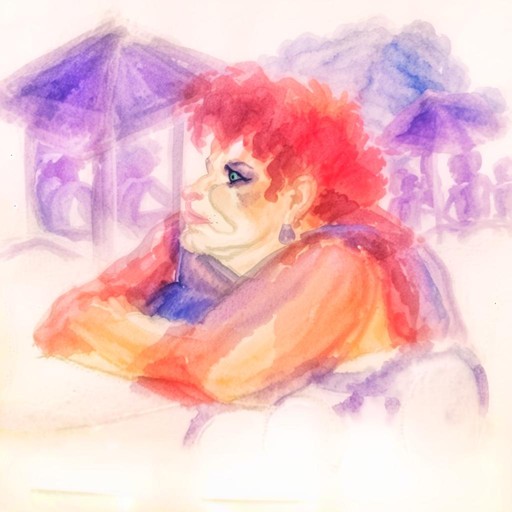} &
        \includegraphics[width=0.15\textwidth]{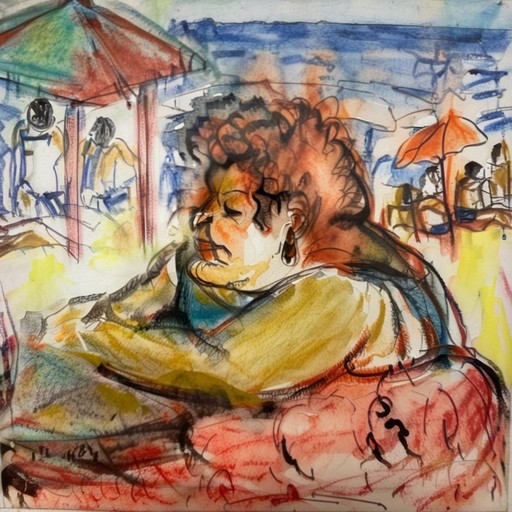} &
        \includegraphics[width=0.15\textwidth]{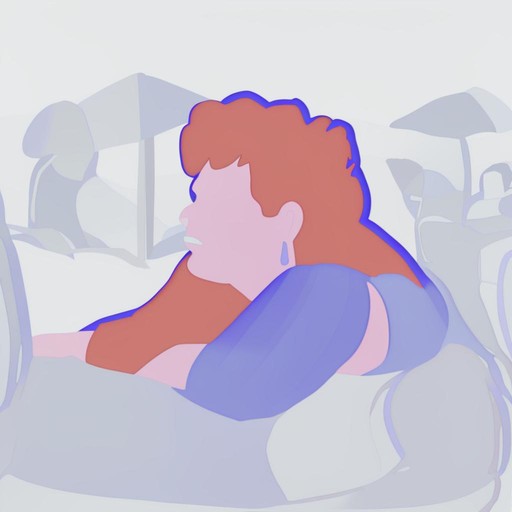} \\
        
        \includegraphics[width=0.15\textwidth]{temp_figs/style_images/working_cartoon.jpg} &
        \hspace{0.11cm}
        \includegraphics[width=0.15\textwidth]{sup_figs/style_mixing/working_cartoon_pen_sketch.jpg} &
        \includegraphics[width=0.15\textwidth]{sup_figs/style_mixing/working_cartoon_drawing3.jpg} &
        \includegraphics[width=0.15\textwidth]{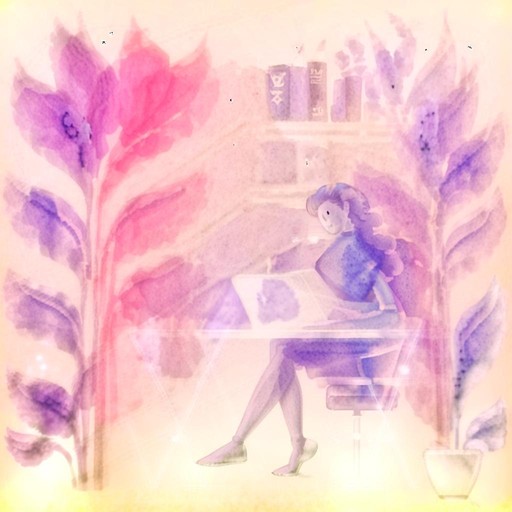} &
        \includegraphics[width=0.15\textwidth]{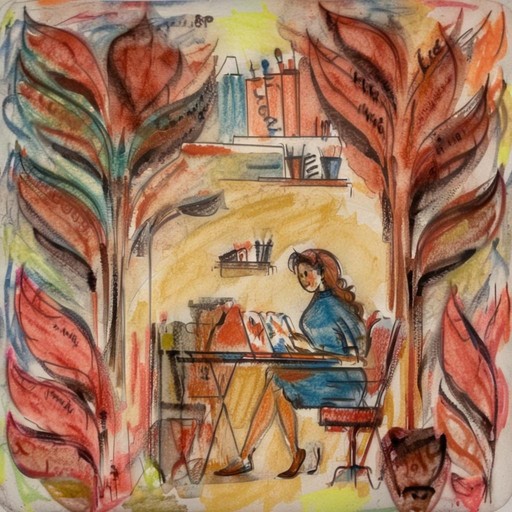} &
        \includegraphics[width=0.15\textwidth]{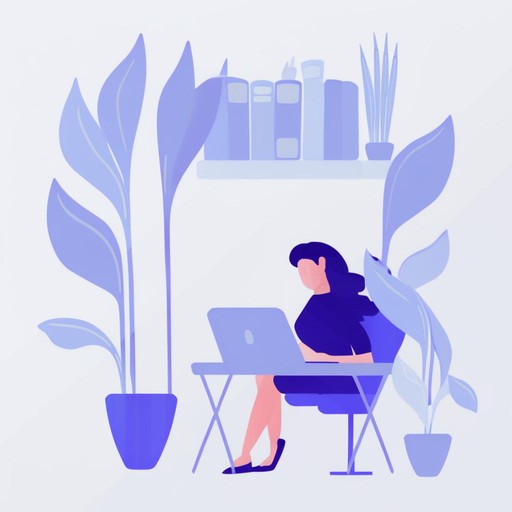} \\[0.3cm]
        
    \end{NiceTabular}
    
    }
    \vspace{0.11cm}
    \caption{Additional results generated using B-LoRA. Our method able to blend content and styles across different style images. Each object in the (i, j) cell is created by combining the $\Delta W^4$ of the i-th row with the $\Delta W^5$ of the j-th column, while the diagonal represents the reconstruction image. \copyright The paintings in the second and third columns (and rows) are by Judith Kondor Mochary.}
    \vspace{-0.2cm}
    \label{fig:style_to_style}
\end{figure*}

\begin{figure*}
    \centering
    \setlength{\tabcolsep}{1.5pt}
    {\small
    \begin{NiceTabular}{c @{\hspace{0.2cm}} c c c c c}

        \diagbox{Input \\ Content}{Style} &
        \hspace{0.11cm}
        \includegraphics[width=0.15\textwidth]{temp_figs/content_images/bull.jpg} &
        \includegraphics[width=0.15\textwidth]{temp_figs/content_images/wolf_plushie.jpg} &
        \includegraphics[width=0.15\textwidth]{temp_figs/content_images/dog2.jpg} &
        \includegraphics[width=0.15\textwidth]{temp_figs/content_images/fat_bird.jpg} &
        \includegraphics[width=0.15\textwidth]{temp_figs/content_images/statue.jpg} \\

        \includegraphics[width=0.15\textwidth]{temp_figs/content_images/bull.jpg} &
        \hspace{0.11cm}
        \includegraphics[width=0.15\textwidth]{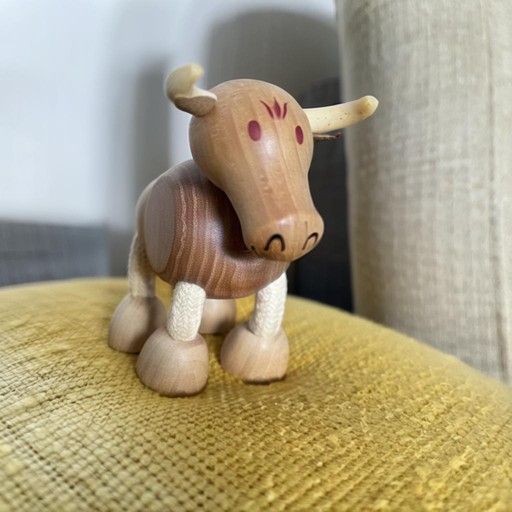} &
        \includegraphics[width=0.15\textwidth]{sup_figs/objects_mixing/bull_wolf_plushie.jpg} &
        \includegraphics[width=0.15\textwidth]{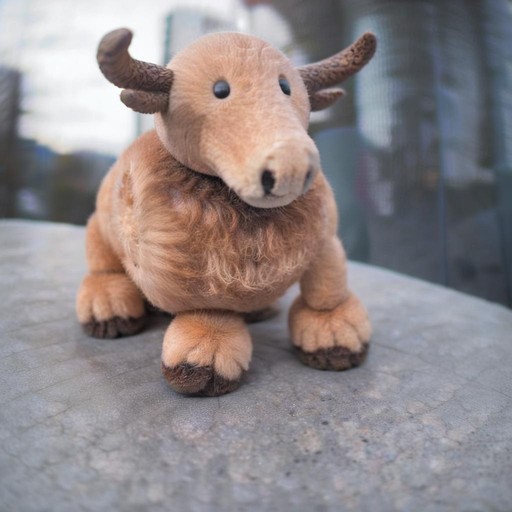} &
        \includegraphics[width=0.15\textwidth]{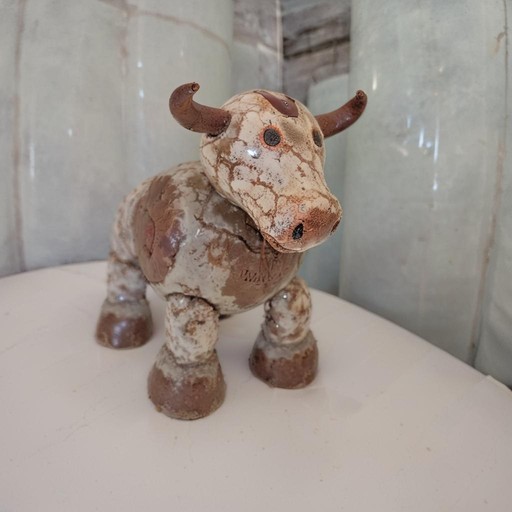} &
        \includegraphics[width=0.15\textwidth]{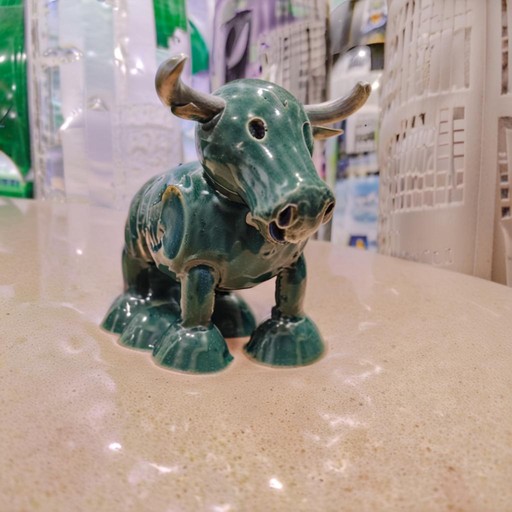} \\

        \includegraphics[width=0.15\textwidth]{temp_figs/content_images/wolf_plushie.jpg} &
        \hspace{0.11cm}
        \includegraphics[width=0.15\textwidth]{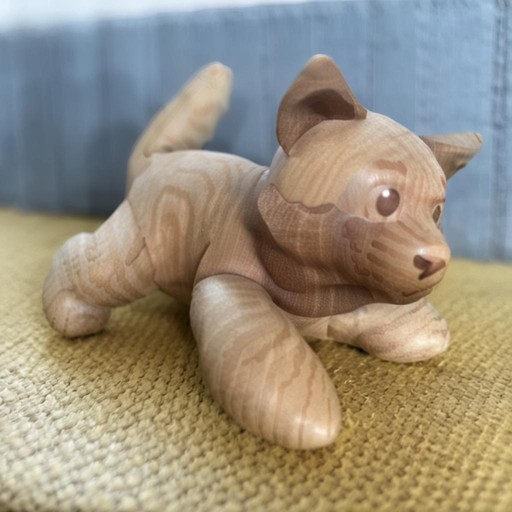} &
        \includegraphics[width=0.15\textwidth]{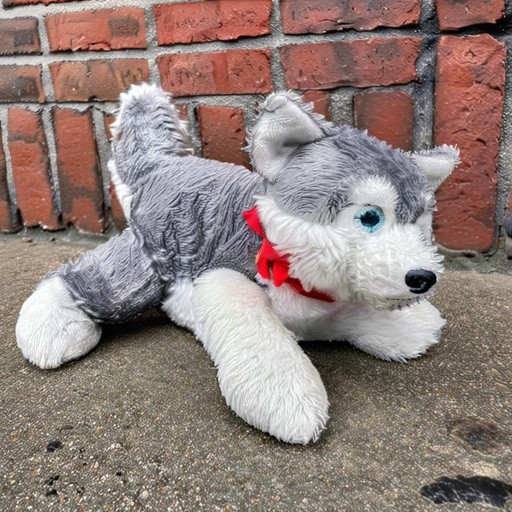} &
        \includegraphics[width=0.15\textwidth]{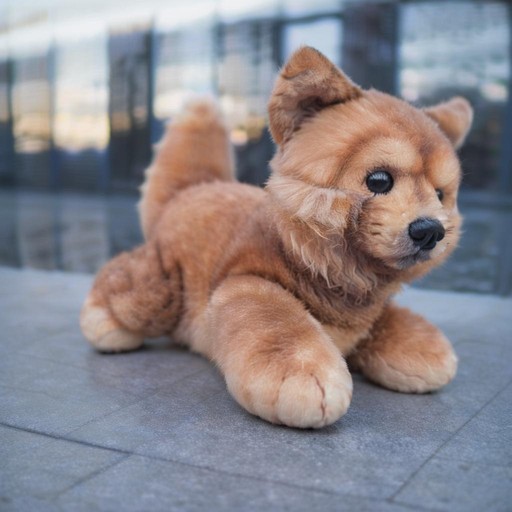} &
        \includegraphics[width=0.15\textwidth]{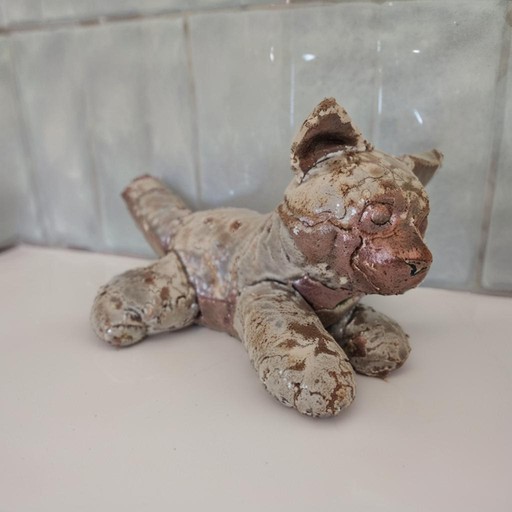} &
        \includegraphics[width=0.15\textwidth]{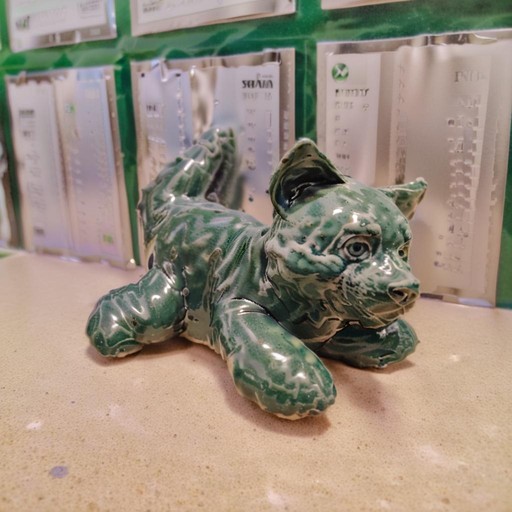} \\

        \includegraphics[width=0.15\textwidth]{temp_figs/content_images/dog2.jpg} &
        \hspace{0.11cm}
        \includegraphics[width=0.15\textwidth]{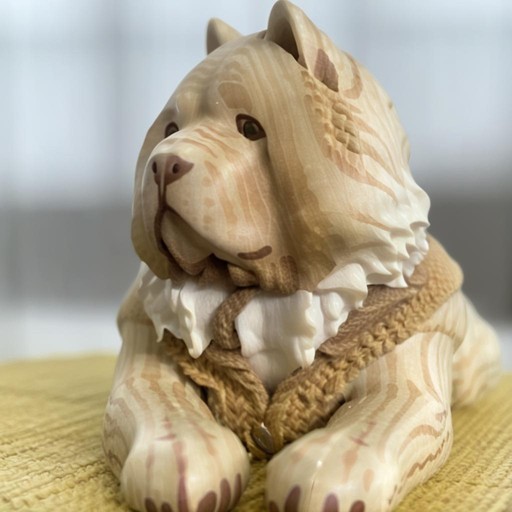} &
        \includegraphics[width=0.15\textwidth]{sup_figs/objects_mixing/dog2_wolf_plushie.jpg} &
        \includegraphics[width=0.15\textwidth]{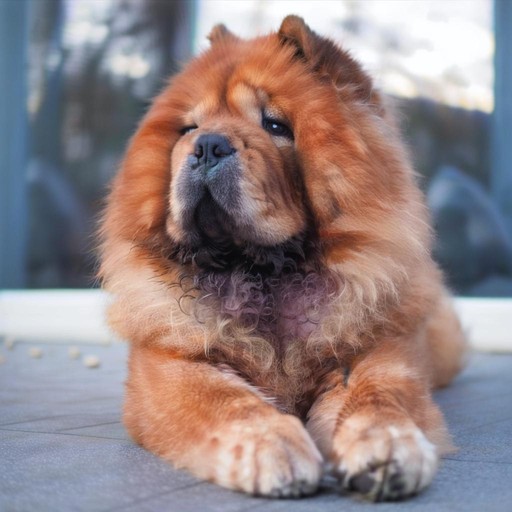} &
        \includegraphics[width=0.15\textwidth]{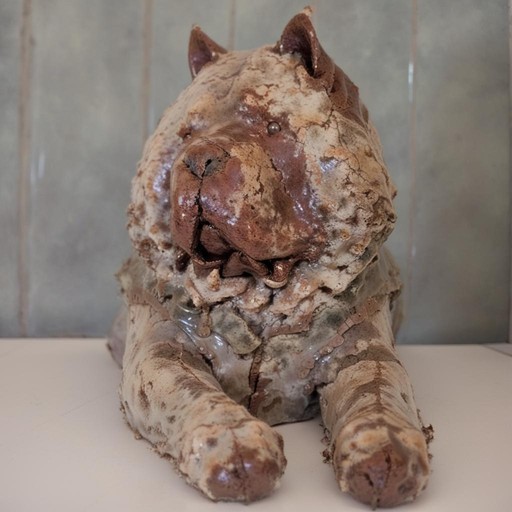} &
        \includegraphics[width=0.15\textwidth]{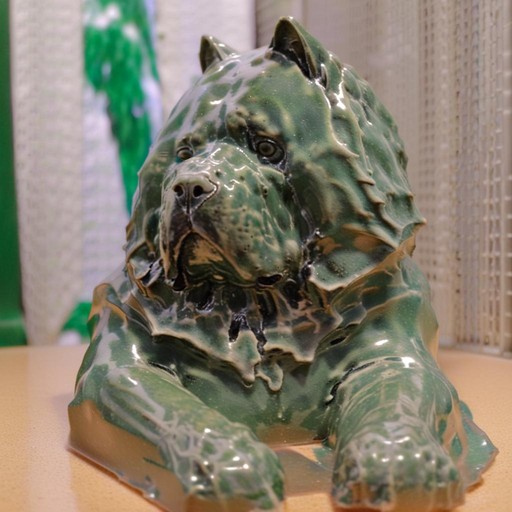} \\
        
        \includegraphics[width=0.15\textwidth]{temp_figs/content_images/fat_bird.jpg} &
        \hspace{0.11cm}
        \includegraphics[width=0.15\textwidth]{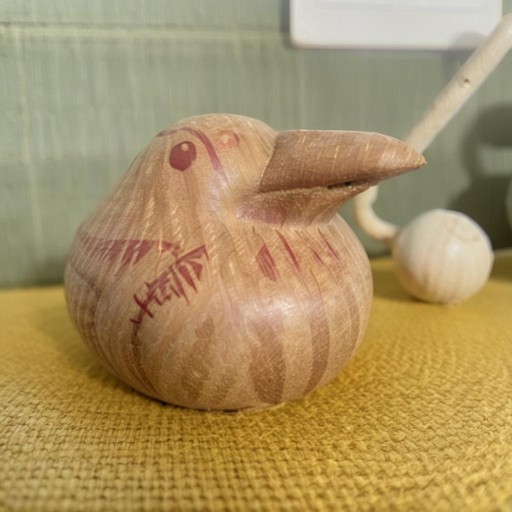} &
        \includegraphics[width=0.15\textwidth]{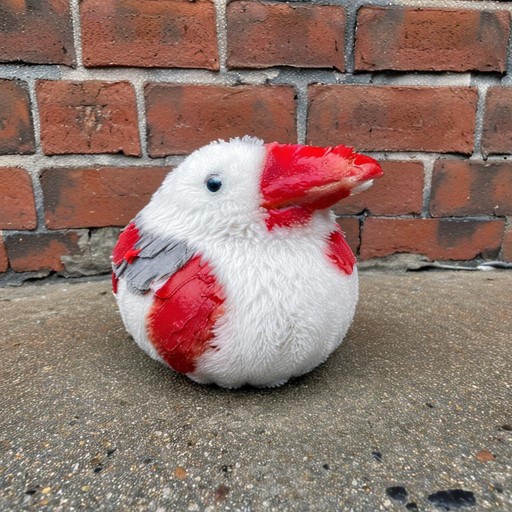} &
        \includegraphics[width=0.15\textwidth]{sup_figs/objects_mixing/fat_bird_dog2.jpg} &
        \includegraphics[width=0.15\textwidth]{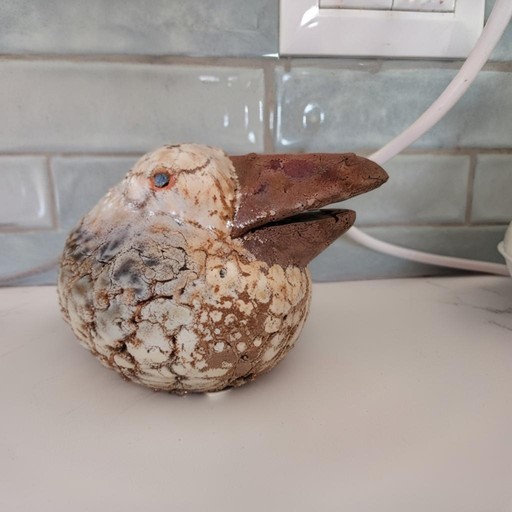} &
        \includegraphics[width=0.15\textwidth]{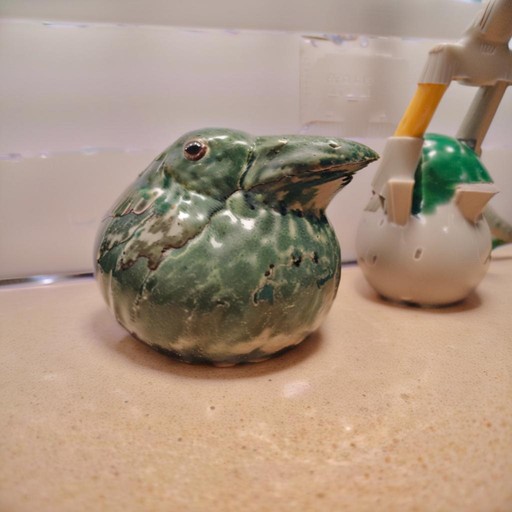} \\
        
        \includegraphics[width=0.15\textwidth]{temp_figs/content_images/statue.jpg} &
        \hspace{0.11cm}
        \includegraphics[width=0.15\textwidth]{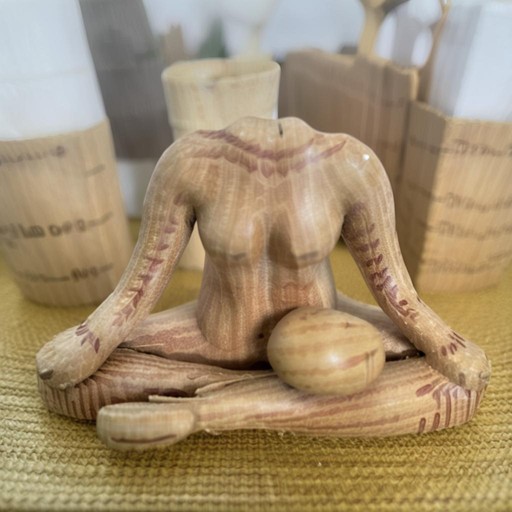} &
        \includegraphics[width=0.15\textwidth]{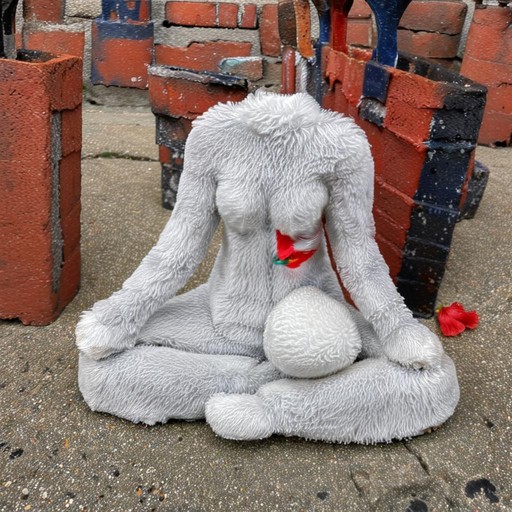} &
        \includegraphics[width=0.15\textwidth]{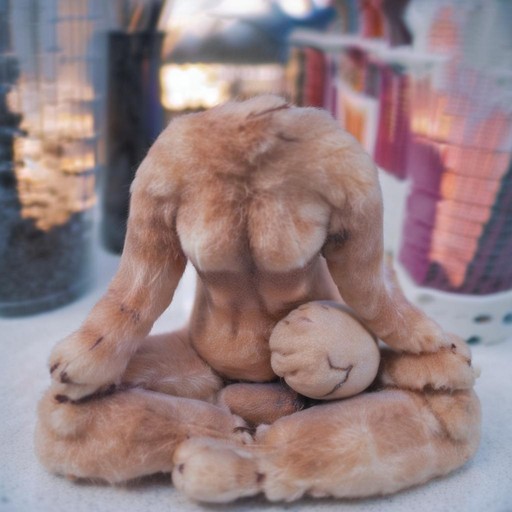} &
        \includegraphics[width=0.15\textwidth]{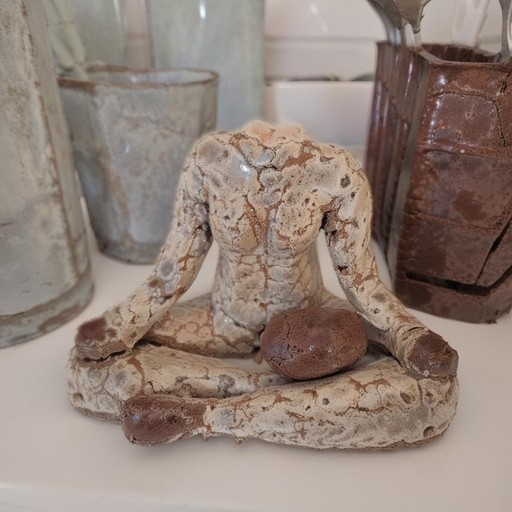} &
        \includegraphics[width=0.15\textwidth]{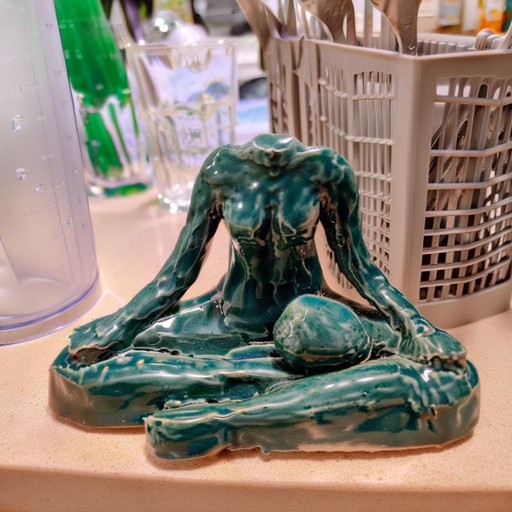} \\[0.3cm]
        
    \end{NiceTabular}
    
    }
    \vspace{0.11cm}
    \caption{Additional results generated using B-LoRA. Our method able to blend content and styles across different objects. Each object in the (i, j) cell is created by combining the $\Delta W^4$ of the i-th row with the $\Delta W^5$ of the j-th column, while the diagonal represents the reconstruction image.}
    \vspace{-0.2cm}
    \label{fig:object_to_object}
\end{figure*}

\begin{figure*}
    \centering
    \setlength{\tabcolsep}{1.5pt}
    {\small
    \begin{NiceTabular}{c | @{\hspace{0.2cm}} c c c c c c}

        \diagbox{Input \\ Content}{Style} &
        \hspace{0.11cm}
        \includegraphics[width=0.13\textwidth]{temp_figs/style_images/cartoon_line.png} &
        \includegraphics[width=0.13\textwidth]{temp_figs/style_images/crayon_drawing.jpg} &
        \includegraphics[width=0.13\textwidth]{temp_figs/style_images/pen_sketch.jpeg} &
        \includegraphics[width=0.13\textwidth]{temp_figs/style_images/kiss.png} &
        \includegraphics[width=0.13\textwidth]{temp_figs/style_images/drawing3.png} &
        \includegraphics[width=0.13\textwidth]{temp_figs/style_images/drawing2.jpg} \\
        \midrule

        \includegraphics[width=0.13\textwidth]{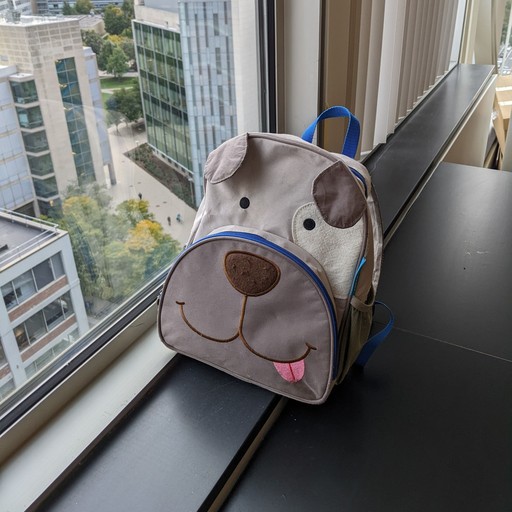} &
        \hspace{0.11cm}
        \includegraphics[width=0.13\textwidth]{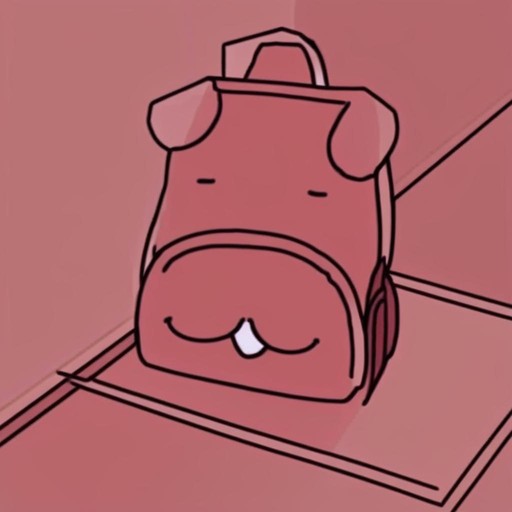} &
        \includegraphics[width=0.13\textwidth]{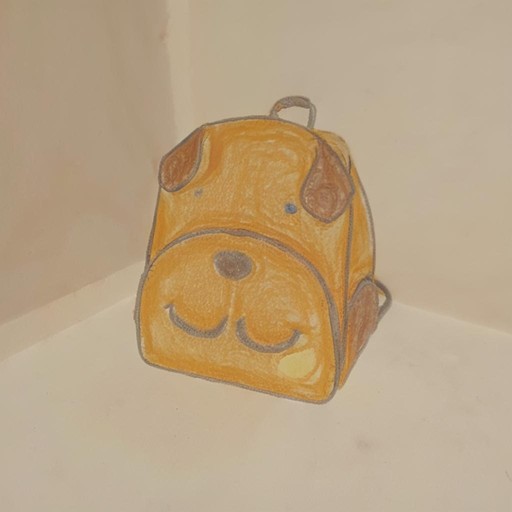} &
        \includegraphics[width=0.13\textwidth]{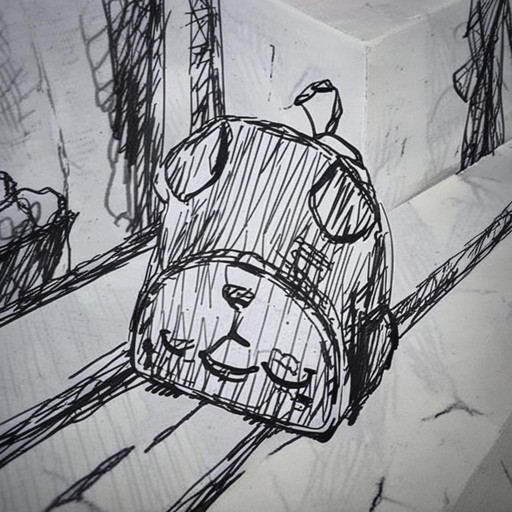} &
        \includegraphics[width=0.13\textwidth]{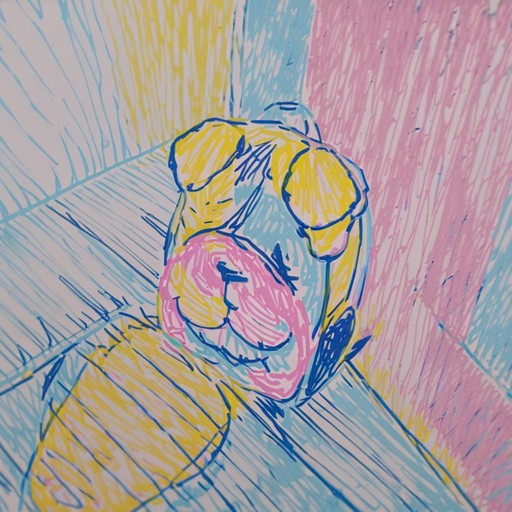} &
        \includegraphics[width=0.13\textwidth]{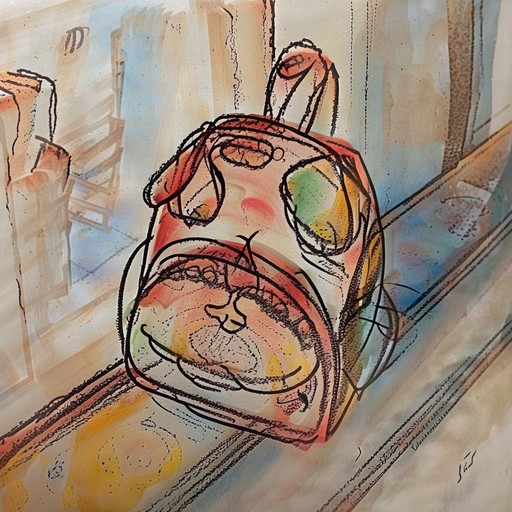} &
        \includegraphics[width=0.13\textwidth]{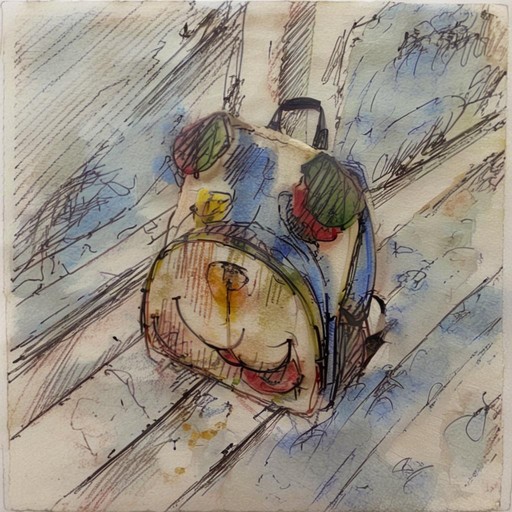} \\

        \includegraphics[width=0.13\textwidth]{temp_figs/content_images/wolf_plushie.jpg} &
        \hspace{0.11cm}
        \includegraphics[width=0.13\textwidth]{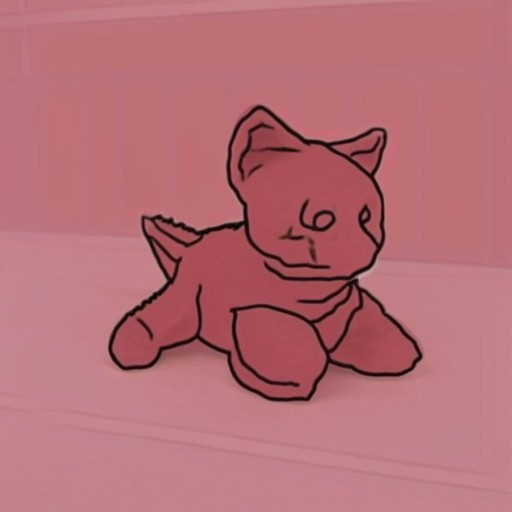} &
        \includegraphics[width=0.13\textwidth]{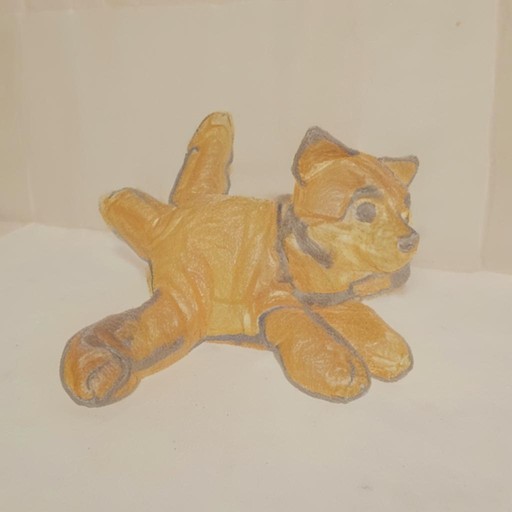} &
        \includegraphics[width=0.13\textwidth]{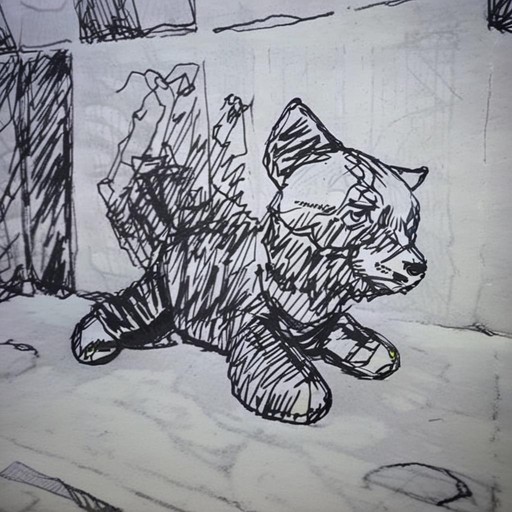} &
        \includegraphics[width=0.13\textwidth]{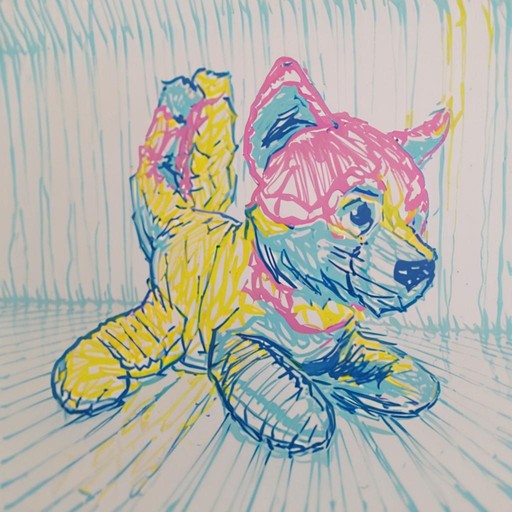} &
        \includegraphics[width=0.13\textwidth]{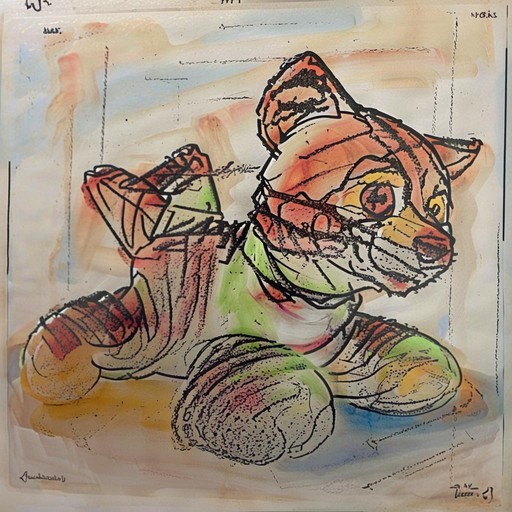} &
        \includegraphics[width=0.13\textwidth]{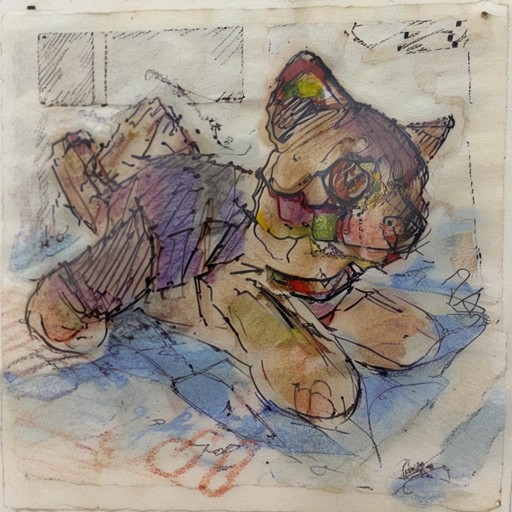} \\

        \includegraphics[width=0.13\textwidth]{temp_figs/content_images/fat_bird.jpg} &
        \hspace{0.11cm}
        \includegraphics[width=0.13\textwidth]{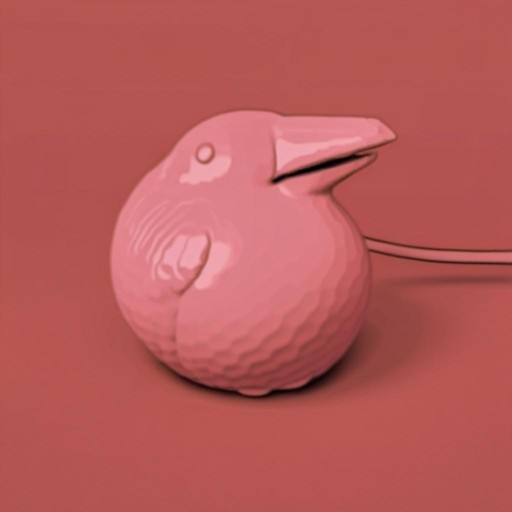} &
        \includegraphics[width=0.13\textwidth]{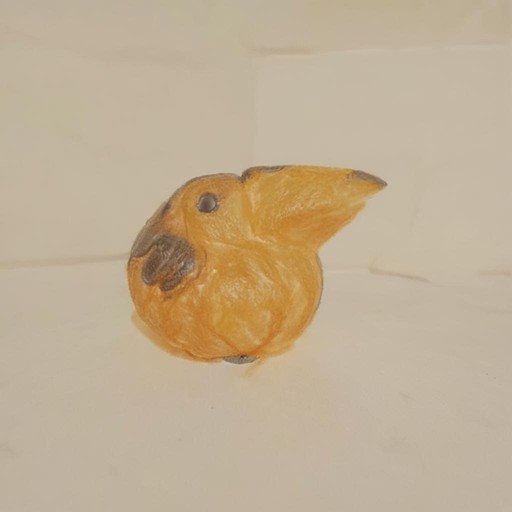} &
        \includegraphics[width=0.13\textwidth]{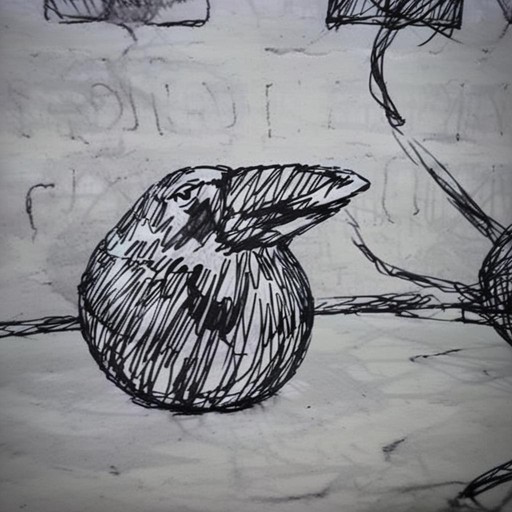} &
        \includegraphics[width=0.13\textwidth]{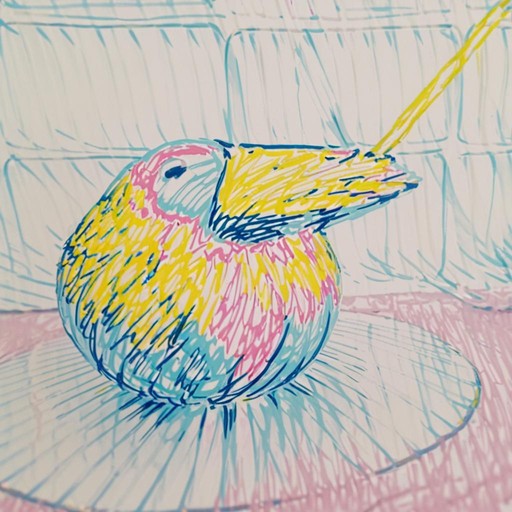} &
        \includegraphics[width=0.13\textwidth]{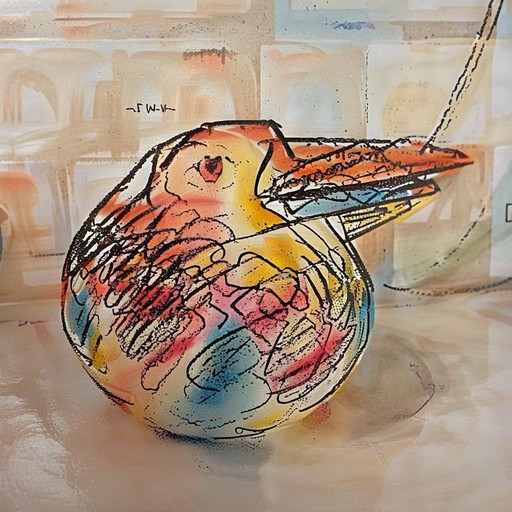} &
        \includegraphics[width=0.13\textwidth]{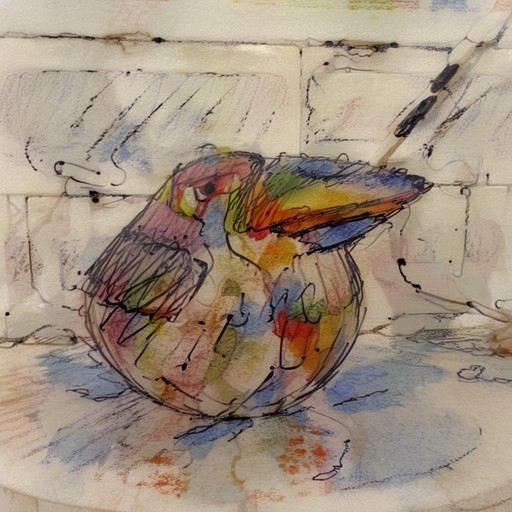} \\

        \includegraphics[width=0.13\textwidth]{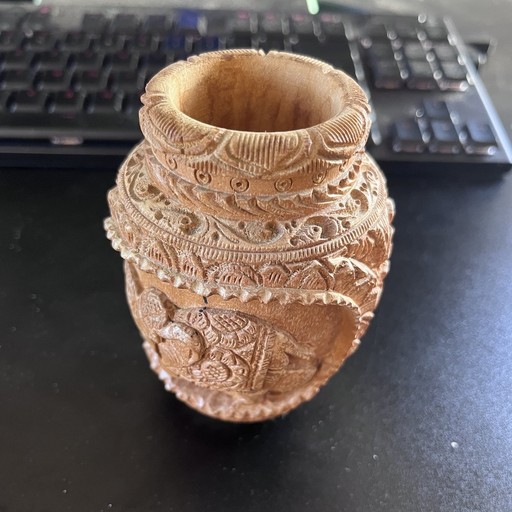} &
        \hspace{0.11cm}
        \includegraphics[width=0.13\textwidth]{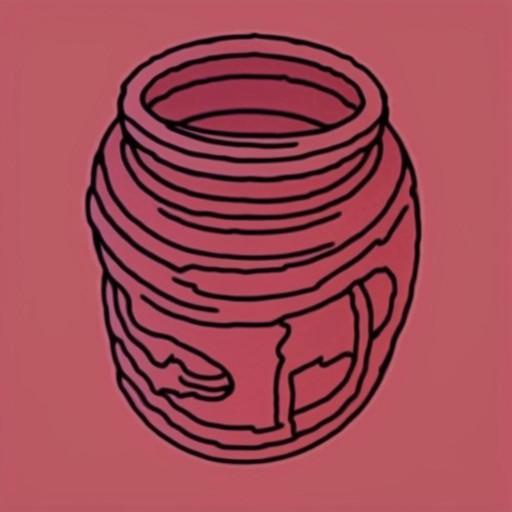} &
        \includegraphics[width=0.13\textwidth]{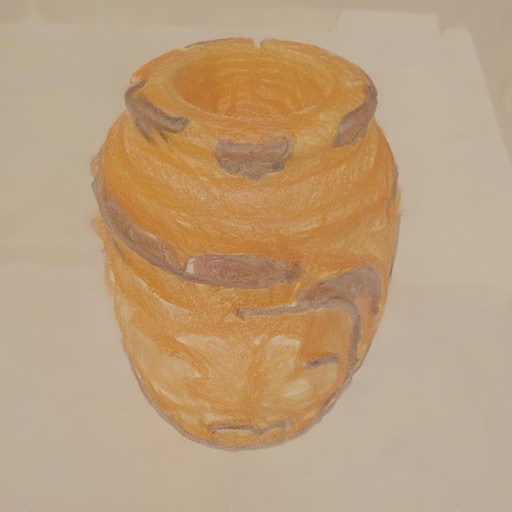} &
        \includegraphics[width=0.13\textwidth]{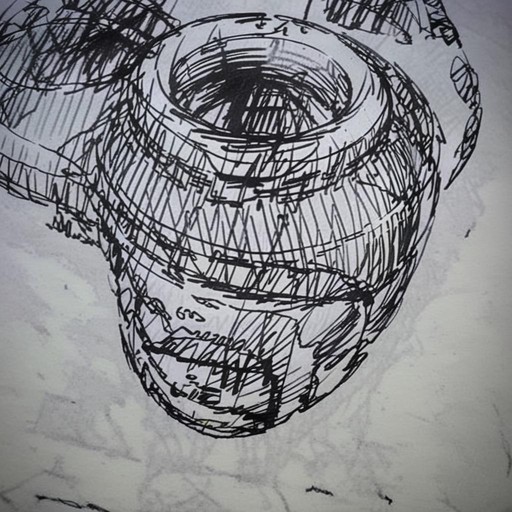} &
        \includegraphics[width=0.13\textwidth]{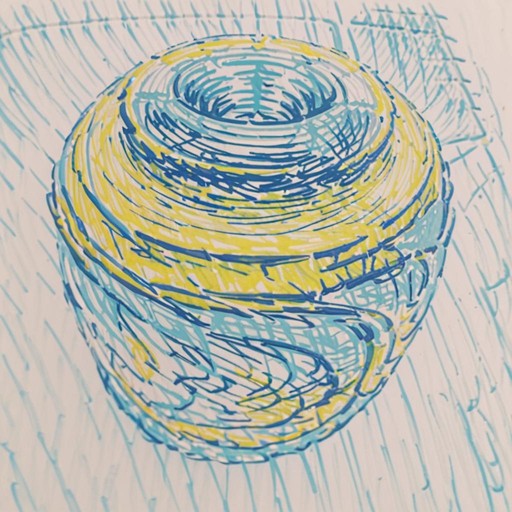} &
        \includegraphics[width=0.13\textwidth]{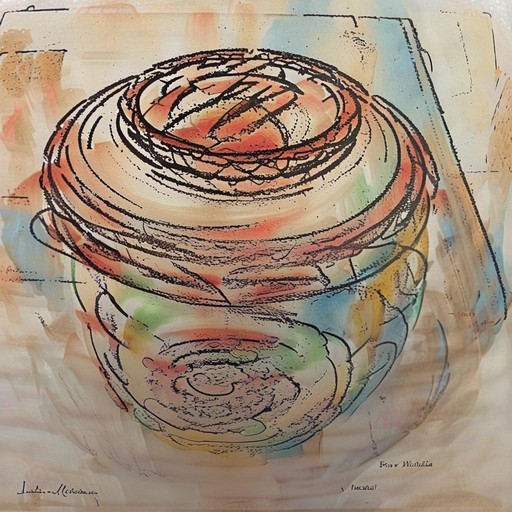} &
        \includegraphics[width=0.13\textwidth]{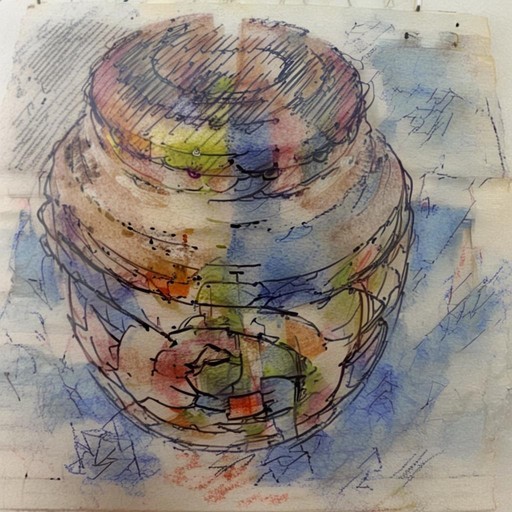} \\

        \includegraphics[width=0.13\textwidth]{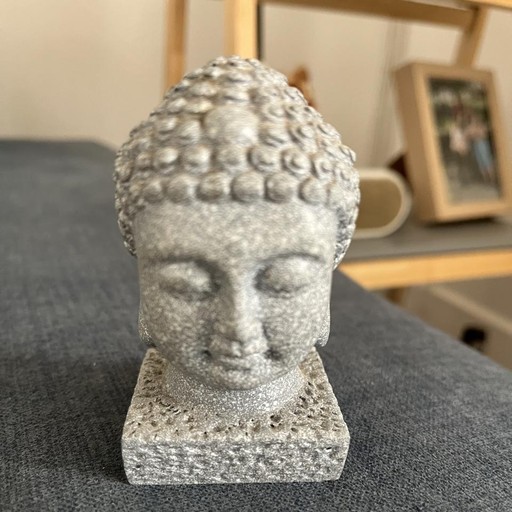} &
        \hspace{0.11cm}
        \includegraphics[width=0.13\textwidth]{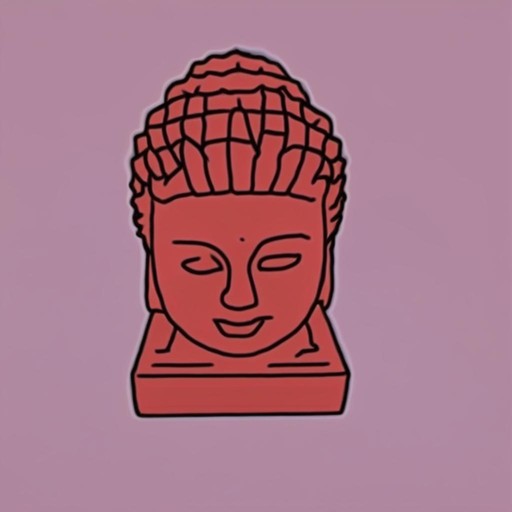} &
        \includegraphics[width=0.13\textwidth]{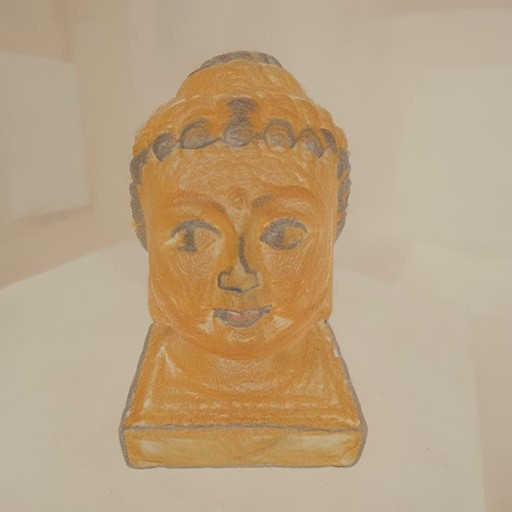} &
        \includegraphics[width=0.13\textwidth]{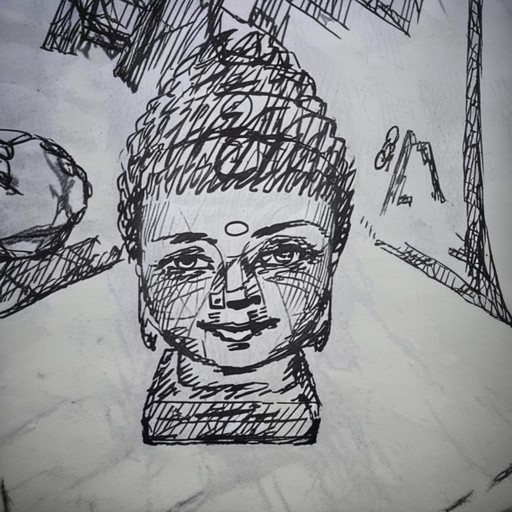} &
        \includegraphics[width=0.13\textwidth]{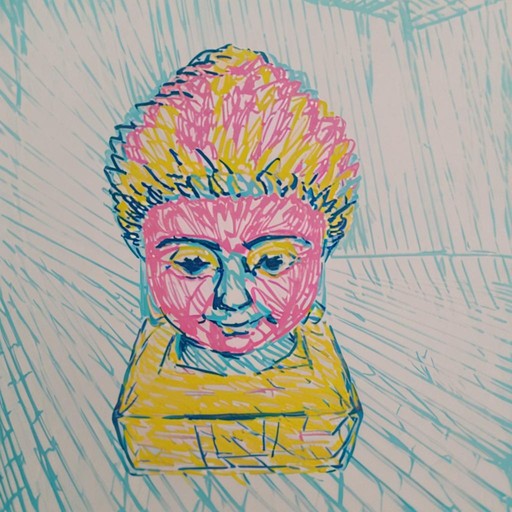} &
        \includegraphics[width=0.13\textwidth]{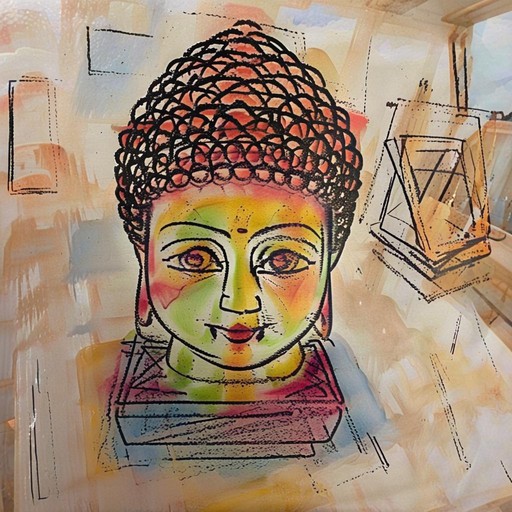} &
        \includegraphics[width=0.13\textwidth]{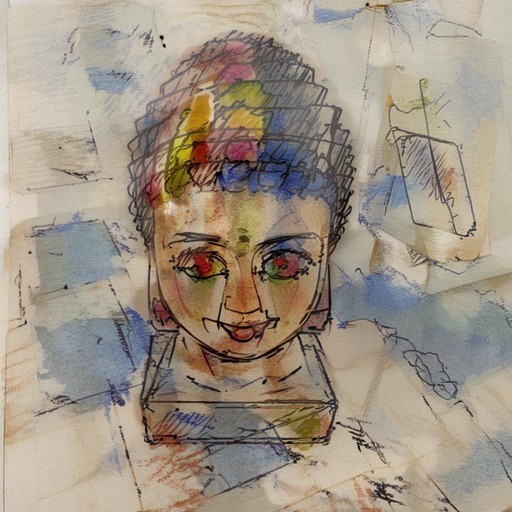} \\

        \includegraphics[width=0.13\textwidth]{temp_figs/content_images/teddybear.jpg} &
        \hspace{0.11cm}
        \includegraphics[width=0.13\textwidth]{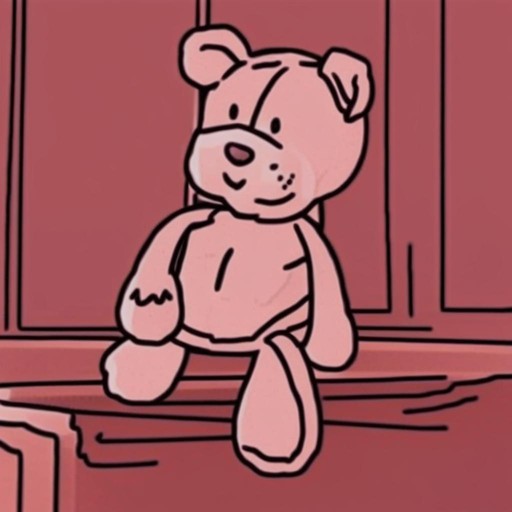} &
        \includegraphics[width=0.13\textwidth]{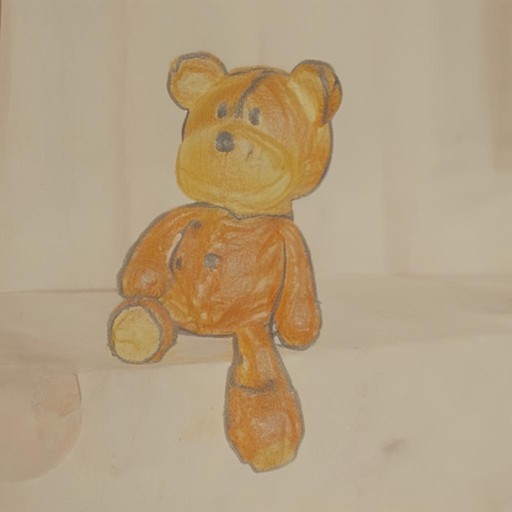} &
        \includegraphics[width=0.13\textwidth]{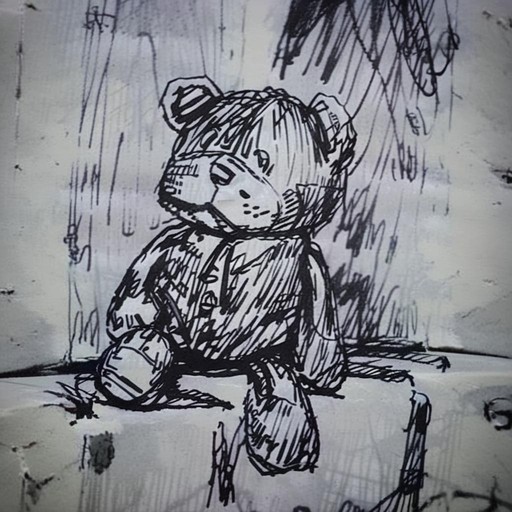} &
        \includegraphics[width=0.13\textwidth]{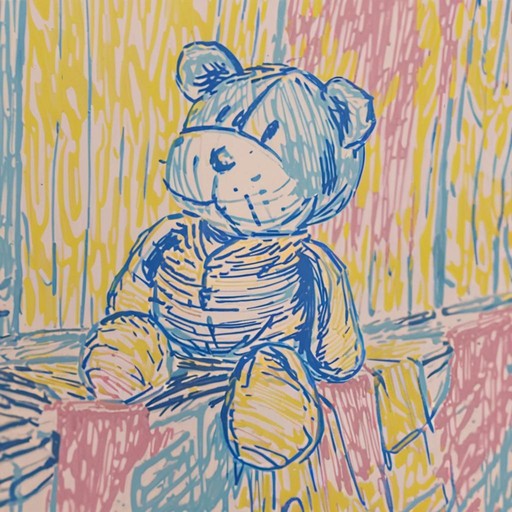} &
        \includegraphics[width=0.13\textwidth]{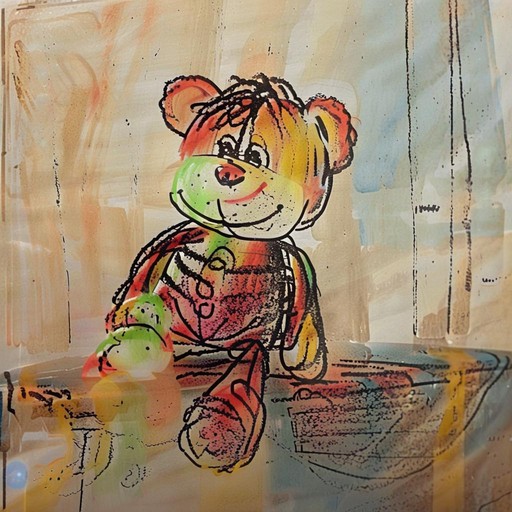} &
        \includegraphics[width=0.13\textwidth]{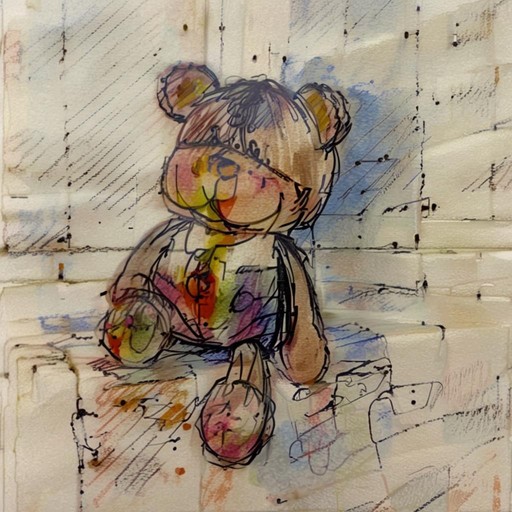} \\

        \includegraphics[width=0.13\textwidth]{temp_figs/content_images/sloth.jpg} &
        \hspace{0.11cm}
        \includegraphics[width=0.13\textwidth]{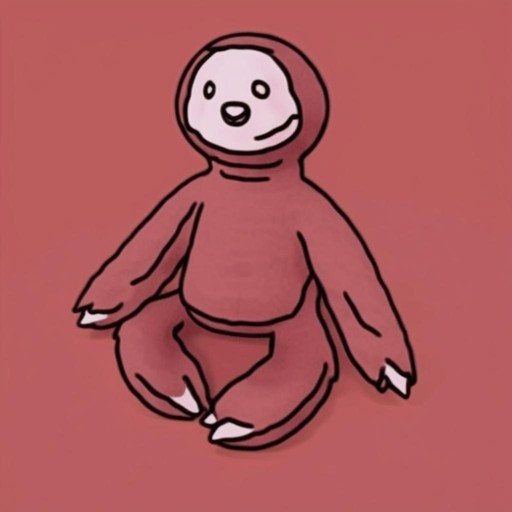} &
        \includegraphics[width=0.13\textwidth]{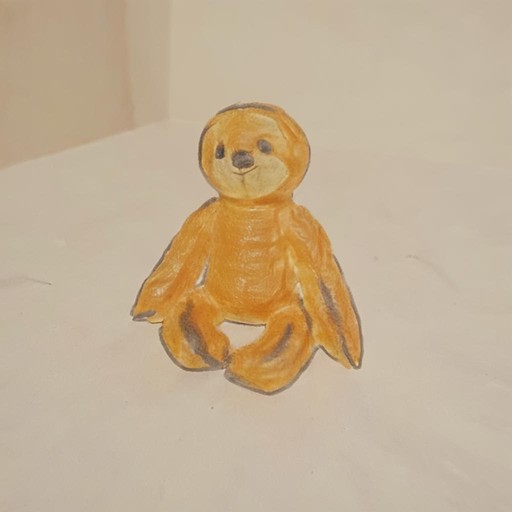} &
        \includegraphics[width=0.13\textwidth]{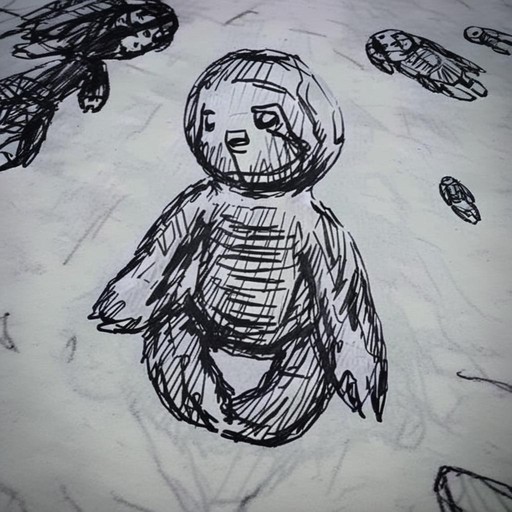} &
        \includegraphics[width=0.13\textwidth]{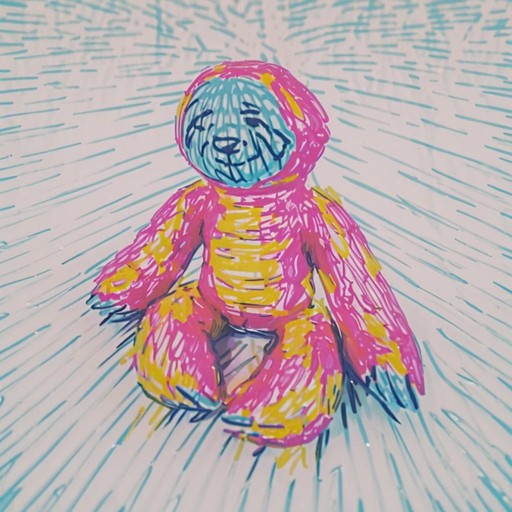} &
        \includegraphics[width=0.13\textwidth]{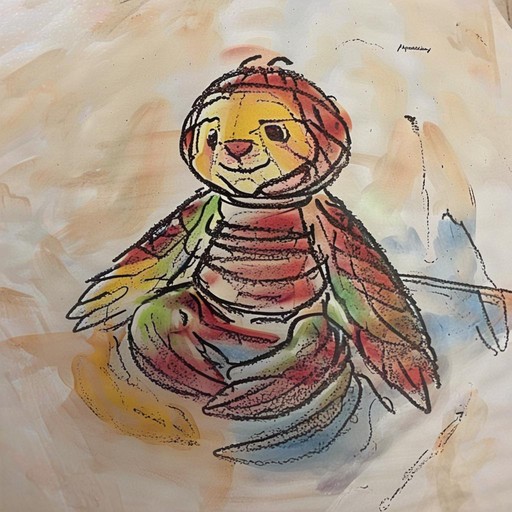} &
        \includegraphics[width=0.13\textwidth]{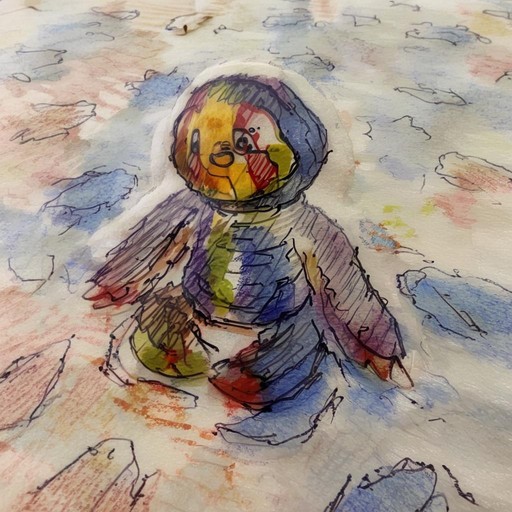} \\

    \end{NiceTabular}
    
    }
    \vspace{0.11cm}
    \caption{Image stylization based on image style reference using B-LoRA for randomly selected objects and styles. \copyright The paintings in the last two columns are by Judith Kondor Mochary.}
    \vspace{-0.2cm}
    \label{fig:random1}
\end{figure*}

\begin{figure*}
    \centering
    \setlength{\tabcolsep}{1.5pt}
    {\small
    \begin{NiceTabular}{c | @{\hspace{0.2cm}} c c c c c c}

        \diagbox{Input \\ Content}{Style} &
        \hspace{0.11cm}
        \includegraphics[width=0.13\textwidth]{temp_figs/style_images/painting.jpg} &

        \includegraphics[width=0.13\textwidth]{temp_figs/style_images/watercolor.png} &
        \includegraphics[width=0.13\textwidth]{temp_figs/style_images/drawing1.jpg} &
        \includegraphics[width=0.13\textwidth]{temp_figs/style_images/working_cartoon.jpg} &
        \includegraphics[width=0.13\textwidth]{temp_figs/style_images/cartoon_line.png} &
        \includegraphics[width=0.13\textwidth]{temp_figs/style_images/ink_sketch.jpeg} \\
        \midrule

        \includegraphics[width=0.13\textwidth]{temp_figs/content_images/child.jpg} &
        \hspace{0.11cm}
        \includegraphics[width=0.13\textwidth]{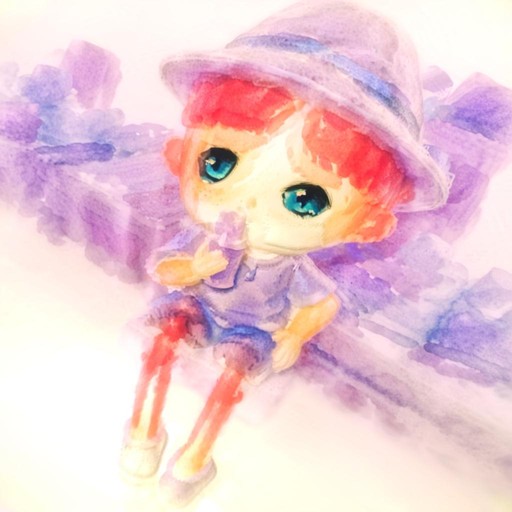} &
        \includegraphics[width=0.13\textwidth]{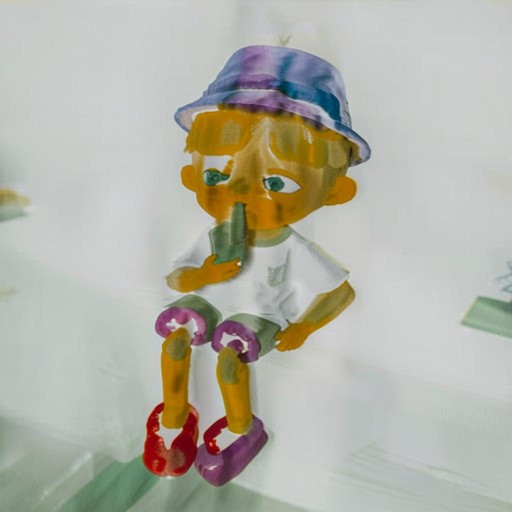} &
        \includegraphics[width=0.13\textwidth]{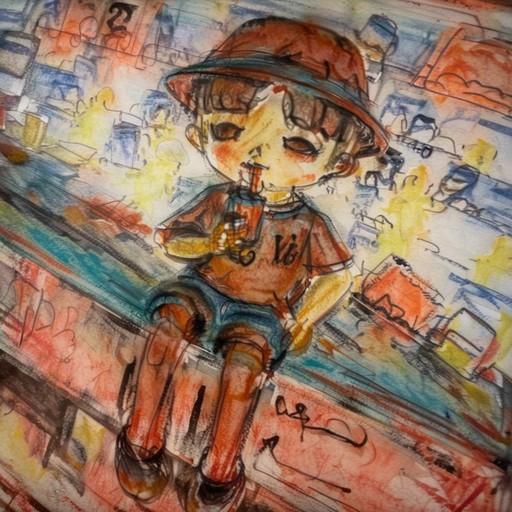} &
        \includegraphics[width=0.13\textwidth]{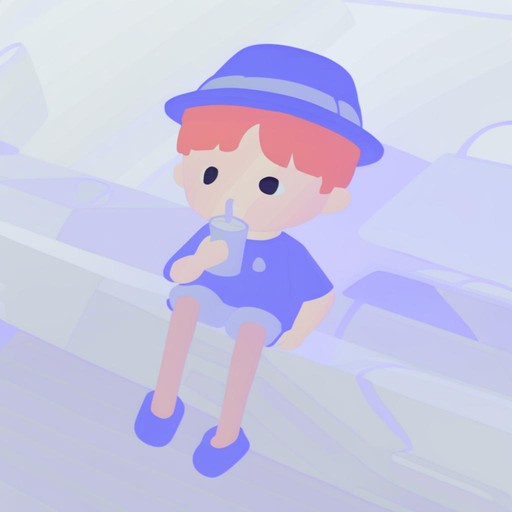} &
        \includegraphics[width=0.13\textwidth]{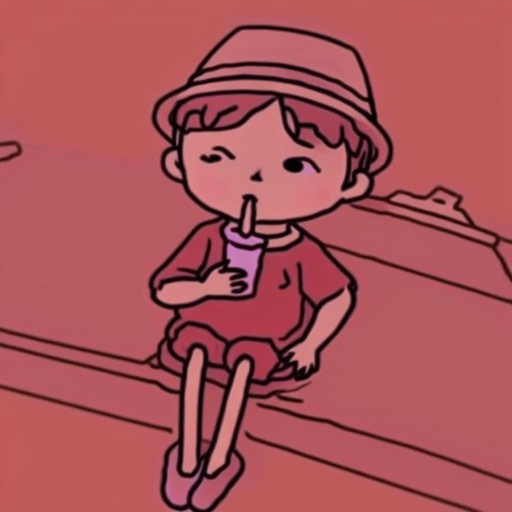} &
        \includegraphics[width=0.13\textwidth]{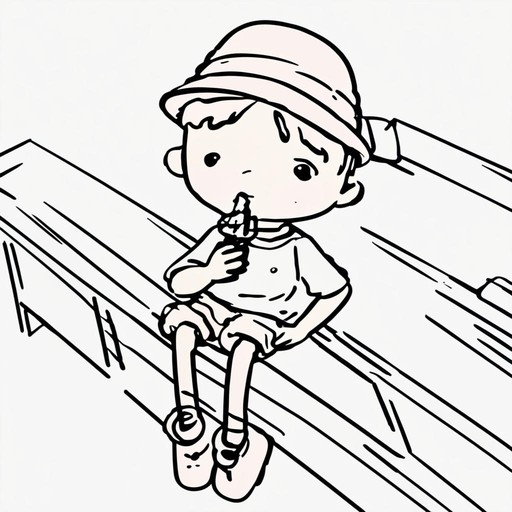} \\

        \includegraphics[width=0.13\textwidth]{temp_figs/content_images/scary_mug.jpg} &
        \hspace{0.11cm}
        \includegraphics[width=0.13\textwidth]{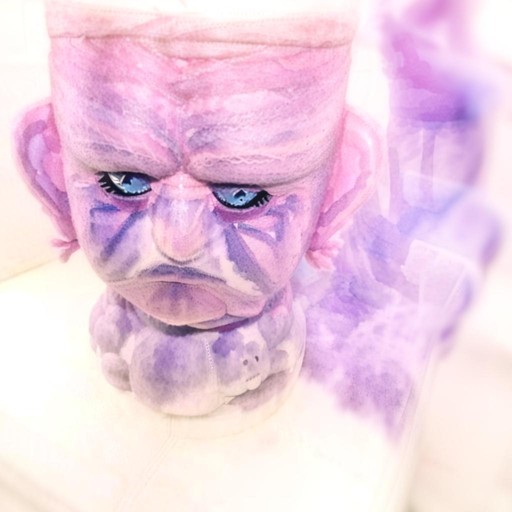} &
        \includegraphics[width=0.13\textwidth]{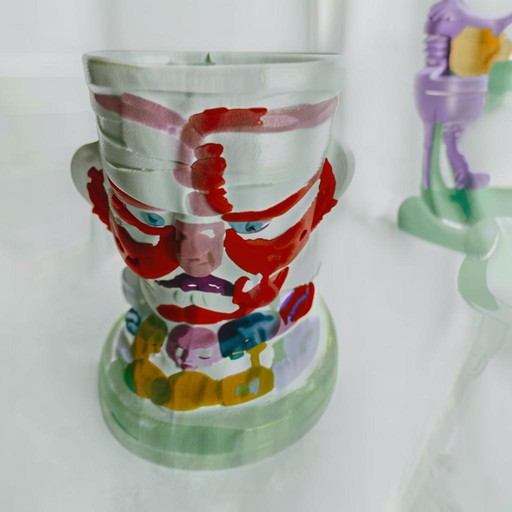} &
        \includegraphics[width=0.13\textwidth]{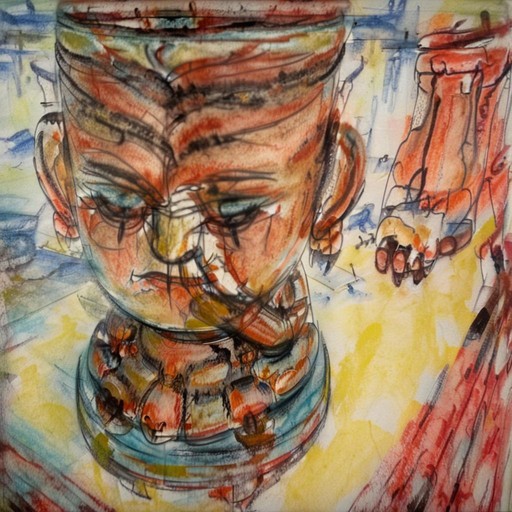} &
        \includegraphics[width=0.13\textwidth]{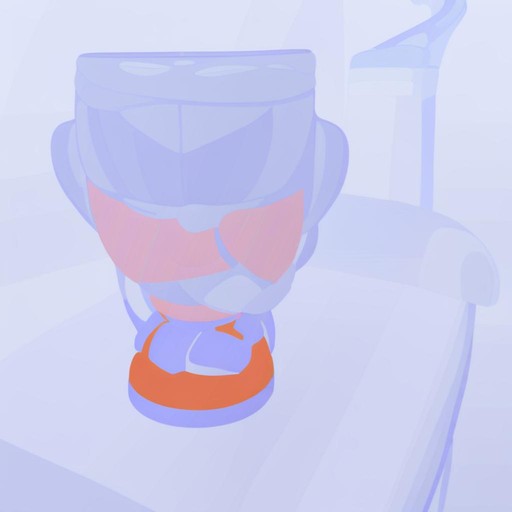} &
        \includegraphics[width=0.13\textwidth]{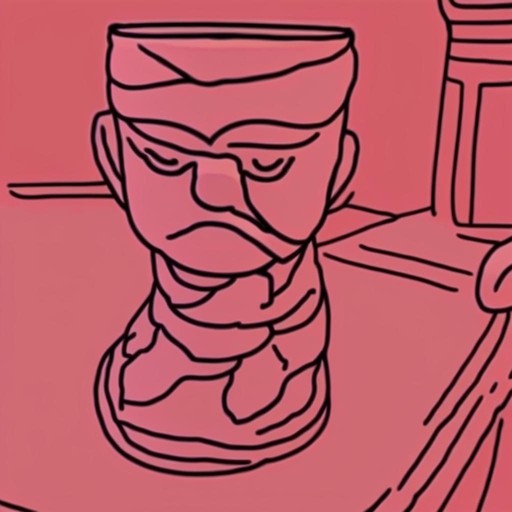} &
        \includegraphics[width=0.13\textwidth]{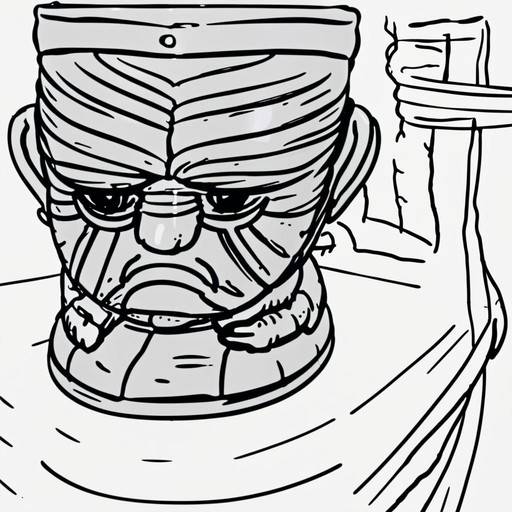} \\

        \includegraphics[width=0.13\textwidth]{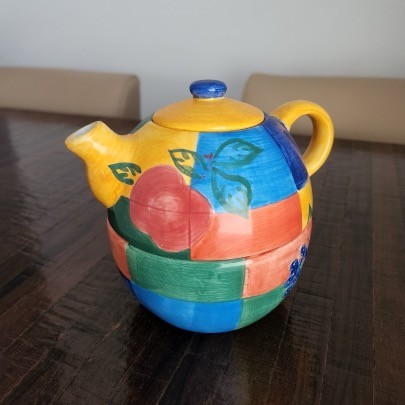} &
        \hspace{0.11cm}
        \includegraphics[width=0.13\textwidth]{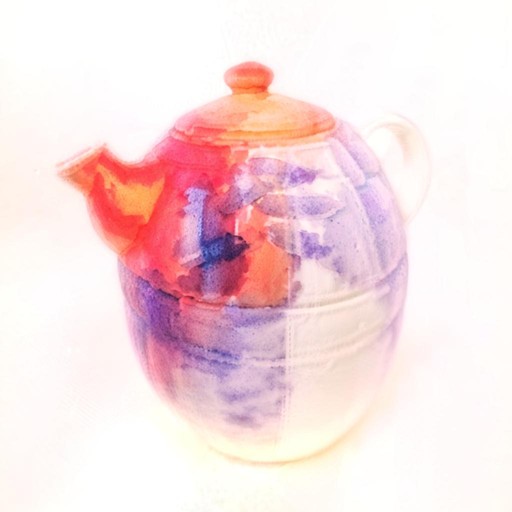} &
        \includegraphics[width=0.13\textwidth]{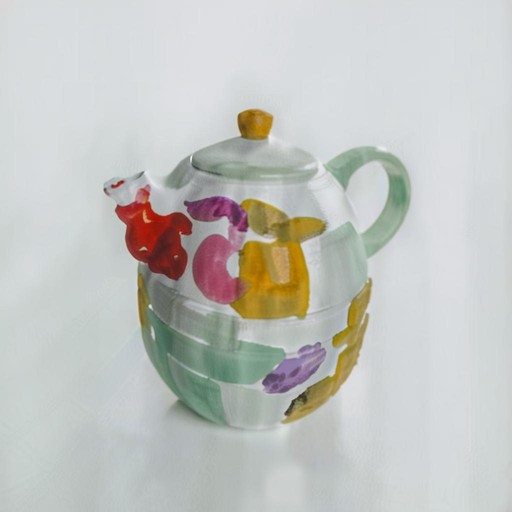} &
        \includegraphics[width=0.13\textwidth]{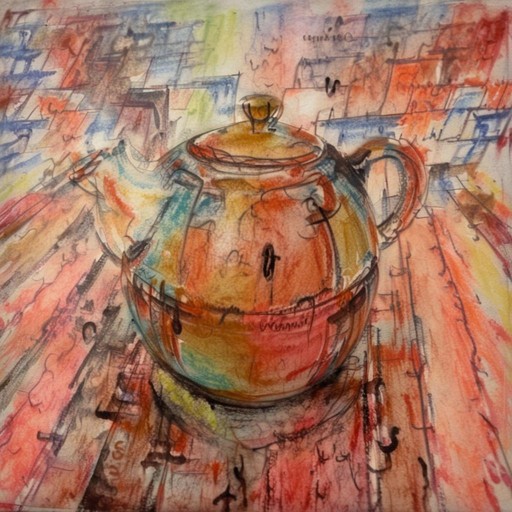} &
        \includegraphics[width=0.13\textwidth]{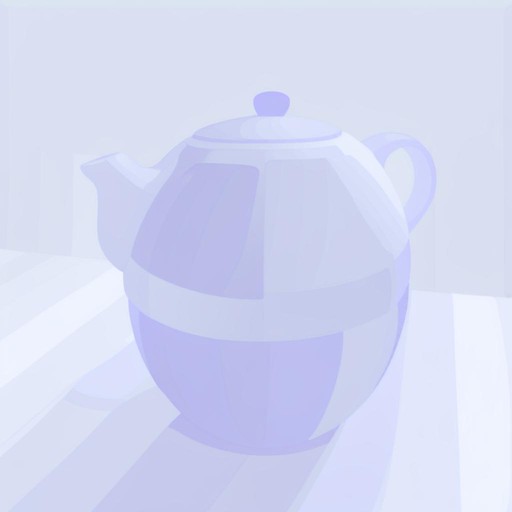} &
        \includegraphics[width=0.13\textwidth]{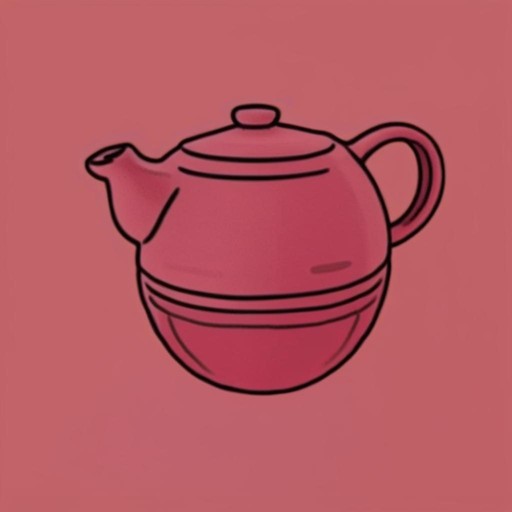} &
        \includegraphics[width=0.13\textwidth]{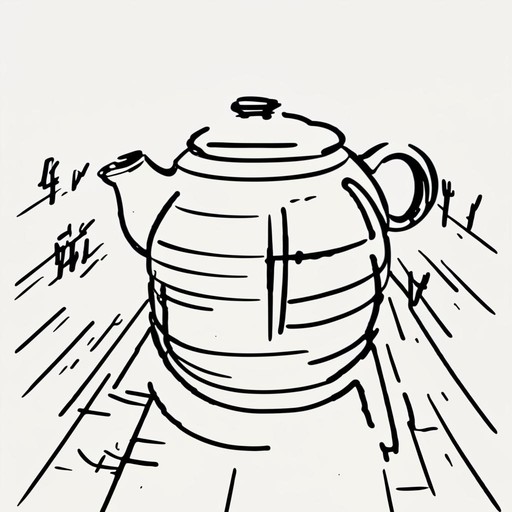} \\

        \includegraphics[width=0.13\textwidth]{temp_figs/content_images/bull.jpg} &
        \hspace{0.11cm}
        \includegraphics[width=0.13\textwidth]{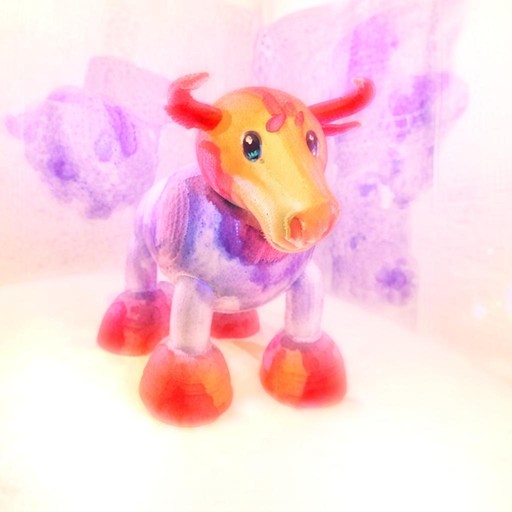} &
        \includegraphics[width=0.13\textwidth]{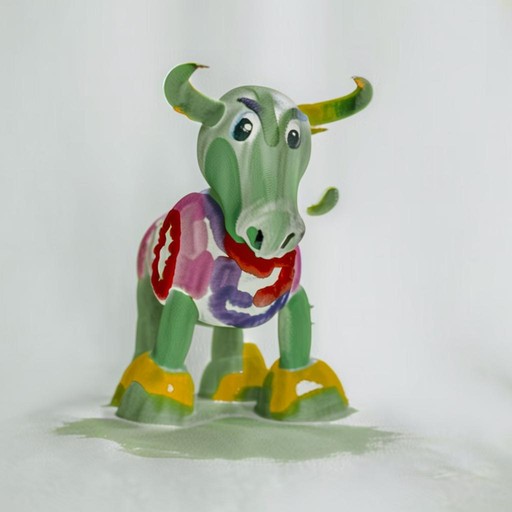} &
        \includegraphics[width=0.13\textwidth]{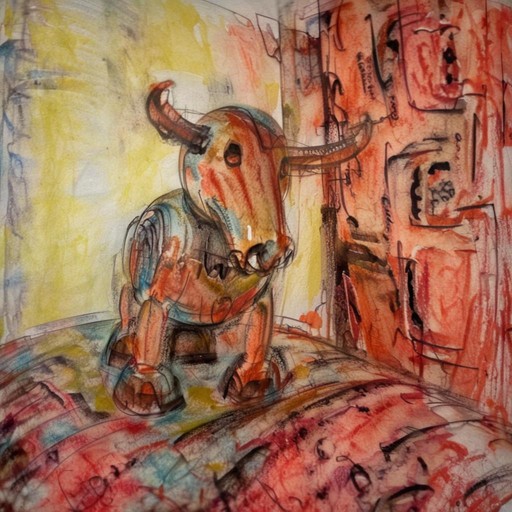} &
        \includegraphics[width=0.13\textwidth]{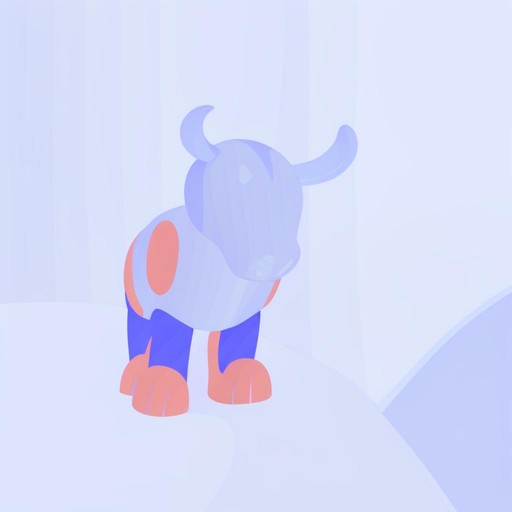} &
        \includegraphics[width=0.13\textwidth]{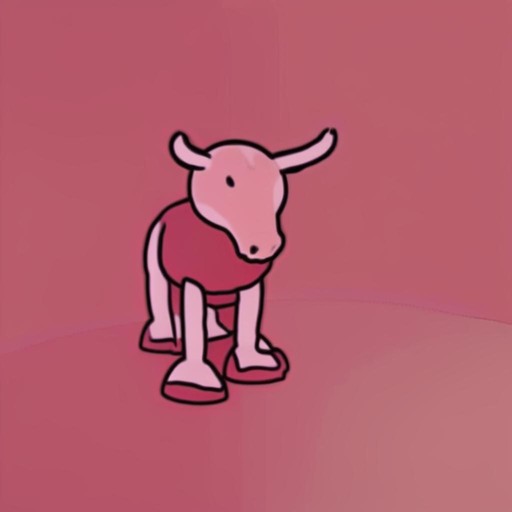} &
        \includegraphics[width=0.13\textwidth]{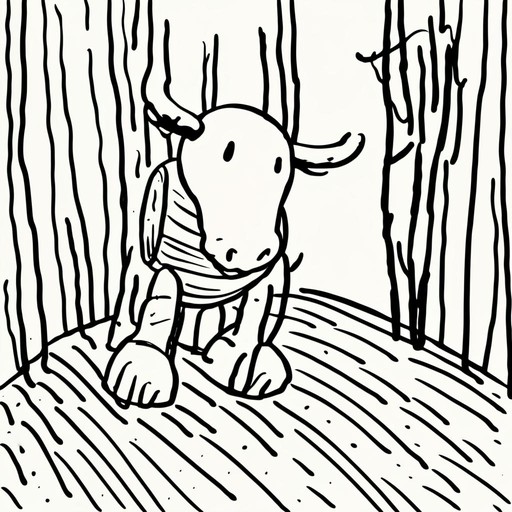} \\

        \includegraphics[width=0.13\textwidth]{temp_figs/content_images/statue.jpg} &
        \hspace{0.11cm}
        \includegraphics[width=0.13\textwidth]{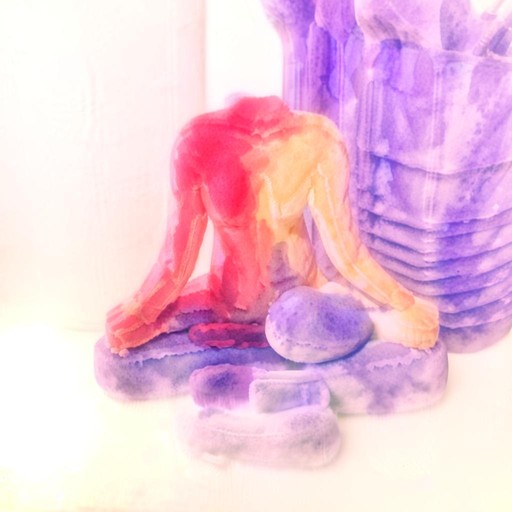} &
        \includegraphics[width=0.13\textwidth]{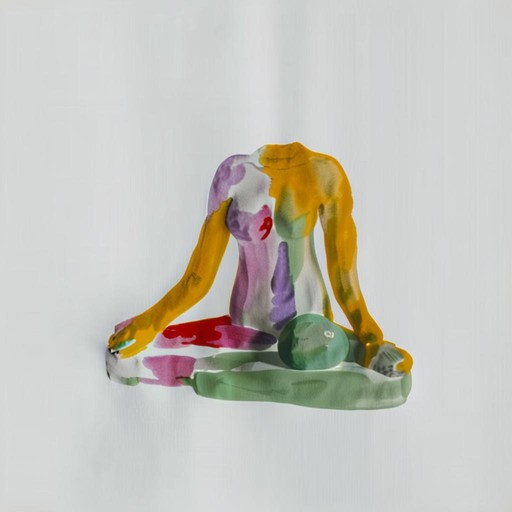} &
        \includegraphics[width=0.13\textwidth]{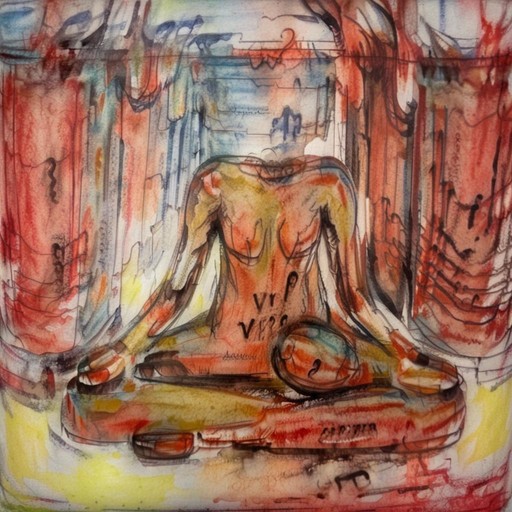} &
        \includegraphics[width=0.13\textwidth]{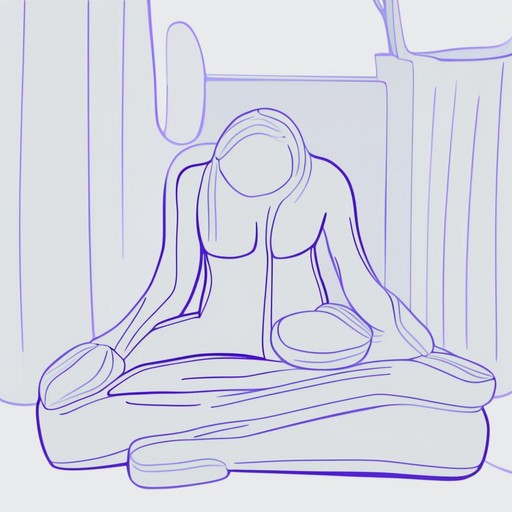} &
        \includegraphics[width=0.13\textwidth]{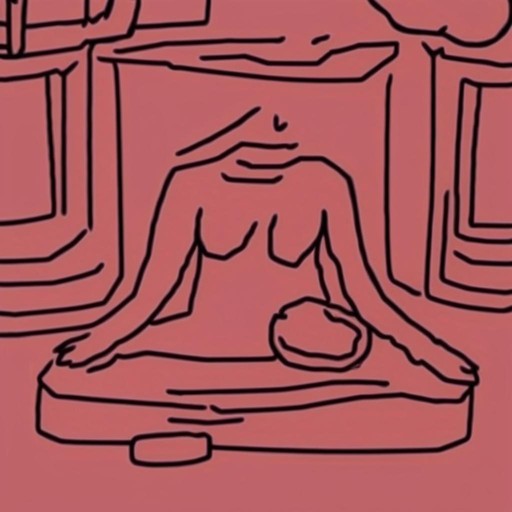} &
        \includegraphics[width=0.13\textwidth]{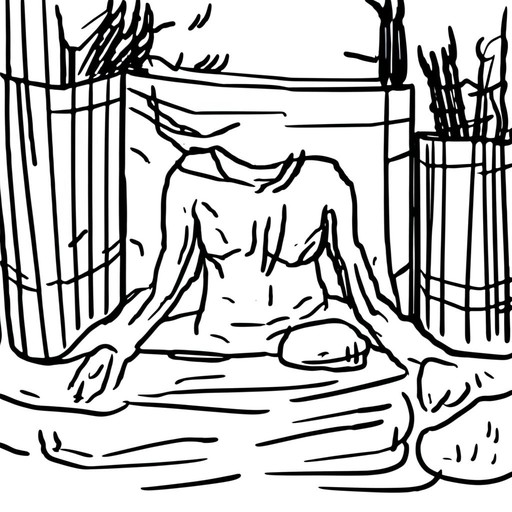} \\

        \includegraphics[width=0.13\textwidth]{temp_figs/content_images/dog2.jpg} &
        \hspace{0.11cm}
        \includegraphics[width=0.13\textwidth]{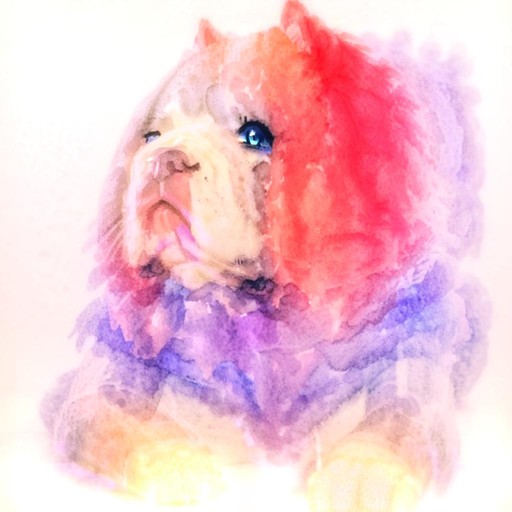} &
        \includegraphics[width=0.13\textwidth]{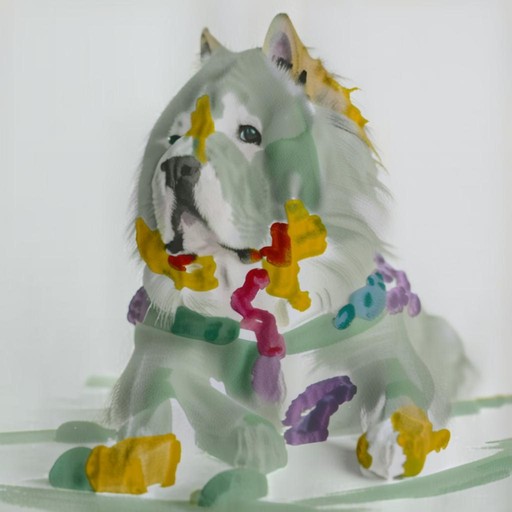} &
        \includegraphics[width=0.13\textwidth]{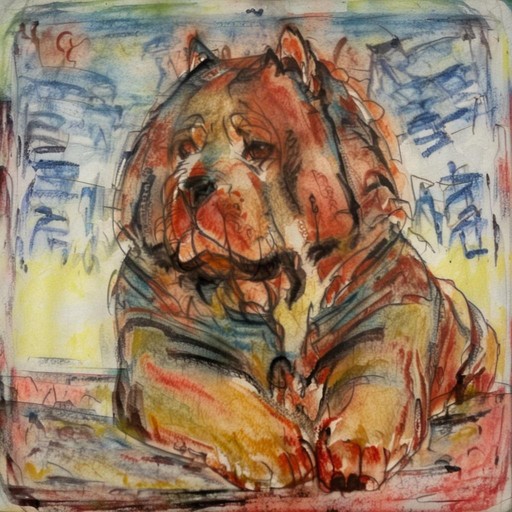} &
        \includegraphics[width=0.13\textwidth]{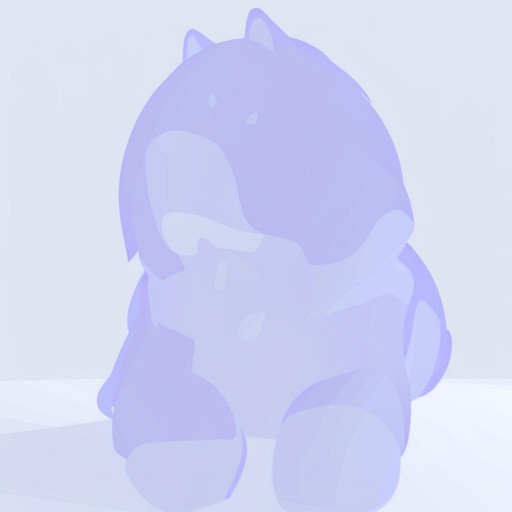} &
        \includegraphics[width=0.13\textwidth]{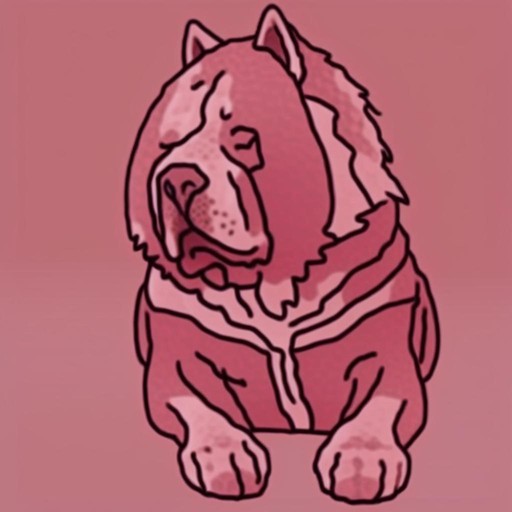} &
        \includegraphics[width=0.13\textwidth]{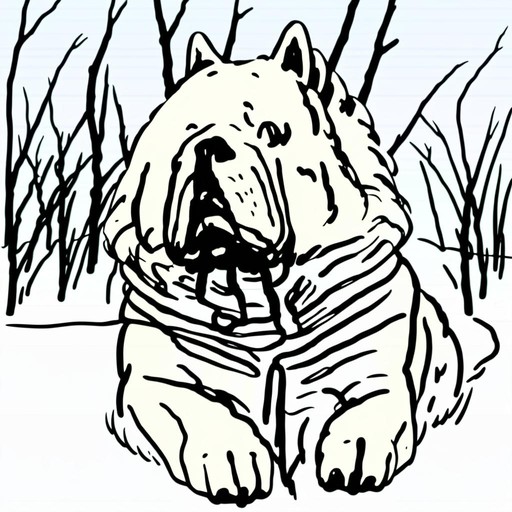} \\

         \includegraphics[width=0.13\textwidth]{temp_figs/content_images/metal_bird.jpg} &
        \hspace{0.11cm}
        \includegraphics[width=0.13\textwidth]{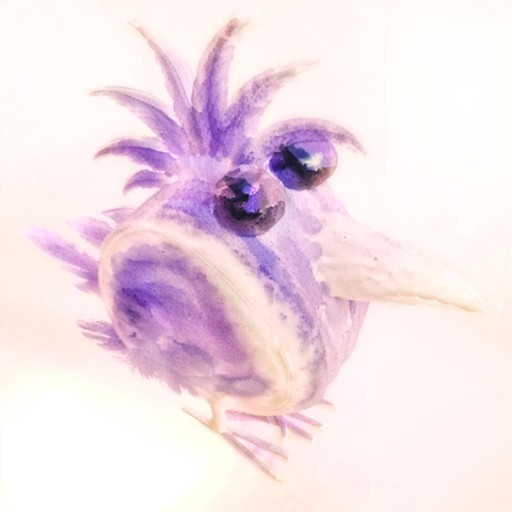} &
        \includegraphics[width=0.13\textwidth]{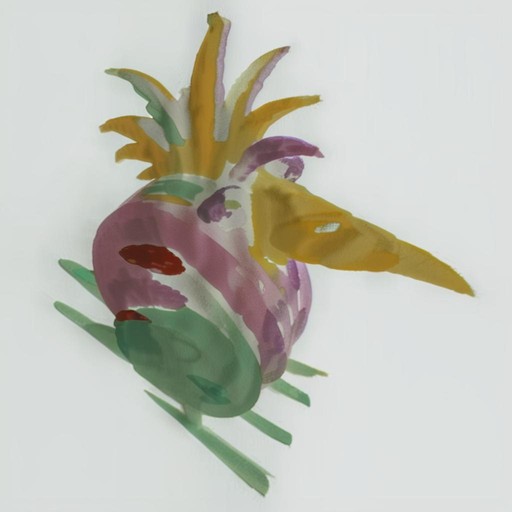} &
        \includegraphics[width=0.13\textwidth]{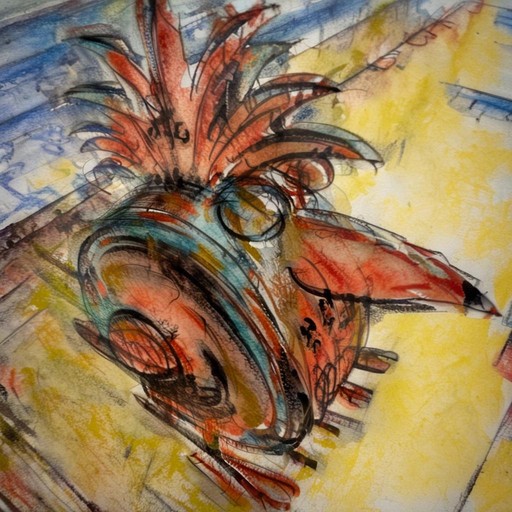} &
        \includegraphics[width=0.13\textwidth]{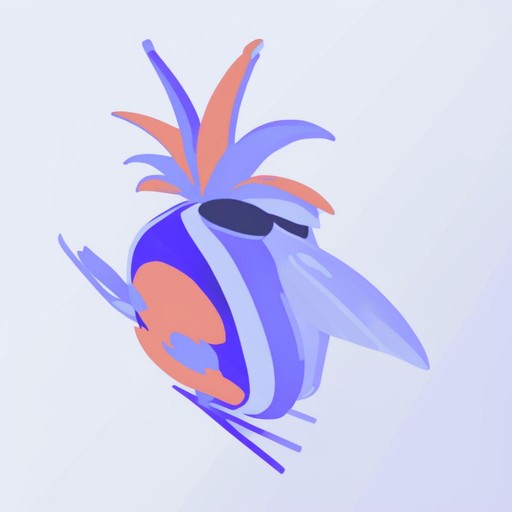} &
        \includegraphics[width=0.13\textwidth]{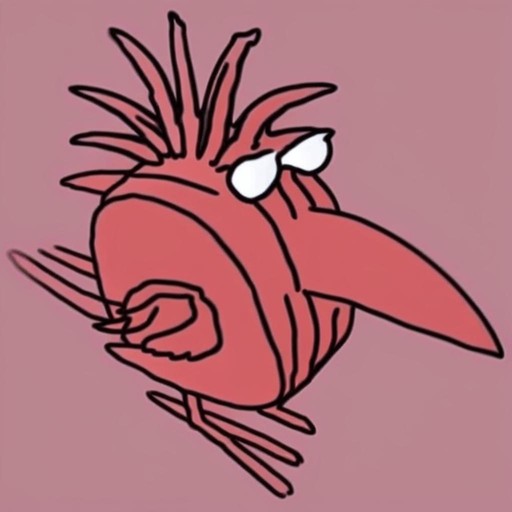} &
        \includegraphics[width=0.13\textwidth]{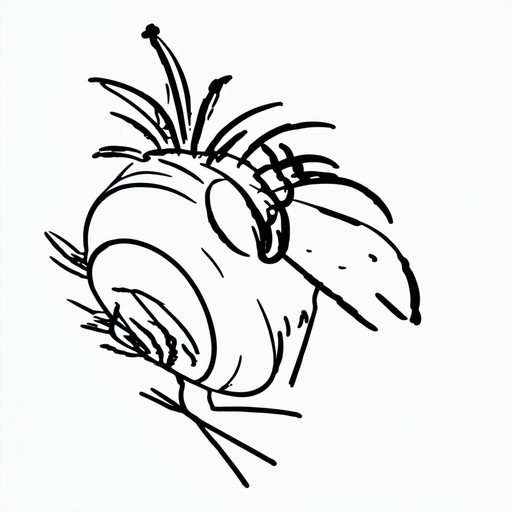} \\

    \end{NiceTabular}
    
    }
    \vspace{0.11cm}
    \caption{Image stylization based on image style reference using B-LoRA for randomly selected objects and styles.}
    \vspace{-0.2cm}
    \label{fig:random2}
\end{figure*}

\begin{figure*}
    \centering
    \setlength{\tabcolsep}{1.5pt}
    {\small
    \begin{tabular}{ c @{\hspace{0.2cm}} c c c c c}
        
        Content & \begin{tabular}[c]{@{}c@{}}``Made of\\ Gold'' \end{tabular} & \ap{... Wood} & \ap{... Glass} & \ap{... Wool} & \ap{... Steel} \\
        \includegraphics[width=0.15\textwidth]{temp_figs/content_images/teddybear.jpg} &
        
        \includegraphics[width=0.15\textwidth]{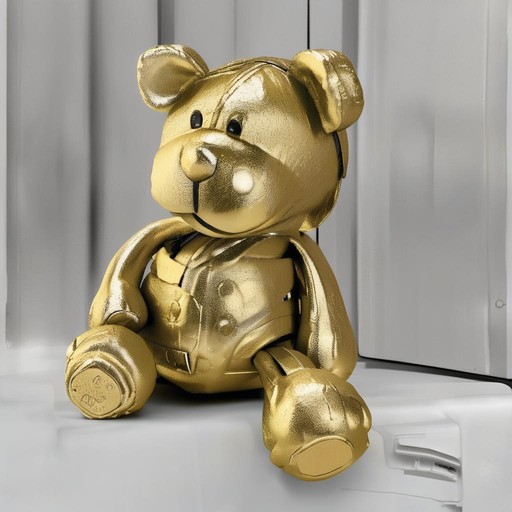} &
        \includegraphics[width=0.15\textwidth]{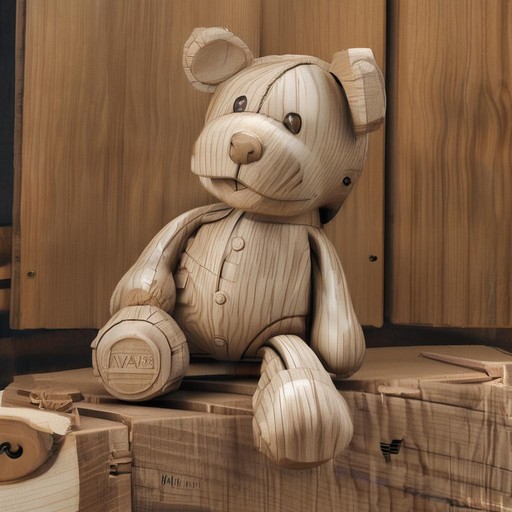} &
        \includegraphics[width=0.15\textwidth]{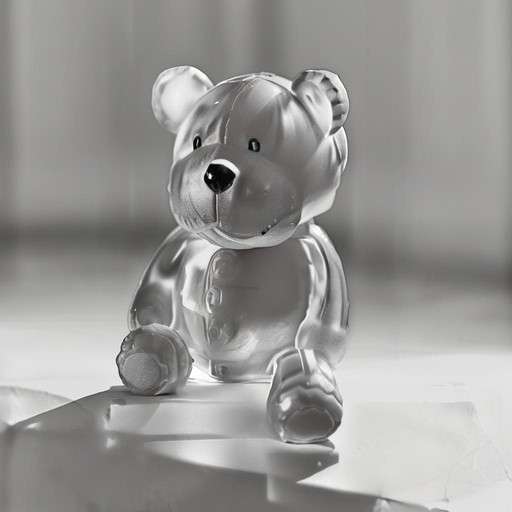} &
        \includegraphics[width=0.15\textwidth]{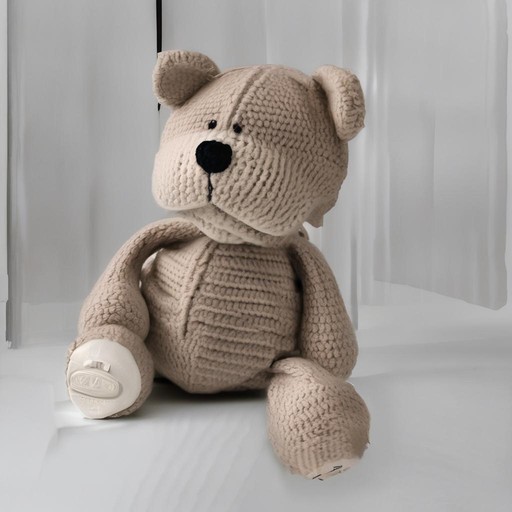} &
        \includegraphics[width=0.15\textwidth]{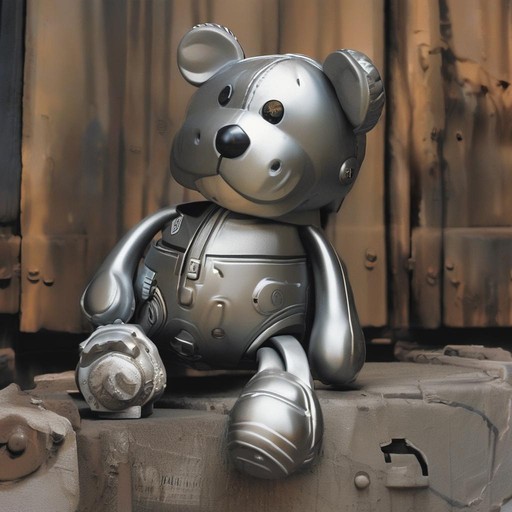} \\

        \includegraphics[width=0.15\textwidth]{temp_figs/content_images/metal_bird.jpg} &
        
        \includegraphics[width=0.15\textwidth]{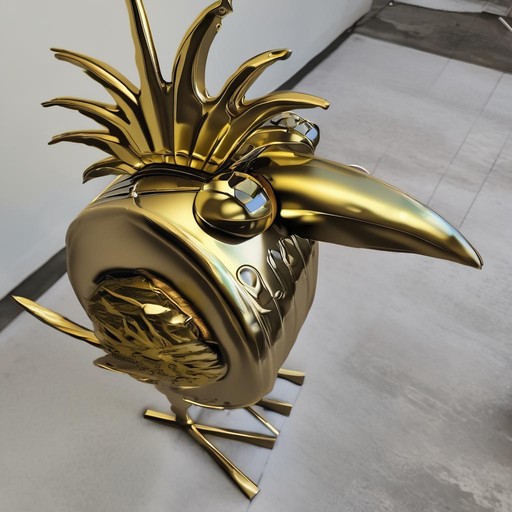} &
        \includegraphics[width=0.15\textwidth]{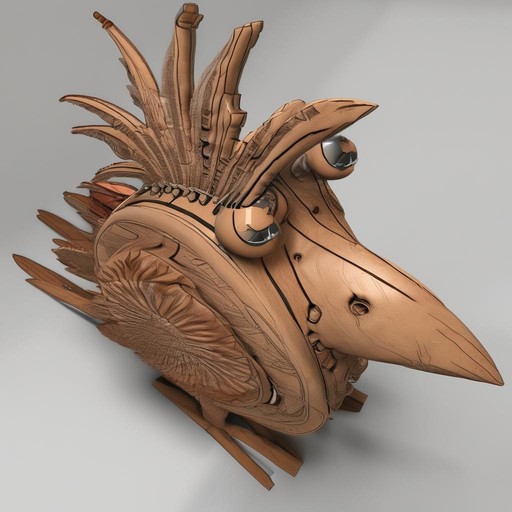} &
        \includegraphics[width=0.15\textwidth]{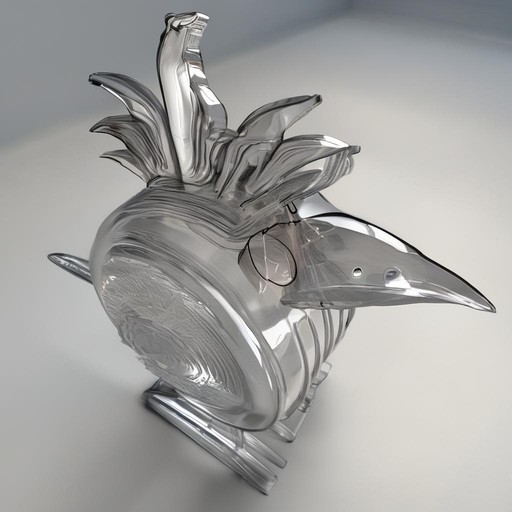} &
        \includegraphics[width=0.15\textwidth]{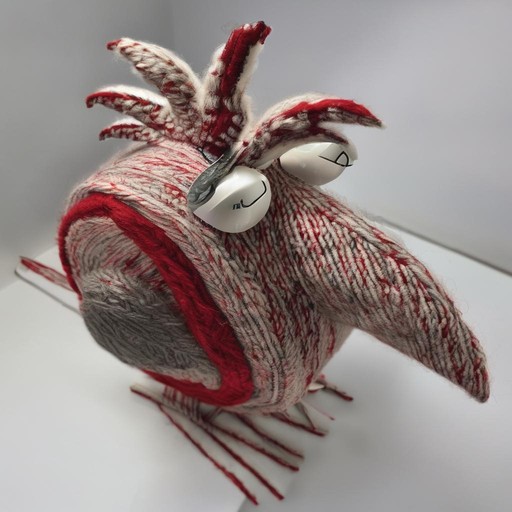} &
        \includegraphics[width=0.15\textwidth]{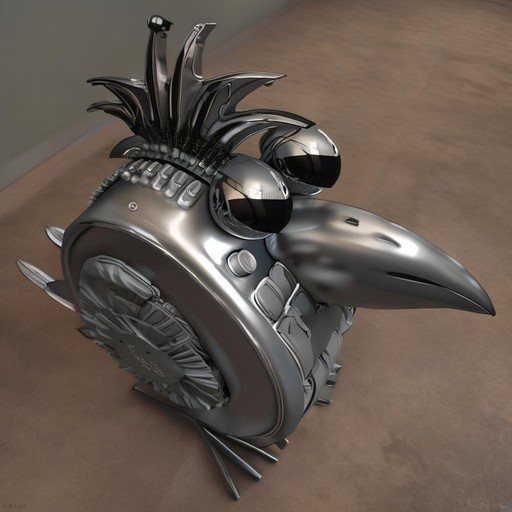} \\

        \includegraphics[width=0.15\textwidth]{temp_figs/content_images/bull.jpg} &
        
        \includegraphics[width=0.15\textwidth]{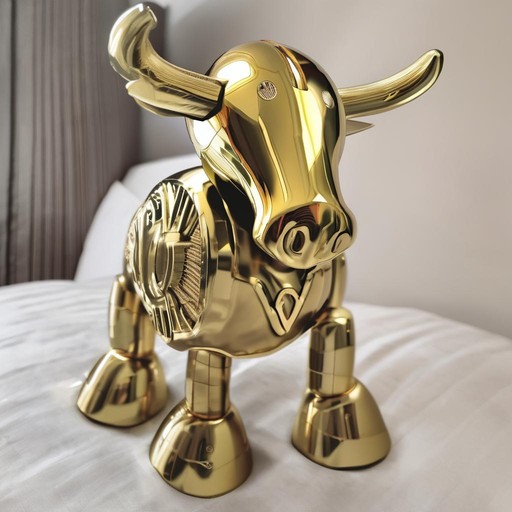} &
        \includegraphics[width=0.15\textwidth]{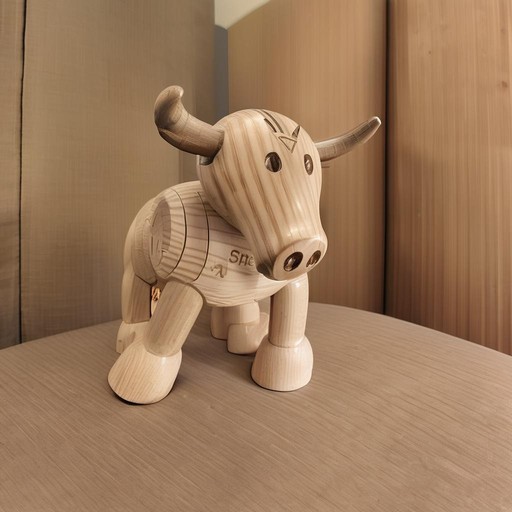} &
        \includegraphics[width=0.15\textwidth]{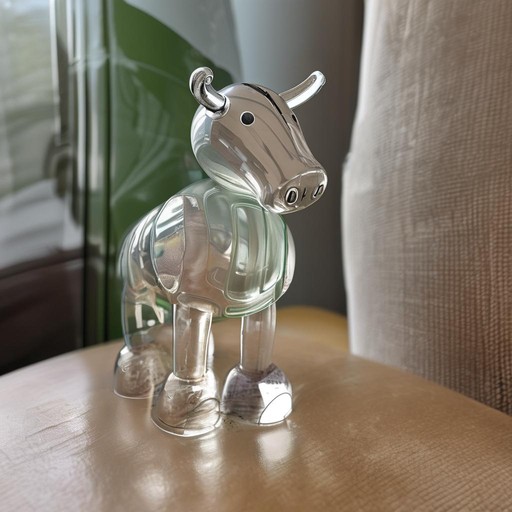} &
        \includegraphics[width=0.15\textwidth]{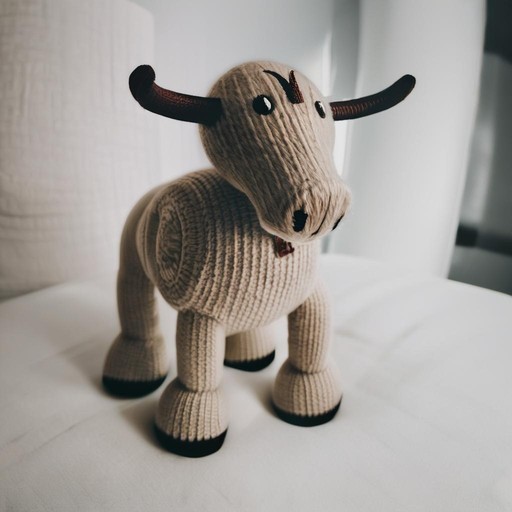} &
        \includegraphics[width=0.15\textwidth]{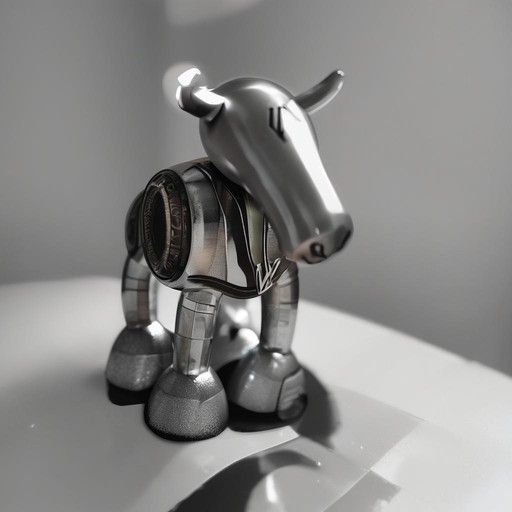} \\
        
        \includegraphics[width=0.15\textwidth]{temp_figs/content_images/fat_bird.jpg} &
        
        \includegraphics[width=0.15\textwidth]{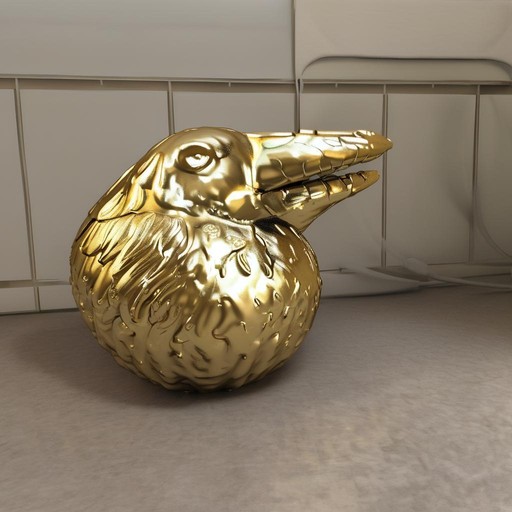} &
        \includegraphics[width=0.15\textwidth]{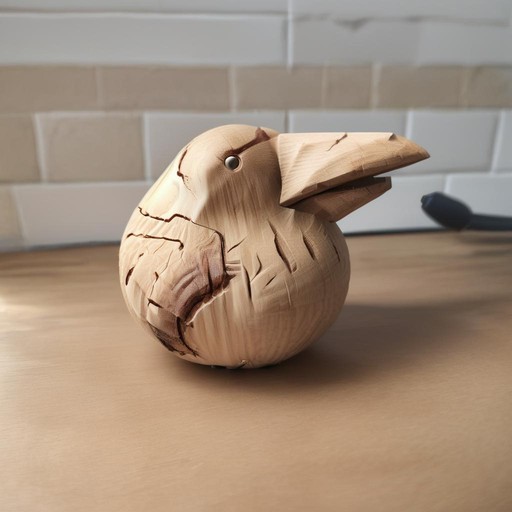} &
        \includegraphics[width=0.15\textwidth]{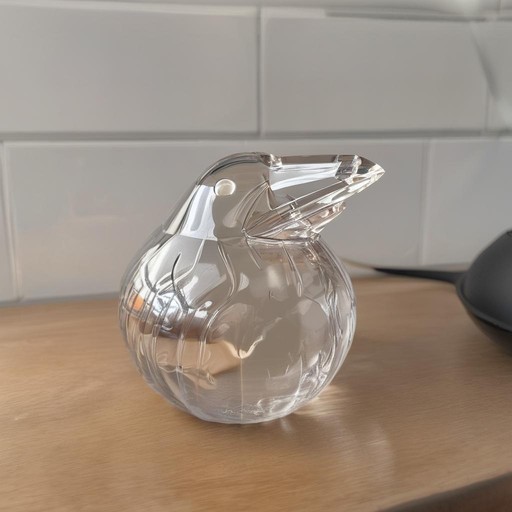} &
        \includegraphics[width=0.15\textwidth]{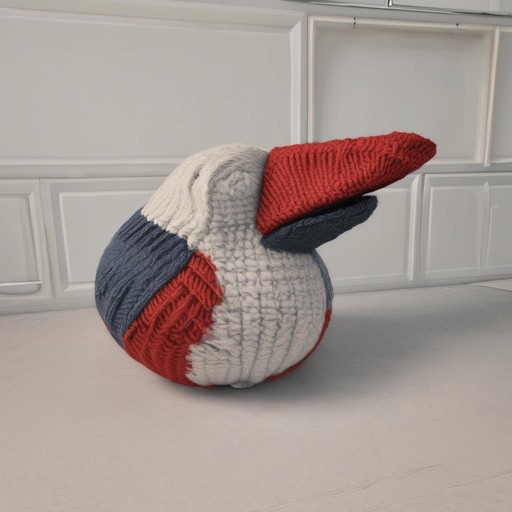} &
        \includegraphics[width=0.15\textwidth]{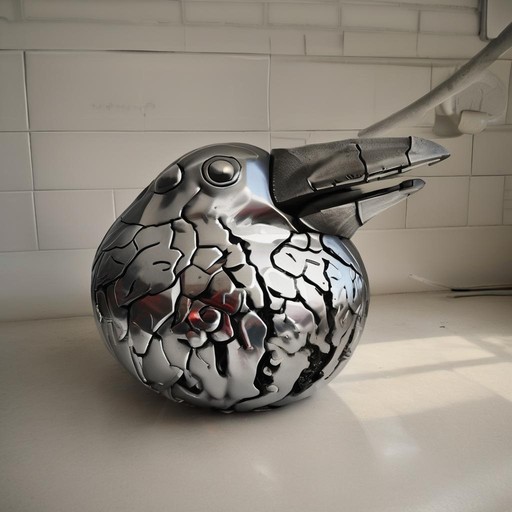} \\

        \includegraphics[width=0.15\textwidth]{temp_figs/content_images/buddha.jpg} &
        
        \includegraphics[width=0.15\textwidth]{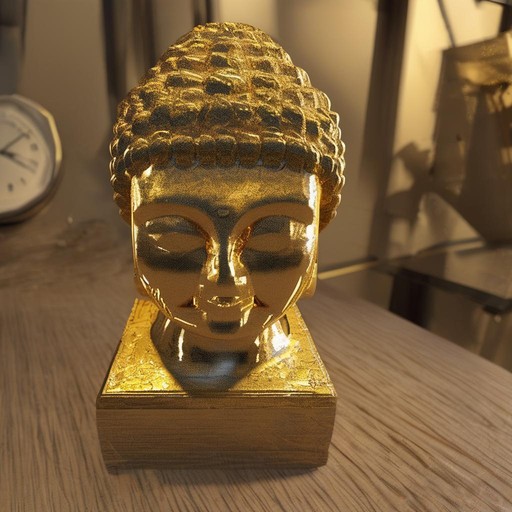} &
        \includegraphics[width=0.15\textwidth]{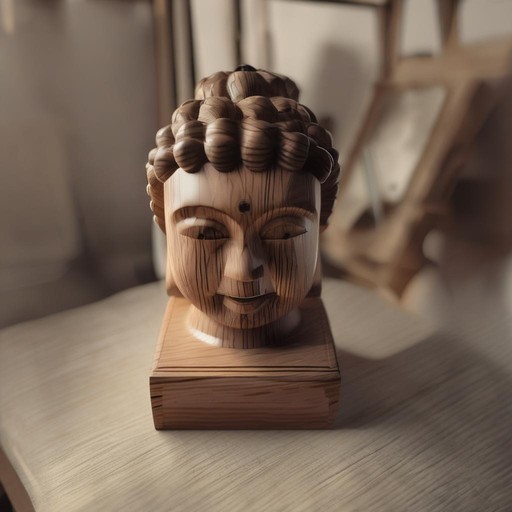} &
        \includegraphics[width=0.15\textwidth]{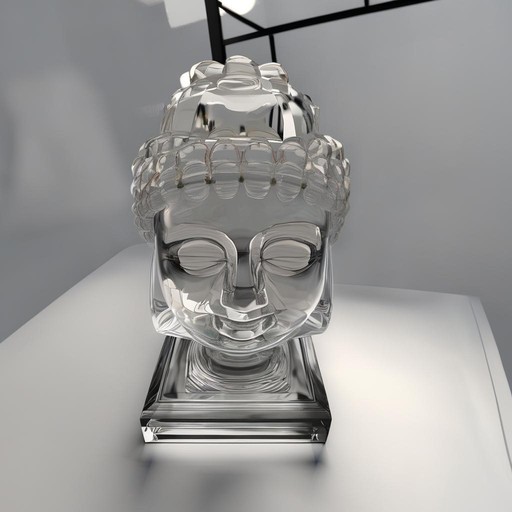} &
        \includegraphics[width=0.15\textwidth]{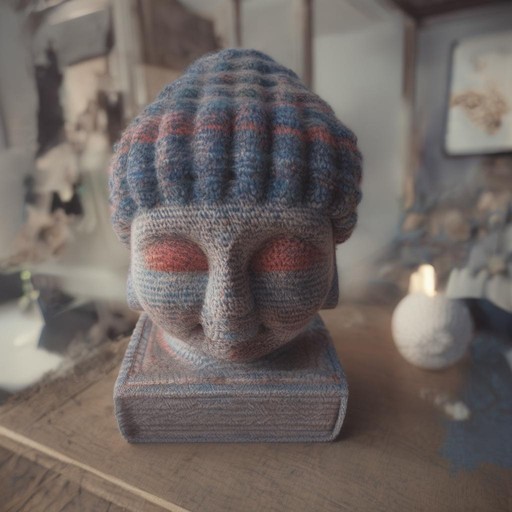} &
        \includegraphics[width=0.15\textwidth]{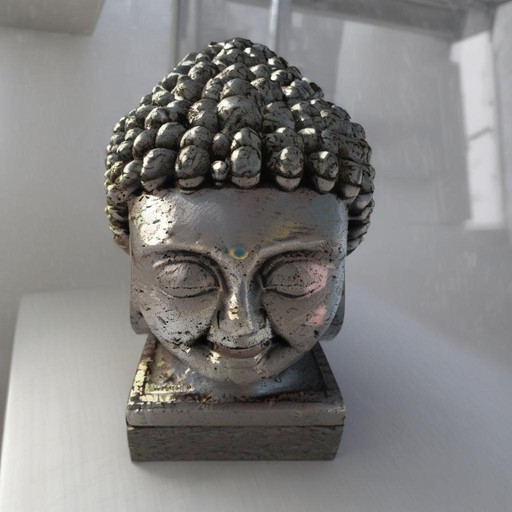} \\
        
        \includegraphics[width=0.15\textwidth]{temp_figs/content_images/scary_mug.jpg} &
        
        \includegraphics[width=0.15\textwidth]{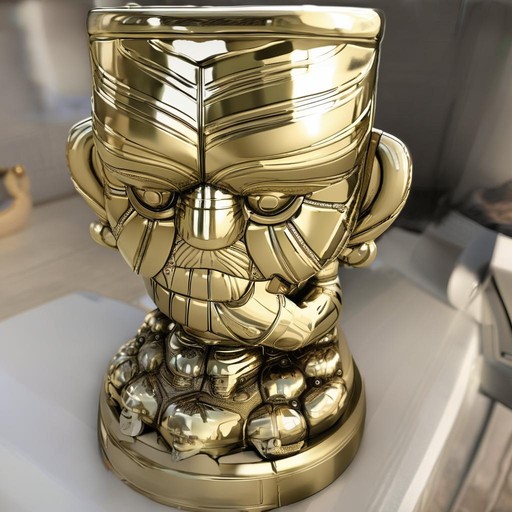} &
        \includegraphics[width=0.15\textwidth]{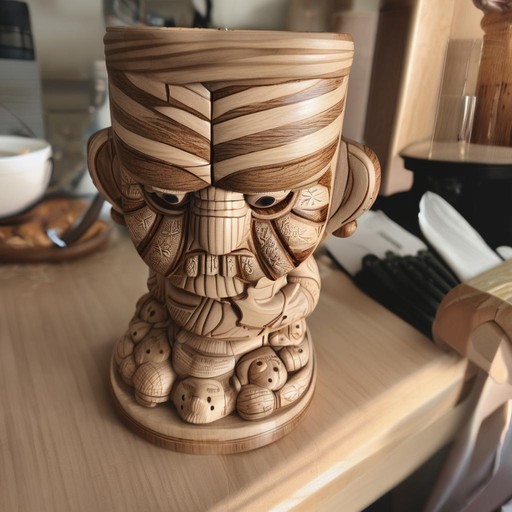} &
        \includegraphics[width=0.15\textwidth]{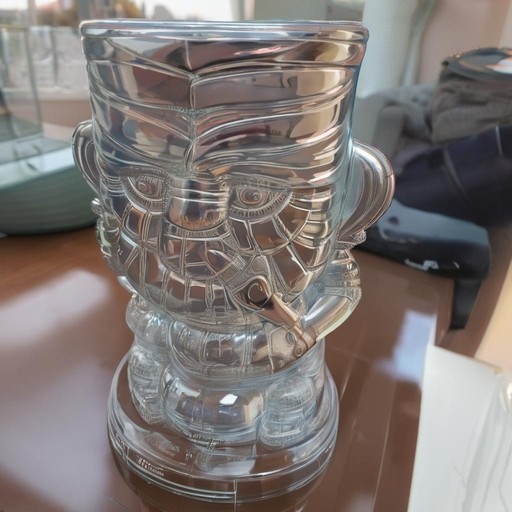} &
        \includegraphics[width=0.15\textwidth]{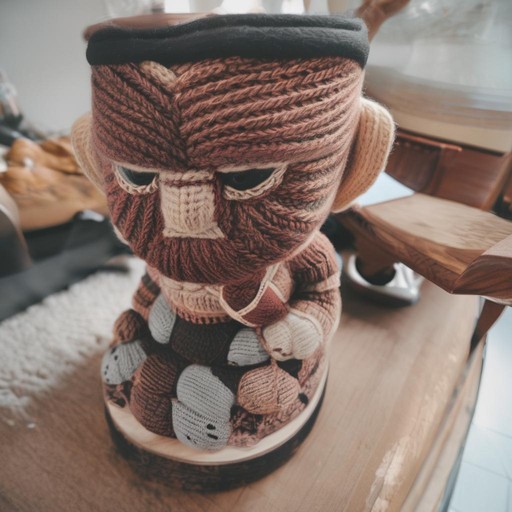} &
        \includegraphics[width=0.15\textwidth]{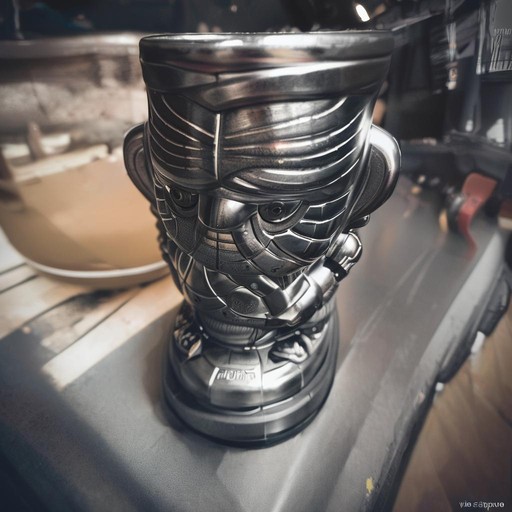} \\

    \end{tabular}
    
    }   
    \caption{Text-based Image stylization using B-LoRA, generated using the prompt \ap{A [v] made of ...}.}
    \label{fig:text_based_image_stylization}
\end{figure*}

Our B-LoRA method focuses on three main applications: image stylization based on image style references, text-based image stylization, and consistent style generation. In \Cref{fig:scene1,fig:scene2}, we present additional results generated by our approach for image stylization based on image style references. The columns represent the style reference images, while the rows correspond to the content reference images. As discussed, our method demonstrates proficiency in extracting content from style images (\Cref{fig:style_to_style}) and extracting style from objects for object mixing tasks (\Cref{fig:object_to_object}).
In \Cref{fig:random1,fig:random2}, we provide qualitative results showcasing our method's performance on randomly selected objects and styles from our evaluation set. These examples further highlight the robustness of our approach to handling diverse content and style references. In \Cref{fig:text_based_image_stylization} we present additional qualitative results for text-based image stylization. As discussed in the paper, by utilizing only the learned B-LoRA weights capturing the content, our method enables text-guided style manipulation while effectively preserving the input object's content and structure. These results demonstrate the flexibility of our approach in allowing challenging style manipulations through textual guidance.

\end{document}